Control Architecture and Design for a Multi-robotic Visual Servoing System in Automated Manufacturing Environment

By

RONGFEI LI

DISSERTATION

Submitted in partial satisfaction of the requirements for the degree of

Doctor of Philosophy

in

Mechanical and Aerospace Engineering

in the

OFFICE OF GRADUATE STUDIES

of the

UNIVERSITY OF CALIFORNIA

DAVIS

Approved:

_________________________________________
Francis Assadian, Chair

_________________________________________
Iman Sotani

_________________________________________
Zhaodan Kong

Committee in Charge

2025

*To my Family.*

# Contents

# List of Figures

# List of Tables

# Abstract

The use of robotic technology has drastically increased in manufacturing in the 21st century. But by utilizing their sensory cues, humans still outperform machines, especially in the micro scale manufacturing, which requires high-precision robot manipulators. These sensory cues naturally compensate for high level of uncertainties that exist in the manufacturing environment. Uncertainties in performing manufacturing tasks may come from measurement noise, model inaccuracy, joint compliance (e.g., elasticity) etc. Although advanced metrology sensors and high-precision microprocessors, which are utilized in nowadays robots, have compensated for many structural and dynamic errors in robot positioning, but a well-designed control algorithm still works as a comparable and cheaper alternative to reduce uncertainties in automated manufacturing. Our work illustrates that a multi-robot control system that simulates the positioning process for fastening and unfastening applications can reduce various uncertainties, which may occur in this process, to a great amount. In addition, all research papers in visual servoing mainly put efforts into developing control and observation architectures in various scenarios, but none of them has discussed the importance of the camera's location in the configuration. In a manufacturing environment, the quality of camera estimations may vary significantly from one observation location to another, as the combined effects of environ-mental conditions result in different noise levels of a single image shot at different locations. Therefore, in this paper, we also propose a novice algorithm for the camera's moving policy so that it explores the camera workspace and searches for the optimal location where the images' noise level is minimized.

# Acknowledgements

First and foremost, I would like to give many thanks to my primary advisor, Professor Farhad Assadian. He is incredibly knowledgeable and experienced in control system design and has provided me with great help in this work. In addition to that, I will never forget the time and effort he has put into discussions every time in his office. We have come up with a lot of interesting ideas that laid the groundwork for this dissertation. It is a quite pleasant and rewarding experience for me to work with him.

I would also like to thank Prof. Iman Soltani and Prof. Zhaodan Kong for serving on my thesis committee for their help, advice, and suggestions with this dissertation.

Moreover, I am deeply grateful to Ford Motor Company for supporting this project through grants. Without them, the success of this research is not possible. I would like to offer my special thanks to Dr Devesh Upadhyay and Dr. Mahdi Naddaf, and the whole AI-ML-QC method team. Devesh gave guidance and advice to me during my summer internship at Ford company. Although it was hard to communicate during the pandemic time, we managed to accomplish many useful works especially in the robotic operational system. I would like to thank Mahdi for serving as a member of my qualification exam committee.

To my parents, you are always supportive and encouraging during my PhD career. We haven't seen each other much during the pandemic, but I want to say I love you both and cannot wait to see you in the future. To my church, you guys are amazing and make me always feel like in a family. Thanks for all the prayers and love you shared with me during my hard time.

Lastly, I would like to thank you, my lord Jesus Christ. I am grateful to you for letting me know you 10 years ago. Now I am reborn as a Christian and live up to the best of my days. I will always humble myself and see you as my Godly father.

# Nomenclature

**Acronyms**

**2D** *Two-dimension.*
3D *Three-dimension.*

**BIBO** *Bounded input and bounded output.*

**DC motor** *Direct current motor.*
**DoF** *Degree of freedom.*
**D-H convention** *Denavit-Hartenberg convention.*

**HIL** *Hardware-In-Loop.*

**IBVS** *Imaged based visual servoing.*

**LDP** *Location determination problem.*

**MIMO** *Multiple input and multiple output.*

**PBVS** *Position based visual servoing.*
**PID** *Proportional Integral Derivative.*
**PnP** *Perspective-n-point problem.*

**SISO** *Singular input and singular output.*
**SVD** *Singular value decomposition.*

**VS** *Visual servoing.*

___________________________________________________________Chapter 1

# Introduction

## 1.1 Research Goal and Scope

It is believed that the rapid emergence of Robotic technology in industry, and specifically in manufacturing, in the 21st century, will have positive impacts in many aspects of our lives. We have already seen many applications of this technology in macro scale, such as pick and place task [1]. However, there are still applications where humans outperform machines, especially in micro scale manufacturing, which requires high-precision robot manipulators.

Accurate positioning of robot arms is very important in automated manufacturing field. Over the past several decades, we have seen great strides in the technology for accurate positioning robots. We have seen researchers tried to implement add-on features such as real-time microprocessors, high precision motors, zero backlash gear set, advanced metrology sensors and so on in today's robots. Indeed, they have compensated for many structural and dynamic errors in robot positioning [2]. However, those add-on features are usually very expensive and unnecessarily increase the cost during the manufacturing process. Robotic systems that employ well- designed sensor-based control strategies can reduce the cost and simultaneously obtain robustness against disturbances and imprecisions from sensors or modeling.

The process of fastening and unfastening a screw is a mundane but a challenging task in the automated manufacturing. We found recent research on this topic only focuses on how robots should generate push/pull force on a driver [3]. The axial forces and torques are first measured through sensors and then controlled to imitate the human approach of fastening and unfastening by applying a similar amount of axial forces and torques. This approach only considers tactile sensing; however, human beings also use the information from visual sensing to help with this task. To replicate visual sensing in robots, for example, a camera system could be utilized to make sure a tool is at the right pose (correct orientation and location where head of a bolt and tail of a driver coincide). A visual system that can provide an accurate and repeatable positional tracking of the tool becomes significantly important and useful not only in this type of application but also in many other applications of automated manufacturing [4].

In this work, we have designed a multi-robotic control system that simulates the positioning process for fastening and unfastening applications and have examined its robustness against various uncertainties which may occur in this process. This control system is a Visual Servoing (VS) system where a camera is mounted on a robot arm manipulator and provides vision data for the motion control of a second robot manipulator with a tool. Both the Position-Based Visual Servoing (PBVS) and the Image-Based Visual Servoing (IBVS) systems have been thoroughly investigated in [5-8]. However, in these related works, in the visual servoing domain, the development of the outer-loop controller is usually achieved with the PID controller, or its simplified variations based only on a kinematic model of camera [5,8]. One improvement in this work is to use Youla robust control design technique [9] that includes both kinematics and dynamics in the model development stage. The increase in the model fidelity for the control design can positively influence the precision of the feature estimation and the control system stability for the high-speed tasks. Benefits of our design are discussed in more details in the following chapters.

Many existing applications for reducing positional errors use high-precision sensors and measurements [10,11]. However, those methods haven't taken into consideration of environment effects on sensor performances. In the sense of camera's performance, many environmental conditions (such as illumination, temperature, etc.) affect the noise level of a single image. In a manufacturing environment, quality of estimations from a camera may vary significantly from one observation location to another. For instance, the precision of an estimation improves greatly if the camera moves from a location where it is placed within the shade of machines to a location that has better illuminations. Also, places near machines may be surrounded with strong electrical signals that also introduce extra noises into the camera sensors. Thus, it is worthwhile to develop an algorithm that searches for camera's workspace to find an optimal location (if its orientation is fixed). A single image taken at this location has the smallest noise level among images taken at all locations in the space. With limited energy for moving the camera, this algorithm also ensures the camera ends up at a suboptimal (if the optimal pose is unreachable) location among the locations already searched. A complete and universal searching algorithm has been developed and simulated in this dissertation.

The primary research objective is to investigate the multi-robot system (includes a visual system and a tool manipulation system) for a specific task of accurate and fast automated manufacturing. In the visual system, a robot arm freely moves a stereo camera so that the system can make precise estimation of tool's pose (including position and orientation). In the tool manipulation system, a robot arm manipulates the tool to finish manufacturing task based on its pose estimated from the visual system. The research objective will be fulfilled with completion of following research goals:

1. Develop mathematical model of robot manipulators and camera systems.
2. Investigate and design a camera movement algorithm that searches for image capturing optimal pose where number of pictures needed is considered as an objective function to be minimized.
3. Investigate and develop modified IBVS controlled visual system that guides the camera precisely to the optimal location.
4. Investigate and develop modified IBVS controlled tool manipulation system that moves the tool to the target location fast and accurately.
5. Analyze how the proposed muti-robot system is robust against model and environmental uncertainties.

## 1.2 Dissertation Outline

**Chapter 1** conducts literature reviews of the metrology techniques in robotic accurate positioning and brief introductions on calibration techniques of stereo cameras and robot manipulators. It also addresses various existing image denoising methodologies and different VS schemes, especially the classical IBVS structure.

**Chapter 2** describes the general topology of the proposed muti-robotic system and control architectures for each sequential movement in the system. This chapter demonstrates the process of accurate positioning throughout different steps by showing control block diagrams.

**Chapter 3** introduces the mathematical models of an elbow robotic manipulator, and the camera utilized in this project. A complete full six-link robotic model with a stereo camera example is first discussed. The mathematical model of the camera is derived. This robotic model includes a total 6 Degree of Freedoms (DoFs) associated with 3 DoFs in positioning and 3 DoFs in orientation. The kinematic relationships between robotic joints are developed. Based on those relationships, a more sophisticated dynamic model of the robot manipulator is built by including the modeling of actuators (DC motors). Finally, a combined model of the robotic kinematic model and the camera model is derived, which relates joints' angle of robotic arm to the observed image coordinates of points. A one-link robotic arm with a monocular camera example is also described.

**Chapter 4** discusses the details of the camera location searching algorithm. Simulations of different scenarios show this algorithm can successfully guide the camera to the optimal location for observation when it is placed in an arbitrary unknown environment.

**Chapter 5** addresses controllers design in the robot positioning and Single Input and Single Output (SISO) outer loops of IBVS architectures. First, positioning control (inner loop control in the IBVS structure) is designed and analyzed. Then, the outer loop controller is developed in SISO IBVS architecture for both camera positioning calibration process and the tool manipulator's movements. Control strategies are developed to integrate the robot dynamic model with a stereo camera model, including when the tool is out of the camera's view range. Simulation results and analysis of SISO cases are also presented in this chapter. The analysis of SISO cases serves as preliminary tests for more advanced six degrees of freedom control design, which will be discussed in later chapters. Both Single Input and Single Output (SISO) and Multiple Inputs and Multiple Outputs (MIMO) cases are analyzed.

**Chapter 6** emphasizes the design of outer controllers for Multiple Inputs and Multiple Outputs (MIMO) IVBS architectures. A combined feedforward and feedback control frame is used to achieve a fast response as well as overcome overdetermined issue in the visual system. Those controllers are updated online so that the systems are stabilized in the entire range of operations.

**Chapter 7** presents simulation results of the entire control architecture combining the prefilters and MIMO controllers presented in Chapters 6. Various scenarios are simulated to test the performance of controllers. Also, robustness tests on model's parameter variation are investigated.

**Chapter 8** concludes this dissertation and discusses the future work in this area.

## 1.3 Literature Review

### 1.3.1 Sensor-based Robotic Positioning

We have seen many efforts been made to improve the positioning accuracy of robotic systems in the past few decades. An effective way to reduce the amount of inaccuracy is to measure it with sensors and compensate is through feedback control loop. Many metrology techniques have been investigated and applied for different kinds of data capturing. Among them, three methods have gained the most popularity in recent research, namely, vision-based methods, tactile-based methods, and vision-tactile integrated methods. In this section, we will briefly review those approaches and their applications.

**a. Vision-based methods**

The vision-based methods have been widely developed in recent years and used to determine position and orientation of target objects in robotic systems. Zhu et al. discussed Abbe errors in a 2D vision system for robotic drilling. Four laser displacement sensors were used to improve the accuracy of the vision-based measurement system [12]. Liu et al. [13,14] proposed a visual servoing method for automative assembly of air-craft components. With the measurements from two CCD cameras and four distance sensors, the proposed method can accurately align the ball-head which is fixed on the aircraft structure in a finite time.

In addition to mentioned applications, there are many contributions to the vision-based methods in robotic manipulation. However, most of those researchers focused on success rate of grasping end-effectors without enough analysis on the positioning accuracy. Du et al. published a study for robotic grasp detection by visually localizing the object and estimating its pose [15]. Avigal et al. proposed a 6-DoFs grasp planning method using fast 3D reconstruction and grasp quality convolutional neural network (CNN) [16]. Wu et al. proposed an end-to-end solution for visual learning [17].

**b. Tactile-based methods**

With the development of tactile sensors in the last few years, we have seen more and more focus on tactile-based methods in robotic positioning domain. The tactile sensors can measure contact states between the end-effector and the object in robotic manipulations. The contact states can be used to determine objects' relative orientations and positions with respect to the gripper. Li et al. designed a tactile sensor of GelSight and generated tactile maps for different poses of a small object in the gripper [18]. They studied the localization and control manipulation for a specific USB connector insertion task. Dong et al. studied the tactile-based insertion task for dense box packing with two GelSlim fingers which are used to estimate object's pose error based on neural network [19]. Furthermore, Hogan et al. developed a tacile-based feedback loop to control a dual-palm robotic system for dexterous manipulations [20]. Those tactile-based methods can only realize relatively accurate positioning of the tool with the end-effector, but the positioning of the robot manipulator itself is not addressed.

**c. Vision-tactile integrated methods**

Vision sensing can provide more environmental information with a wide measurement range, while tactile sensing can provide more detailed information of robotic manipulations. Therefore, the vision–tactile integrated methods effectively complement each other. Fazeli et al. proposed a hierarchical learning method for complex manipulation skills with multisensory fusion in seeing and touching [21]. Gregorio et al. developed a manipulation system for automatic electric wires insertion performed by an industrial robot with a camera and tactile sensors implemented on a commercial gripper [22].

According to the analysis, all the integrated sensory applications have achieved accurate robotic manipulation tasks such as insertion and their performances have been verified in experiments. However, the error space in those references is usually small and none of them has considered all the translational and rotational errors in 6 DoFs. Moreover, tactile-based or vision-tactile integrated methods will increase the expense of mass-produced manufacturing because tactile or vision-tactile sensors are more expensive to purchase and maintain compared to visual sensors alone. Based on that, our work is to explore the capability of vision-based methods and design a new method to improve the accuracy of positioning of the robot manipulation system.

## 1.3.2 Overview of a Stereo Camera Model and Its Calibration

A camera model (i.e., the pin-hole model [23]) has been adopted in the VS techniques to generate an interaction matrix [5]. The object depth, the distance between a point on the object and the camera center, needs to be either estimated or approximated by an interaction matrix [5]. One of the methods is to directly measure the depth by a stereo (binocular) camera (Figure 1.1) with the use of two image planes [24]. The camera model uncertainties may arise from the estimation of camera intrinsic parameter values, such as focal lengths, and baseline. Camera calibration techniques can be used to precisely estimate these values.

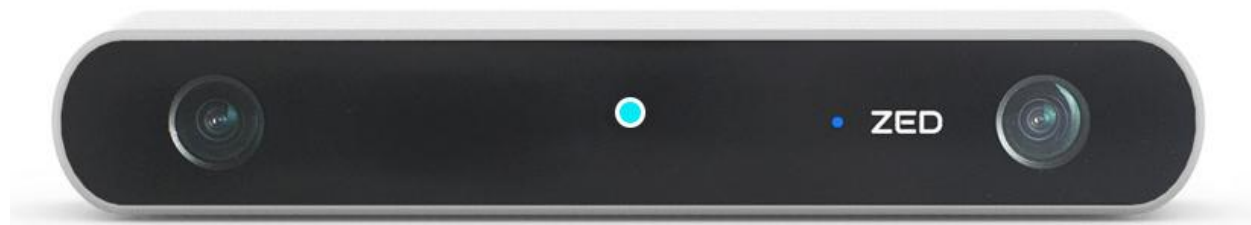

**Figure 1.1:** An example of stereo camera (Zed 2 stereo camera).

The stereo camera calibration has been well studied in [25-28]. As summarized in [25], the calibration method can be divided into two broad categories: the photogrammetric calibration and the self-calibration. In the photogrammetric calibration [26], the camera is calibrated by observing a calibration object whose geometry is well known in 3D space. These methods are very accurate but require an expensive apparatus and elaborate setups [25]. The self-calibration [25,27-28] is performed by finding the equivalences between the captured images of a static scene from different perspectives. Although cheap and flexible, these methods are not always reliable [25]. The author in [25] proposed a new self-calibration technique that observed planar pattern at different orientations and showed improved results.

## 1.3.3 Overview of a Robot Manipulator Model and Its Calibration

In this work, we consider the elbow manipulators [29] with the spherical wrist in the multi-robot system to move an end-effector freely in 6 DoFs. This model of the robot has six links with three for the arms and the other three for the wrist. The robot arms freely move the end effector to any position in the reachable space with 3 DoFs while the robot spherical wrists allow the end effector to orient in any direction with another 3 DoFs. For the elbow manipulators, a joint is connected between each of the two adjacent links and there are in total six convolutional joints. The specific industry model of this type is ABB IRB 4600 [30]. Figure 1.2 shows an image of ABB IRB 4600 and the rotational axis of all six joints.

The estimation of the robot parameters, such as link length, etc., determines the accuracy of the kinematic models of the robot manipulators. The paper [31] provides a good summary of the current robot calibration methods. The author of [32] states that over 90% of the position errors are due to the errors in the robot zero position (the kinematic parameter errors). As a result, most researchers focus on kinematic robot calibration (or level 2 calibration [31]) to enhance the robot's absolute positioning accuracy [33-36]. Generally, the kinematic model-based calibration involves four sequential steps: Modeling, Measurement, Identification, and Correction. Modeling is the development of a mathematical model of geometry and robot motion. The most popular one is D-H convention [37] and other alternatives include S-model [38] and zero-reference model [39]. At the measurement step, the absolute position of the end-effector is measured from the sensors, e.g., the acoustic sensors [38], the visual sensors [34], etc. In the identification step, the parameter errors for the robot are identified by minimizing the residual position errors with different techniques [39, 40]. This final step is to implement the new model with the corrected parameters.

On the other hand, the non-kinematic calibration modeling (level 3 calibration [31]) [41, 42], which includes dynamic factors such as the joint and the link flexibility in the calibration, increase accuracy of the robot calibration, but complicates the mathematical functions that govern the parameters relationship.

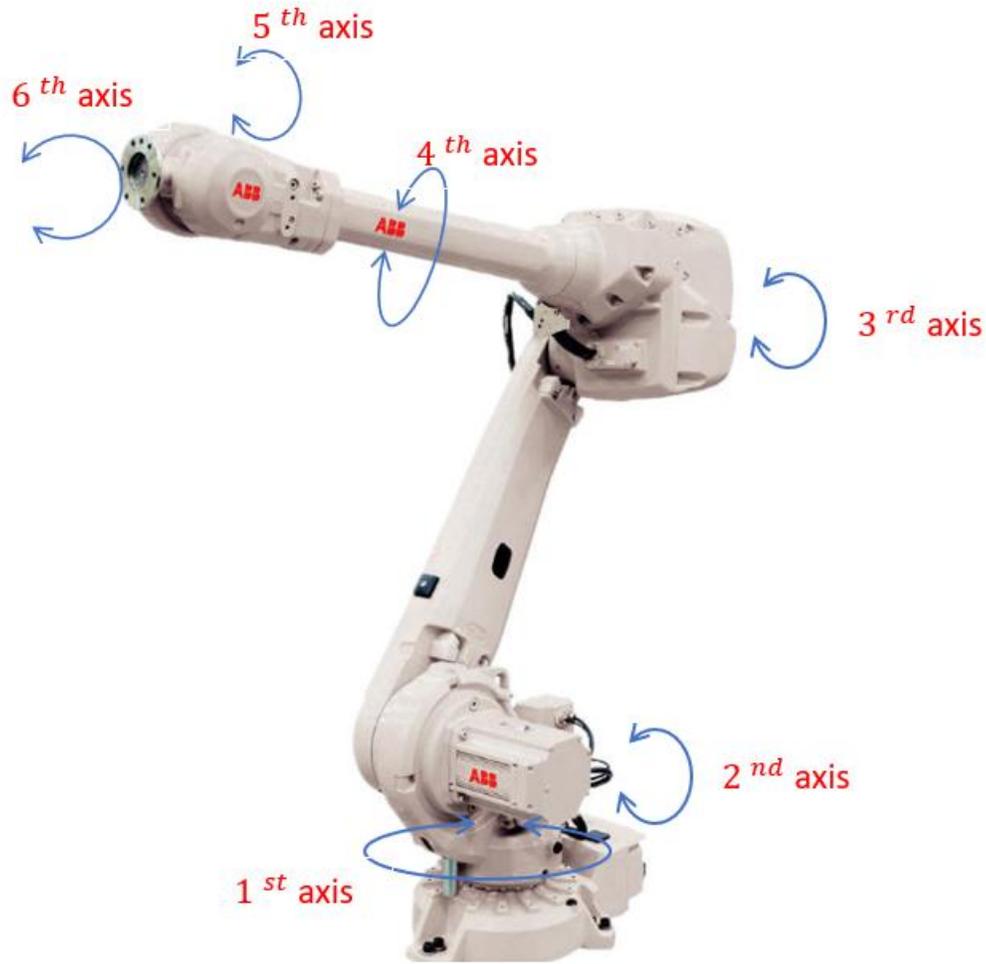

**Figure 1.2:** ABB IRB 4600 model.

### 1.3.4 Noise in Image Processing

Noises are inevitably introduced in the image processing. In this section, we briefly review existing work on image denoising and noise estimation.

   A. **Image noise Model**

Noise in digital images is a random variation of brightness in the image. Image noise is an undesirable by-product of image capture that obscures the desired information. Noise inevitably exists on all images even when they are taken by the most advanced digital cameras at the most optimal conditions. The noise that occurs in the image can be described by an additive noise model [43], i.e.

$$v(i,j) = w(i,j) + \eta(i,j) \tag{1.1}$$

where the observed image $v(i,j)$ is the sum of the true image $w(i,j)$ and the noise $\eta(i,j)$. The denotation of $v(i,j)$, $w(i,j)$ and $\eta(i,j)$ are pixel-wise intensity at coordinate $(i,j)$. In computer vision, a widespread approximation of noise $\eta(i,j)$ as a Poissonian-Gaussian model [44]. That is:

$$\eta(i,j) = \eta_p(i,j) + \eta_g(i,j) \tag{1.2}$$

In this model, the noise term $\eta$ is composed of two mutually independent parts a Poissionian component $\eta_p$ and a Gaussian component $\eta_g$. The Poissionian component mainly results from photon noise, which comes from the randomness of light photons collected by the sensor. The photon counts follow a classic Poisson process. But in practice, with large photon counts, the central limit theorem ensures photon noise is modeled as Gaussian distribution whose variance depends on the expected photon count [44,45]. The Gaussian component is the main result of read (or detector) noise. This kind of noise is due to the discrete nature at the stage of quantifying the number of photons detected for each pixel. This noise can be modeled with an independent identical distributed (i.i.d.) white Gaussian noise. In summary, if we can assume enough photon counts in the sensor (i.e., sufficient illumination), the noises existed in the image pixel $\eta(i,j)$ can be modeled with an i.i.d. white normal distribution.

### B. Overview of Image Denoising Techniques

Various denoising techniques have been proposed so far. A good noise removal algorithm ought to remove as much noise as possible while safeguarding edges. Gaussian white noise has been dealt with spatial filters, e.g., Gaussian filter, Mean filter, and Wiener filter [46]. The image can be viewed as a matrix with variant intensity values at each pixel. The spatial filter is a square matrix that performs convolutional production with the Image matrix to produce filtered (noise-reduced) image, i.e.

$$g(i,j) = k * v(i,j) \tag{1.3}$$

Where $v(i,j)$ is the original (noisy) image matrix entry at the coordinate $(i,j)$. $k$ is the spatial filter (or kernel) and $*$ accounts for convolution. $g(i,j)$ is filtered (noisy-reduced) image matrix entry at coordinate $(i, j)$. All the spatial filters have the benefit of being computationally efficient but with a shortcoming of blurring edges [47].

Also, image Gaussian noise reduction can be approached in the wavelet domain. In the conventional Fourier Transform, only frequency information is provided while the temporal information is lost in the transformation process. However, Wavelet transformation will keep both the frequency and temporal information of the image [47]. The most widely used wavelet noise decreasing method is the wavelet threshold method, as it can not only get an approximate optimal estimation of the original image, but also calculate speedily and adapt widely [47]. The method can be divided into three steps. The first step is to decompose the noise polluted image by wavelet transform. Then wavelets of useful signal will be retained while most wavelets of noise will be set to zero according to the set threshold. The last step is to synthesize the new noise reduced image by inverse wavelet transform of cleaned wavelets in the previous step [47,48]. The Wavelet transformation method has the benefits of keeping more useful details but is more computationally complex than the spatial filters. Threshold affects the performance of the filter. Soft thresholding provides smoother results while hard threshold provides better edge preservation [47]. However, whatever threshold is selected, filters that operate in the wavelet domain still filter out (or blur) some important high frequency useful details in the original image, even though more edges are preserved than spatial filters.

All the aforementioned methods present ways to reduce noise in the image processing starting from a noisy image. We can approach this problem with the multiple noisy images taken from the same perspective. Assuming the same perspective ensures the same environmental conditions (illumination, temperature, etc.) that affect the image noise level. Given the same conditions, an image noise level taken at a particular time should be very similar to another image taken at a different time. This redundancy can be used to improve image precision estimation in the presence of noise. The method that uses this redundancy to reduce noise is called signal averaging (or image averaging in the application of image processing) [49]. Image averaging has the natural advantage of retaining all the image details as well as reducing unwanted noises, given that all the images for the averaging technique are taken from the same perspective. The robot's rigid end effector that holds a camera minimizes shaking and drift when shooting pictures. Furthermore, in the denoising process, the precise estimations require that the original image details be retained. Considering these previous statements, we decided to choose image averaging over all other denoising techniques in this work.

The image averaging technique is illustrated in Figure 1.3. Assume a random, unbiased noise signal, and additionally, assume that this noise signal is completely uncorrelated with the image signal itself. As noisy images are averaged, the original true image remains unchanged and the magnitude of the noise signal is compressed thus improving the signal-to-noise ratio.

In Figure 1.3, we generated two random signals with the same standard deviation, and they are respectively represented by the blue and red lines. The black line is the average of the two signals, with its magnitude significantly reduced compared to each of the original signals.

In general, we can establish a mathematical relationship between the noise level reduction and the sample size for averaging. Assume we have **N** numbers of Gaussian white noise samples with the standard deviation **σ**. Each sample is denoted as $z_i$, where **i** represents $i^{th}$ sample signal. Therefore, we derive the following:

$$var(z_i) = E(z_i^2) = \sigma^2 \qquad (1.4)$$

where $E(\cdot)$ is the expectation value and $\sigma$ is the standard deviation of the noise signal. By averaging the $N$ Gaussian white noise signals, we can write:

$$var(z_{\text{avg}}) = var(\frac{1}{N}\sum_{i=1}^{N} z_i) = \frac{1}{N^2} N\sigma^2 = \frac{1}{N}\sigma^2 = (\frac{1}{\sqrt{N}}\sigma)^2 \qquad (1.5)$$

where $z_{\text{avg}}$ is the average of the N noise signals.

Equation (1.5) demonstrates that a total number of N samples are required to reduce the signal noise level by $\sqrt{N}$. Since our goal is to reduce the image noise to within a fixed threshold (a constant number expressed as a standard deviation), a smaller variation in the original image requires fewer samples to make an equivalent noise-reduction estimation. Thus, it is worthwhile for the camera to move around, rather than being stationary, to find the best locations where the image noise level estimation is small. In general, we can reduce the noise level as much as needed by taking more samples.

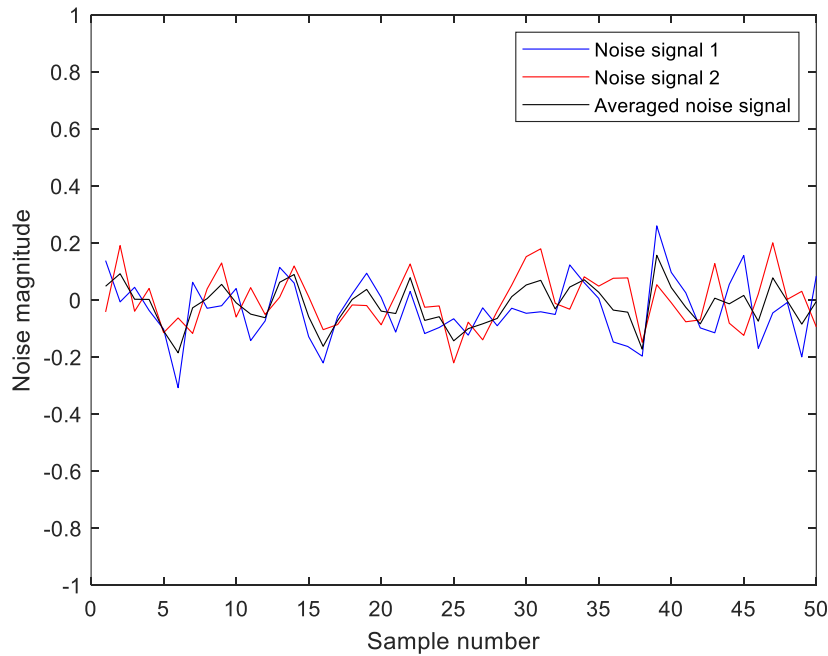

**Figure 1.3**: An example of a noise level reduction by image averaging.

A comparison of the above denoising methods is presented below. In Figure 1.4, a default picture from MATLAB is polluted with a Gaussian-distributed white noise with SNR of 5dB. A spatial filter (Gaussian filter), a wavelet filter (wavelet threshold method), and the image averaging technique (1000 averaged sample) have been applied to denoise the noisy image. Figure 1.4 shows all techniques succeed in reducing the noise to some extent while the wavelet filter and the image averaging technique performing better than the spatial filter regarding denoising. The difference between the wavelet filter and the image averaging technique is not obvious in Figure 1.4 but becomes more significant in the frequency domain.

Figure 1.5 compares the performance in the frequency domain of image averaging methods with different samples and Figure 1.6 shows the difference in the frequency domain after applying the above three denoising methods. The power intensity is separated from low to high as in those two figures. By comparing the original image and the noisy image, it can be observed that noise intensity is larger in the higher frequency domain. Figure 1.5 shows that by averaging more samples, the noise level decreases in all frequencies. Especially, the averaged image is almost the same as the original one when 1000 samples are averaged.

Figure 1.6 shows the image generated from the Gaussian filter has a lower power intensity than that of the original image in high frequency, which indicates that a lot of high-frequency data in the original image has been filtered out as well. Although it performs better than a Gaussian filter, the wavelet filter still removes some details in the original image. Only the averaging method can reduce the noise level and retain all information.

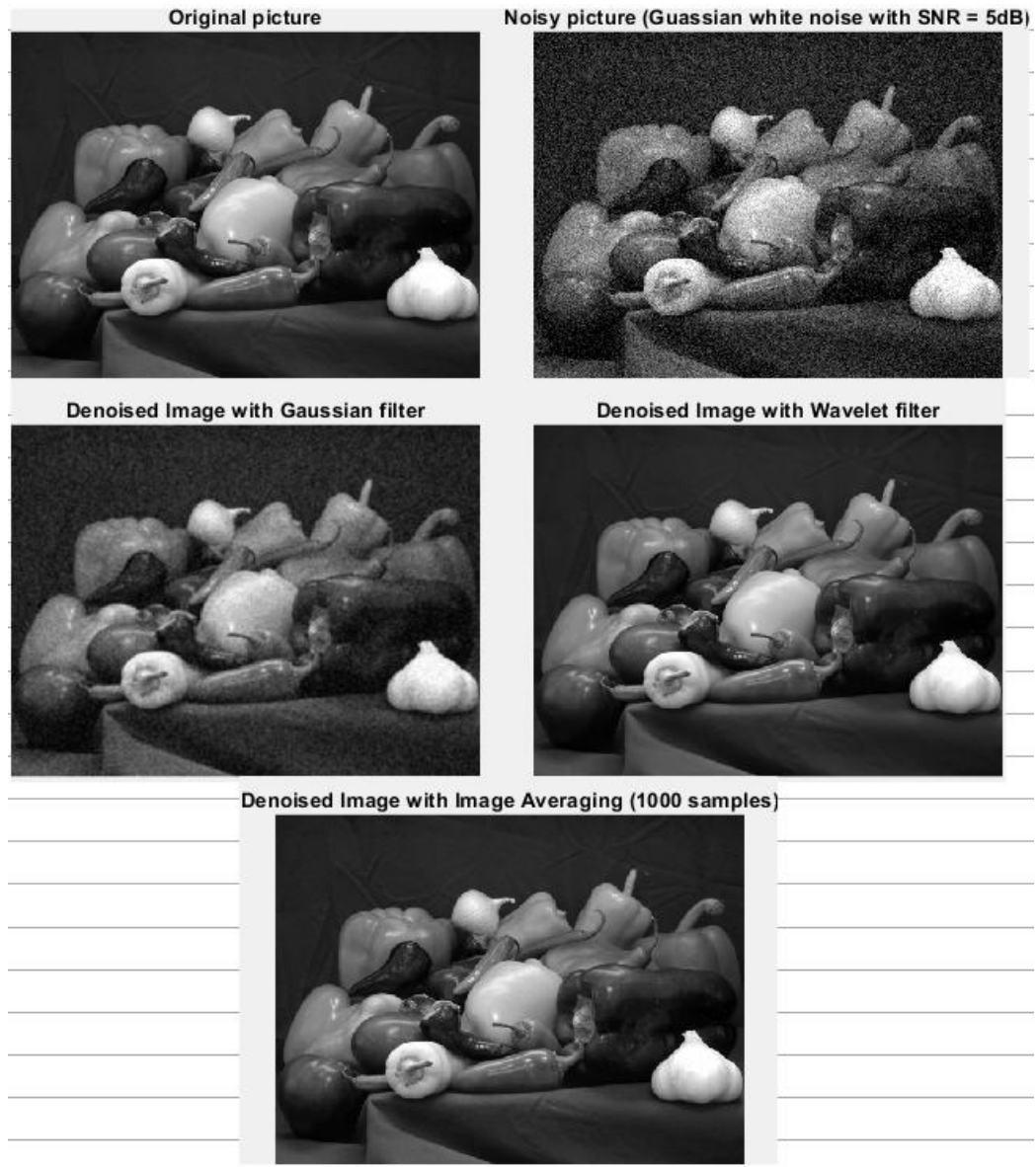

**Figure 1.4:** Original image, noisy image and denoised images.

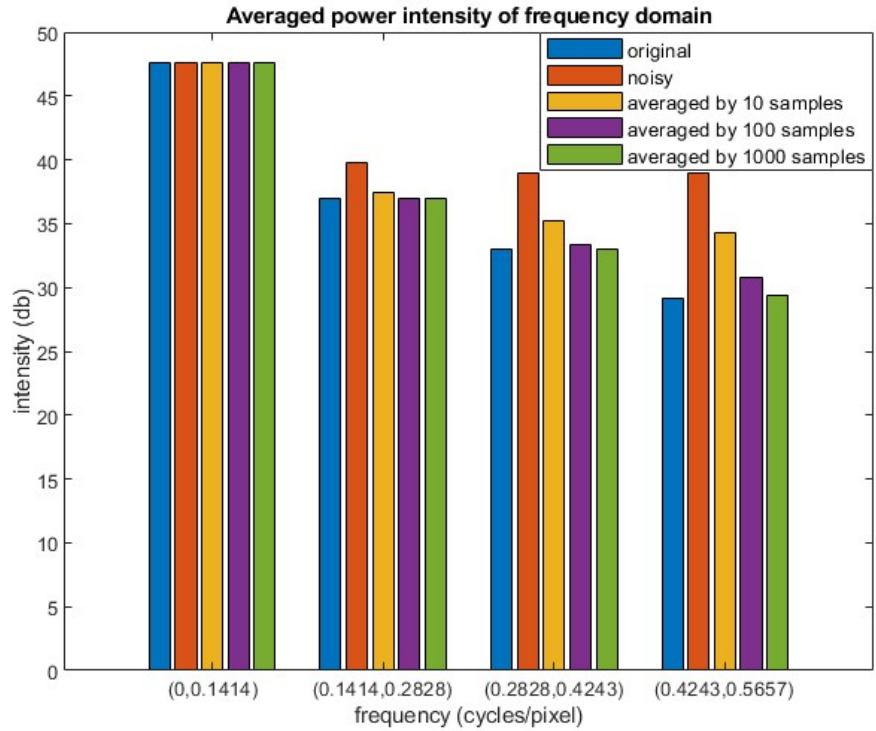

**Figure 1.5:** Frequency analysis of image averaging technique with varying samples.

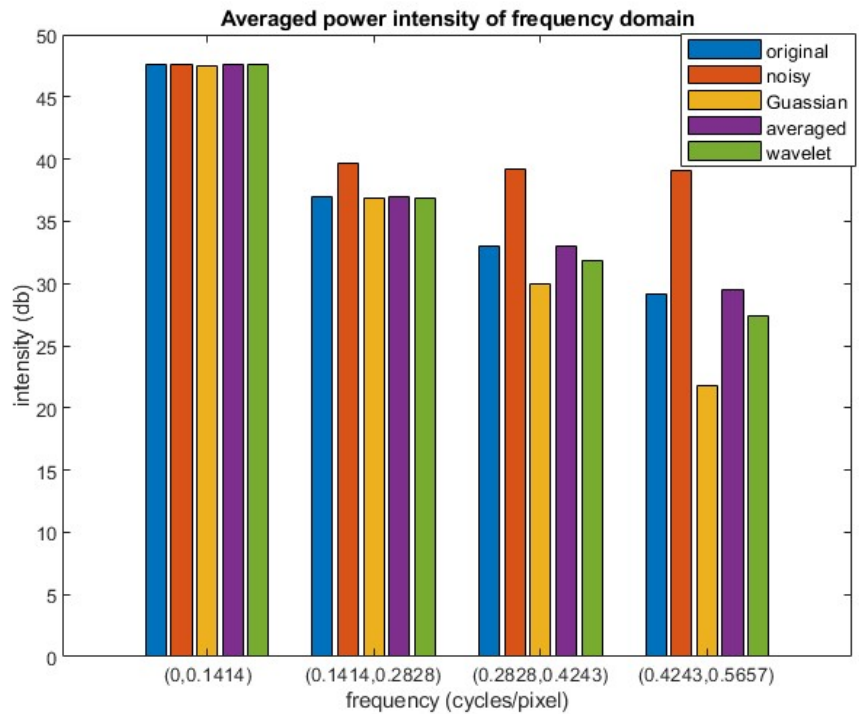

**Figure 1.6:** Frequency analysis of different denoising techniques.

## C. Image noise level estimation

Noise level, measured by noise variance, is a critical parameter in image denoising [50-52]. Numerous methods have been proposed to estimate the noise level. Many methods [53-55] assume that a sufficient amount of flat area can be obtained in the processed image; however, this assumption is often not achievable in practical image processing. Two more recent papers [56-57] using Principal Component Analysis (PCA) claim that their methods can accurately measure the noise level without homogenous image areas. Despite this, they tend to underestimate the noise level as they take the smallest eigenvalues of the covariance of selected low-rank patches as the estimated noise estimates [58].

A more accurate algorithm is introduced in [58], based on the observation that patches from noiseless images typically lie within a low-dimensional subspace, while those from noisy images are distributed across the ambient subspace. Consequently, noise level estimation can be derived from the eigenvalues of these redundant dimensions. The redundant dimensions set are selected by comparing the mean and median of the eigenvalues within the set. The author in [58] claims that their algorithm can effectively estimate noise variance from a single noisy image, achieving superior accuracy and reduced execution time compared to the methods in [56-57]. The result in [58] has been validated both theoretically and experimentally. The algorithm from [58] is illustrated in Figure 1.7.

**Algorithm 1** Estimating Image Noise Level
―――――――――――――――――――――――――――――――――――――――
**Require:** Observed Image $I \in \mathbb{R}^{M \times N \times c}$, Patch Size $d$.
1: Generating dataset $X = \{x_t\}_{t=1}^{s}$, which contains $s = (M - d + 1)(N - d + 1)$ patches with size $r = cd^2$ from the image $I$.
2: $\mu = \sum_{t=1}^{s} x_t$
3: $\Sigma = \frac{1}{s}\sum_{t=1}^{s}(x_t - \mu)(x_t - \mu)^T$
4: Calculating the eigenvalues $\{\lambda_i\}_{i=1}^{r}$ of the covariance matrix $\Sigma$ with $r = d^2$ and order $\lambda_1 \geq \lambda_2 \geq \ldots \geq \lambda_r$
5: **for** i = 1:r **do**
6: $\quad \tau = \frac{1}{r-i+1}\sum_{j=i}^{r}\lambda_j$
7: $\quad$ **if** $\tau$ is the median of the set $\{\lambda_j\}_{j=i}^{r}$ **then**
8: $\quad\quad \sigma = \sqrt{\tau}$ and **break**
9: $\quad$ **end if**
10: **end for**
11: **return** noise level estimation $\sigma$
―――――――――――――――――――――――――――――――――――――――

**Figure 1.7:** Image noise level estimation algorithm.

### 1.3.5 Visual Servoing Techniques

VS refers to controlling the motion of a robot with the aid of computer vision data. In most related work [5-7], a camera is used to acquire vision data whether the camera is mounted on a robot manipulator, or the camera is fixed in the workspace to observe from a stationary configuration. In this part, we have a brief review of the developed control scheme in this area.

1) **Eye-in-hand and Eye-to-hand configuration**

An eye-in-hand configuration is a visual system that has one or more cameras attached rigidly to the end effector of a robot arm. In an eye-to-hand configuration, a camera or cameras are fixed in the workspace to have a complete observation of robot motion [6]. The two configurations [59] are illustrated in the following Figure 1.8. The eye-in-hand configuration allows the camera to move closer to the object so that accuracy can be guaranteed with a limited view. The eye-to-hand configuration ensures a panoramic view of the workspace but with lower accuracy.

Some researchers have worked on the design of a visual servoing system based on hybrid eye-in-hand and eye-to-hand configurations. In [60], an eye-to-hand camera is applied to observe the translational movement of the robot tool while an eye-in-hand camera takes charge of the robot tool's rotational movement. In [59], a camera mounted on the end effector of a robot provides visual information as an eye-to-hand camera to guide the work of another robot. The author of [61] integrates cameras in two configurations and applies the Extended Kalman filter to estimate the robot's position in the event of occlusion.

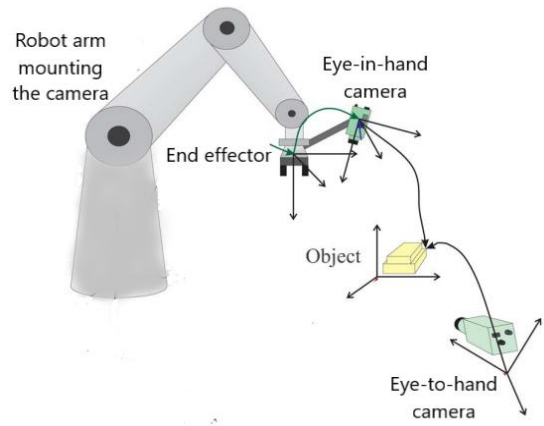

**Figure 1.8:** Eye-in-hand/ Eye-to-hand configurations.

## 2) "Look and Move" Scheme and Direct Servoing Scheme

The earliest work in VS involves robot systems and computer vision in an open loop (Figure 1.9). This Scheme is known as "look and move" [62]. In other words, a computer vision system works as a pose estimator and generates the required motion commands to the robot. This approach doesn't require a visual system to check the positions of the robot during and in the final position. Thus, "look and move" scheme is very sensitive to disturbances that alter the object positions, which are different from what the visual system planned at the beginning.

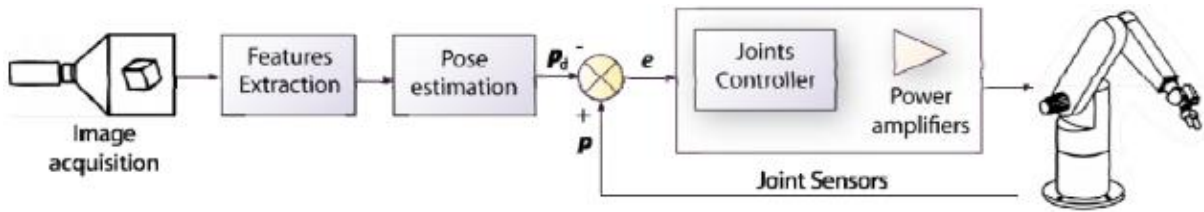

**Figure 1.9**: "Look and Move" scheme.

An alternative to the previous approach is the direct servoing scheme (Figure 1.10). Instead of implementing an internal joint controller, this scheme uses the vision system's data directly to control and stabilize the robot [63]. The vision controller directly generates forces and torques applied to the motors in the robot's joints.

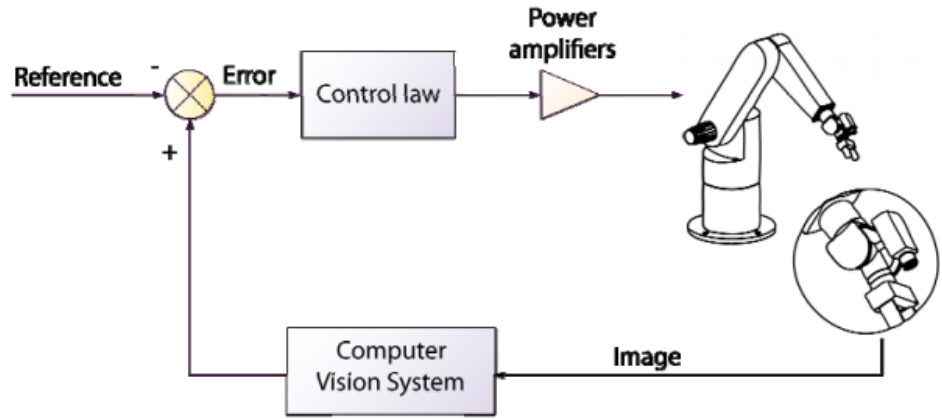

**Figure 1.10:** Direct visual servoing scheme.

3) **Position-Based Visual Servoing and Image-Based Visual Servoing**

Many Position-Based Visual Servoing (PBVS) and Image-Based Visual Servoing (IBVS) systems are using an indirect servoing approach [5-7] (Figure 1.11 and Figure 1.12). Contrary to direct visual servoing, in the indirect servoing configuration, the robot's external controller generates control actions specified as velocities applied at the robot's end-effector, and the internal controller translates these velocities into forces and torques exerted on the robot's joints. However, the interaction matrix in traditional indirect PBVS and IBVS primarily involve kinematic models of robots and cameras. It has not taken into account the dynamics of robot arms [5]. As a result, imperfections in dynamic positioning systems may dramatically result in inaccuracies in estimation and even instability. In this section, PBVS and IBVS are discussed in the context of indirect servoing approach although they can also be applied in the direct approach.

In a PBVS system, the camera is a sensor that estimates the 3D pose of an object. It extracts features in an image and estimates the pose, including $P^C$ (position) and $\varphi^C$ (angle), with respect to a 3D coordinate frame in the workspace. The control error is defined as the difference between the current pose and the desired pose $(P_d^C, \varphi_d^C)$. Computing poses requires the knowledge of the camera's intrinsic parameters and the 3D model of the object. This is also a problem of 3D localization.

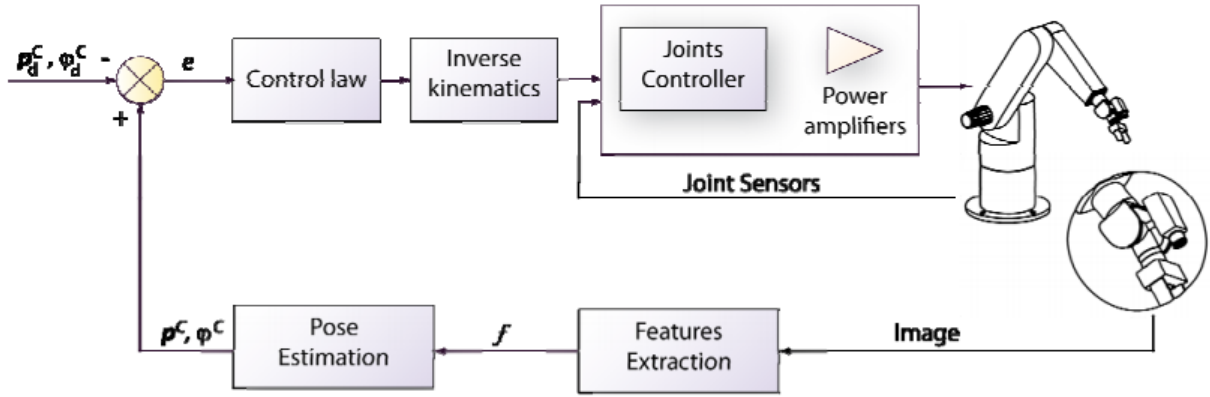

**Figure 1.11:** Position-based visual servoing scheme.

Compared to the PBVS system, the IBVS system computes the control action in the 2D space of the image. It tries to directly eliminate differences between 2D visual features $s$ (e.g. points, lines and corners) at current positions and the features $s_d$ at the goal positions.

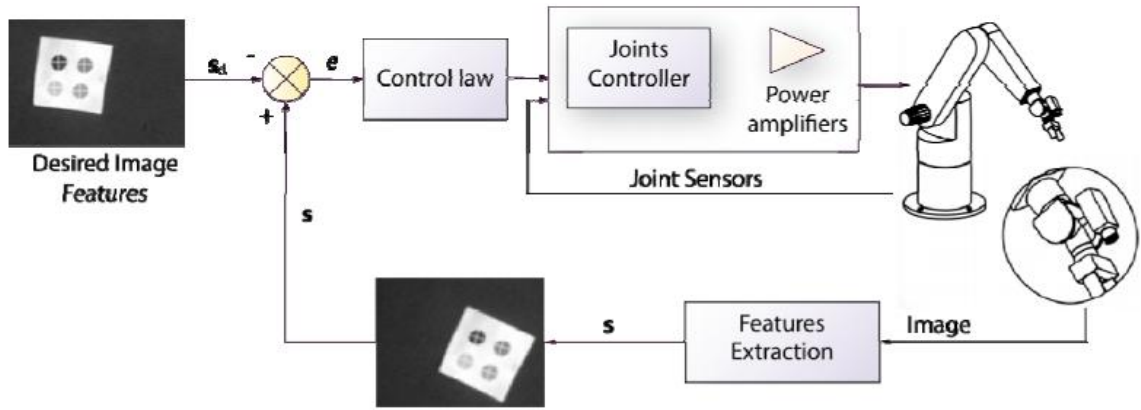

**Figure 1.12:** Image-based visual servoing scheme.

In [64], a feedforward neural network is used to learn implicit relationships between robot poses and observed variations in global descriptors in the image, e.g. geometric moments and Fourier descriptors. Reinforcement learning is implemented in VS [65]. If the robot reaches the final desired position by observing the scene, the system receives positive rewards, while failures to reach a goal pose result in negative rewards.

### 1.3.6 Overview of the Classical IBVS Architecture

This work focuses on the IBVS structure. Figure 1.13 shows the control block diagram for a classical IBVS architecture. In Figure 1.13, $s = [u, v]^T$ is the image feature position vector, $s^* = [u^*, v^*]^T$ is the target image feature position vector, and their difference $e = s^* - s$ is the error vector. $L_e$ is the so-called interaction matrix [5], which is a 2-by-6 matrix, and relates the time derivative of the image feature $s$ to the spatial velocity of the camera $V_c$, a column vector of six-elements, by the following:

$$\dot{s} = L_e V_c \tag{1.6}$$

We can design a proportional controller to force the error to exponentially converge to zero, i.e.:

$$\dot{e} = -ke, k > 0 \tag{1.7}$$

Suppose the target image feature is a constant; that is $\dot{s}^* = 0$, hence, we can derive from Equation (1.7):

$$\dot{e} = \dot{s}^* - \dot{s} = -L_e V_c \tag{1.8}$$

From Equations (1.7) and (1.8), it follows:

$$V_c = k L_e^+ e \tag{1.9}$$

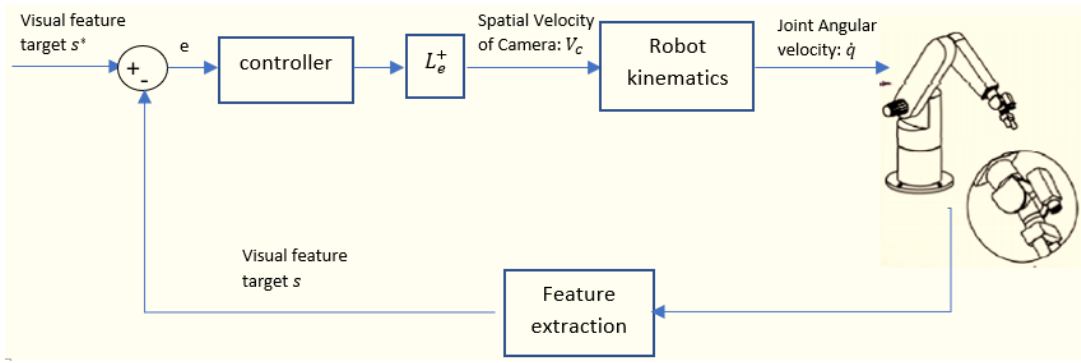

**Figure 1.13:** The block diagram of the classical IBVS control architecture.

where $L_e^+$ is the pseudoinverse of $L_e$. The derivation of the interaction matrix with a monocular camera is further explained next. We assume a point with the three-dimension (3D) coordinates in the camera frame is given as $P^C = [X^C, Y^C, Z^C]$.

Assume this point is projected to an image in a monocular camera, then the image feature coordinate $s = [u, v]^T$ can be expressed as:

$$\begin{bmatrix} u \\ v \end{bmatrix} = \begin{bmatrix} \dfrac{f_u X^C}{Z^C} + u_0 \\ \dfrac{f_v Y^C}{Z^C} + v_0 \end{bmatrix} \tag{1.10}$$

Where $f_u$ and $f_v$ are the horizontal and the vertical focal lengths, and $u_0, v_0$ are coordinates offsets of $u$ and $v$ axis.

Taking the time derivative of Equation (1.10), it follows:

$$\begin{bmatrix} \dot{u} \\ \dot{v} \end{bmatrix} = \begin{bmatrix} \dfrac{f_u(\dot{X}^C Z^C - \dot{Z}^C X^C)}{(Z^C)^2} \\ \dfrac{f_v(\dot{Y}^C Z^C - \dot{Z}^C Y^C)}{(Z^C)^2} \end{bmatrix} \tag{1.11}$$

The rigid body motion of a 3D point in the camera model can be derived as:

$$\dot{P}^C = v^C + \omega^C \times P^C \Leftrightarrow \begin{cases} \dot{X}^C = v_X^C - \omega_Y^C Z^C + \omega_Z^C Y^C \\ \dot{Y}^C = v_Y^C - \omega_Z^C X^C + \omega_Z^C Z^C \\ \dot{Z}^C = v_Z^C - \omega_X^C Y^C + \omega_y^C X^C \end{cases} \tag{1.12}$$

Substituting (1.12) into (1.11), and rearranging the terms, we obtain:

$$\begin{bmatrix} \dot{u} \\ \dot{v} \end{bmatrix} = \begin{bmatrix} \dfrac{f_u}{Z^C} & 0 & -\dfrac{u-u_0}{Z^C} & -\dfrac{(u-u_0)(v-v_0)}{f_v} & \dfrac{f_u^2 + (u-u_0)^2}{f_u} & -\dfrac{f_u(v-v_0)}{f_v} \\ 0 & \dfrac{f_v}{Z^C} & -\dfrac{v-v_0}{Z^C} & -\dfrac{f_v^2 + (v-v_0)^2}{f_v} & \dfrac{(u-u_0)(v-v_0)}{f_u} & \dfrac{f_v(u-u_0)}{f_u} \end{bmatrix} \begin{bmatrix} v_X^C \\ v_Y^C \\ v_Z^C \\ \omega_X^C \\ \omega_Y^C \\ \omega_Z^C \end{bmatrix} \tag{1.13}$$

Equation (1.13) can be simply written as:

$$\dot{s} = L_e V_c = L_e \begin{bmatrix} v^C \\ \omega^C \end{bmatrix} \quad (1.14)$$

Some drawbacks of the classical IBVS are summarized as follows. To compute the interaction matrix $L_e$ from Equation (1.14), the depth $Z^C$ needs to be estimated. This can usually be approximated as either the depth of the initial position, the depth of the target position, or their average value [5].

A careless estimation of the depth may lead to system instability. In addition, the design of the proportional controller is based on Equation (1.9) and the camera kinematic relationships, which means that no dynamics are considered in this model. The kinematic model is sufficient for a very slow responding system; however, for faster responses, one must take into account the manipulator dynamics along with the camera model.

In this work, we propose a new controller algorithm similar to the classical IBVS structure, where the controller is designed with the complete dynamic and kinematic models of the robot manipulator and the camera. Furthermore, this algorithm doesn't require any depth estimation; therefore, it will not be necessary to use the interaction matrix. The development of this new algorithm is presented in Chapter 5.

___________________________________________________________Chapter 2

# Control Architectures Overview

This chapter begins by proposing a topology for the multi-robot system, which comprises both a visual system and a tool manipulation system. It then outlines the sequential control procedures of the system, including the optimal camera pose determination, the camera movement adjustment control process, and the high-accuracy tool manipulation control process. Control architectures and flowcharts for each of these processes are illustrated in the following sections. Additionally, this chapter introduces kinematic models along with their Hardware-In-the-Loop equivalent models for both the visual and tool manipulation systems. It is anticipated that the accurate positioning of tools can be achieved by eliminating various uncertainties at different stages of the process.

To achieve accurate positioning, it is essential to develop methodologies that eliminate uncertainties arising in the manufacturing process. These uncertainties can stem from various sources and can be categorized into three main types: sensor measurement noise, and dynamic and kinematic modeling errors related to both the measurement system and the robot manipulators.

Kinematic modeling errors, which represent estimation inaccuracies in parameter values, can be significantly reduced through various established calibration processes. Previous research on calibrations in the domains of cameras and robot manipulators has been summarized in Sections 1.3.2 and 1.3.3, respectively.

Dynamic errors in robot manipulators, on the other hand, are associated with non-static joint errors. Among these sources of error, deviations between the actual joint rotations and their measured values—caused by unmodeled dynamic uncertainties such as backlash, friction, gear compliance, joint or link flexibility, and thermal effects—have the most substantial impact on the accuracy of robot position control. As discussed in Section 1.3.2, these errors can be addressed in the dynamic modeling phase by developing a high-fidelity dynamic model, which allows for the identification of these parameter values through calibration. Alternatively, these dynamic errors can be treated as disturbances in the manipulator control system. The control system is designed to ensure that the plant output follows the desired input (the reference signal) while simultaneously rejecting these disturbances. This chapter will delve into the design of robot manipulator control systems and demonstrate the effectiveness of feedback loops in mitigating the aforementioned dynamic errors.

Lastly, sensor noise, which originates from image noise during camera capture, can be minimized using denoising methods such as the image averaging technique discussed in Section 1.3.4. This chapter will also demonstrate the application of this method in one of the control processes.

## 2.1 Topology of the Multi-robotic System

In this section, we explore the proposed control architectures for a multi-robot system designed to facilitate high-precision tool movement across various manufacturing scenarios by minimizing process uncertainties. We assume that all cameras and robot manipulators are calibrated at the outset using one or more of the methods discussed earlier, ensuring that the initial parameters for both cameras and manipulators are accurately identified. Consequently, the primary uncertainties in this context are sensor noise and dynamic modeling errors. Figure 2.1 illustrates the overall topology of the multi-robot system.

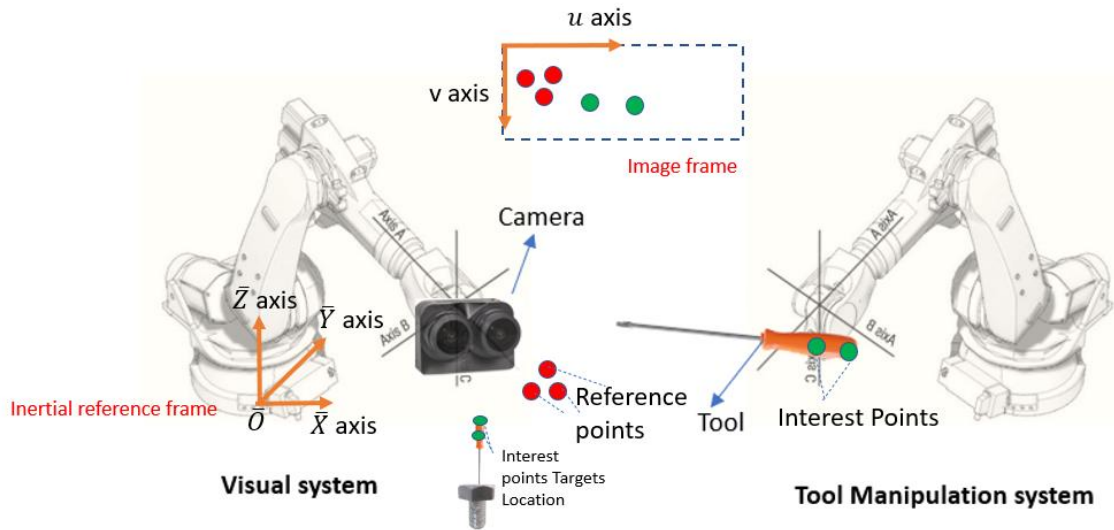

**Figure 2.1:** The topology of the multi-robotic system for accurate positioning control.

The multi-robot system consists of a visual system and a tool manipulation system, as illustrated in Figure 2.1. In the visual system, a camera is mounted on an elbow robot arm, while the tool is held by the end-effector of the robot manipulator. The primary objective of the visual system is to provide precise estimation of the tool's pose, enabling the tool manipulator to control its position with guidance from the visual system. To assist in detecting the position and orientation of the tool, two fiducial markers (green circles, or interest points) are placed on the tool.

The absolute coordinates of reference points (red circles) are established within the inertial reference frame. These reference points are positioned near the tool's target location (the target interest points), ensuring that both the reference points and the target interest points can be captured in the camera frame as the tool approaches its target pose.

Three reference points are selected to be close to one another in space. In a visual servoing context, a unique location in space from which an image is captured by a monocular camera requires at least four points: three points to establish the specific location and one additional point to determine the orientation. This situation is known as the Location Determination Problem (LDP) using image recognition [66].

In this case, however, we consider utilizing only three reference points to determine the pose of a stereo camera in 3D space. Appendix B proves that at least three points are sufficient for solving the LDP for stereo cameras.

Conversely, when the camera pose is fixed and known in space, the stereo camera can accurately detect depth, allowing it to provide the distinct 3D location of a point based on the image coordinates. Consequently, two interest points (resulting in a total of six 3D coordinates) are sufficient to determine the pose of the tool in real time, accounting for all six degrees of freedom (6 DoFs).

## 2.2 Multi-robotic System Sequential Control Procedure

The movement control of the robot manipulators operates asynchronously within the visual and tool manipulation systems. Figure 2.2 illustrates a flowchart that demonstrates this sequential control process.

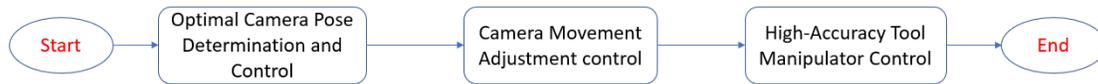

**Figure 2.2:** The flow chart of the sequential control procedure.

The first stage involves determining and controlling the optimal camera pose. During this stage, the camera moves to find the best position by minimizing the number of images needed while keeping the image noise of the reference points within an acceptable threshold.

In the second stage, known as camera movement adjustment control, any uncertainties arising from the camera's movement in the previous stage—such as unmodeled joint compliance—are addressed using a hand-in-eye visual servoing (VS) controller. Once the movement has been adjusted, the camera remains static to provide precise estimations of the tool's position.

Finally, the last stage focuses on high-accuracy tool manipulation control, where the tool's movement is guided to the target pose location using a hand-to-eye VS controller. Each control method and architecture will be discussed in the following sections.

## 2.3 Optimal Camera Pose Determination Process

The proposed algorithm aims to identify and move the camera to a location where the least number of pictures are required to reduce the noise level in the averaged image to a target degree, denoted as $\sigma_{noise\_reduced}$. A flowchart illustrating the algorithm is presented in Figure 2.3.

The camera, mounted on a 6-joint robot manipulator, can move freely in six DoFs within its operational space. For each candidate's optimal location, we assume the camera's orientation is fixed; for example, the camera will only point towards the face of interest on the object when capturing images. Additionally, the camera can only reach locations within the maximum reach of the robot manipulator, and the object must be detectable within the camera's field of view. These constraints define a camera operational space, which is then divided into a grid with a specified resolution. The intersections of the grid create a set of nodes ($S_{total}$), representing all potential locations where the camera can move and search using the algorithm.

The algorithm iteratively directs the camera to capture an image at one location, generates the next target location, and moves to that location until it identifies the optimal position that requires the fewest pictures. In each iteration, the camera takes a photo at the current candidate location, denoted as $P^C$. After image processing, it produces a noisy image with an intensity matrix $I_a$. A previously developed algorithm [55] estimates the noise level $\sigma_a$ across the image. The number of pictures $N^C$ required to reduce the noise level to $\sigma_{noise\_reduced}$ is calculated using the equation: $N^C = (\frac{\sigma_a}{\sigma_{noise\_reduced}})^2$. Using this information (specifically $N^C$ at $P^C$), the algorithm generates the next target location, $P^{Next}$. If the next generated target position $P^{Next}$ is the same as the current position $P^C$, then $P^C$ is determined to be the optimal location, and the camera remains there to observe the object. Otherwise, the robot joint controller (details in Chapter 5) adjusts the robot's joint angles to the target $\bar{q}$ (calculated using forward kinematics) so that the camera moves to the next target position $P^{Next}$. This algorithm will be explained and analyzed in detail in Chapter 4.

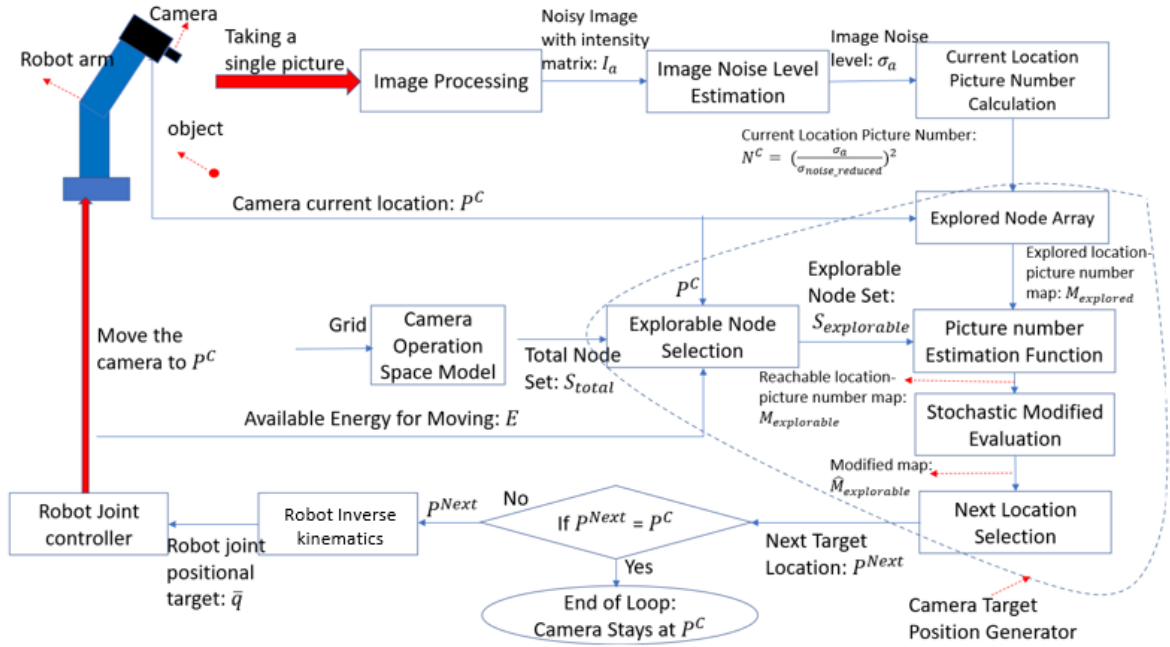

**Figure 2.3:** The optimal camera's pose determination process and its control architecture.

## 2.4 Camera Movement Adjustment Control Block Diagram

In the previous stage, the camera is mounted on a 6 DoFs robot arm and its movement is controlled by adjusting the joint angles of the robot manipulator. Ideally, by the end of that stage, the camera should be positioned at the optimal location $\overline{P^C}$ with the corresponding joint angles $\overline{q_V}$. However, the actual location of the camera $\widetilde{P^C}$ may deviate from the target optimal location $\overline{P^C}$ due to uncertainties such as sensor noise from the encoders measuring the joint positions and unmodeled dynamic components, such as joint compliance. To reduce the camera's position error caused by these uncertainties, we have developed a cascaded control architecture (Figure 2.4.).

As mentioned, three reference points with known absolute coordinates $\overline{P_R^V}$ measured in the inertial system attached to the visual system, are selected to uniquely determine the camera's pose. Fiducial markers are placed on these reference points, enabling their locations to be recognized and estimated in a 2D image coordinate frame using computer vision. From the kinematic model of the robot arm and the camera, the image coordinates of the reference points $\overline{p_R}$ can be calculated from the joint angles at the optimal position $\overline{q_V}$, These coordinates serve as targets for the cascaded control loop illustrated in Figure 2.4.

Assuming that the deviation between the actual and optimal camera positions $\widetilde{P^C}$ and $\overline{P^C}$ is small, we consider the number of images $N$ needed for averaging to be the same for both positions. After averaging $N$ images, we obtain a precise measurement of the current image coordinates of the reference points, denoted as $\widehat{p_R}$. The Image-Based Visual Servoing (IBVS) cascaded control system then minimizes the image coordinate errors between $\overline{p_R}$ and $\widehat{p_R}$ to adjust the camera to the optimal position. The inner-loop control strategy mirrors the closed joint control loop used in the previous stage (Figure 2.3.). Additionally, a feedforward controller has been designed to enhance the performance of the controlled system, which will be discussed in later chapters.

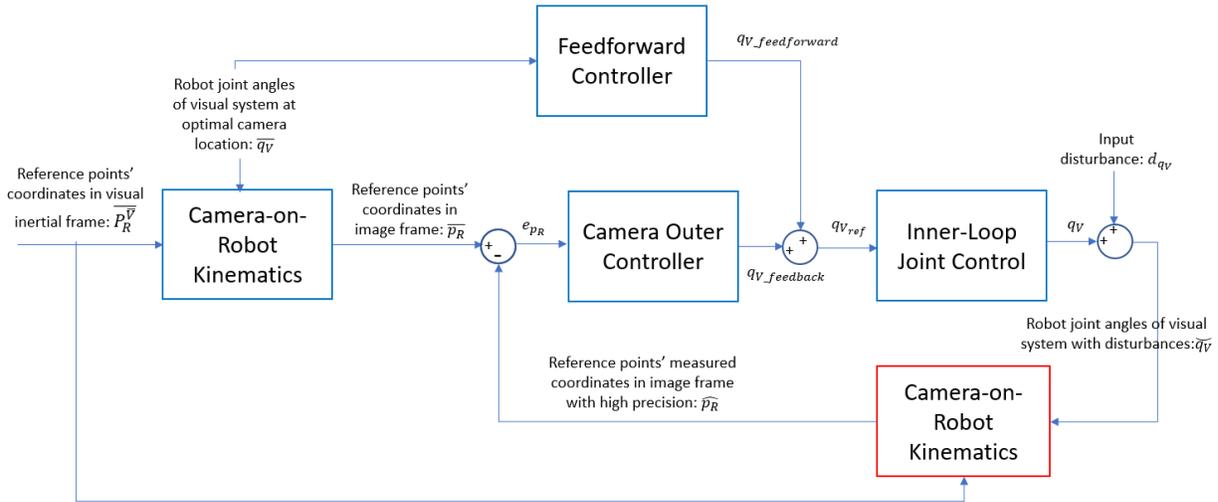

**Figure 2.4:** The camera movement adjustment control block diagram.

*(Note: Blocks in red lines are referred to HIL Models)*

## 2.5 High-accuracy Tool Manipulation Control Block Diagram

We propose a control strategy for the tool manipulation system, accompanied by a control block diagram as shown in Figure 2.5. The control algorithm in this block diagram integrates both feedforward and feedback control mechanisms.

The feedforward control loop operates as an open loop, bringing the tool as close as possible to the target position despite the presence of input disturbances, denoted as $d_{q_T}$. In the inner joint control loop, noise may arise from low-fidelity, inexpensive encoder joint sensors, as well as dynamic errors from the joints, such as compliance. All sources of noise within the joint control loop are modeled as an input disturbance $d_{q_T}$ to the outer control loop. Consequently, the final disturbed angles of the robot manipulators are denoted as $\widetilde{q_T}$, which define the tool's 6 DoFs pose as $\widetilde{P_I^V}$, measured in the inertial coordinate system attached to the visual system.

The outputs of the feedforward controller are the reference joint angles of rotation $q_{T\_feedforward}$. These are combined with the outputs from the outer feedback controller $q_{T\_feedback}$, to set the targets for the inner joint control loop. The forward kinematics function transforms the current joint angles of the tool manipulator into the current pose of the tool on the end effector, using the kinematic model of the robot arm.

The movement of the tool can be fine-tuned by the adaptive feedback control loop. The adaptive controller is computed online based on the current image coordinates $\hat{p}_I$ (indicated by the red line process in Figure 2.6). The feedback control loop rejects the input disturbance $d_{q_T}$ by minimizing the error between the target coordinates of two makers $\bar{p}_I$ he estimated image coordinates $\hat{p}_I$.

Both the feedforward and feedback controllers work in cooperation to move the tool to the target pose location within the tool manipulation system. The combined target $q_{T_{ref}}$ serves as the inputs to the joint control loop, allowing both controllers to manipulate the tool's pose effectively. The advantage of implementing both feedback and feedforward controls in the manipulation system is the reduction in time duration required for movement. Relying solely on feedback control would necessitate multiple images for pose estimation from the visual system, resulting in slow tool movement. We can divide the task of tool movement control into two stages. In the first stage, the feedforward control directs the tool to an approximate location near the desired destination. In the second stage, the feedback controller refines the tool's position to the precise target location using pose estimation from the camera. Furthermore, the camera has a limited range of view and can only detect the tool and measure its 2D features as $\hat{p}_I$ when the tool is within this range. When the tool is outside the camera's field of view, we must estimate its features as $\tilde{p}_I$ until it moves into the detectable range.

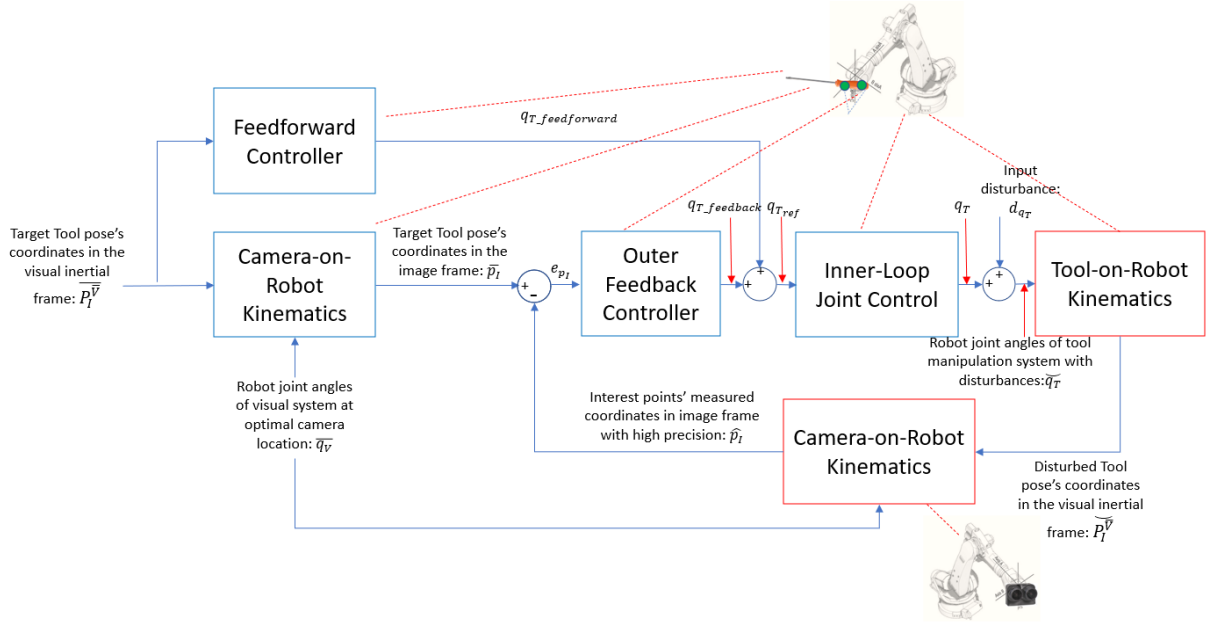

**Figure 2.5:** The high-accuracy tool manipulation control block diagram.

In Figure 2.6., we have developed an additional feedback loop to estimate the 2D coordinates of the tool using the same mathematical model as the real Hardware-In-Loop (HIL) models, particularly when real measurements are unavailable. The normal feedback loop (indicated by blue lines) is maintained when the tool is within the camera's field of view, allowing the camera to measure the tool's 2D feature as $\hat{p}_I$. However, when the tool moves outside this range, the 2D feature can only be estimated as $\tilde{p}_I$ (represented by green dashed lines). To ensure a seamless transition between these modes of operation, we implement a bump-less switch that smoothly toggles between the two feedback loops. The switching signal is activated when the tool enters or exits the camera's range of view.

**Figure 2.6:** The high-accuracy tool manipulation control block diagram.

*(Note: Blocks in red lines are referred to HIL Models)*

This control topology is analogous to the current industrial trend of macro-micro manipulation, where large-scale robots are employed for approximate positioning, while small-scale robots are used for precise positioning [2].

## 2.6 Camera-on-Robot Kinematics Model Block Diagram

As illustrated in Figure 2.7, the Camera-on-Robot Kinematics model provides high-precision estimations within the feedback loop of the camera movement adjustment control. This model also serves as a target generator, transforming targets from the inertial frame, which is attached to the visual system, into the image frame. Mathematically, the model combines a robot arm kinematics model (distinct from the forward kinematics model) with a stereo camera model.

Figure 2.7 presents the details of the Camera-on-Robot Kinematics model alongside its equivalent Hardware-In-the-Loop (HIL) model. The upper configuration represents the mathematical model used in simulations to generate image coordinates and to design the outer-loop controller within the robot arm control loop. In contrast, the lower HIL configuration replaces this mathematical model in practical applications. In the HIL model, image processing accurately estimates the locations of reference points and the tool pose with high precision.

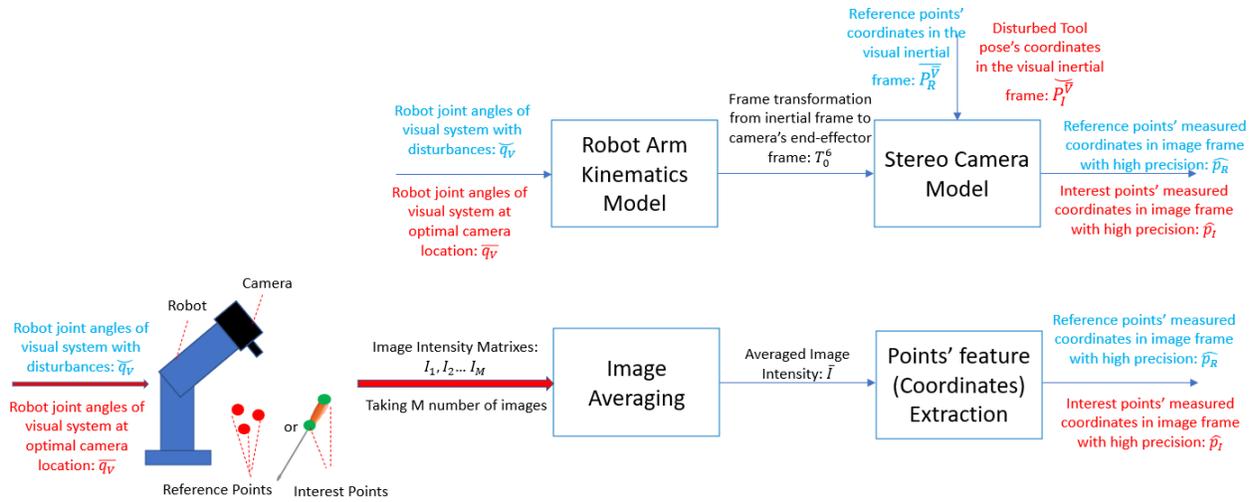

**Figure 2.7:** The Camera-on-Robot kinematic model.

*Note: Blue are signals used in the control system of Figure 2.4 and red are signals used in the control system of Figure 2.5 and Figure 2.6.*

## 2.7 Tool-On-Robot Kinematics Model Block Diagram

The Tool-on-Robot arm kinematics model represents the forward kinematic model of the robotic manipulator equipped with the tool. It transforms the combined effects of all joint revolutions into the 3D coordinates of interest points in the inertial frame. Figure 2.8 provides details of the tool robot arm kinematic model along with its equivalent Hardware-In-the-Loop (HIL) model.

The mathematical equations governing this model are developed in detail in Chapter 3.

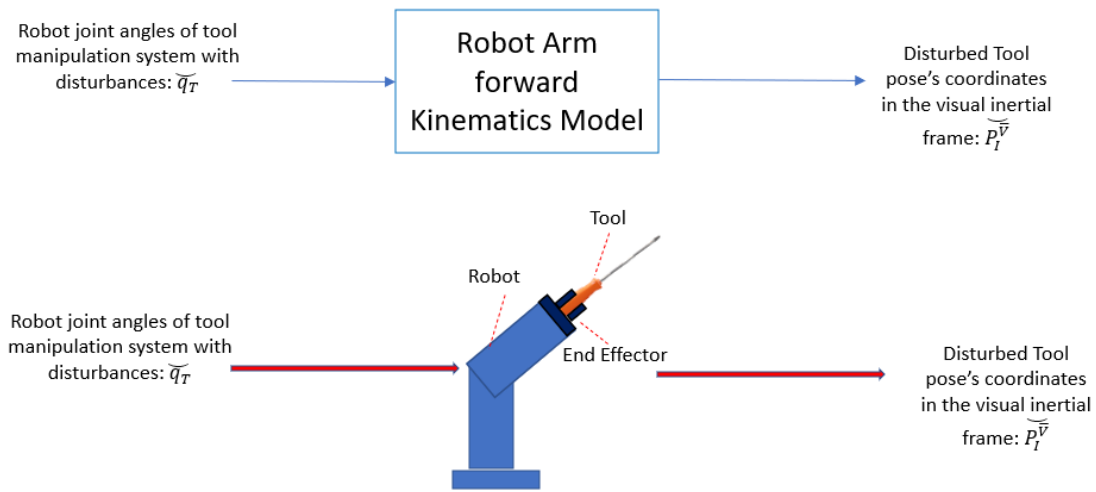

**Figure 2.8:** The tool robot arm kinematic model.

## 2.8 Camera-and-Tool Combined Model Block Diagram

The tool manipulation control architecture demonstrates that both the Tool-on-Robot kinematics model and the Camera-on-Robot kinematics model are incorporated within the feedback loop. By combining these two models, we can simplify the development of a cascaded outer controller. This integrated model is referred to as the Camera-and-Tool Combined Model, as illustrated in Figure 2.9.

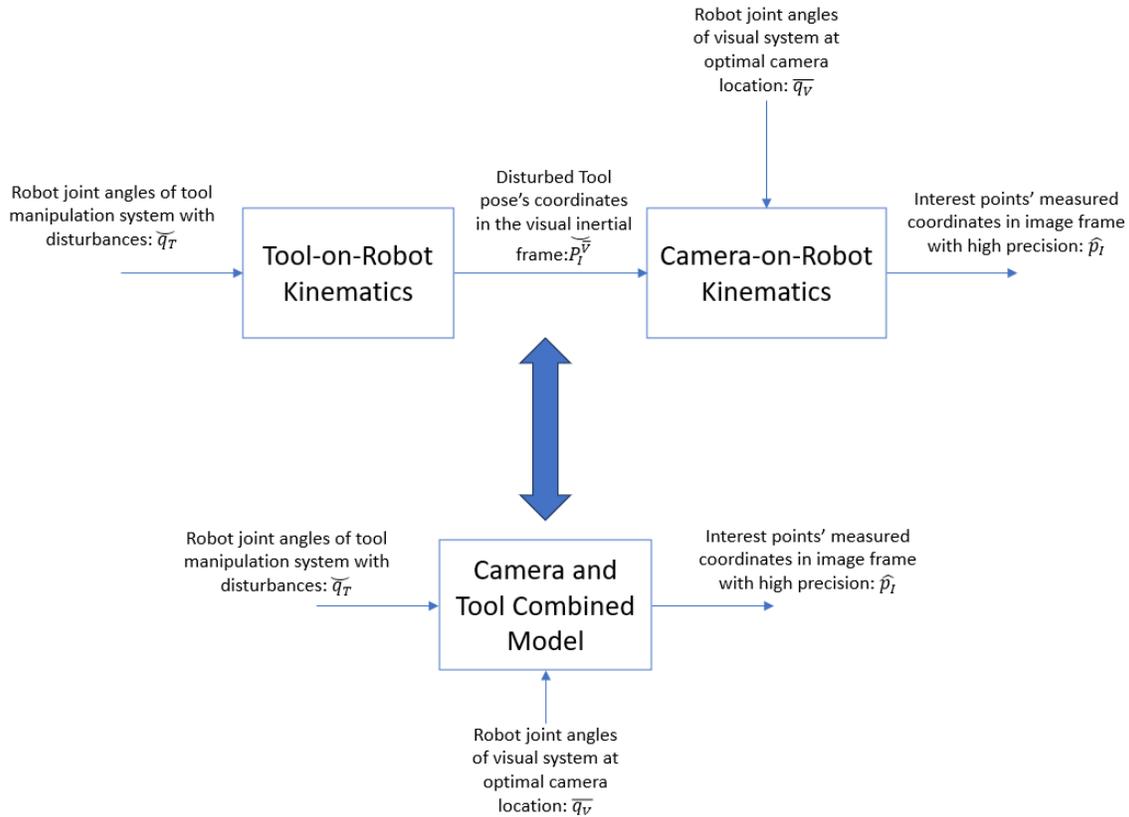

**Figure 2.9**: The Camera-and-Tool Combined model.

## 2.9 Conclusion

In this chapter, we established a multi-robot system that utilizes fiducial markers, with their 2D features employed in the Image-Based Visual Servoing (IBVS) technique to accurately control the pose of both the camera and the tool. A flowchart illustrates an algorithm designed to guide the camera to search for an optimal position in an unknown environment. Additionally, we presented control architectures that adjust the camera pose in response to uncertainties, achieving precise tool positioning. The control strategies incorporate both feedforward and feedback loops, ensuring rapid and accurate movements, which are essential for manufacturing applications.

Furthermore, we discussed the kinematic models used in both the visual and tool manipulation systems, with a particular emphasis on Hardware-in-the-Loop (HIL) models as applied in real-world applications. The mathematical formulations for these models will be provided in Chapter 3. Chapter 4 will provide the algorithm for determining optimal camera placements. Controller development will be addressed in Chapter 5 for the SISO case and Chapter 6 for the MIMO case.

⎯⎯⎯⎯⎯⎯⎯⎯⎯⎯⎯⎯⎯⎯⎯⎯⎯⎯⎯⎯⎯⎯⎯⎯⎯⎯⎯⎯⎯⎯⎯⎯⎯⎯⎯⎯⎯⎯⎯⎯⎯⎯⎯⎯⎯⎯⎯⎯⎯⎯Chapter 3

# System Modeling

This chapter explores the development of mathematical models for a multi-robot system. The first section introduces mathematical representations of camera models, progressing from the pinhole model to the stereo vision model, and explains the process for extracting point coordinates in computer vision. The second and third sections present the derivations of the forward and inverse kinematic models, respectively, for a specific robot manipulator, the ABB IRB 4600. Section 3.4 then details the derivation of the dynamic model for this robot manipulator system, including its actuators. Section 3.5 first discusses the mathematical model of the visual system, which relates joint angles to the image coordinates of three reference points in space, and then focuses on the model for the tool manipulation system, linking joint angles to the image coordinates of two key points on the tool. Finally, section 3.6 introduces simplified one-degree-of-freedom models for the systems described in sections 3.5.

⎯⎯⎯⎯⎯⎯⎯⎯⎯⎯⎯⎯⎯⎯⎯⎯⎯⎯⎯⎯⎯⎯⎯⎯⎯⎯⎯⎯⎯⎯⎯⎯⎯⎯⎯⎯⎯⎯⎯⎯⎯⎯⎯⎯⎯⎯⎯⎯⎯⎯⎯⎯⎯⎯⎯⎯⎯⎯⎯⎯⎯⎯⎯⎯⎯⎯⎯⎯⎯

The first step in control design is to develop accurate system models. These models enable analysis of the system's dynamics, facilitate model-based control design, and serve as representations for simulations. The model should capture the essential plant dynamics while remaining simple enough to support control strategy development. First, a complete six-DoFs model is created, encompassing both the robotic manipulator's degrees of freedom and a stereo camera. Next, a dynamic model of robot arm, including actuators, is developed to support joint positioning control design. Finally, simplified one-DoF models are derived and examined for single-input, single-output (SISO) controllers design.

## 3.1 Coordinate System Setup and Notations

Section 2.1 and Figure 2.1 introduced the topology of the multi-robot system. As described, three reference points ($R_1, R_2, R_3$) are positioned near the tool's target location, and two interest points ($I_1, I_2$) are placed on the tool. In this chapter, we develop mathematical functions to calculate the pose of the stereo camera and the tool in 3D space based on the coordinates of these points. These points are measured in different Cartesian coordinate systems during the calculations.

Figure 3.1 illustrates five distinct Cartesian coordinate systems: two inertial coordinate systems attached to the base of the visual system and the tool manipulation system, respectively, two body-fixed coordinate systems attached to the end-effectors of the visual system and the tool manipulation system, and another body-fixed coordinate attached to the camera. Character $V$ indicates coordinate systems attached to the visual system, while Character $T$ indicates coordinate systems attached to the tool manipulation system. In addition, a bar symbol ($\bar{\ }$) represents inertial coordinate systems, and a superscript symbol ($e$) represents body fixed coordinate systems attached to the end-effector and a superscript symbol ($c$). To summarize, five coordinate systems are expressed as follows:

- Inertial coordinate system attached to the visual system: $\bar{V}$ ($O_{\bar{V}} X_{\bar{V}} Y_{\bar{V}} Z_{\bar{V}}$).
- Body fixed coordinate system attached to the end-effector of visual system: $V^e$ ($O_{V^e} X_{V^e} Y_{V^e} Z_{V^e}$).
- Inertial coordinate system attached to the tool manipulation system: $\bar{T}$ ($O_{\bar{T}} X_{\bar{T}} Y_{\bar{T}} Z_{\bar{T}}$).
- Body fixed coordinate system attached to the end-effector of tool manipulation system: $T^e$ ($O_{T^e} X_{T^e} Y_{T^e} Z_{T^e}$).

- Body fixed coordinate system attached to the camera system: $V^c$ ($O_{V^c}X_{V^c}Y_{V^c}Z_{V^c}$).

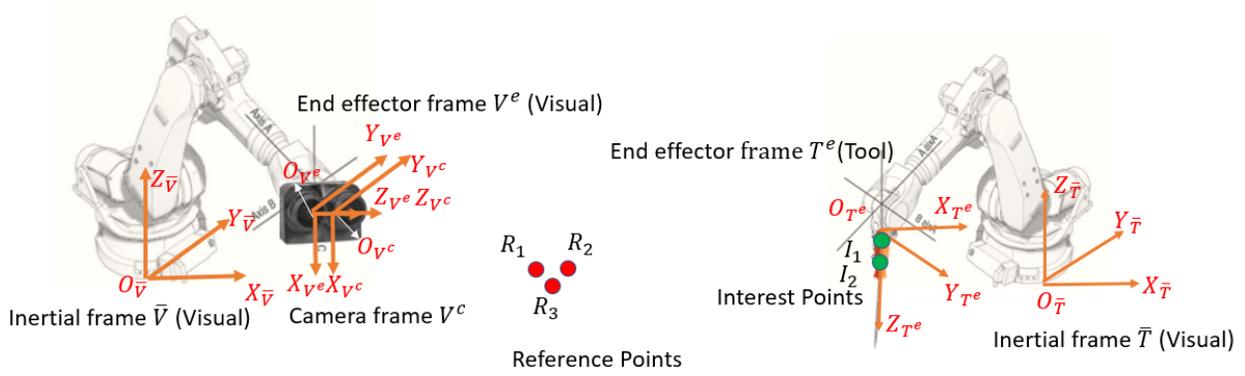

**Figure 3.1:** Coordinate systems in the multi-robotic system.

To represent a point's coordinate measured within a specific coordinate system in space, we use the characters X, Y, and Z to represent x-coordinate, y-coordinate, and z-coordinate, respectively. The superscript indicates the coordinate system in which the coordinates are measured, while the subscript specifies the point being measured.

For example, the 3D coordinates of an interest point $I_1$ measured in the body-fixed coordinate system attached to the visual system's end-effector can be expressed as:

$$P_{I_1}^{V^e} = [X_{I_1}^{V^e}, Y_{I_1}^{V^e}, Z_{I_1}^{V^e}]^T \tag{3.1}$$

The control architecture of the tool manipulator is designed to continuously adjust the current pose of the tool, driving it toward the target pose by minimizing the error between the target 3D coordinates of points and the current disturbed 3D coordinates obtained from the feedback loop. In this notation, an accent ( ¯ ) above the point coordinate expression represents the target coordinates, while an accent ( ˘ ) indicates the current disturbed coordinates of the points.

For example, current disturbed coordinates of an interest point $I_1$ measured in the body-fixed coordinate system attached to the visual system's end-effector can be expressed as:

$$\widetilde{P_{I_1}^{V^e}} = [\widetilde{X_{I_1}^{V^e}}, \widetilde{Y_{I_1}^{V^e}}, \widetilde{Z_{I_1}^{V^e}}]^T \tag{3.2}$$

In a stereo camera model, three coordinates are expressed in the image frame for each point. These coordinates are defined as follows:

- $ul$: The u-coordinate of the point as seen by the left lens.
- $ur$: The u-coordinate of the point as seen by the right lens.
- $v$: The common $v$-coordinate shared by both lenses.

For example, the 2D coordinates of an interest point $R_1$ expressed in the image frame can be expressed as:

$$p_{R_1} = [ul_{R_1}, ur_{R_1}, v_{R_1}]^T \qquad (3.3)$$

The target 2D coordinates of a point are computed using the Camera-on-Robot Kinematics model. Within the control loop, the 2D coordinates of this point can either be directly measured by the stereo camera or estimated using the mathematical model. In this notation, an accent ( ¯ ) above the point coordinate expression represents the target coordinates, while an accent ( ^ ) indicates the measured coordinates, and an accent ( ~ ) indicates the estimated coordinates.

For example, the 2D coordinates of an interest point $R_1$ measured in the image frame can be expressed as:

$$\widehat{p_{R_1}} = [\widehat{ul_{R_1}}, \widehat{ur_{R_1}}, \widehat{v_{R_1}}]^T \qquad (3.4)$$

In the expression of Equations (3.1), (3.2), (3.3) and (3.4), the transpose symbol '$T$' indicates that the vector is represented as a column vector rather than a row vector.

## 3.2 Camera Model

### 3.2.1 Pin-hole Camera Model

The pinhole camera model, illustrated in Figure 3.2, commonly represents a monocular camera setup. In this model, $S_1$ is the distance from the object to the lens, $S_2$ is the distance from the image to the lens, $F$ is the focal length, $D$ is the object's dimension, and $d$ is the image's dimension. According to the thin lens formula:

$$\frac{1}{S_1} + \frac{1}{S_2} = \frac{1}{F} \tag{3.5}$$

which rearranges to:

$$S_2 = \frac{F \cdot S_1}{S_1 - F} \tag{3.6}$$

Additionally, the dimensional relationship is given by:

$$\frac{D}{S_1} = \frac{d}{S_2} \tag{3.7}$$

Combining Equations (3.6) and (3.7), we find:

$$d = \frac{D \cdot F}{S_1 - F} \tag{3.8}$$

The specific camera used in this project is the Zed Stereo camera, which has a fixed focal length. Fixed focal length cameras cannot produce sharp images if positioned too close to an object. According to the Zed camera specifications, the range for $S_1$ to maintain sharp focus is $0.5m \leq S_1 \leq 25m$, and the physical focal length is $F = 2.8mm$. Since $S_1 \gg F$, we can approximate Equation (3.8) by neglecting $F$ in the denominator. Consider an object point $O$ and whose coordinates $P_O^{V^c} = [X_O^{V^c}, Y_O^{V^c}, Z_O^{V^c}, 1]^T$, measured in the camera frame, is projected onto the image plane. Let $D^X$ and $D^Y$ represent the object's dimensions in the $X-$ and $Y-$ directions of the camera frame, respectively. We can substitute $(D^X, D^Y, S_1)$ with the object coordinates $(X_O^{V^c}, Y_O^{V^c}, Z_O^{V^c})$ in the camera frame, resulting in the 2D image coordinates $(u_O, v_O)$ measured in mm:

$$u_O = \frac{F \cdot X_O^{V^c}}{Z_O^{V^c}} \tag{3.9}$$

$$v_O = \frac{F \cdot Y_O^{V^c}}{Z_O^{V^c}} \tag{3.10}$$

We can also express images image coordinates $(u_O, v_O)$ measured in pixels:

$$u_O = \frac{f_u \cdot X_O^{V^c}}{Z_O^{V^c}} \tag{3.11}$$

$$v_O = \frac{f_v \cdot Y_O^{V^c}}{Z_O^{V^c}} \tag{3.12}$$

Here, $f_u$ and $f_v$ measured in pixels are the focal lengths in the *u*- and *v*- directions, respectively and can be precisely determined through camera calibration [25-26]. From Zed 2 specifications tables in Appendix A, we can estimate $f_u$ and $f_v$ from the following equations:

$$f_u = F \frac{N_u}{W} = 2.8mm \frac{1920 pixels}{5.23mm} = 1028 \; pixels \tag{3.13}$$

$$f_v = F \frac{N_v}{H} = 2.8mm \frac{1080 pixels}{2.94mm} = 1028 \; pixels \tag{3.14}$$

Here, W = 5.23 mm and H = 2.94 mm are width and height of Zed 2 camera's image sensor. $N_u$ = 1920 $pixels$ and $N_v$ = 1080 $pixels$ are resolutions in the direction of width and height respectively.

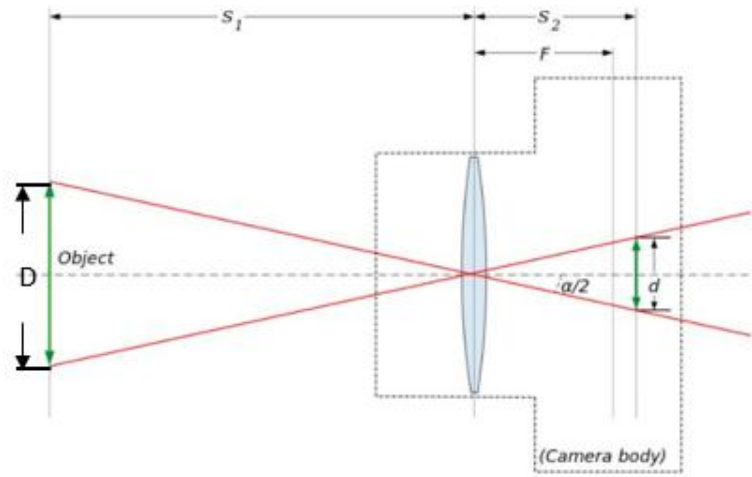

**Figure 3.2**: The pinhole camera model.

## 3.3.2 Stereo Camera Model

Depth between the objects to the camera plane is either approximated or estimated in the IBVS for generating the interaction matrix [5]. Using a stereo camera system in IBVS eliminates the inaccuracies associated with monocular depth estimation, as it directly measures depth by leveraging the disparity between the left and right images.

The stereo camera model is illustrated in Figure 3.3. A stereo camera consists of two lenses separated by a fixed baseline $b$. Each lens has a focal length $F$ (measured in mm) which is the distance from the image plane to the focal point. Assuming the camera is calibrated, the intrinsic parameters: $b, F$ is accurately estimated. A scene point $O$ is measured in the 3D coordinate frame $\{V^c\}$ centered at the middle of the baseline with its coordinates as $[X_O^{V^c}, Y_O^{V^c}, Z_O^{V^c}]^T$. The stereo camera model maps the 3D coordinates of this point to its 2D coordinates projected on the left and right image plane as $[u_{l_O}, v_O]^T$ and $[u_{r_O}, v_O]^T$, respectively. The full camera projection map, incorporating both intrinsic and extrinsic parameters, is given by:

$$s \cdot p_O = K \cdot [R|T] \cdot P_O^{V^c} \qquad (3.15)$$

Where $p_O$ are the image coordinates of the point and $P_O^{V^c}$ are the 3D coordinates measured in the camera frame $\{V^c\}$. $s$ is the scale factor that ensures correct projection between 2D and 3D features. $K$ is the intrinsic matrix with a size of 3×3, and the mathematical expression is presented as:

$$K = \begin{bmatrix} F & k & u_0 \\ 0 & F & v_0 \\ 0 & 0 & 1 \end{bmatrix} \qquad (3.16)$$

Where $k$ is the skew factor, which represents the angle between the image axes ($u$ and $v$ axis). $u_0$ and $v_0$ are coordinates offsets in image planes.

In Equation (3.15), $R$ is the rotational matrix from camera frame $\{V^c\}$ to each image coordinate frame, and $T$ is the translation matrix from camera frame $\{V^c\}$ to each camera lens center. Since there is no rotation between the camera frame $\{V^c\}$ and image frames but only a translation along the $X_{V^c}$ axis occurs, the transformation matrices for the left and right image planes are expressed as:

$$[R|T]_{Left} = \begin{bmatrix} 1 & 0 & 0 & -b/2 \\ 0 & 1 & 0 & | & 0 \\ 0 & 0 & 1 & 0 \end{bmatrix} \qquad (3.17)$$

$$[R|T]_{Right} = \begin{bmatrix} 1 & 0 & 0 & b/2 \\ 0 & 1 & 0 & | & 0 \\ 0 & 0 & 1 & 0 \end{bmatrix} \quad (3.18)$$

Assume the u and v axis are perfectly perpendicular (take k = 0), and there are no offsets in the image coordinates (take $u_0 = v_0 = 0$) for both lens. Also, set factor $s = Z_O^{V^c}$ accounts for perspective depth scaling. The projection equations for the left and right image planes can be rewritten in homogeneous coordinates as:

$$Z_O^{V^c} \cdot \begin{bmatrix} u_{lO} \\ v_O \\ 1 \end{bmatrix} = \begin{bmatrix} F & 0 & 0 \\ 0 & F & 0 \\ 0 & 0 & 1 \end{bmatrix} \cdot \begin{bmatrix} 1 & 0 & 0 & b/2 \\ 0 & 1 & 0 & | & 0 \\ 0 & 0 & 1 & 0 \end{bmatrix} \cdot \begin{bmatrix} X_O^{V^c} \\ Y_O^{V^c} \\ Z_O^{V^c} \\ 1 \end{bmatrix} \quad (3.19)$$

$$Z_O^{V^c} \cdot \begin{bmatrix} u_{rO} \\ v_O \\ 1 \end{bmatrix} = \begin{bmatrix} F & 0 & 0 \\ 0 & F & 0 \\ 0 & 0 & 1 \end{bmatrix} \cdot \begin{bmatrix} 1 & 0 & 0 & -b/2 \\ 0 & 1 & 0 & | & 0 \\ 0 & 0 & 1 & 0 \end{bmatrix} \cdot \begin{bmatrix} X_O^{V^c} \\ Y_O^{V^c} \\ Z_O^{V^c} \\ 1 \end{bmatrix} \quad (3.20)$$

We can also generate expressions for image coordinates in pixels similar to the monocular camera model in Equations (3.13) and (3.14):

$$Z_O^{V^c} \cdot \begin{bmatrix} u_{lO} \\ v_O \\ 1 \end{bmatrix} = \begin{bmatrix} f_u & 0 & 0 \\ 0 & f_v & 0 \\ 0 & 0 & 1 \end{bmatrix} \cdot \begin{bmatrix} 1 & 0 & 0 & b/2 \\ 0 & 1 & 0 & | & 0 \\ 0 & 0 & 1 & 0 \end{bmatrix} \cdot \begin{bmatrix} X_O^{V^c} \\ Y_O^{V^c} \\ Z_O^{V^c} \\ 1 \end{bmatrix} \quad (3.21)$$

$$Z_O^{V^c} \cdot \begin{bmatrix} u_{rO} \\ v_O \\ 1 \end{bmatrix} = \begin{bmatrix} f_u & 0 & 0 \\ 0 & f_v & 0 \\ 0 & 0 & 1 \end{bmatrix} \cdot \begin{bmatrix} 1 & 0 & 0 & -b/2 \\ 0 & 1 & 0 & | & 0 \\ 0 & 0 & 1 & 0 \end{bmatrix} \cdot \begin{bmatrix} X_O^{V^c} \\ Y_O^{V^c} \\ Z_O^{V^c} \\ 1 \end{bmatrix} \quad (3.22)$$

Equations (3.19) - (3.22) establish the mathematical relationship between the 3D coordinates of a point in the camera frame $\{V^c\}$ and its 2D projections on the left and right image planes. The image coordinates' value along the $v$-axis remain the same for both images. In this thesis, let's consider measuring image coordinates in mm, and we can expand Equations (3.19) and (3.20) to the following nonlinear equations:

$$v_O = \frac{Y_O^{V^c}}{Z_O^{V^c}} F \quad (3.23)$$

$$u_{lO} = \frac{2X_O^{V^c} - b}{2Z_O^{V^c}} F \quad (3.24)$$

$$u_{ro} = \frac{2X_0^{V^c} + b}{2Z_0^{V^c}} F \qquad (3.25)$$

In conclusion, a scene point's 3D coordinates can be mapped to a set of three image coordinates in the stereo camera system, expressed as:

Stereo-camera mapping $M$ : $P_O^{V^c} = [X_O^{V^c}, Y_O^{V^c}, Z_O^{V^c}]^T \xRightarrow{M} p_O = [u_{lo}, u_{ro}, v_O]^T$ (3.26)

The mapping function $M$ is nonlinear and depends on the stereo camera parameters $P_{a_{camera}}$, specifically $b$, and $F$. The specifications of the Zed 2 camera model are summarized in Appendix A Table A3.

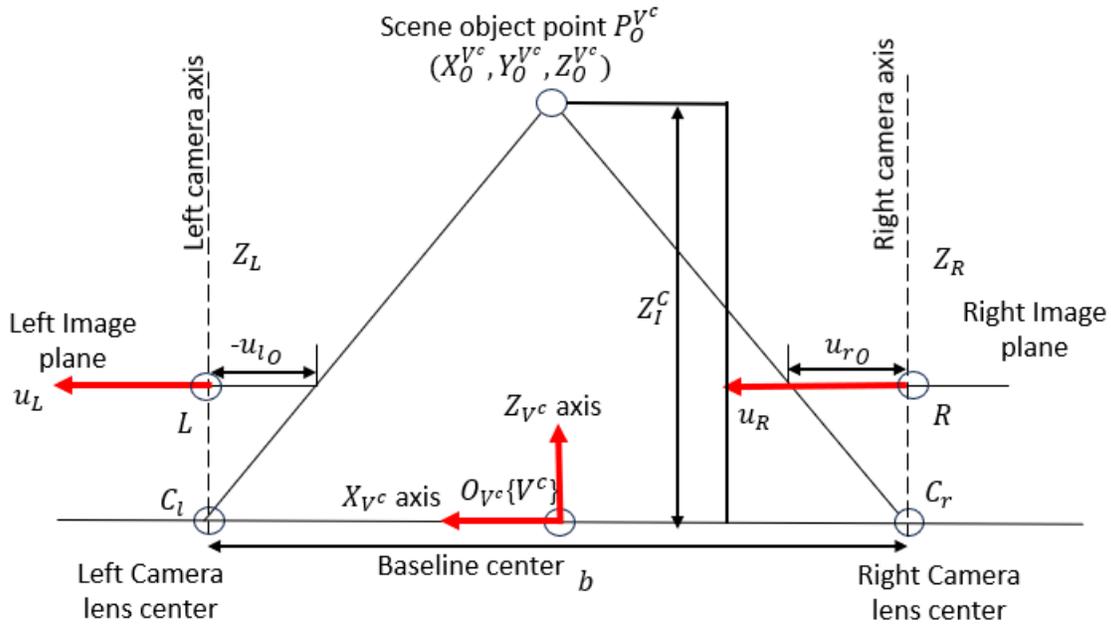

**Figure 3.3:** Projection of a scene object onto the image planes of a stereo camera. *Note: The v-coordinate on each image plane is not displayed in this plot but is measured along the axis that is perpendicular to and pointing out of the plot.*

## 3.2.3 Inverse Stereo Camera Model

In this section, we define the inverse mapping of the stereo projection function $M$, denoted as $M^{-1}$, which transforms image coordinates of a scene point into 3D coordinates expressed in the camera-centered coordinate frame.

That is:

Inverse Stereo-camera mapping $M^{-1}$ :

$$p_O = [u_{l_O}, u_{r_O}, v_O]^T \xRightarrow{M^{-1}} P_O^{V^c} = [X_O^{V^c}, Y_O^{V^c}, Z_O^{V^c}]^T \tag{3.27}$$

This inverse mapping $M^{-1}$ is nonlinear and can be derived by rearranging Equations (3.23)–(3.25). The resulting expressions are:

$$X_O^{V^c} = \frac{(u_{l_O} + u_{r_O})b}{2(u_{r_O} - u_{l_O})} \tag{3.28}$$

$$Y_O^{V^c} = \frac{v \cdot b}{(u_{r_O} - u_{l_O})} \tag{3.29}$$

$$Z_O^{V^c} = \frac{b \cdot F}{(u_{r_O} - u_{l_O})} \tag{3.30}$$

The mapping is unique as for every valid triplet of image coordinates $(u_{l_O}, u_{r_O}, v_O)$, the corresponding 3D point $(X_O^{V^c}, Y_O^{V^c}, Z_O^{V^c})$ is uniquely defined.

Moreover, $M^{-1}$ is undefined only when the disparity $u_{r_O} - u_{l_O} = 0$, since the expressions for $X_O^{V^c}$, $Y_O^{V^c}$, and $Z_O^{V^c}$ involve division by this quantity. In this case, the depth $Z_O^{V^c}$ tends to infinity, corresponding to a point located at infinity along the viewing direction—i.e., where the two rays from the stereo cameras are parallel and do not intersect.

However, such a situation does not occur in practical scenarios, as it implies the cameras cannot triangulate a finite 3D point. Therefore, as long as the stereo system is able to detect and localize a point in both views with non-zero disparity, the inverse mapping exists and yields a unique 3D representation of the point in the camera-centered coordinate frame.

### 3.2.4 Point Feature Extraction

To capture and locate points in the image, circular fiducial markers are placed, allowing the Hough transform in computer vision to detect and localize their centers. The general Hough transform process is outlined below.

A circle with a radius $R$ and center $(a, b)$ can be defined by the equation:
$$(x - a)^2 + (y - b)^2 = R^2 \tag{3.31}$$
Where $(x, y)$ represents points on the circle's perimeter. The three-element tuple $(a, b, R)$ uniquely characterizes a circle in an image.

To illustrate this mapping, we only consider a single circle on the tool. For each point $(x, y)$ on the circle's perimeter, the set of possible centers $(a, b)$ with radius $R$ forms a circle centered at $(x, y)$ the parameters space. This process is shown in Figure 3.4., where three points are used as an example. Each point in the geometric space generates a corresponding circle in the parameter space. If the points belong to the same circle, the circles in the parameter space intersect at the center, which represents the actual center of the original circle in geometric space.

During the Hough transform calculation, each image point $(x, y)$ on the circle's perimeter generates a new circle in parameter space, with $(x, y)$ as the center and radius iterated over a predefined range. For a given radius $R$, the generated circles in parameter space intersect at various points. Among these intersections, the point $(a, b)$ with the highest number of overlaps is identified and form the tuple $(a, b, R)$ which represents a candidate for the circle's center and radius in geometric space. By exploring different radii and generating corresponding tuples, the point $(a, b)$ with the maximum number of intersections across all tuples provides the best estimate for the circle's center, while the corresponding $R$ represents the estimated radius in geometric space.

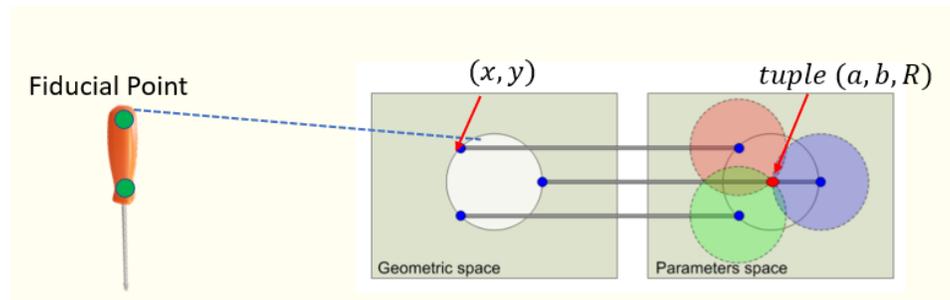

**Figure 3.4:** Feature extraction by Hough transformation.
*Note: the red point in parameters space represents a possible tuple $(a, b, R)$.*

For the tool, two circular fiducial markers are placed to capture its pose (position and orientation). The stereo camera then provides the image coordinates of the centers of these fiducial markers as follows:

$$\widehat{p_{I_1}} = [\widehat{ul_{I_1}}, \widehat{ur_{I_1}}, \widehat{v_{I_1}}]^T \tag{3.32}$$

$$\widehat{p_{I_2}} = [\widehat{ul_{I_2}}, \widehat{ur_{I_2}}, \widehat{v_{I_2}}]^T \tag{3.33}$$

where $\widehat{p_{I_1}^T}$ and $\widehat{p_{I_2}^T}$ represent the image coordinates of the centers of the first and second fiducial markers, respectively. Here, $\widehat{ul}$ and $\widehat{ur}$ are the $u$- axis coordinates measured on the left and right image planes of the stereo camera, while $\hat{v}$ represents the $v$- axis coordinates.

## 3.3 Forward Kinematics of the Elbow Manipulator

A widely used method for defining and generating reference frames in robotic applications is the Denavit-Hartenberg (D-H) convention [37]. In this approach, each robotic link is associated with a Cartesian coordinate frame $O_i X_i Y_i Z_i$. According to the D-H convention, the homogeneous transformation matrix $A_i^{i-1}$ which represents the transformation from frame $i-1$ to frame $i$, can be decomposed into a sequence of four fundamental transformations:

$$A_i^{i-1} = Rot_{z,q_i} Trans_{z,d_i} Trans_{x,a_i} Rot_{x,\alpha_i}$$

$$= \begin{bmatrix} c_{q_i} & -s_{q_i} & 0 & 0 \\ s_{q_i} & c_{q_i} & 0 & 0 \\ 0 & 0 & 1 & 0 \\ 0 & 0 & 0 & 1 \end{bmatrix} \begin{bmatrix} 1 & 0 & 0 & 0 \\ 0 & 1 & 0 & 0 \\ 0 & 0 & 1 & d_i \\ 0 & 0 & 0 & 1 \end{bmatrix} \begin{bmatrix} 1 & 0 & 0 & a_i \\ 0 & 1 & 0 & 0 \\ 0 & 0 & 1 & 0 \\ 0 & 0 & 0 & 1 \end{bmatrix} \begin{bmatrix} 1 & 0 & 0 & 0 \\ 0 & c_{\alpha_i} & -s_{\alpha_i} & 0 \\ 0 & s_{\alpha_i} & c_{\alpha_i} & 0 \\ 0 & 0 & 0 & 1 \end{bmatrix} \tag{3.34}$$

$$= \begin{bmatrix} c_{q_i} & -s_{q_i} c_{\alpha_i} & s_{q_i} s_{\alpha_i} & a_i c_{q_i} \\ s_{q_i} & c_{q_i} c_{\alpha_i} & -c_{q_i} s_{\alpha_i} & a_i s_{q_i} \\ 0 & s_{\alpha_i} & c_{\alpha_i} & d_i \\ 0 & 0 & 0 & 1 \end{bmatrix} \tag{3.35}$$

Note: $c_{\theta_i} \equiv \cos(q_i), c_{\alpha_i} \equiv \cos(\alpha_i), s_{\theta_i} \equiv \sin(q_i), s_{\alpha_i} \equiv \sin(\alpha_i)$ \hfill (3.36)

Here, $q_i, a_i, \alpha_i$ and $d_i$ are parameters of link $i$ and joint $i$, $a_i$ is the link length, $q_i$ is the rotational angle, $\alpha_i$ is the twist angle and $d_i$ is the offset length between the previous $(i-1)^{th}$ and the current $i^{th}$ robot links. The values of these parameters in Equation (3.35) are calculated following the procedure outlined in [37].

To determine the transformation from the end-effector frame $O_6X_6Y_6Z_6$ (**denoted as $P^6$**) to the base frame $O_0X_0Y_0Z_0$ (denoted as $P^0$), we compute the cumulative transformation matrix $T_6^0$ as follows:

$$T_6^0 = A_1^0 A_2^1 A_3^2 A_4^3 A_5^4 A_6^5 \qquad (3.37)$$

If a point $P^6$ is defined in the end-effector frame, we can find its coordinates in the base frame $P^0$ by applying this transformation:

$$P^0 = T_6^0 \, P^6 \qquad (3.38)$$

Furthermore, the transformation matrix from the base frame $P^0$ to the end-effector frame $P^6$ can be derived by taking the inverse of $T_6^0$:

$$T_0^6 = (T_6^0)^{-1} \qquad (3.39)$$

which is used to generate the image coordinates of a point captured by a camera with its center attached to the end effector, from the 3D coordinates of a point in the base frame.

Equation (3.38) shows that the position of the end-effector, denoted as $P^{end}$, in the base frame (where $P^{end}$ is located at the origin of the end-effector frame $P^6$) depends on the joint angles $q = [q_i | i \in 1, 2, 3, 4, 5, 6]$ and the robot parameters $P_{a_{robot}} = [a_i, \alpha_i, d_i | i \in 1, 2, 3, 4, 5, 6]$. This relationship can be expressed as:

$$P^{end} = \mathcal{F}(q, P_{a_{robot}}) \qquad (3.40)$$

Equation (3.40) defines the forward kinematic model of the robot manipulator. This model is used to calculate the position of the end effector in the base frame based on the joint angles and specific robot parameters.

Figure 3.5. illustrates the coordinate systems attached to an ABB IRB 4600 manipulator [38], an elbow-type robot equipped with a spherical wrist. This robot has six links in total: three that comprise the arm and three that form the wrist. Each pair of adjacent links is connected by a joint, with five in total. The joint axes $Z_0, \cdots Z_5$ represent the rotational directions of each joint, with corresponding rotational angles $q_1, \cdots q_6$. A key advantage of the spherical wrist is that the center of the wrist (where $Z_3$, $Z_4$ and $Z_5$ intersect) remains fixed regardless of wrist orientation. This configuration decouples the wrist's orientation from its position, allowing the orientation task to be completed without affecting the wrist center's position. Each frame's coordinate system is established according to the steps outlined in Appendix C, using the Denavit-Hartenberg convention for deriving forward kinematics. Specifically, for the coordinate frame $O_6X_6Y_6Z_6$ attached to the end-effector:

$\widehat{k_6} = \hat{a}$ is aligned with the direction of $z_5$,

$\widehat{j_6} = \hat{s}$ points in the direction of gripper closure,

$\widehat{i_6} = \hat{n} = \hat{s} \times \hat{a}$ forming a right-handed coordinate system.

In the base frame $O_0X_0Y_0Z_0$, $\hat{a}, \hat{s}$ and $\hat{n}$ defined the orientation of the tool or camera mounted on the end-effector.

The parameters $q_i, a_i, \alpha_i$ and $d_i$ for each link, as given in Equation (3.31), are listed in Table 3.1. and calculated according to the procedures in [37].

**Table 3.1**: DH-Parameter for elbow manipulator with spherical wrist.

| Link | $a_i$ | $\alpha_i$(rad) | $d_i$ | $q_i$(rad) |
| --- | --- | --- | --- | --- |
| 1 | $a_1$ | $-\pi/2$ | $L_1$ | $q_1^*$ |
| 2 | $L_2$ | 0 | 0 | $q_2^* - \pi/2$ |
| 3 | $L_3$ | $-\pi/2$ | 0 | $q_3^*$ |
| 4 | 0 | $\pi/2$ | $L_4$ | $q_4^*$ |
| 5 | 0 | $-\pi/2$ | 0 | $q_5^*$ |
| 6 | 0 | 0 | $L_t$ | $q_6^* + \pi$ |

\* accounts for variables

As shown in Figure 3.4, the $x_2$-axis is obtained by rotating the $x_1$-axis around $z_1$-axis by $-\frac{\pi}{2}$, and the $x_6$-axis is obtained by rotating the $x_5$-axis around $z_5$-axis by $\pi$. According to the procedures for constructing coordinate frames based on the D-H convention outlined in Appendix C, the angles $-\pi/2$ and $\pi$ are added to the joint angle column to represent the directional changes of the x-axes.

Using the parameters provided in Table 1.1. and applying Equation (3.35), we can calculate the transformation matrix for each coordinate frame.

$$A_1^0 = \begin{bmatrix} c_{q_1^*} & 0 & -s_{q_1^*} & a_1 c_{q_1^*} \\ s_{q_1^*} & 0 & c_{q_1^*} & a_1 s_{q_1^*} \\ 0 & -1 & 0 & L_1 \\ 0 & 0 & 0 & 1 \end{bmatrix} \tag{3.41}$$

$$A_2^1 = \begin{bmatrix} s_{q_2^*} & c_{q_2^*} & 0 & L_2 s_{q_2^*} \\ -c_{q_2^*} & s_{q_2^*} & 0 & -L_2 c_{q_2^*} \\ 0 & 0 & 1 & 0 \\ 0 & 0 & 0 & 1 \end{bmatrix} \tag{3.42}$$

$$A_3^2 = \begin{bmatrix} c_{q_3^*} & 0 & -s_{q_3^*} & L_3 c_{q_3^*} \\ s_{q_3^*} & 0 & c_{q_3^*} & L_3 s_{q_3^*} \\ 0 & -1 & 0 & 0 \\ 0 & 0 & 0 & 1 \end{bmatrix} \tag{3.43}$$

$$A_4^3 = \begin{bmatrix} c_{q_4^*} & 0 & s_{q_4^*} & 0 \\ s_q & 0 & -c_{q_4^*} & 0 \\ 0 & 1 & 0 & L_4 \\ 0 & 0 & 0 & 1 \end{bmatrix} \tag{3.44}$$

$$A_5^4 = \begin{bmatrix} c_{q_5^*} & 0 & -s_{q_5^*} & 0 \\ s_{q_5^*} & 0 & c_{q_5^*} & 0 \\ 0 & -1 & 0 & 0 \\ 0 & 0 & 0 & 1 \end{bmatrix} \tag{3.45}$$

$$A_6^5 = \begin{bmatrix} -c_{q_6^*} & s_{q_6^*} & 0 & 0 \\ -s_{q_6^*} & -c_{q_6^*} & 0 & 0 \\ 0 & 0 & 1 & L_t \\ 0 & 0 & 0 & 1 \end{bmatrix} \tag{3.46}$$

Then the transformation from frame $O_6 X_6 Y_6 Z_6$ to frame $O_0 X_0 Y_0 Z_0$ is:

$$T_6^0 = A_1^0 A_2^1 A_3^2 A_4^3 A_5^4 A_6^5 \tag{3.47}$$

$$T_6^0 = \begin{bmatrix} n_x & s_x & a_x & d_x \\ n_y & s_y & a_y & d_y \\ n_z & s_z & a_z & d_z \\ 0 & 0 & 0 & 1 \end{bmatrix} \tag{3.48}$$

By multiplying each term in Equation (3.47) and simplify with trigonometry, each entry in matrix Equation (3.48) can be expressed as follows:

$$n_x = c_1 s_{2,3}(s_4 s_6 - c_4 c_5 c_6) - s_1(s_4 c_5 c_6 + c_4 s_6) - c_1 c_{2,3} s_5 c_6$$
$$n_y = s_1 s_{2,3}(s_4 s_6 - c_4 c_5 c_6) + c_1(s_4 c_5 c_6 + c_4 s_6) - s_1 c_{2,3} s_5 c_6$$
$$n_z = c_{2,3}(s_4 s_6 - c_4 c_5 c_6) + s_{2,3} s_5 c_6 \quad (3.49)$$

$$s_x = c_1 s_{2,3}(s_4 c_6 + c_4 c_5 c_6) + s_1(s_4 c_5 s_6 - c_4 c_6) + c_1 c_{2,3} s_5 s_6$$
$$s_y = s_1 s_{2,3}(s_4 c_6 + c_4 c_5 c_6) - c_1(s_4 c_5 s_6 - c_4 c_6) + s_1 c_{2,3} s_5 s_6$$
$$s_z = c_{2,3}(s_4 c_6 + c_4 c_5 c_6) - s_{23} s_5 s_6 \quad (3.50)$$

$$a_x = -c_1 s_{2,3} c_4 s_5 - s_1 s_4 s_5 + c_1 c_{2,3} c_5$$
$$a_y = -s_1 s_{2,3} c_4 s_5 + c_1 s_4 s_5 + s_1 c_{2,3} c_5$$
$$a_z = c_{2,3} c_4 s_5 - s_{2,3} c_5 \quad (3.51)$$

$$d_x = L_t(-c_1 s_{2,3} c_4 s_5 - s_1 s_4 s_5 + c_1 c_{2,3} c_5) + c_1(L_4 c_{2,3} + L_3 s_{2,3} + L_2 s_2 + a_1)$$
$$d_y = L_t(-s_1 s_{2,3} c_4 s_5 + c_1 s_4 s_5 + s_1 c_{2,3} c_5) + s_1(L_4 c_{2,3} + L_3 s_{2,3} + L_2 s_2 + a_1)$$
$$d_z = -L_t(c_{2,3} c_4 s_5 + s_{2,3} c_5) - L_4 s_{2,3} + L_3 c_{2,3} + L_2 c_2 + L_1 \quad (3.52)$$

Note:
$$c_i \equiv \cos(q_i),\ s_i \equiv \sin(q_i)$$
$$c_{i,j} \equiv \cos(q_i + q_j),\ s_{i,j} \equiv \sin(q_i + q_j)$$
$$i,j \in \{1,2,3,4,5,6\} \quad (3.53)$$

Here, the vectors $[n_x, n_y, n_z]^T$, $[s_x, s_y, s_z]^T$ and $[a_x, a_y, a_z]^T$ represent the end-effector's directional vectors for Yaw, Pitch, and Roll, respectively, in the base frame $O_0 X_0 Y_0 Z_0$. These vectors define the orientation of the end-effector in terms of its alignment relative to each principal axis in the base frame. Additionally, the vector $[d_x, d_y, d_z]^T$ denotes the absolute position of the center of the end-effector in the base frame $O_0 X_0 Y_0 Z_0$, providing the exact spatial coordinates of the end-effector's origin relative to the base. For a specific model ABB IRB 4600-45/2.05 (Handling capacity: 45 kg/ Reach 2.05m) [30], further details regarding its dimensions and mass are provided in Appendix Table A1.

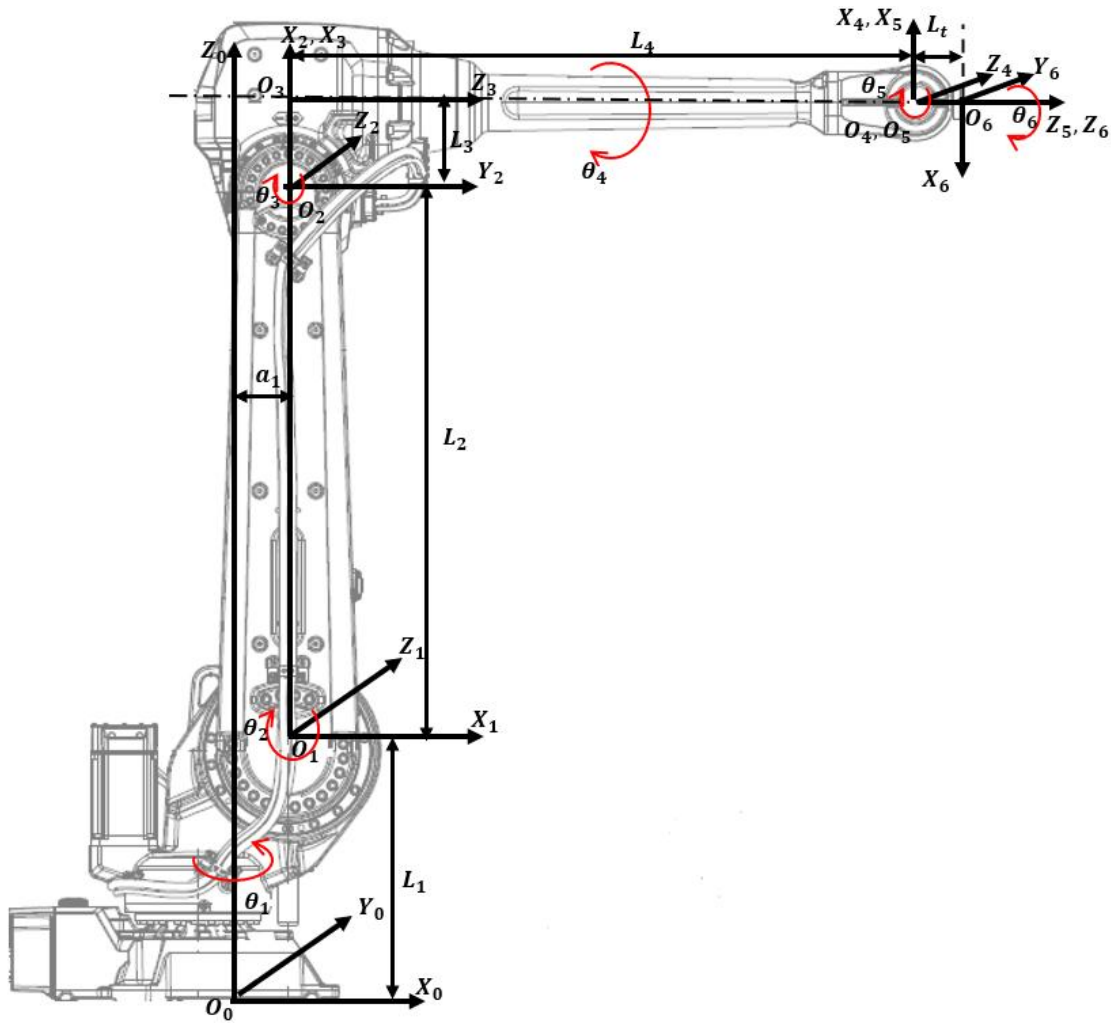

**Figure 3.5:** IRB ABB 4600 model with attached frames.

## 3.4 Inverse Kinematics of the Robot Manipulator

In the previous section, we explored how to determine the pose of the end-effector in the inertia frame based on the joint angles and link parameters. The reverse process, known as inverse kinematics, involves computing the joint angles from a given end-effector position and orientation relative to the inertia frame.

For robot arms equipped with a spherical wrist, the inverse kinematics problem can often be simplified by decoupling it into inverse position kinematics and inverse orientation kinematics. This simplification is possible because the last three joints intersect at a single point, known as the wrist center. The first three joints of the manipulator control the 3D position of the wrist center (and, consequently, the end-effector), while the last three joints independently determine the orientation of the end-effector. In contrast, for robot arms without a defined wrist center, this decoupling is not generally possible. In such cases, the motion of the last three joints can also influence the position of the end-effector, leading to a more complex and coupled kinematics problem. In applications involving elbow manipulators with spherical wrists, this allows us to solve the inverse kinematics in two steps:

- Determine the position of the wrist center relative to the inertial base frame, which provides the first set of joint variables.
- Determine the orientation of the wrist based on this wrist center position, yielding the remaining joint variables.

The inverse kinematics problem does not always yield a unique solution. For a given end-effector pose matrix, as defined in Equation (3.44), there may exist multiple valid joint angle configurations that achieve the same position and orientation. This multiplicity arises from the nonlinear structure of the manipulator's kinematics and the existence of geometrically distinct configurations, such as elbow-up (above) and elbow-down (down) postures. At singular configurations, the problem becomes even more complex: the robot may admit infinitely many joint angle solutions corresponding to the same pose. This occurs when the manipulator loses a degree of freedom due to joint axis alignment or workspace boundaries, making the mapping from pose to joint space non-unique or underdetermined.

Below, we present a structured approach to solving the inverse kinematics for an elbow-type manipulator with a spherical wrist. The first subsection addresses general solutions in non-singular configurations, while the second subsection focuses on singular configurations and their implications for joint angle determination. Finally, we discuss the strategy for selecting a unique joint configuration from among the multiple possible solutions, ensuring consistency and stability in control applications.

## 3.4.1 Inverse Kinematics Solutions of Non-singularity Pose

In Equation (3.48), we can reform the transformation matrix $T_6^0$ from the end-effector frame to the base frame as:

$$T_6^0 = \begin{bmatrix} R & d \\ 0 & 1 \end{bmatrix} \tag{3.54}$$

where

$$R = \begin{bmatrix} n_x & s_x & a_x \\ n_y & s_y & a_y \\ n_z & s_z & a_z \end{bmatrix} \tag{3.55}$$

$$d = \begin{bmatrix} d_x \\ d_y \\ d_z \end{bmatrix} \tag{3.56}$$

In this context, $R$ and $d$ represent the orientation and position of the end-effector in the base frame, respectively. The position of the wrist center $P = [P_x, P_y, P_z]^T$ can then be calculated using the end-effector position $d$ and orientation component $[a_x, a_y, a_z]^T$ as follows:

$$P_x = d_x - L_t a_x \tag{3.57}$$
$$P_y = d_y - L_t a_y \tag{3.58}$$
$$P_z = d_z - L_t a_z \tag{3.59}$$

where $L_t$ is the distance from the wrist center to the end-effector along the a-axis. This provides the coordinates of the wrist center.

In general, there are four distinct joint configurations that can result in the same wrist center position: the right-above configuration, right-below configuration, left-above configuration, and left-below configuration. These configurations arise from combinations of shoulder (left or right) and elbow (above or below) positions that yield the same end-effector location. Figure 3.6. illustrates these four joint configurations for a given wrist center position.

In addition, in the spherical wrist, for a given orientation, there exist two distinct sets of joint angles $\{q_4, q_5, q_6\}$ that produce the same end-effector orientation. As illustrated in Figure 3.7., these two sets— $\{q_4, q_5, q_6\}$ and $\{q_4 + \pi, -q_5, q_6 + \pi\}$— are kinematically equivalent and yield the same rotation matrix for the end-effector.

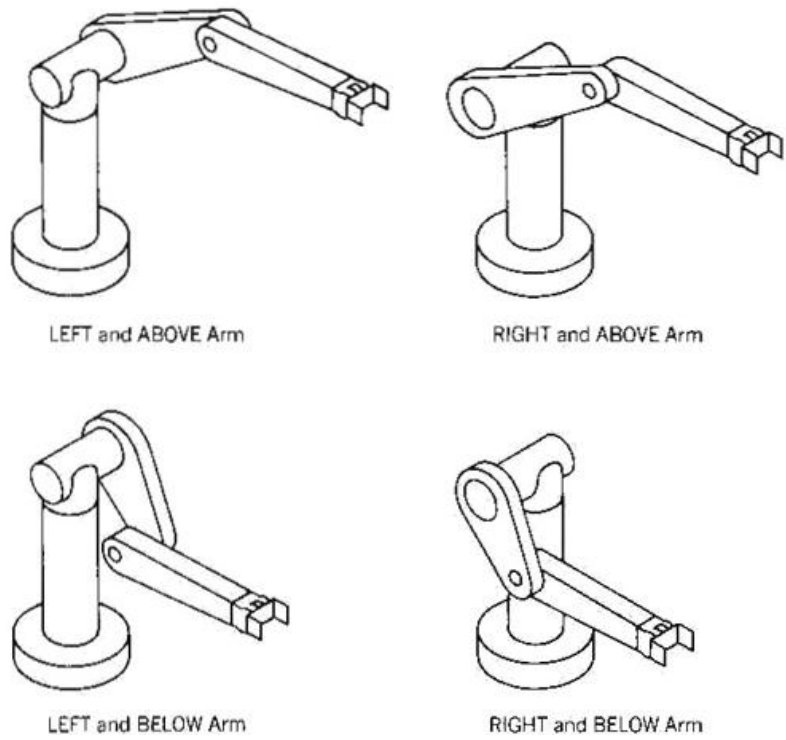

**Figure 3.6:** Four solutions of the inverse kinematics for a specific location.

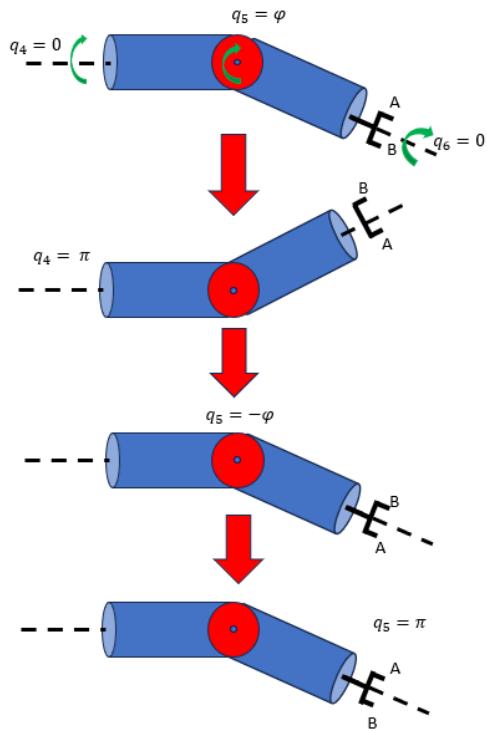

**Figure 3.7:** Demonstration of two equivalent joint configurations for the orientation.

The joint angles for the manipulator can be computed using the following expressions:

## Joint angles that relate to the wrist center locations $(q_1, q_2, q_3)$:

### Right above Configuration:

$$q_1 = arctan(\frac{P_y}{P_x}) \qquad (3.60.A)$$

$$q_2 = \frac{\pi}{2} - arccos\left(\frac{(r^2 + Z^2) + L_2^2 - (L_3^2 + L_4^2)}{2 \cdot \sqrt{r^2 + Z^2} \cdot L_2}\right) - arctan\left(\frac{Z}{r}\right) \qquad (3.61.A)$$

$$q_3 = \pi - arccos\left(\frac{(L_3^2 + L_4^2) + L_2^2 - (r^2 + Z^2)}{2 \cdot \sqrt{L_3^2 + L_4^2} \cdot L_2}\right) - arctan\left(\frac{L_4}{L_3}\right) \qquad (3.62.A)$$

### Right down Configuration:

$$q_1 = arctan(\frac{P_y}{P_x}) \qquad (3.60.B)$$

$$q_2 = \frac{\pi}{2} + arccos\left(\frac{(r^2 + Z^2) + L_2^2 - (L_3^2 + L_4^2)}{2 \cdot \sqrt{r^2 + Z^2} \cdot L_2}\right) - arctan\left(\frac{Z}{r}\right) \qquad (3.61.B)$$

$$q_3 = \pi + arccos\left(\frac{(L_3^2 + L_4^2) + L_2^2 - (r^2 + Z^2)}{2 \cdot \sqrt{L_3^2 + L_4^2} \cdot L_2}\right) - arctan\left(\frac{L_4}{L_3}\right) \qquad (3.62.B)$$

### Left above Configuration:

$$q_1 = arctan\left(\frac{P_y}{P_x}\right) - \pi \qquad (3.60.C)$$

$$q_2 = -\frac{\pi}{2} + arccos\left(\frac{(r^2 + Z^2) + L_2^2 - (L_3^2 + L_4^2)}{2 \cdot \sqrt{r^2 + Z^2} \cdot L_2}\right) + arctan\left(\frac{Z}{r}\right) \qquad (3.61.C)$$

$$q_3 = -\pi + arccos\left(\frac{(L_3^2 + L_4^2) + L_2^2 - (r^2 + Z^2)}{2 \cdot \sqrt{L_3^2 + L_4^2} \cdot L_2}\right) - arctan\left(\frac{L_4}{L_3}\right) \qquad (3.62.C)$$

**Left down Configuration:**

$$q_1 = arctan\left(\frac{P_y}{P_x}\right) - \pi \tag{3.60.D}$$

$$q_2 = -\frac{\pi}{2} - \arccos\left(\frac{(r^2 + Z^2) + L_2{}^2 - (L_3{}^2 + L_4{}^2)}{2 \cdot \sqrt{r^2 + Z^2} \cdot L_2}\right) + \arctan\left(\frac{Z}{r}\right) \tag{3.61.D}$$

$$q_3 = -\pi - \arccos\left(\frac{(L_3{}^2 + L_4{}^2) + L_2{}^2 - (r^2 + Z^2)}{2 \cdot \sqrt{L_3{}^2 + L_4{}^2} \cdot L_2}\right) - \arctan\left(\frac{L_4}{L_3}\right) \tag{3.62.D}$$

## Joint angles that relate to the end-effector orientations ($q_4, q_5, q_6$):

**The first solution set:**

$$q_4 = arctan\left(\frac{s_1 a_x - c_1 a_y}{c_1 s_{2,3} a_x + s_1 s_{2,3} a_y + c_{2,3} a_z}\right) \tag{3.63.A}$$

$$q_5 = \arccos\left(c_1 c_{2,3} a_x + s_1 c_{2,3} a_y - s_{2,3} a_z\right) \tag{3.64.A}$$

$$q_6 = -arctan\left(\frac{c_1 c_{2,3} s_x + s_1 c_{2,3} s_y - s_{2,3} s_z}{c_1 c_{2,3} n_x + s_1 c_{2,3} n_y - s_{2,3} n_z}\right) \tag{3.65.A}$$

**The second solution set:**

$$q_4 = arctan\left(\frac{s_1 a_x - c_1 a_y}{c_1 s_{2,3} a_x + s_1 s_{2,3} a_y + c_{2,3} a_z}\right) + \pi \tag{3.63.B}$$

$$q_5 = -\arccos\left(c_1 c_{2,3} a_x + s_1 c_{2,3} a_y - s_{2,3} a_z\right) \tag{3.64.B}$$

$$q_6 = -arctan\left(\frac{c_1 c_{2,3} s_x + s_1 c_{2,3} s_y - s_{2,3} s_z}{c_1 c_{2,3} n_x + s_1 c_{2,3} n_y - s_{2,3} n_z}\right) + \pi \tag{3.65.B}$$

where
$$c_i \equiv \cos(q_i), s_i \equiv \sin(q_i)$$
$$c_{i,j} \equiv \cos(q_i + q_j), s_{i,j} \equiv \sin(q_i + q_j)$$
$$i,j \in \{1,2,3\} \tag{3.66}$$

The derivation of those equations is elaborated in Appendix D.

## 3.4.2 Singularity Pose and its Inverse Kinematics Solution

In robotic manipulators, a singularity refers to a configuration in which the manipulator loses one or more degrees of freedom, resulting in a loss of control in certain directions of motion. At these configurations, the Jacobian matrix becomes rank-deficient, meaning that it no longer provides a full-rank mapping between joint velocities and end-effector velocities. This leads to critical consequences, such as infinite joint velocities required to produce finite end-effector motion, or the inability to move the end-effector in specific directions. In Chapter 6, we will address strategies for handling singularities from a control perspective, including methods to maintain stability and ensure continuous end-effector motion near singular configurations.

Singularities in robotic manipulators can generally be classified into two categories: boundary singularities and interior singularities. A boundary singularity occurs when the end-effector reaches the edge of the robot's reachable workspace, such as when the arm is fully extended or folded, causing a loss of mobility in certain directions. In these configurations, the Jacobian matrix becomes rank-deficient, and the robot cannot generate motion along directions that extend beyond the boundary, resulting in limited controllability and infinite joint velocities for specific tasks. On the other hand, interior singularities arise within the interior of the workspace due to specific geometric alignments of joint axes—for example, when two rotational axes become parallel, as seen in wrist singularities. Unlike boundary singularities, interior singularities do not involve physical limits of reach but rather reflect intrinsic kinematic constraints of the manipulator. Furthermore, interior singularities also complicate the inverse kinematics problem, making it ill-posed or yielding an infinite number of joint angle solutions for a given pose.

In this section, we first examine three common singularities encountered in robotic manipulators: the elbow singularity, wrist singularity, and shoulder singularity, each of which exemplifies a distinct mechanism by which singular configurations arise.

> **Elbow Singularity**

The elbow singularity is a specific type of boundary singularity that occurs when the robot's upper arm (shoulder-to-elbow segment) and forearm (elbow-to-wrist segment) become aligned. This singularity condition typically arises when the robotic arm is fully extended or fully folded. However, many robot manufacturers impose mechanical limits to prevent the fully folded configuration, including the ABB IRB 4600 model used in this dissertation. Consequently, in this section, we focus exclusively on the fully extended configuration of the robotic arm, as illustrated in Figure 3.8.

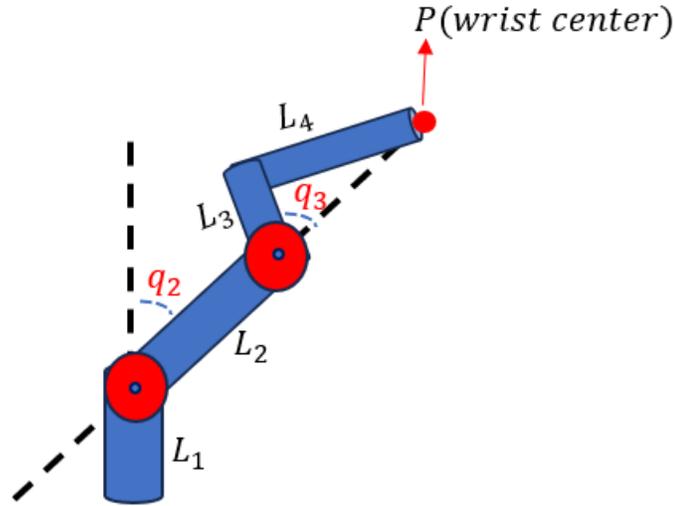

**Figure 3.8:** Demonstration of elbow singularity.

In this configuration, the wrist center lies precisely along the line connecting joints 2 and 3. This placement corresponds to the farthest possible position of the wrist center along this direction, indicating that the robotic arm has reached its maximum extension.

Therefore, the elbow singularity is defined by the following conditions:

$$q_3 = -\arctan\left(\frac{L_4}{L_3}\right) \text{ for all configurations} \qquad (3.67)$$

Additionally, from the perspective of the end-effector's position and orientation relative to the inertial (base) frame, the configuration satisfies the following condition:

$$(L_2 + \sqrt{L_3^2 + L_4^2})^2 = (\sqrt{P_x^2 + P_y^2} - a_1)^2 + (P_z - L_1)^2 \qquad (3.68)$$

Where $P_x$, $P_y$, and $P_z$ represent the position of the wrist center in the base frame, as defined in Equations (3.53)−(3.55).

➢ **Wrist Singularity**

The wrist singularity is a specific type of interior singularity that occurs when the rotational axes of joints $q_4$ and $q_6$ become aligned, as illustrated in Figure 3.9. This alignment leads to a loss of one degree of rotational freedom in the end-effector, resulting in an infinite number of joint angle combinations for $q_4$ and $q_6$ that yield the same end-effector orientation. Mathematically, this singularity occurs when $q_5= 0$ or $q_5= \pi$, corresponding to the wrist being either fully aligned or fully flipped.

However, as shown in Appendix A, Table A2, the ABB IRB 4600 model constrains the joint angle of $q_5$ to the range [-125°, 120°]. Therefore, only the configuration $q_5= 0$ is physically achievable in this robot.

At the wrist singularity, the yaw and roll directions of the end-effector become indistinguishable. As a result, if the target rotation about the combined yaw/roll direction is $\delta$, then any pair of angles $(q_4, q_6)$ satisfying the condition $q_4+ q_6 = \delta$ will produce the same orientation.

Thus, the wrist singularity is defined by the following conditions:

$$q_5 = 0 \tag{3.69}$$

In addition, the condition from the end-effector's orientation relative to the inertial frame is given by:

$$c_1 c_{2,3} a_x + s_1 c_{2,3} a_y - s_{2,3} a_z = 1 \tag{3.70}$$

At this configuration, the inverse kinematics solution becomes non-unique, resulting in an infinite number of valid joint angle combinations. Specifically, the following relation holds:

$$q_4 + q_6 = \delta \tag{3.71}$$

And
$$\delta = \arctan\left(\frac{s_1 n_x - c_1 n_y}{c_1 s_{2,3} n_x + s_1 s_{2,3} n_y + c_{2,3} n_z}\right) \tag{3.72}$$

where $\delta$ represents the desired rotation about the end-effector's yaw (or roll) axis.

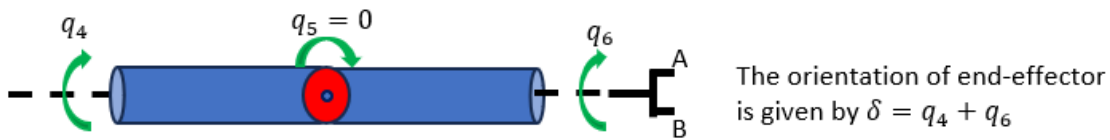

**Figure 3.9**: Demonstration of wrist singularity.

- **Shoulder Singularity**

The shoulder singularity is a specific type of interior singularity that occurs when the wrist center lies along the rotational axis of joint $q_1$. As illustrated in Figure 3.10., in this configuration, any rotation of $q_1$ results in no change to the position of the wrist center. This leads to an infinite number of solutions for $q_1$, rendering the inverse kinematics problem underdetermined.

The occurrence of this singularity is governed by a geometric relationship between the rotational angles $q_2$ and $q_3$. Specifically, the projection of the upper arm and forearm onto the horizontal plane must cancel out, placing the wrist center directly above or below the base axis. From planar triangle geometry, the shoulder singularity condition is defined as:

$$L_2 \cos(q_2) + \sqrt{L_3^2 + L_4^2} \cos\left(q_2 + q_3 + arctan\left(\frac{L_4}{L_3}\right)\right) = 0 \text{ for all configurations} \quad (3.73)$$

In addition, from the perspective of end-effector position relative to the inertial frame, the wrist center must lie on the $Z_0$-axis (the axis of joint $q_1$). Therefore, the following condition must also be satisfied:

$$P_x = 0, \text{ and } P_y = 0 \quad (3.74)$$

where $P_x$ and $P_y$ are the $X$ and $Y$ coordinates of the wrist center, as defined in Equations (3.57) and (3.58).

Under these conditions, the inverse kinematics solution becomes non-unique due to the loss of control authority over the base rotation. As a result:

$$q_1 = \text{any value} \quad (3.75)$$

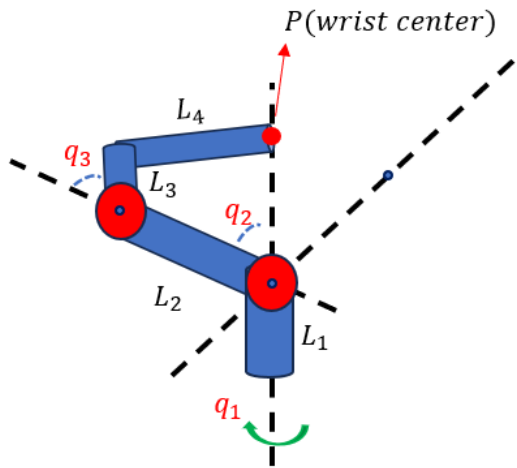

**Figure 3.10:** Demonstration of shoulder singularity.

## 3.4.3 Unique Selection of Joint Configurations in the Inverse Kinematic process

The preceding sections have developed analytical solutions to the inverse kinematics problem for both non-singular and singular configurations. In non-singular poses, the kinematic structure of a 6-DOF robot with a spherical wrist allows for multiple valid solutions:

- There are typically four distinct solutions for the wrist center position, based on different elbow configurations (e.g., elbow-up, elbow-down, left/right).
- For each wrist center position, there are two possible solutions for the wrist orientation, depending on the alignment of the wrist joints.

Due to the decoupled structure of position and orientation in spherical wrist robots, these combinations result in eight total solutions for each non-singular pose.

In this dissertation, Jacobian linearization and feedback linearization methods are employed to design the outer-loop controllers, which generate the target joint configurations for the robotic system. The joint angle updates produced by these controllers are computed locally, evolving around the current joint configuration.

In the first stage of the control architecture (illustrated in Figure 2.3), an analytical inverse kinematics model is used to compute the target joint angles corresponding to the desired position of the camera. Additionally, in the adaptive MIMO controller design discussed in Chapter 6, the inverse kinematics model is applied at each control iteration to estimate the current joint angles from task-space feedback. Given that the inverse kinematics problem generally admits multiple solutions—especially in non-singular configurations—it is essential to select a unique and consistent joint configuration throughout the control process. This ensures stable system behavior and avoids undesirable jumps between different solution branches, which could lead to erratic motion or instability.

To maintain consistency and predictability in motion, and to ensure the system follows the shortest and smoothest joint trajectory, the following configuration conventions are adopted in this dissertation:

- For wrist center position, we select the right-above configuration (i.e., elbow-up and shoulder-left).

- For wrist orientation, we choose the solution in which $q_5 > 0$.

In singular configurations, the inverse kinematics solution becomes non-unique due to the loss of one or more degrees of freedom. Two types of singularities are considered:

◆ Wrist Singularity:

In this case, the rotational axes of joints $q_4$ and $q_6$ become aligned, and any combination of $q_4$ and $q_6$ satisfying the following condition yields the same end-effector orientation:

$$q_4 + q_6 = \arctan\left(\frac{s_1 n_x - c_1 n_y}{c_1 s_{2,3} n_x + s_1 s_{2,3} n_y + c_{2,3} n_z}\right) \tag{3.76}$$

To ensure a unique solution in the target joint generation process, we fix:

$$q_4 = 0, \tag{3.77}$$

And
$$q_6 = \arctan\left(\frac{s_1 n_x - c_1 n_y}{c_1 s_{2,3} n_x + s_1 s_{2,3} n_y + c_{2,3} n_z}\right) \tag{3.78}$$

However, when inverse kinematics is used for estimating the current joint configuration, such as during real-time control or observer-based state reconstruction, special care must be taken near singular configurations. To preserve temporal smoothness and avoid sudden jumps between equivalent solutions, we retain the previous value of $q_4$ from the last time step:

$$q_4 = q_4^{prev} \tag{3.79}$$

And
$$q_6 = \arctan\left(\frac{s_1 n_x - c_1 n_y}{c_1 s_{2,3} n_x + s_1 s_{2,3} n_y + c_{2,3} n_z}\right) - q_4^{prev} \tag{3.80}$$

◆ Shoulder Singularity:

In this case, the wrist center lies directly along the axis of joint $q_1$, making $q_1$ undefined. Any value of $q_1$ is valid and leads to the same end-effector position. To ensure a unique solution during target joint generation, we impose the following condition:

$$q_1 = 0 \tag{3.81}$$

However, when using inverse kinematics to estimate the current joint configuration in real time, it is important to preserve continuity and avoid discontinuities caused by arbitrary changes in $q_1$. Therefore, we adopt a strategy similar to the one used in wrist singularities by referencing the previously known value of $q_1$:

$$q_1 = q_1^{prev} \qquad (3.82)$$

The selection of ambiguous joint states in singular configurations—as defined in Equations (3.79), (3.80), and (3.82)—ensures a smooth and consistent evolution of the estimated joint trajectories over time, particularly when the system encounters or passes through singular poses. This approach prevents discontinuities in the control process and maintains stability in joint space estimation. The chosen joint configurations for both singular and non-singular conditions are summarized in Table 3.2.

**Table 3.2:** Joint Configuration Selection Under Different Conditions.

| Configuration Type | Description | End-Effector Pose-Based Condition | Selection Strategy |
|---|---|---|---|
| Non-Singular Pose | Regular configuration with full degrees of freedom | No special constraint | - **Wrist center:** Right-above configuration (Eqs. 3.60A-3.62A) <br> - **Orientation:** $q_5 > 0$ (Eqs. 3.63A-3.65A) |
| Wrist Singularity | Axes of joints $q_4$ and $q_6$ aligned | (Eq. 3.70): $c_1 c_{2,3} a_x + s_1 c_{2,3} a_y - s_{2,3} a_z = 1$ | -**Target generation**: $q_4 = 0$, $q_6 = \delta$ (Eqs. 3.77-3.78) <br> -**Estimation**: $q_4 = q_4^{prev}$, $q_6 = \delta - q_4^{prev}$ (Eqs. 3.79–3.80) |
| Shoulder Singularity | Wrist center lies on base axis $Z_0$ | (Eq. 3.74): $P_x = 0$, and $P_y = 0$ | - **Target generation**: $q_1 = 0$ (Eq. 3.81) <br> - **Estimation**: $q_1 = q_1^{prev}$ (Eq. 3.82) |
| Elbow Singularity | Upper arm and forearm are collinear (fully extended) | (Eq. 3.68): $(L_2 + \sqrt{L_3^2 + L_4^2})^2 = (\sqrt{P_x^2 + P_y^2} - a_1)^2 + (P_z - L_1)^2$ | Not appliable |

$\delta = \arctan \left( \dfrac{s_1 n_x - c_1 n_y}{c_1 s_{2,3} n_x + s_1 s_{2,3} n_y + c_{2,3} n_z} \right)$.

$c_1 \equiv \cos(q_1), s_1 \equiv \sin(q_1), c_{2,3} \equiv \cos(q_2 + q_3), s_{2,3} \equiv \sin(q_2 + q_3)$.

$a_x, a_y, a_z$: components of the end-effector's approach vector $\vec{a}$.

$P_x, P_y, P_z$: wrist center coordinates in base frame.

## 3.5 Dynamic Model of Robot the Arm and the Actuator

This section presents the dynamic model of the robot manipulator and its actuators. The dynamic equations governing the time evolution of the manipulator are derived using the Euler-Lagrange formulation.

### 3.5.1 Manipulator Jacobian Matrix

From the forward kinematic transformation, we can express both the angular and linear velocities at any point on the manipulator as follows:

Consider a point on the $n^{th}$ link of the manipulator:

$$V_0^n = J_{V_n} \dot{q} \qquad (3.83)$$

$$\omega_0^n = J_{\omega_n} \dot{q} \qquad (3.84)$$

where $V_0^n$ and $\omega_0^n$ are the linear velocity and angular velocities of a point on the $n^{th}$ link measured in the base inertial frame, $\dot{q}$ is the time derivative of the joint variables $q$, and $J_{V_n}$ and $J_{\omega_n}$ are $3 \times n$ matrices. Equations (3.83) and (3.84) can be combined as:

$$\begin{bmatrix} V_0^n \\ \omega_0^n \end{bmatrix} = J_0^n \dot{q} \qquad (3.85)$$

where $J_0^n$ is defined as:

$$J_0^n = \begin{bmatrix} J_{V_n} \\ -- \\ J_{\omega_n} \end{bmatrix} \qquad (3.86)$$

The $6 \times n$ matrix $J_0^n$ is known as the manipulator Jacobian. We will now derive the Jacobian matrix expression. In the case of an elbow manipulator with a spherical wrist, all six joints are revolute.

For a revolute $i^{th}$ joint ($i \leq n$), the $i^{th}$ column of $J_0^n$ is given by:

$$J_0^n(i) = \begin{bmatrix} z_{i-1} \times (d_0^n - d_0^{i-1}) \\ z_{i-1} \end{bmatrix} \qquad (3.87)$$

And:

$$z_{i-1} = R_0^{i-1} k \qquad (3.89)$$

Here, for any integer $i \in (1,2,3\ldots,n+1)$, $d_0^{i-1}$ and $R_0^{i-1}$ represent the translational and rotational components of the transformation matrix $T_0^{i-1}$, which defines the transformation from the base frame to the frame attached to the $(i-1)^{th}$ joint. The vector $k$ is the unit vector $[0,0,1]^T$.

In Equation (3.87) the upper entry represents the $i^{th}$ column of the linear velocity Jacobian component and the lower entry represents the $i^{th}$ column of the angular velocity Jacobian component.

### 3.5.2 Robot Manipulator Dynamic Model

The dynamics of a robot manipulator can be described using the Euler-Lagrange equations of motion:

$$\frac{\delta}{\delta t}\frac{\delta L}{\delta \dot{q}_j} - \frac{\delta L}{\delta q_j} = \tau_j \tag{3.89}$$

where the Lagrangian $L$ is defined as:

$$L = K - V \tag{3.90}$$

Here, L represents the Lagrangian, which is the difference between kinematic energy $K$ and potential energy $V$. The variable $q_j$ is the generalized joint coordinate (convolute variable) with $j \in (1,2,3,4,5,6)$ for the application, and $\tau_j$ is the external force or torque applied to each joint. Figure 3.11. illustrates the ABB IRB 4600 Robot with red points marking the center of mass for each link. The mass, center of mass positions, and moment of inertia matrices for each link have been calculated and summarized in Table A1 (Appendix A). It is important to note that the moment of inertia matrix for each link is diagonal.

The total kinetic energy of the manipulator is the sum of the translational and rotational energies of all its links:

$$K = \frac{1}{2}\dot{q}^T \sum_{i=1}^{6} [m_i J_{V_{ci}}(q)^T J_{V_{ci}}(q) + J_{\omega_i}(q)^T R_0^i(q) I_i R_0^i(q)^T J_{\omega_i}(q)]\dot{q} \tag{3.91}$$

where $m_i$ and $I_i$ are the mass and the moment of inertial matrix of link $i$. $J_{V_{ci}}(q)$ is the upper part of the Jacobian matrix for the center of mass of link $i$ and $J_{\omega_i}(q)$ is the lower part of the Jacobian matrix for link $i$. $R_0^i(q)$, as the function of joint angles $q$, is the rotational matrix that transforms vectors from the coordinate frame attached to link $i$ $(O_i X_i Y_i Z_i)$ to the base frame $(O_0 X_0 Y_0 Z_0)$.

Equation (3.91) can be expressed in a simplified form:

$$K = \frac{1}{2}\dot{q}^T D(q)\dot{q} \tag{3.92}$$

where $D(q)$ is a $6 \times 6$ symmetric positive definite matrix, known as the inertia matrix. It is a function of the current joint variation $q$. For a rigid body, the only source of potential energy is gravity. The potential energy is expressed as:

$$V = \hat{g}^T \sum_{i=1}^{6} r_{ci}(q)\, m_i \tag{3.93}$$

where $\hat{g} = [0, 0,\ 9.8\ m/s^2]^T$ is the gravitational acceleration vector in the base frame, $r_{ci}(q)$ is the position vector of the center of mass of link $i$ in the base frame, and $m_i$ is the mass of link $i$. Using Equations (3.92) and (3.93), the Lagrangian can be rewritten as:

$$L = K - V = \frac{1}{2}\sum_{i,j=1}^{6} d_{i,j}(q)\dot{q}_i\dot{q}_j - V(q) \tag{3.94}$$

Here $d_{i,j}$ is the $(i,j)$-th element of the inertia matrix $D(q)$, and $i, j \in (1, 2, 3, 4, 5, 6)$. From Equation (3.94), the Euler-Lagrange Equation (3.89) can be derived as:

$$\sum_{j=1}^{6} d_{k,j}(q)\ddot{q}_j + \sum_{i,j}^{6} c_{i,j,k}(q)\dot{q}_i\dot{q}_j + \Phi_k(q) = \tau_k,\ k = 1,2,3,4,5,6 \tag{3.95}$$

$c_{i,j,k}(q)$ are the Christoffel symbols, given by:

$$c_{i,j,k} = \frac{1}{2}\left(\frac{\delta d_{k,j}}{\delta q_i} + \frac{\delta d_{k,i}}{\delta q_j} - \frac{\delta d_{i,j}}{\delta q_k}\right) \tag{3.96}$$

$\Phi_k(q)$ is the derivative of potential energy with respective to $q_k$:

$$\Phi_k = \frac{\delta V}{\delta q_k} \tag{3.97}$$

For a fixed $k$, which is the joint angle index, the Christoffel symbols satisfy $c_{i,j,k} = c_{j,i,k}$. Equation (3.95) can be rewritten in matrix form as:

$$D(q)\ddot{q} + C(q,\dot{q})\dot{q} + g(q) = \tau \tag{3.98}$$

Here, $D(q)$ is the inertia matrix, $C(q,\dot{q})$ is the Coriolis and centrifugal matrix, where the $(k,j)$-th element is defined as:

$$c_{k,j} = \sum_{i=1}^{6} c_{i,j,k}(q)\dot{q}_i$$

$$= \sum_{i=1}^{6} \frac{1}{2}\left(\frac{\delta d_{k,j}}{\delta q_i} + \frac{\delta d_{k,i}}{\delta q_j} - \frac{\delta d_{i,j}}{\delta q_k}\right)\dot{q}_i \quad (3.99)$$

and, $g(q)$ is the gravitational vector, where $\Phi_k$ corresponds to the $k^{th}$ element of $g(q)$.

Equation (3.99) represents a second-order differential equation system. The vectors $q$, $\dot{q}$, $\ddot{q}$ and $\tau$ are all $1 \times 6$ vectors, where each element corresponds to the respective joint variables. This compact form facilitates the analysis and computation of the manipulator's dynamics.

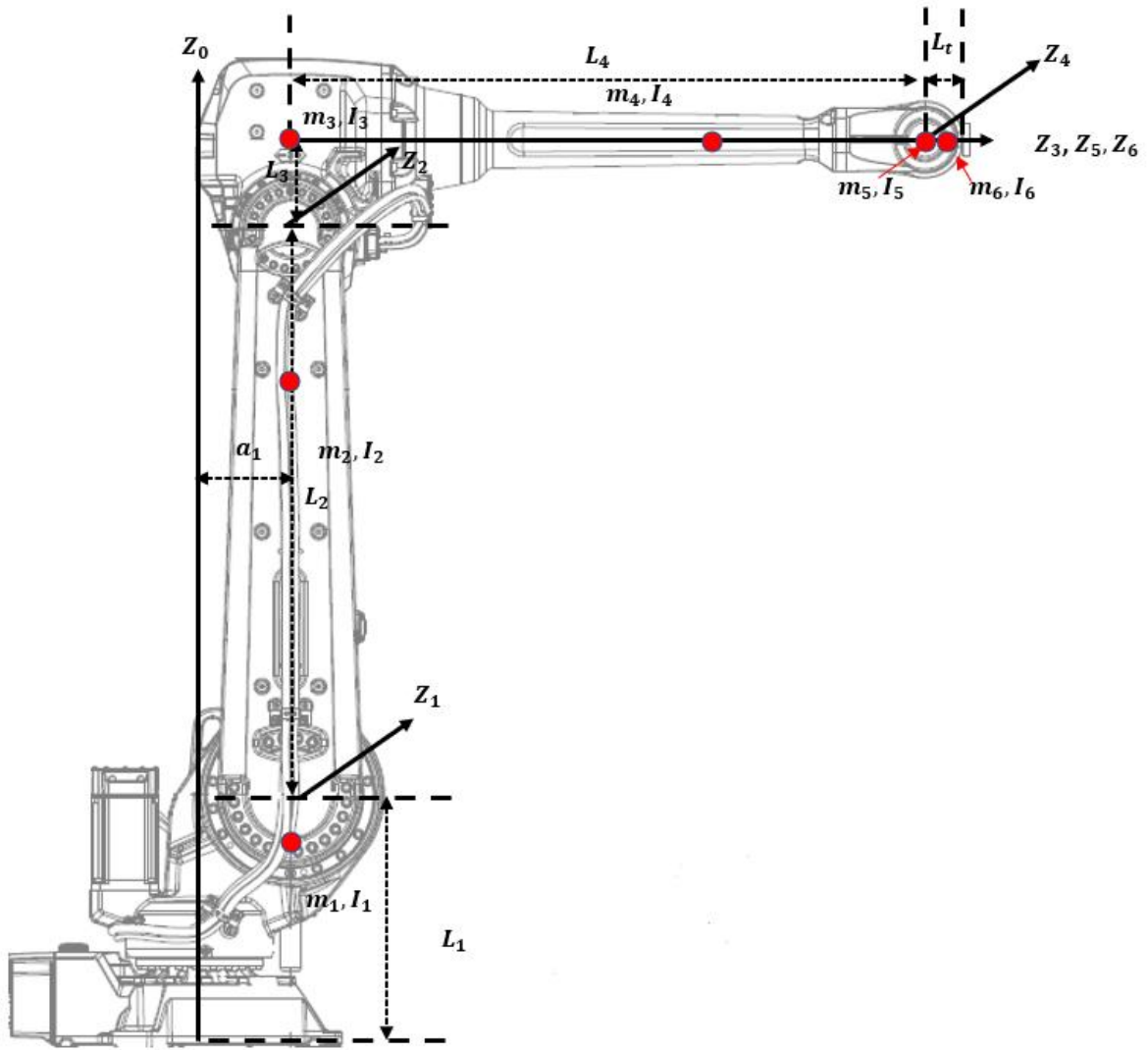

**Figure 3.11:** IRB ABB 4600 manipulator model with center of masses.

### 3.5.3 Actuator Dynamic Model

This section addresses the dynamics of actuators responsible for generating the generalized force $\tau$. The analysis assumes the use of permanent magnet DC motors in the joints, as they are widely employed in modern robotic systems. A schematic diagram of a DC motor is presented in Figure 3.11, and the parameters used in the model, along with their respective symbols, are summarized in Table 3.3.

Table 3.3: Parameters of DC motor.

| Parameter Name | Symbol |
| --- | --- |
| Armature Voltage | $V(t)$ |
| Armature Inductance | $L$ |
| Armature Resistance | $R$ |
| Back Emf | $V_b$ |
| Armature Current | $i_a$ |
| Rotor Position | $\vartheta_m$ |
| Generated Torque | $\tau_m$ |
| Load Torque | $\tau_l$ |

The dynamics of the actuator can be described by the following equations. The differential equation for the armature current, derived from Figure 3.12., is:

$$L\frac{di_a}{dt} + Ri_a = V - V_b \tag{3.100}$$

The torque developed by the motor, $\tau_m$ is given by:

$$\tau_m = K_m i_a \tag{3.101}$$

where $K_m$ is the torque constant measured in $N - m/amp$.

The back emf, $V_b$, can be calculated as:

$$V_b = K_m \frac{d\theta_m}{dt} \tag{3.102}$$

Where $K_m$ is the same constant defined in Equation (3.101), but in Equation (3.102) it is defined as back emf constant measured in $Volt - s/rad$.

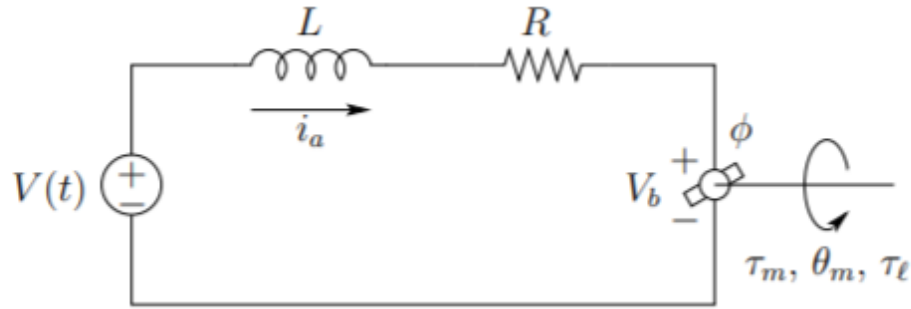

**Figure 3.12:** Circuit diagram for amature controlled DC motor.

Consider a DC motor in series with a gear train, with a gear ratio of $1:r$, connected to a manipulator link, as shown in Figure 3.12. $J_a, J_g$ and $J_l$ represent the moment of inertia of the motor, the gear train, and link of the manipulator. The equation of motion for the system, derived from Figure 3.13., is expressed as:

$$(J_a + J_g)\frac{d^2\theta_m}{dt^2} + B_m\frac{d\theta_m}{dt} = \tau_m - r\tau_l \tag{3.103}$$

Where $B_m$ is the damping ratio of the gear train,

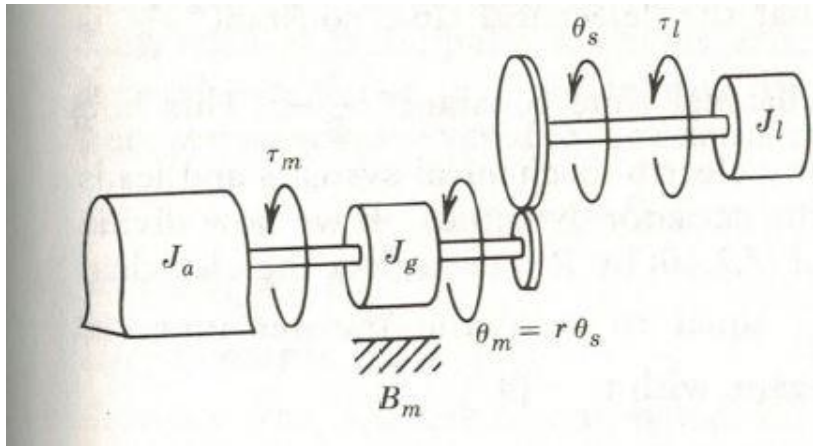

**Figure 3.13:** Lumped model of a single link with actuator/gear train.

The dynamics of the system can be analyzed in the Laplace domain by combining (3.97) and (3.99) as well as (3.103) and (3.101). This results in the following equations:

$$(Ls + R)I_a(s) = V(s) - K_m s \Theta_m(s) \tag{3.104}$$

$$((J_a + J_g)s^2 + B_m s)\Theta_m(s) = K_m I_a(s) - r\tau_l(s) \tag{3.105}$$

Taking $V(s)$ (applied voltage) and $\tau_l(s)$ (load torque) as the two inputs to the system. Then the transfer function from $V(s)$ to $\Theta_m(s)$ is given by:

$$\frac{\Theta_m(s)}{V(s)} = \frac{K_m}{s[(Ls + R)(J_a + J_g)s + B_m) + K_m^2]} \tag{3.106}$$

The transfer function from $\tau_l(s)$ to $\Theta_m(s)$ is given by:

$$\frac{\Theta_m(s)}{\tau_l(s)} = \frac{-r(Ls + R)}{s[(Ls + R)(J_a + J_g)s + B_m) + K_m^2]} \tag{3.107}$$

Assume that the electrical time constant: $\frac{L}{R}$ is much smaller than mechanical time constant $\frac{J_a+J_g}{B_m}$, i.e., $\frac{L}{R} \ll \frac{J_a+J_g}{B_m}$). This is a reasonable assumption for many electro-mechanical systems. Under this assumption, we can divide the numerator and denominator of Equations (3.106) and (3.107) by $R$ and neglect the term $\frac{L}{R}$ by setting it to zero. By superposition and transforming back to the time domain, the system's dynamics can be expressed as a second-order differential equation:

$$(J_a + J_g)\ddot{\theta}_m + (B_m + \frac{K_m^2}{R})\dot{\theta}_m = \frac{K_m}{R}V - r\tau_l \tag{3.108}$$

Setting $J_m = J_a + J_g$, the sum of the actuator and gear inertias, and $B = B_m + \frac{K_m^2}{R}$, Equation (3.108) can be simplified as:

$$J_m \ddot{\theta}_m + B\dot{\theta}_m = \frac{K_m}{R}V - r\tau_l \tag{3.109}$$

This equation represents the dynamics of the rotor position $\theta_m$ for a single motor.

By incorporating the robot manipulator dynamics in Equation (3.98) and the actuator dynamics in Equation (3.109), and using the relationship $\theta_m = \frac{1}{r_k} q_k$ for each joint $k$, where $r_k$ is the gear ratio of the $k^{th}$ joint, the overall dynamics equation governing the $k^{th}$ joint becomes:

$$\frac{1}{r_k}J_{m_k}\ddot{q}_k + \sum_{j=1}^{6} d_{j,k}\ddot{q}_j + \sum_{i,j}^{6} c_{i,j,k}(q)\dot{q}_i\dot{q}_j + \frac{1}{r_k}B\dot{q}_k + \Phi_k(q) = \frac{K_m}{r_k R}V \tag{3.110}$$

where 
$$J_m = J_a + J_g, \quad B = B_m + \frac{K_m^2}{R} \tag{3.111}$$

The dynamics in Equation (3.110) can be expressed in matrix form as:

$$(D(q) + J)\ddot{q} + (C(q,\dot{q}) + \frac{B}{r})\dot{q} + g(q) = u \qquad (3.112)$$

Where $D(q)$ is the 6 × 6 inertial matrix of the manipulator as defined in Equation (3.92) and $J$ is a diagonal matrix with diagonal elements $\frac{1}{r_k}J_{m_k}$, representing the contributions from the actuator inertias. The vector $C(q,\dot{q})$, already defined in Equations (3.98) and (3.99), captures the Coriolis and centrifugal effects, while $g(q)$, also defined earlier, represents the gravitational forces. The gear ratios are grouped into the vector $r$, and the input vector $u$ contains components: $u_k = \frac{K_m}{r_k R}V$, where $k \in (1,2,3,4,5,6)$.

## 3.6 MIMO Models

Chapter 2 introduced control architectures designed to reduce uncertainties in the manufacturing process. These architectures involve two key control processes:

- Camera movement adjustment control, which utilizes the Camera-on-Robot Kinematics Model.
- High-accuracy tool manipulator control, which incorporates the Camera-and-Tool Combined Model.

In this section, the mathematical equations for these models are derived.

### 3.6.1 Camera-on-Robot Kinematics Model

This model takes the joints' angle of the manipulator as inputs and outputs the coordinates of three references points $R_1$, $R_2$, and $R_3$ in the image frame as observed by a camera mounted on the manipulator. In Figure 3.14., an inertial frame is attached to the base of the robot manipulator. The homogeneous coordinates of the reference points $R_1$, $R_2$, and $R_3$ in the inertial frame are assumed to be pre-determined as: $\overline{P_{R_1}^V} = (\overline{X_{R_1}^V}, \overline{Y_{R_1}^V}, \overline{Z_{R_1}^V}, 1)$, $\overline{P_{R_2}^V} = (\overline{X_{R_2}^V}, \overline{Y_{R_2}^V}, \overline{Z_{R_2}^V}, 1)$, and $\overline{P_{R_3}^V} = (\overline{X_{R_3}^V}, \overline{Y_{R_3}^V}, \overline{Z_{R_3}^V}, 1)$. An end-effector frame is attached to the center of the camera, with its Z axis always aligned with the principal axis of the camera. The transformation of the reference points' coordinates from the inertial frame to the end-effector frame is achieved using a transformation matrix, denoted as $T_0^6\{visual\}$.

The transformation matrix $T_0^6\{visual\}$ represents the cumulative transformation from the base frame to the end-effector frame. Similar to the form given in Equation (3.43), it can be calculated as:

$$T_0^6\{visual\} = A_5^6 \, A_4^5 \, A_3^4 \, A_2^3 \, A_1^2 \, A_0^1 \tag{3.113}$$

Here, each $A_i^{i+1}$ ($i \in [0,1,2,3,4,5,6]$) represents the homogeneous transformation matrix from frame $i$ to frame $i+1$. It is important to note that each $A_i^{i+1}$ is the inverse matrix of $A_{i+1}^i$, which is derived from Equations (3.43)–(3.48). These matrices encode both rotational and translational transformations based on the manipulator's joint configuration.

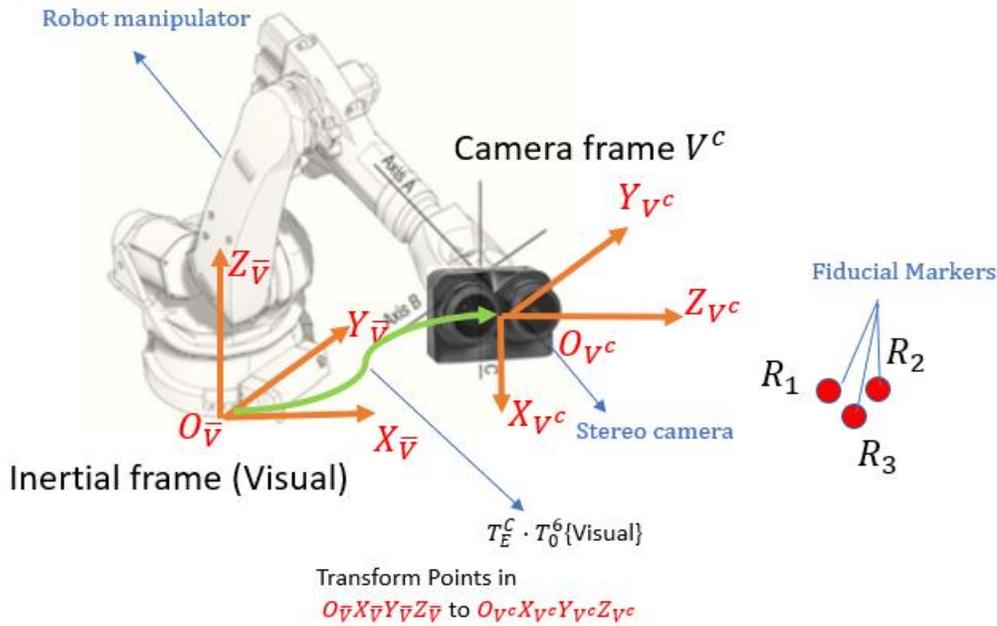

**Figure 3.14**: Coordinates' transformation in Camera-on-Robot model.

To explicitly express $T_0^6\{visual\}$ as a 4×4 transformation matrix and in terms of the six joint angles $(q_1, \ldots, q_6)$ of the camera robot arm, we need to derive the homogeneous transformation matrix step-by-step using the Denavit-Hartenberg (DH) parameters and compose it as per Equation (3.50). Substituting the DH parameters of the manipulator and joint angles $(q_1, \ldots, q_6)$, the transformation matrix $T_0^6\{visual\}$ becomes:

$$T_0^6\{visual\} = \begin{bmatrix} n_x & s_x & a_x & d_x \\ n_y & s_y & a_y & d_y \\ n_z & s_z & a_z & d_z \\ 0 & 0 & 0 & 1 \end{bmatrix} \tag{3.114}$$

$$n_x = c_1(s_{23}s_4s_6 - s_{23}c_4c_5c_6 - c_{23}s_5c_6) - s_1(c_4s_6 + s_4c_5c_6)$$
$$n_y = s_1(s_{23}s_4s_6 - s_{23}c_4c_5c_6 - c_{23}s_5c_6) + c_1(c_4s_6 + s_4c_5c_6)$$
$$n_z = c_{23}s_4s_6 - c_{23}c_4c_5c_6 + s_{23}s_5c_6 \tag{3.115}$$

$$s_x = c_1(s_{23}s_4c_6 + c_{23}c_4c_5s_6 + c_{23}s_5s_6) - s_1(c_4c_6 - s_4c_5s_6)$$
$$s_y = s_1(s_{23}s_4c_6 + c_{23}c_4c_5s_6 + c_{23}s_5s_6) + c_1(c_4c_6 - s_4c_5s_6)$$
$$s_z = c_{23}s_4c_6 + c_{23}c_4c_5s_6 - s_{23}s_5s_6 \tag{3.116}$$

$$a_x = c_1(c_{23}c_5 - s_{23}c_4s_5) - s_1s_4s_5$$
$$a_y = s_1(c_{23}c_5 - s_{23}c_4s_5) + c_1s_4s_5$$
$$a_z = -s_{23}c_5 - c_{23}c_4s_5 \tag{3.117}$$

$$d_x = L_1(s_{23}s_4s_6 - c_{23}c_4c_5c_6 + s_{23}s_5c_6) - L_2(c_3s_4s_6 - c_3c_4c_5c_6 + s_3s_5c_6)$$
$$\quad - L_3(s_4s_6 - c_4c_5c_6) + L_4s_5c_6 - a_1(s_{23}s_4s_6 - s_{23}c_4c_5c_6 - c_{23}s_5c_6)$$
$$d_y = -L_1(c_{23}s_4c_6 + c_{23}c_4c_5s_6 - s_{23}s_5s_6) - L_2(c_3s_4c_6 + c_3c_4c_5s_6 - s_3s_5s_6)$$
$$\quad - L_3(s_4c_6 + c_4c_5s_6) - L_4s_5s_6 - a_1(s_{23}s_4c_6 + s_{23}c_4c_5s_6 + c_{23}s_5s_6)$$
$$d_z = L_1(s_{23}c_5 + c_{23}c_4s_5) + L_2(s_3c_5 + c_3c_4s_5) + L_3c_4s_5 - L_4c_5 - a_1(c_{23}c_5 - s_{23}c_4s_5) - L_t \tag{3.118}$$

$$c_i \equiv \cos(q_i),\ s_i \equiv \sin(q_i)$$
$$c_{i,j} \equiv \cos(q_i + q_j),\ s_{i,j} \equiv \sin(q_i + q_j)$$
$$i, j \in \{1,2,3,4,5,6\} \tag{3.119}$$

Assuming that the camera remains static relative to the end-effector, we introduce a constant transformation matrix $\overline{T_E^C}$ that maps points from the end-effector frame $\{V^e\}$ to the camera frame $\{V^c\}$, which is located at the center of the stereo baseline, as shown in Figure 3.2. The distance between the origin of the camera frame and the end-effector frame in the Z-direction is denoted as $L_{CE}$. Then, the matrix $T_E^C$ can be expressed as:

$$\overline{T_E^C} = \begin{bmatrix} 1 & 0 & 0 & 0 \\ 0 & 1 & 0 & 0 \\ 0 & 0 & 1 & L_{CE.} \\ 0 & 0 & 0 & 1 \end{bmatrix} \tag{3.120}$$

The coordinates of the reference points $R_1$, $R_2$, and $R_3$ in the camera frame can be obtained by applying the transformation $\overline{T_E^C} \cdot T_0^6$ {visual} to their coordinates in the inertial frame. Using homogeneous coordinates, this transformation is expressed as:

$$P_{R_1}^{V^c} = \overline{T_E^C} \cdot T_0^6 \{visual\} [P_{R_1}^{\overline{V}}]^T = [X_{R_1}^{V^c}, Y_{R_1}^{V^c}, Z_{R_1}^{V^c}, 1]^T \tag{3.121}$$

$$P_{R_2}^{V^c} = \overline{T_E^C} \cdot T_0^6 \{visual\} [P_{R_2}^{\overline{V}}]^T = [X_{R_2}^{V^c}, Y_{R_2}^{V^c}, Z_{R_2}^{V^c}, 1]^T \tag{3.122}$$

$$P_{R_3}^{V^c} = \overline{T_E^C} \cdot T_0^6 \{visual\} [P_{R_3}^{\overline{V}}]^T = [X_{R_3}^{V^c}, Y_{R_3}^{V^c}, Z_{R_3}^{V^c}, 1]^T \tag{3.123}$$

To summarize, the total transformation matrix from the camera's base inertia frame to the camera's end-effector frame is denoted as $T_{total}^{Visual}$, and can be expressed as:

$$T_{total}^{Visual} = \overline{T_E^C} \cdot T_0^6 \{visual\} \tag{3.124}$$

Using the stereo camera model as described in Equations (3.23) − (3.25), the coordinates of $R_1$, $R_2$, and $R_3$ in the image frame can be expressed in terms of the transformed coordinates in the visual frame:

$$p_{R_1} = \left[ \frac{2X_{R_1}^{V^c} - b}{2Z_{R_1}^{V^c}} F, \frac{2X_{R_1}^{V^c} + b}{2Z_{R_1}^{V^c}} F, \frac{Y_{R_1}^{V^c}}{Z_{R_1}^{V^c}} F, 1 \right]^T = [ul_{R_1}, ur_{R_1}, v_{R_1}, 1]^T \tag{3.125}$$

$$p_{R_2} = \left[ \frac{2X_{R_2}^{V^c} - b}{2Z_{R_2}^{V^c}} F, \frac{2X_{R_2}^{V^c} + b}{2Z_{R_2}^{V^c}} F, \frac{Y_{R_2}^{V^c}}{Z_{R_2}^{V^c}} F, 1 \right]^T = [ul_{R_2}, ur_{R_2}, v_{R_2}, 1]^T \tag{3.126}$$

$$p_{R_3} = \left[ \frac{2X_{R_3}^{V^c} - b}{2Z_{R_3}^{V^c}} F, \frac{2X_{R_3}^{V^c} + b}{2Z_{R_3}^{V^c}} F, \frac{Y_{R_3}^{V^c}}{Z_{R_3}^{V^c}} F, 1 \right]^T = [ul_{R_3}, ur_{R_3}, v_{R_3}, 1]^T \tag{3.127}$$

We define a nonlinear transformation $\Psi$, which maps reference points $R_1$, $R_2$, and $R_3$ measured in the base frame $\{\overline{V}\}$ to their image coordinates as captured by the stereo camera. This transformation is expressed as:

Camera-on-robot mapping $\Psi$ : $\overline{P_R^{\overline{V}}} = [\overline{P_{R_1}^{\overline{V}}}, \overline{P_{R_2}^{\overline{V}}}, \overline{P_{R_3}^{\overline{V}}}] \overset{\Psi}{\Rightarrow} p_R = [p_{R_1}, p_{R_2}, p_{R_3}]$ (3.128)

Here, $\overline{P_R^{\overline{V}}} \in \mathbb{R}^{4\times 3}$ is the matrix formed by stacking the 3D homogenous coordinates of the reference points in the base frame, and $p_R \in \mathbb{R}^{4\times 3}$ is the corresponding set of 2D homogenous stereo image coordinates of those points.

The Mapping $\Psi$ is nonlinear and depends on variables current joint angles of the visual robot manipulator: $q_V$ and parameters $P_{a_{visual}}$, which includes stereo camera parameter $P_{a_{camera}}$, robot geometric parameter $P_{a_{robot}}$, and transformation matrix $\overline{T_E^C}$. In other words:

$$\forall t \geq 0, p_R(t) = \Psi(q_V(t), P_{a_{visual}}, \overline{P_R^{\overline{V}}})$$ (3.129)

$$P_{a_{visual}} = [P_{a_{camera}}, P_{a_{robot}}, \overline{T_E^C}]$$ (3.130)

If the joint angles corresponding to the camera's optimal poses ($\overline{q_V}$) are provided, this function can be used to compute the target image coordinates of the reference points ($\overline{p_R}$). Similarly, for any disturbed current joint angles of the robot manipulator ($\widetilde{q_V}$), the same nonlinear function can mathematically compute the measured image coordinates of the reference points ($\widehat{p_R}$) for simulation purposes. In other words, the relationships can be expressed as:

$$\forall t \geq 0, \overline{p_R}(t) = \Psi(\overline{q_V}(t), P_{a_{visual}}, \overline{P_R^{\overline{V}}})$$ (3.131)

$$\forall t \geq 0, \widehat{p_R}(t) = \Psi(\widetilde{q_V}(t), P_{a_{visual}}, \overline{P_R^{\overline{V}}})$$ (3.132)

In addition, during the process of tool pose control, this nonlinear function is used to obtain the target image coordinates of the interest points ($\overline{p_I}$). This is achieved by using the optimal pose of the camera ($\overline{q_V}$) and the target 3D coordinates of the interest points in the inertial frame attached to the visual system ($\overline{P_I^{\overline{V}}}$). In other words, this relationship can be expressed as:

$$\forall t \geq 0, \overline{p_I}(t) = \Psi(\overline{q_V}(t), P_{a_{visual}}, \overline{P_R^{\overline{V}}})$$ (3.133)

### 3.6.2 Inverse Camera-on-Robot Kinematics Model

We define the inverse process of the nonlinear mapping $\Psi$ as $\Psi^{-1}$, which maps a set of reference points $R_1$, $R_2$, and $R_3$ from their measured image coordinates to their corresponding positions in the robot base frame $\{\overline{V}\}$. This transformation is described as:

Inverse Camera-on-robot mapping $\Psi^{-1}$ : (3.134)

$$p_R = [p_{R_1}, p_{R_2}, p_{R_3}] \xRightarrow{\Psi^{-1}} \overline{P_R^{\bar{V}}} = [\overline{P_{R_1}^{\bar{V}}}, \overline{P_{R_2}^{\bar{V}}}, \overline{P_{R_3}^{\bar{V}}}]$$

To establish the existence and uniqueness of $\Psi^{-1}$, we adopt a modular approach by decomposing the nonlinear function $\Psi$ into a sequence of sub-functions, and then examining the invertibility of each component.

From Equations (3.121) − (3.130), the original forward mapping $\Psi$ in Equation (3.128) can be written as:

$$\overline{P_R^{\bar{V}}} \xRightarrow{T_{total}^{Visual}} P_R^{V^c} = [P_{R_1}^{V^c}, P_{R_2}^{V^c}, P_{R_3}^{V^c}] \xRightarrow{M} p_R \qquad (3.135)$$

Here $T_{total}^{Visual}$ is a transformation matrix from the base frame $\{\bar{V}\}$ to the camera frame $\{V^C\}$, dependent on the joint configuration $q_V$, as defined in Equation (3.124). $P_R^{V^c} \in \mathbb{R}^{4 \times 3}$ is the matrix formed by stacking the 3D homogenous coordinates of the reference points in the camera end-effector frame.

$M$ is the stereo-camera projection function that maps 3D camera frame coordinates to image coordinates, as defined in Equation (3.26).

By inverting each sub-function, we obtain the inverse mapping $\Psi^{-1}$ as:

$$p_R \xRightarrow{M^{-1}} P_R^{V^c} \xRightarrow{(T_{total}^{Visual})^{-1}} \overline{P_R^{\bar{V}}} \qquad (3.136)$$

The first stage, $M^{-1}$, maps image coordinates $p_R$ to camera-frame coordinates $P_R^{V^c}$, using the nonlinear stereo triangulation expressions in Equations (3.28) − (3.30). As discussed in Section 3.2.3, this inverse is well-defined and unique for all image points with non-zero disparity.

The second stage, $(T_{total}^{Visual})^{-1}$, is a rigid-body transformation matrix derived from forward kinematics and is defined as:

$$(T_{total}^{Visual})^{-1} = (\overline{T_E^C} \cdot T_0^6\{visual\})^{-1} = (T_0^6\{visual\})^{-1} \cdot (\overline{T_E^C})^{-1} \qquad (3.137)$$

Also, we have:

$$\overline{P_R^{\bar{V}}} = (T_{total}^{Visual})^{-1} \cdot P_R^{V^c} \qquad (3.138)$$

$T_0^6\{visual\}$ is a transformation matrix defined in Equation (3.114). $\overline{T_E^C}$ is defined in Equation (3.120). Both $T_0^6\{visual\}$ and $\overline{T_E^C}$ belong to the Special Euclidean group SE(3) [67], and are thus always invertible. As a result, $T_{total}^{Visual}$ is invertible, and thus $(T_{total}^{Visual})^{-1}$ exist.

Substituting Equation (3.137) into (3.138), and rearranging yields:

$$T_0^6\{\text{visual}\} = \left(\overline{T_E^C}\right)^{-1} \cdot P_R^{V^c} \cdot (\overline{P_R^{\overline{V}}})^{\dagger} \qquad (3.139)$$

Where $(\overline{P_R^{\overline{V}}})^{\dagger}$ is Moore-Penrose pseudoinverse of $\overline{P_R^{\overline{V}}}$. Given $T_0^6\{\text{visual}\}$, one can recover the joint configuration $q_V$ through an inverse kinematics process, which, as discussed in Section 6.4, is generally non-unique.

We define a nonlinear function $\varphi$ representing the inverse camera-on-robot kinematics, which computes $q_V$ from the reference points' image coordinates $p_R$, their known base frame positions $\overline{P_R^{\overline{V}}}$, and the visual system parameters $P_{a_{visual}}$:

$$\forall\, t \geq 0,\; q_V(t) = \varphi(p_R(t), P_{a_{visual}}, \overline{P_R^{\overline{V}}}) \qquad (3.140)$$

In general, the function $\varphi$ exists due to the invertibility of its component mappings, but it is not unique because of the inherent redundancy and multiple solution branches in the robot's inverse kinematics.

### 3.6.3 Camera-and-Tool Combined Model

The second control process introduced in Chapter 2 focuses on high-accuracy tool manipulation control. This process employs the Camera-and-Tool Combined Model, where the camera's robotic arm remains static, while the tool's robotic arm moves freely with six degrees of freedom (DoFs). The purpose of this model is to relate the joint angles of the tool manipulator to the image coordinates of two interest points, $I_1$ and $I_2$, placed on the tool.

As depicted in Figure 3.15., inertial frames are attached to the base of both the camera robot manipulator and the tool robot manipulator. The distance between the origins of these two inertial frames in the X-direction is denoted as $L_{VT}$. Body-fixed frames are attached to the center of the camera and to one end of the tool.

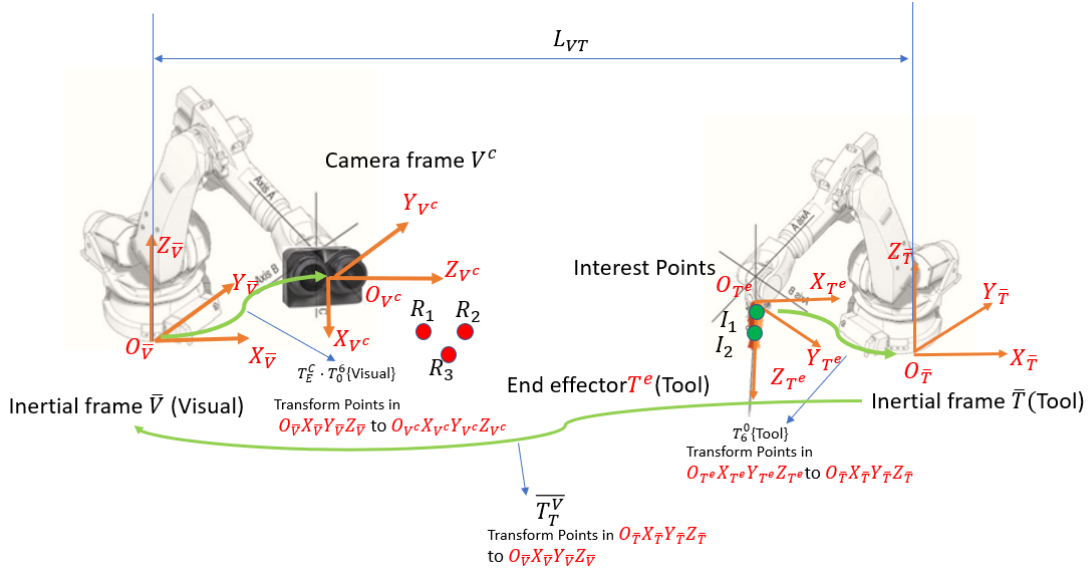

**Figure 3.15**: Coordinates' transformation in Camera-and-Tool combined model.

Assume that the interest points $I_1$ and $I_2$ are located along the $Z_{T^e}$-axis of the tool's end-effector frame. The 3D homogeneous coordinates of these points in the tool's end-effector frame are given as $\overline{P_{I_1}^{T^e}} = [0, 0, Z_{I_1}^{T^e}, 1]^T$, $\overline{P_{I_2}^{T^e}} = [0, 0, Z_{I_2}^{T^e}, 1]^T$. To transform these points into the tool manipulator's inertial frame, the $4 \times 4$ transformation matrix $T_6^0\{\text{Tool}\}$ is used. This matrix represents the forward kinematics of the tool robot arm as described in Equation (3.48). The transformed coordinates in the inertial frame of the tool manipulator are:

$$P_{I_1}^{\overline{T}} = T_6^0\{\text{Tool}\}[\overline{P_{I_1}^{T^e}}]^T = [X_{I_1}^{\overline{T}}, Y_{I_1}^{\overline{T}}, Z_{I_1}^{\overline{T}}, 1]^T \tag{3.141}$$

$$P_{I_2}^{\overline{T}} = T_6^0\{\text{Tool}\}[\overline{P_{I_2}^{T^e}}]^T = [X_{I_2}^{\overline{T}}, Y_{I_2}^{\overline{T}}, Z_{I_2}^{\overline{T}}, 1]^T \tag{3.142}$$

Then, the coordinates of the interest points $I_1$ and $I_2$, initially in the tool inertial frame, can be transformed to the camera inertial frame using the transformation matrix $\overline{T_T^V}$. The transformation is given by:

$$P_{I_1}^{\overline{V}} = \overline{T_T^V}[P_{I_1}^{\overline{T}}, 1]^T = [X_{I_1}^{\overline{V}}, Y_{I_1}^{\overline{V}}, Z_{I_1}^{\overline{V}}, 1]^T \tag{3.143}$$

$$P_{I_2}^{\overline{V}} = \overline{T_T^V}[P_{I_2}^{\overline{T}}, 1]^T = [X_{I_2}^{\overline{V}}, Y_{I_2}^{\overline{V}}, Z_{I_2}^{\overline{V}}, 1]^T \tag{3.144}$$

The transformation matrix $\overline{T_T^V}$ is defined as:

$$\overline{T_T^V} = \begin{bmatrix} 1 & 0 & 0 & L_{VT} \\ 0 & 1 & 0 & 0 \\ 0 & 0 & 1 & 0 \\ 0 & 0 & 0 & 1 \end{bmatrix} \quad (3.145)$$

Next, the coordinates of $I_1$ and $I_2$ in the camera inertial frame can be transformed to the camera end-effector frame using the transformation matrix $\overline{T_E^C} \cdot T_0^6\{\text{Visual}\}$. This transformation is expressed as:

$$P_{I_1}^{V^c} = \overline{T_E^C} \cdot T_0^6\{\text{Visual}\}[P_{I_1}^{\overline{V}}, 1]^T = [X_{I_1}^{V^c}, Y_{I_1}^{V^c}, Z_{I_1}^{V^c}, 1]^T \quad (3.146)$$

$$P_{I_2}^{V^c} = \overline{T_E^C} \cdot T_0^6\{\text{Visual}\}[P_{I_2}^{\overline{V}}, 1]^T = [X_{I_2}^{V^c}, Y_{I_2}^{V^c}, Z_{I_2}^{V^c}, 1]^T \quad (3.147)$$

Where the $4 \times 4$ matrix $T_E^C$ is the transformation matrix from the end-effector frame of the visual system to the camera frame as defined in Equation (3.120). The $4 \times 4$ matrix $T_0^6\{\text{Visual}\}$ represents the transformation from the inertial frame of the visual system to the end-effector frame of the visual system, as defined in Equation (3.114) and it is a function of camera's joint angles at the optimal position $\overline{q_V}$.

To summarize, the total transformation matrix from the tool's end-effector frame to the camera's end-effector frame is denoted as $T_{total}$. Using this matrix, the coordinates of the interest points $I_1$ and $I_2$ in the camera's end-effector frame are expressed as:

$$P_{I_1}^{V^c} = T_{total}^{Tool}[\overline{P_{I_1}^{Te}}, 1]^T = [X_{I_1}^{V^c}, Y_{I_1}^{V^c}, Z_{I_1}^{V^c}, 1]^T \quad (3.148)$$

$$P_{I_2}^{V^c} = T_{total}^{Tool}[\overline{P_{I_2}^{Te}}, 1]^T = [X_{I_2}^{V^c}, Y_{I_2}^{V^c}, Z_{I_2}^{V^c}, 1]^T \quad (3.149)$$

The total transformation matrix $T_{total}$ is expressed as:

$$T_{total}^{Tool} = \overline{T_E^C} \cdot T_0^6\{\text{Visual}\} \cdot \overline{T_T^V} \cdot T_6^0\{\text{Tool}\} \quad (3.150)$$

Here, $T_6^0\{\text{Tool}\}$ represents the transformation from the tool's end-effector frame to its inertial frame. $\overline{T_T^V}$ denotes the transformation between the tool manipulator's and camera manipulator's inertial frames. $\overline{T_E^C} \cdot T_0^6\{\text{Visual}\}$ denotes the transformation from the camera manipulator's inertial frame to the camera frame.

Since the camera remains static during this process, the combined matrix $\overline{T_E^C} \cdot T_0^6\{\text{Visual}\} \cdot \overline{T_T^V}$ is constant.

Finally, using the stereo camera model described in Equations (3.23) − (3.25), the image coordinates of $I_1$ and $I_2$ are computed as follows:

$$p_{I_1} = \left[\frac{2X_{I_1}^{V^c} - b}{2Z_{I_1}^{V^c}}F, \frac{2X_{I_1}^{V^c} + b}{2Z_{I_1}^{V^c}}F, \frac{Y_{I_1}^{V^c}}{Z_{I_1}^{V^c}}F, 1\right]^T = [ul_{I_1}, ur_{I_1}, v_{I_1}, 1]^T \quad (3.151)$$

$$p_{I_2} = \left[\frac{2X_{I_2}^{V^c} - b}{2Z_{I_2}^{V^c}}F, \frac{2X_{I_2}^{V^c} + b}{2Z_{I_2}^{V^c}}F, \frac{Y_{I_2}^{V^c}}{Z_{I_2}^{V^c}}F, 1\right]^T = [ul_{I_2}, ur_{I_2}, v_{I_2}, 1]^T \quad (3.152)$$

Thus, in the Camera-and-Tool Combined Model, the image coordinates of $I_1$ and $I_2$ are ultimately expressed as functions of the six joint angles ($q_{T_1}$,..., $q_{T_6}$) of the tool robotic manipulator. Since the camera arm is static at the optimal pose during this process, the joint angles of the camera robotic manipulator ($\overline{q_V}$) are treated as constant parameters within the nonlinear functions. This relationship enables precise mapping of the interested point's coordinates measured in the effector frame $\overline{P_I^{T^e}}$ to the observed measurements of the interest points in the image frame $p_I$.

We define a nonlinear transformation $\mathcal{H}$, which maps reference points $I_1$, and $I_2$ measured in the in the tool end-effector frame $\{T^e\}$ to their image coordinates as captured by the stereo camera. This transformation is expressed as:

$$\text{Camera-and-Tool mapping } \mathcal{H}: \overline{P_I^{T^e}} = [\overline{P_{I_1}^{T^e}}, \overline{P_{I_2}^{T^e}}] \xrightarrow{\Psi} p_I = [p_{I_1}, p_{I_2}] \quad (3.153)$$

Here, $\overline{P_I^{T^e}} \in \mathbb{R}^{4\times 2}$ is the matrix formed by stacking the 3D homogenous coordinates of the interests points in the tool end-effector frame, and $p_I \in \mathbb{R}^{4\times 2}$ is the corresponding set of 2D homogenous stereo image coordinates of those points.

The Mapping $\mathcal{H}$ is nonlinear and depends on variables current joint angles of the visual robot manipulator: $q_T$ and parameters $P_{a_{tool}}$, which includes stereo camera parameter $P_{a_{camera}}$, robot geometric parameter $P_{a_{robot}}$, and transformation matrix $\overline{T_E^C}$, $\overline{T_T^V}$ and joint angles $\overline{q_V}$. In other words:

$$\forall t \geq 0, p_I = \mathcal{H}(q_T(t), P_{a_{tool}}, \overline{P_I^{T^e}}) \quad (3.154)$$

$$P_{a_{tool}} = [P_{a_{camera}}, P_{a_{robot}}, \overline{T_E^C}, \overline{T_T^V}, \overline{q_V}] \quad (3.155)$$

If the tool is outside the view of the camera, the functions relate the current expected joint angles of tool manipulator ($q_T$) to the estimated image coordinates of interest points ($\widetilde{p_I}$). Similarly, if the tool is within the view of the camera, the same nonlinear functions compute the measured image coordinates of the interest points ($\tilde{p}_I$) based on the disturbed current joint angles ($\widetilde{q_T}$). In other words, the relationships can be expressed as:

$$\forall t \geq 0, \tilde{p}_I(t) = \mathcal{H}(q_T(t), P_{a_{tool}}, \overline{P_I^{T^e}}) \tag{3.156}$$

$$\forall\, t \geq 0, \hat{p}_I(t) = \mathcal{H}(\widetilde{q_T}(t), P_{a_{tool}}, \overline{P_I^{T^e}}) \tag{3.157}$$

### 3.6.4 Inverse Camera-and-Tool Combined Model

Similar to section 3.6.2, we define the inverse process of the nonlinear mapping $\mathcal{H}$ as $\mathcal{H}^{-1}$, which maps a set of interests points $I_1$, and $I_2$ from their measured image coordinates to their corresponding positions in the tool end-effector frame $\{T^e\}$. This transformation is described as:

Inverse Camera-on-robot mapping $\mathcal{H}^{-1}$:

$$p_I = [p_{I_1}, p_{I_2}] \overset{\mathcal{H}^{-1}}{\Longrightarrow} \overline{P_I^{T^e}} = [\overline{P_{I_1}^{T^e}}, \overline{P_{I_2}^{T^e}}] \tag{3.158}$$

From Equations (3.141) – (3.153), the original forward mapping $\mathcal{H}$ in Equation (3.153) can be written as:

$$\overline{P_I^{T^e}} \overset{T_{total}^{Tool}}{\Longrightarrow} P_I^{V^c} = [P_{I_1}^{V^c}, P_{I_2}^{V^c}] \overset{M}{\Longrightarrow} p_I \tag{3.159}$$

Here $T_{total}^{Tool}$ is a transformation matrix from the end-effector frame $\{T^e\}$ to the camera frame $\{V^C\}$, dependent on the joint configuration $q_T$, as defined in Equation (3.150). $P_I^{V^c} \in \mathbb{R}^{4\times 2}$ is the matrix formed by stacking the 3D homogenous coordinates of the interests points in the camera frame. $M$ is the stereo-camera projection function that maps 3D camera frame coordinates to image coordinates, as defined in Equation (3.26).

By inverting each sub-function, we obtain the inverse mapping $\mathcal{H}^{-1}$ as:

$$p_I \overset{M^{-1}}{\Longrightarrow} P_I^{V^c} \overset{(T_{total}^{Tool})^{-1}}{\Longrightarrow} \overline{P_I^{T^e}} \tag{3.160}$$

The first stage, $M^{-1}$, is a unique process that maps image coordinates $p_I$ to camera-frame coordinates $P_I^{V^c}$.

The second stage, $(T_{total}^{Tool})^{-1}$, is a rigid-body transformation matrix derived from forward kinematics and is defined as:

$$(T_{total}^{Tool})^{-1} = (\overline{T_E^C} \cdot T_0^6\{\text{Visual}\} \cdot \overline{T_T^V} \cdot T_6^0\{\text{Tool}\})^{-1}$$
$$= (T_6^0\{\text{Tool}\})^{-1} \cdot (\overline{T_T^V})^{-1} \cdot (T_0^6\{\text{visual}\})^{-1} \cdot (\overline{T_E^C})^{-1} \tag{3.161}$$

Also, we have:
$$\overline{P_I^{Te}} = (T_{total}^{Tool})^{-1} \cdot P_I^{Vc} \tag{3.162}$$

All matrices expressed in $T_{total}^{Tool}$ are always invertible as they all belong to the Special Euclidean group SE(3). As a result, $T_{total}^{Tool}$ is invertible, and thus $T_{total}^{Tool})^{-1}$ exist.

Substituting Equation (3.161) into (3.162), and rearranging yields, by taking $(T_6^0\{Tool\})^{-1} = T_0^6\{Tool\})$:

$$T_0^6\{Tool\} = \overline{P_I^{Te}} \cdot (P_I^{Vc})^\dagger \cdot \overline{T_E^C} \cdot T_0^6\{Visual\} \cdot \overline{T_T^V} \tag{3.163}$$

Where $(P_I^{Vc})^\dagger$ is Moore-Penrose pseudoinverse of $P_I^{Vc}$. Given $T_0^6\{Tool\}$, one can recover the joint configuration $q_T$ through an inverse kinematics process, which, as discussed in Section 6.4, is generally non-unique.

We define a nonlinear function $\hbar$ representing the inverse camera-and-Tool Combined kinematics, which computes $q_T$ from the interest points' image coordinates $p_I$, their known end-effector frame positions $\overline{P_I^{Te}}$, and the tool system parameters $P_{a_{tool}}$:

$$\forall\, t \geq 0,\, q_T(t) = \hbar(p_I(t), P_{a_{tool}}, \overline{P_I^{Te}}) \tag{3.164}$$

In general, the function $\hbar$ exists due to the invertibility of its component mappings, but it is not unique because of the inherent redundancy and multiple solution branches in the robot's inverse kinematics.

## 3.7 SISO Models

In this section, simpler models with one degree of freedom (DoF) will be developed and analyzed. These models are streamlined representations of the system, consisting of a single link for each robot manipulator and a monocular camera. Unlike the full models, only one reference point is placed in space and one interest point is located on the tool, as illustrated in Figure 3.16. and Figure 3.17. These simplified models facilitate a focused investigation of the system's behavior while retaining the essential kinematics of the interaction between the camera, tool, and observed points.

### 3.7.1 SISO Camera-on-Robot Model

In the simplified one degree of freedom (DoF) model, as shown in Figure 3.16., the robot arm has a single link of length $L_1$. The monocular camera is attached at the end of the link and is constrained to rotate around the $Z_{\bar{V}}$-axis, which is fixed in the base inertial frame. The initial orientation of the camera is such that its $Z_{Vc}$-axis is parallel to the $X_{\bar{V}}$-axis in the inertial frame. The angle of rotation of the camera from its initial position is denoted as $\widetilde{q_{V_1}}$. This angle determines the camera's final orientation relative to the inertial frame. The reference point $R_1$ is assumed to lie on the $Z_{Vc} - Y_{Vc}$ plane, with its coordinates in the inertial frame given as: $\overline{P_{R_1}^{\bar{V}}} = [\overline{X_{R_1}^{\bar{V}}}, \overline{Y_{R_1}^{\bar{V}}}, L_1, 1]^T$.

Using the D-H convention outlined in Equations (3.34) – (3.36), the transformation matrix $T_0^1\{Visual\}$ from the inertial frame to the camera end-effector frame is derived as:

$$T_0^1\{Visual\} = \begin{bmatrix} 0 & 0 & -1 & L_1 \\ \sin(\widetilde{q_{V_1}}) & \cos(\widetilde{q_{V_1}}) & 0 & 0 \\ \cos(\widetilde{q_{V_1}}) & -\sin(\widetilde{q_{V_1}}) & 0 & 0 \\ 0 & 0 & 0 & 1 \end{bmatrix} \quad (3.165)$$

Where $\widetilde{q_{V_1}}$ is the rotational angle of the camera about the $Z_{\bar{V}}$-axis, measured positively in the clockwise direction. Assume the camera coordinate frame is set at the end-effector frame of the visual system.

Then, the coordinates of $R_1$ in the camera's end-effector frame are calculated as:

$$\overline{P_{R_1}^{Vc}} = T_0^1\{Visual\} \left[\overline{P_{R_1}^{\bar{V}}}\right]^T = [\overline{X_{R_1}^{Vc}}, \overline{Y_{R_1}^{Vc}}, \overline{Z_{R_1}^{Vc}}, 1]^T$$

$$= \begin{bmatrix} 0 \\ \overline{X_{R_1}^{\bar{V}}}\sin(\widetilde{q_{V_1}}) + \overline{Y_{R_1}^{\bar{V}}}\cos(\widetilde{q_{V_1}}) \\ \overline{X_{R_1}^{\bar{V}}}\cos(\widetilde{q_{V_1}}) - \overline{Y_{R_1}^{\bar{V}}}\sin(\widetilde{q_{V_1}}) \\ 1 \end{bmatrix} \quad (3.166)$$

By applying the monocular camera model equations (3.9) and (3.10), the coordinates of the reference point $R_1$ in the image frame can be derived as follows:

$$\widehat{p_{R_1}} = [\widehat{u_{R_1}}, \widehat{v_{R_1}}]^T = \left[\frac{\overline{X_{R_1}^{Vc}}}{\overline{Z_{R_1}^{Vc}}}F, \frac{\overline{Y_{R_1}^{Vc}}}{\overline{Z_{R_1}^{Vc}}}F\right]^T \quad (3.167)$$

Substituting the transformed coordinates ($\overline{X_{R_1}^{Vc}}, \overline{Y_{R_1}^{Vc}}, \overline{Z_{R_1}^{Vc}}$) from Equation (3.166), the image coordinates become:

$$\widehat{p_{R_1}} = \begin{bmatrix} \widehat{u_{R_1}} \\ \widehat{v_{R_1}} \end{bmatrix} = \begin{bmatrix} 0 \\ \dfrac{\overline{X_{R_1}^V}\sin(\widetilde{q_{V_1}}) + \overline{Y_{R_1}^V}\cos(\widetilde{q_{V_1}})}{\overline{X_{R_1}^V}\cos(\widetilde{q_{V_1}}) - \overline{Y_{R_1}^V}\sin(\widetilde{q_{V_1}})} F \end{bmatrix} \quad (3.168)$$

Let us rewrite the coordinates $(\overline{X_{R_1}^V}, \overline{Y_{R_1}^V})$ of the reference point $R_1$ in polar form for clarity. The radial distance $R$ and angle $\varphi$ are defined as:

$$R = \sqrt{\overline{X_{R_1}^V}^2 + \overline{Y_{R_1}^V}^2}$$

and
$$\overline{X_{R_1}^V} = R\cos(\varphi), \ \overline{Y_{R_1}^V} = R\sin(\varphi),$$

where
$$\varphi = \tan^{-1}(\overline{Y_{R_1}^V}/\overline{X_{R_1}^V}) \quad (3.169)$$

Substituting these expressions into Equation (3.167), the vertical coordinate $\widehat{v_{R_1}}$ in the image frame can be rewritten in terms of polar coordinates. Using trigonometric identities, we have:

$$\widehat{v_{R_1}} = F \frac{R\sin(\varphi + \widetilde{q_{V_1}})}{R\cos(\varphi + \widetilde{q_{V_1}})} = F\tan(\varphi + \widetilde{q_{V_1}}) \quad (3.170)$$

The angle $\varphi$ represents the initial orientation of the camera (axis $Z_{V^c}$) with respect to the $X_{\bar{V}}$-axis, and angle $\widetilde{q_{V_1}}$ represents the rotation of the camera link from its initial position. Equation (3.170) provides a mathematical expression for the 1-DoF Camera-on-Robot Model that relates the measured 1-Dof image coordinate to the 1-Dof disturbed joint angle of the visual system.

Similarly as shown in Equation (3.131), the model that relates the target reference coordinate to the optimal joint angle of the camera manipulator can be expressed as:

$$\overline{v_{R_1}} = F\tan(\varphi + \overline{q_{V_1}}) \quad (3.171)$$

where
$$\varphi = \tan^{-1}(\overline{Y_{R_1}^V}/\overline{X_{R_1}^V}) \quad (3.172)$$

Also, as shown in Equation (3.133), in the high precise tool manipulation control process, the model that relates the target tool coordinate to the target 3D coordinate of the tool can be expressed as:

$$\overline{v_{I_1}} = f_v tan(\varphi + \widetilde{q_{V_1}}) \tag{3.173}$$

where
$$\varphi = tan^{-1}(\overline{Y_{I_1}^V}/\overline{X_{I_1}^V}) \tag{3.174}$$

For lenses projecting rectilinear (non-spatially distorted) images, the angle of field $\alpha$ is determined by the focal length $F$ and the width $m$ of the imbedded image sensor, which is used for converting light into electrical charge and transferring it pixel by pixel for processing. The angle of field $\alpha$ of Zed 2 camera is calculated using the formula:

$$\alpha = 2\,tan^{-1}\frac{m/2}{F} = 86.09° \tag{3.175}$$

From the specifications of the Zed 2 camera (Appendix A Table A3), the focal length and sensor width are given as: $F = 2.8mm$, and $m = 5.23mm$.

The camera's rotation angle $\widetilde{q_{V_1}}$ must be constrained to ensure that the point $R_1$ remains within the camera's field of view. Considering the initial angle $\varphi$, it is evident that the total angle $|\varphi + \widetilde{q_{V_1}}|$ must not exceed half of the camera's angle of field $\alpha/2$. This can be expressed as:

$$|\varphi + \widetilde{q_{V_1}}| \leq \frac{\alpha}{2} = 43.045° = 0.751\ \text{rad} \tag{3.176}$$

Therefore, the allowable range for the rotation angle $\widetilde{q_{V_1}}$ is given by:

$$-\varphi - 43.045° \leq \widetilde{q_{V_1}} \leq -\varphi + 43.045° \tag{3.177}$$

Additionally, to ensure that $R_1$ can be projected onto the sensor frame at the camera's initial position ($\widetilde{q_{V_1}} = 0°$), the initial angle $\varphi$ must satisfy:

$$-43.045° \leq \varphi \leq 43.045 \tag{3.178}$$

The vertical coordinate of the reference point in the image frame, $\widehat{v_{R_1}}$, must also lie within the bounds of the sensor height. Given the width of the sensor $m = 5.23\ mm$, the bound on $\widehat{v_{R_1}}$ is:

$$|\widehat{v_{R_1}}| \leq m/2 = 2.615\text{mm} \tag{3.179}$$

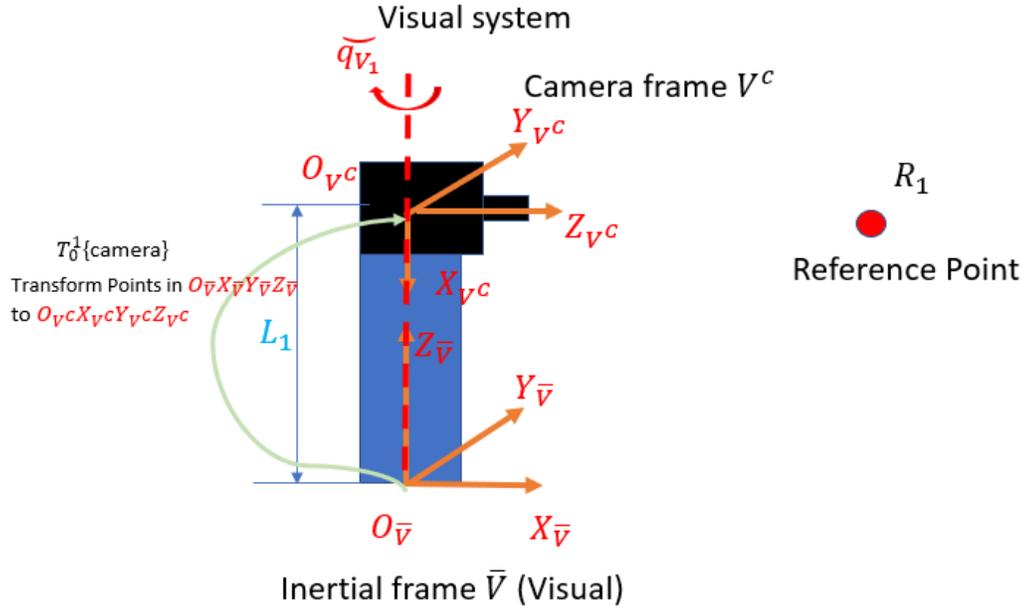

**Figure 3.16:** The SISO Camera-on-Robot model.

### 3.7.2 SISO Camera-and-Tool Combined Model

In Figure 3.17., the setup involves two one-link robot arms, each of length $L_1$, separated by a distance $L_{VT}$. The camera is mounted on one arm, which rotates about the $Z_{\bar{V}}$ -axis of the inertial frame by an angle $\overline{q_{V_1}}$. Once rotated, the camera is held static. The tool is attached to the second arm, which rotates around the $Z_{\bar{T}}$ - axis of its inertial frame by an angle $\widetilde{q_{T_1}}$. The tool has a length $L_t$, and the interest point $I_1$ is located at one of its ends, with its coordinates in the tool's end-effector frame given as: $\overline{P_{I_1}^{Te}} = (0, 0, L_t, 1)$.

Using the D-H convention outlined in Equations (3.34) – (3.36), the transformation matrix $T_1^0\{Tool\}$, which maps coordinates from the tool's end-effector frame to its inertial frame, is expressed as:

$$T_1^0\{Tool\} = \begin{bmatrix} 0 & -\sin(\widetilde{q_{T_1}}) & -\cos(\widetilde{q_{T_1}}) & 0 \\ 0 & -\cos(\widetilde{q_{T_1}}) & \sin(\widetilde{q_{T_1}}) & 0 \\ -1 & 0 & 0 & L_1 \\ 0 & 0 & 0 & 1 \end{bmatrix} \quad (3.180)$$

Where $\widetilde{q_{T_1}}$ is the rotation angle of the tool about the $Z_{\bar{V}}$-axis, measured positively in the clockwise direction.

By applying the joint configuration of the visual system, specifically the rotation angle $\overline{q_{V_1}}$, to the transformation matrix in Equation (3.165), the matrix $T_0^1\{Visual\}$ can be expressed as:

$$T_0^1\{Visual\} = \begin{bmatrix} 0 & 0 & -1 & L_1 \\ \sin(\overline{q_{V_1}}) & \cos(\overline{q_{V_1}}) & 0 & 0 \\ \cos(\overline{q_{V_1}}) & -\sin(\overline{q_{V_1}}) & 0 & 0 \\ 0 & 0 & 0 & 1 \end{bmatrix} \quad (3.181)$$

Where $\overline{q_{V_1}}$ is the rotation angle of the tool about the $Z_{\bar{V}}$-axis, measured positively in the clockwise direction.

The total transformation matrix from the tool's end-effector frame to the camera's end-effector frame, as expressed in Equation (3.150), is calculated as follows:

$$T_{total} = T_0^1\{Visual\} * \overline{T_T^V} * T_1^0\{Tool\}$$

$$= \begin{bmatrix} 0 & 0 & 0 & 0 \\ 0 & -\sin(\overline{q_{V_1}})\sin(\widetilde{q_{T_1}}) - \cos(\overline{q_{V_1}})\cos(\widetilde{q_{T_1}}) & -\sin(\overline{q_{V_1}})\cos(\widetilde{q_{T_1}}) + \cos(\overline{q_{V_1}})\sin(\widetilde{q_{T_1}}) & L_{VT}\sin(\overline{q_{V_1}}) \\ 0 & -\cos(\overline{q_{V_1}})\sin(\widetilde{q_{T_1}}) + \sin(\overline{q_{V_1}})\cos(\widetilde{q_{T_1}}) & -\cos(\overline{q_{V_1}})\cos(\widetilde{q_{T_1}}) - \sin(\overline{q_{V_1}})\sin(\widetilde{q_{T_1}}) & L_{VT}\cos(\overline{q_{V_1}}) \\ 0 & 0 & 0 & 1 \end{bmatrix} \quad (3.182)$$

Using the total transformation matrix $T_{total}$, the coordinates of the interest point $I_1$ in the camera's frame are calculated as:

$$P_{I_1}^{Vc} = T_{total}\left[\overline{P_{I_1}^{Te}}\right]^T = \left[X_{I_1}^{Vc}, Y_{I_1}^{Vc}, Z_{I_1}^{Vc}, 1\right]^T \quad (3.183)$$

Substituting $\overline{P_{I_1}^{Te}} = (0, 0, L_t, 1)$ the resulting coordinates are:

$$P_{I_1}^{Vc} = \begin{bmatrix} 0 \\ [-\sin(\overline{q_{V_1}})\cos(\widetilde{q_{T_1}}) + \cos(\overline{q_{V_1}})\sin(\widetilde{q_{T_1}})]L_t + L_{VT}\sin(\overline{q_{V_1}}) \\ [-\cos(\overline{q_{V_1}})\cos(\widetilde{q_{T_1}}) - \sin(\overline{q_{V_1}})\sin(\widetilde{q_{T_1}})]L_t + L_{VT}\cos(\overline{q_{V_1}}) \\ 1 \end{bmatrix} \quad (3.184)$$

The monocular camera model, as defined in Equations (3.9) and (3.10), projects the 3D coordinates of a point from the camera's end-effector frame into the 2D image frame. The relationship is given by:

$$\widehat{p_{I_1}} = \left[\frac{X_{I_1}^{Vc}}{Z_{I_1}^{Vc}}F, \frac{Y_{I_1}^{Vc}}{Z_{I_1}^{Vc}}F\right]^T = [\widehat{u_{I_1}}, \widehat{v_{I_1}}]^T \quad (3.185)$$

Substituting the transformed coordinates $(X_{I_1}^{V^c}, Y_{I_1}^{V^c}, Z_{I_1}^{V^c})$ from Equation (3.184), the image coordinates become:

$$\widehat{p_{I_1}} = \begin{bmatrix} \widehat{u_{I_1}} \\ \widehat{v_{I_1}} \end{bmatrix}$$

$$= \begin{bmatrix} 0 \\ \dfrac{[-\sin(\overline{q_{V_1}})\cos(\widetilde{q_{T_1}}) + \cos(\overline{q_{V_1}})\sin(\widetilde{q_{T_1}})]L_t + L_{VT}\sin(\overline{q_{V_1}})}{[-\cos(\overline{q_{V_1}})\cos(\widetilde{q_{T_1}}) - \sin(\overline{q_{V_1}})\sin(\widetilde{q_{T_1}})]L_t + L_{VT}\cos(\overline{q_{V_1}})} F \end{bmatrix} \quad (3.186)$$

Using trigonometric identities, Equation (3.186) can be simplified as:

$$\widehat{v_{I_1}} = F \frac{Q(\widetilde{q_{T_1}}) + \tan(\overline{q_{V_1}})}{1 - Q(\widetilde{q_{T_1}})\tan(\overline{q_{V_1}})}$$

where

$$Q(\widetilde{q_{T_1}}) = \frac{L_t \sin(\widetilde{q_{T_1}})}{L_{VT} - L_t \cos(\widetilde{q_{T_1}})} \quad (3.187)$$

This shows that the one DoF Camera-and-Tool Combined Model maps the tool's rotational angle $\widetilde{q_{T_1}}$ to the measured image coordinate $\widehat{v_{I_1}}$. The parameters $L_{VT}$, $L_t$, $F$, and $\overline{q_{V_1}}$ constant for the system. This equation captures the essential kinematics of the system in a compact mathematical form.

Similarly, as shown in Equation (3.156), the model that relates the estimated coordinate to the disturbed joint angle of the tool manipulator can be expressed as:

$$\widetilde{v_{I_1}} = F \frac{Q(q_{T_1}) + \tan(\overline{q_{V_1}})}{1 - Q(q_{T_1})\tan(\overline{q_{V_1}})}$$

where

$$Q(q_{T_1}) = \frac{L_t \sin(q_{T_1})}{L_{VT} - L_t \cos(q_{T_1})} \quad (3.188)$$

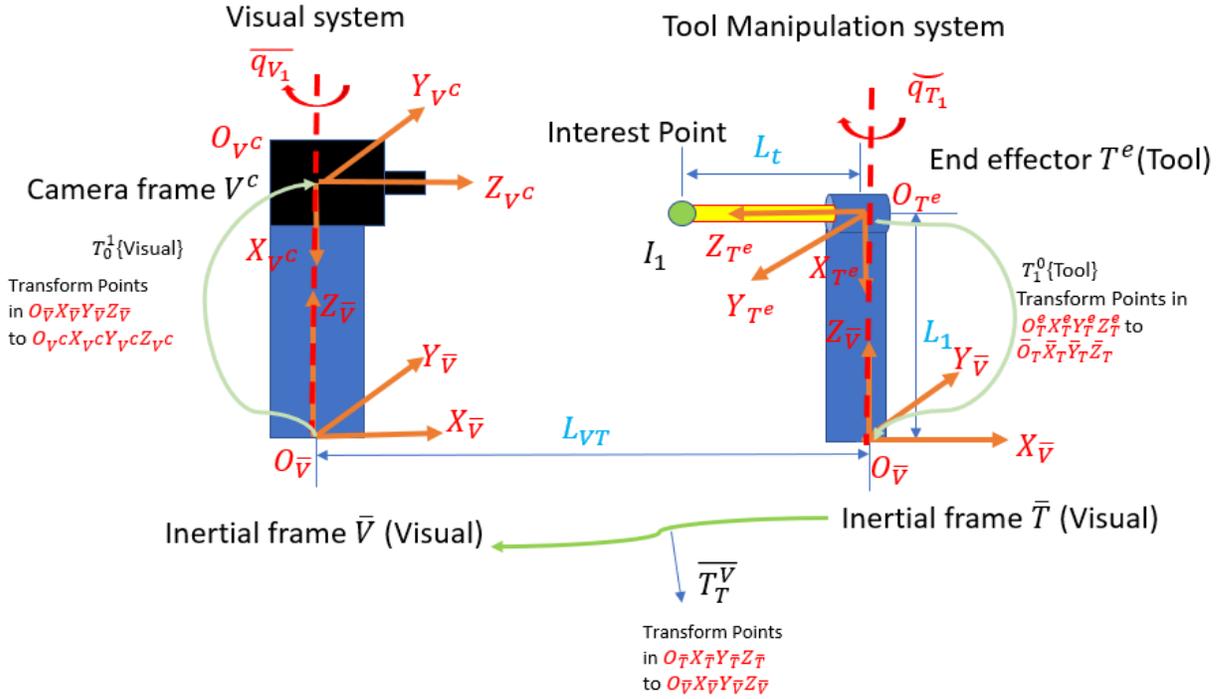

**Figure 3.17:** The SISO Camera-and-Tool combined model.

## 3.8 Conclusion

In this chapter, we developed and analyzed mathematic models of the camera system and robotic manipulators used in the development of control architectures. MIMO models were explicitly derived for both the visual system and the tool manipulation systems, demonstrating the relationship between joint configurations and the observed image coordinates of points. Additionally, SISO models were constructed for one DoF robotic setups, facilitating the development and analysis of SISO Youla controllers in Chapter 5.

These models establish a mathematical foundation for precise control architectures in Chapter 2. By explicitly linking joint angles and image projections, the chapter paves the way for the development of the image-based visual servoing control strategies in the following chapters. Furthermore, the derived frameworks are essential for evaluating robustness against uncertainties arising from model variations.

___________________________________________________________Chapter 4
# The Camera Location Search Algorithm

In a manufacturing environment, the quality of camera estimations may vary significantly from one observation location to another, as the combined effects of environmental conditions result in different noise levels of a single image shot at different locations. In this chapter, we propose an algorithm for the camera's moving policy so that it explores the camera workspace and searches for the optimal location where the images' noise level is minimized. Also, this algorithm ensures the camera ends up at a suboptimal (if the optimal one is unreachable) location among the locations already searched, with limited energy available for moving the camera. Unlike a simple brute force approach, the algorithm enables the camera to explore space more efficiently by adapting the search policy from learning the environment. With the aid of an image averaging technique, this algorithm, in use of a solo camera, achieves the observation accuracy in eye-to-hand configurations to a desirable extent without filtering out high-frequency information in the original image. A virtual scenario of camera searching and locating has been simulated and the results show the success of this algorithm's improvement of observation precision with limited energy.

## 4.1 Motivation

Real-world products are usually manufactured in a confined and complicated environment where complex factors may significantly degrade the quality of observed signals given from even the most precise sensors. For example, Illumination changes in space [68] could be a serious issue for camera's estimation in real manufacturing process, which is not normally addressed in research experiments where illumination is usually sufficient and equally distributed.

In this chapter, we would like to address the importance of cameras' location in the ETH and discuss the benefits of developing algorithms placing the camera in an optimal location. To observe a single point in the workspace, there are an infinite number of poses that a fixed camera can be placed in the space. Among all possible poses of the camera that keep the same target within the field of view limit, the image noise at different locations may play a great role in the precision of observations.

Many environmental conditions (such as illumination, temperature, etc.) affect the noise level of a single image [69]. In a manufacturing environment, the quality of estimations from a camera may vary significantly from one observation location to another. For instance, the precision of an estimation improves greatly if the camera moves from a location where it is placed within the shade of machines to a location that has better illumination. Also, places near machines may be surrounded by strong electrical signals that also introduce extra noise into the camera sensors.

In section 1.3.4, we have discussed and compared various image denoising methods. The image averaging approach, averaging multiple noisy images taken from the same perspective, has the advantages of reducing the unwanted noise as well as retaining all original image details. However, filters that operate in either spatial or wavelet domain inevitably filter out (or blur) some important high-frequency useful information of the original image, even though they are fast in processing. Especially at locations where the signal-to-noise ratio (SNR) in an image is low, no matter what filters are applied, it is difficult to safeguard the edges of the noise-free image as well as reduce the noise to a desirable level. Thus, developing an algorithm that searches for locations of the camera where SNR is high is beneficial.

Moreover, the number of images required for averaging has a quadratic dependence on the ratio of an image's original noise level to the reduced level that is desirable for observations. Therefore, the number of images required can serve as an indicator of the noise level in a single image.

Furthermore, in the denoising process, the precise estimations require that the original image details are retained as much as possible. Based on previous statements, we decided to choose image averaging over all other denoising techniques in this work.

Frame-based cameras, which detect, track and match visual features by processing images at consecutive frames, cause a fundamental problem of delays in image processing and the consequent robot action [70]. By averaging more images for an observation, less noises are maintained in the averaged image, but more processing time is required for acquiring visual features; it is a typical tradeoff between high-rate and high-resolution in cameras [68]. The control loop frequency is limited in many cases due to low rate of imaging, which undermines the capability/usage of visual servoing technique in high-speed operations. Therefore, it is rewarding for an eye-in-hand configuration to find the location of the camera where original image noise is lowest and thus the least number of images is needed in order to facilitate the manufacturing speed.

Brute force search [71] is a simple and straightforward approach to look for the optimal location of the camera. It systematically searches all possible locations in space and gives the best solution. This method is only feasible when the size of search space is small enough. Otherwise, brute force which checks every possibility, is usually inefficient, especially for large datasets or complex problems, where the number of possibilities can grow exponentially. An intelligent algorithm must be developed to optimize energy efficiency and reduce time consumption.

In this chapter, we propose an algorithm that searches efficiently for the camera's workspace to find an optimal location (if its orientation is fixed) of the camera so that a single image taken at this location has the smallest noise level among images taken at all locations in the space. With limited energy for moving the camera, this algorithm also ensures the camera ends up at a suboptimal (if the optimal pose is unreachable) location among the locations already searched.

## 4.2 Core Algorithm

A camera, mounted on a 6-joint robot manipulator, moves freely in 6 degrees of freedom (6 DoFs) in space. Assuming the camera is fixed in orientation towards the face of the object of interest, only the position (3 DoFs) of the camera can vary from the movement of the robot arm. The camera can only move to locations within the maximum reach of the robot manipulator and the object can only be detectable within the view angle of the camera. Those constraints create a camera operational space (Figure 4.1. a), which is then gridded with a certain resolution of distance. All the intersections of grids result in a set of nodes ($S_{total}$), which are all possible candidates of locations that the camera may move to and search for an optimal/sub-optimal pose utilizing the algorithm. Each node is indexed and denoted as $Node_{index}$.

Figure 2.3. has illustrated the iterative process of the camera's search and movement algorithm. The core of this algorithm is to generate the target locations to be explored in the camera's operational space. The details of this algorithm are shown in the area circled with dashed lines in Figure 2.3. In this section, we will put efforts into the development of the number of images estimation function, the explorable nodes set, the stochastic modified evaluation function, the next location selection process, and the estimated energy cost function.

### 4.2.1 Picture Number Estimation Function

In each iteration of the algorithm, the camera moves to a new node $Node_{index}$ with location $P^C$, and the algorithm evaluates the number of images $N^C$ needed at this location. The node ($Node_{index}$) is then set to be explored and saved in a set $S_{explored}$, and its index is saved in another set $S_{index-exp}$. Each node $Node_{index}$ (where $index \in S_{index-exp}$) that has been explored in iterations is one-on-one paired with the values: location $P_{index}$, and image numbers of that node $N_{index}$. Mathematically, pairs are expressed as $Node_{index}$: ($P_{index}$, $N_{index}$) and all pairs are saved in a hash map $M_{explored}$. Then $M_{explored}$ is used to develop an estimation function that can estimate the number of images needed for the rest unexplored nodes. The estimated image number calculated for all nodes in space is saved in a set $S_{\widetilde{N}}$. Expressions of any element in $S_{\widetilde{N}}$ are:

*For any node index $x$ in the Total Node Set: $S_{total} = \{Node_x | x \in (1,2, \ldots m)\}$, where $m$ is the total number of nodes, and $\widetilde{N}_x$ is estimated image numbers at $Node_x$:*

$$\widetilde{N}_x = \begin{cases} \sum_{i \in S_{index-exp}} \frac{\overline{ED}(P_i, P_x)^{-K_{est}}}{\sum_{i \in S_{index-exp}} \overline{ED}(P_i, P_x)^{-K_{est}}} N_i, & if\ x \notin S_{index-exp} \\ N_x, & if\ x \in S_{index-exp} \end{cases} \quad (4.1)$$

$$Then: S_{\widetilde{N}} = \{\widetilde{N}_x | x \in (1,2, \ldots m)\} \quad (4.2)$$

where $K_{est}$ is a positive constant parameter

$$S_{index-exp} = \{index | Node_{index} \in S_{explored}\}, \quad (4.3)$$

$$\overline{ED}(P_a, P_b) = \sqrt{(P_{ax} - P_{bx})^2 + (P_{ay} - P_{by})^2 + (P_{az} - P_{bz})^2}, \quad (4.4)$$

with $P_a, P_b$ are 3D locations of $Node_a$ and $Node_b$:

$P_a = (P_{ax}, P_{ay}, P_{az})$ and $P_b = (P_{bx}, P_{by}, P_{bz})$

Equation (4.1) is a weighted function concerning each explored node. If one explored node $Node_i$ (where $i \in S_{index-exp}$) is closer to the node $Node_x$, its image number $N_i$ has a larger weight (effect) on the estimation of image number of node $Node_x$: $\tilde{N}_x$. Thus, a minus sign of $K_{est}$ is utilized to indicate the negative correlation between the Euclidean distance and weight.

Equation (4.1) shows that at least some initial nodes are required to be explored before their values can be used to estimate image numbers in other nodes. For the estimation function to work properly by covering all nodes in the camera operational space, four initial nodes are selected in the following steps:

1. Find the first two nodes on the boundary of operational space so that the Euclidean distance between these two nodes is the largest.

2. Select other two nodes whose connection line is perpendicular to the previous connection line and the Euclidean distance between these two nodes is also the largest among all possible node pairs.

3. Move the camera to those four nodes in space with proper orientation. Take a single image at each location and estimate the number of images required at those locations. And save all four nodes in $S_{explored}$ and their values in $M_{explored}$.

Figure 4.1. b shows the selection of four initial nodes in a camera operational space and shows an example of the use of the estimation function to estimate the picture number in one node.

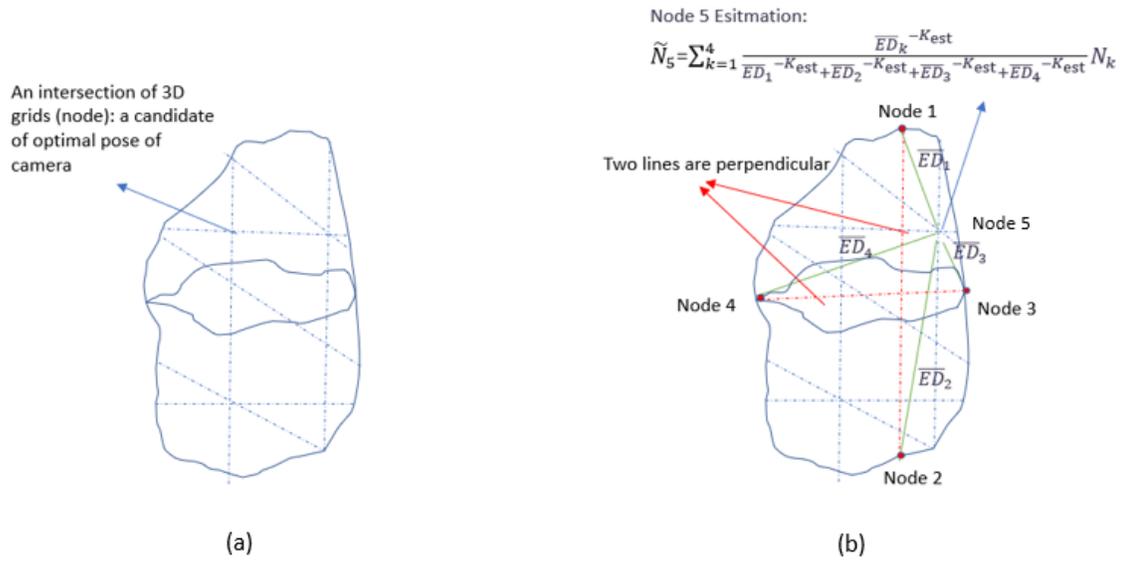

**Figure 4.1:** The operational space of camera and an example of estimation function. Note: (a): The operational space of the camera is an arbitrary closed 3D geometry in the space shown in the plot. The gridded area is in 3D (blue dashed lines). (b): Four initial nodes are chosen so that their connection lines (red dashed line) are perpendicular. The picture number of Node 5 can be estimated by Equation (4.1) from four initial explored nodes. $\overline{ED}$ represents Euclidean distance (green line).

### 4.2.2 Explorable Node Selection

A stage in the algorithm, Explorable Node Selection, generates a set of nodes $S_{explorable}$, which contains all the camera's next explorable locations from the current location $P^C$ if available current energy for moving $E_{bound}$ is given. The explorable set is found in the following steps (Figure 4.2):

With current energy bound $E_{bound}$, find all nodes in the space that can be reached from the current node $P^c$.

In other words, all feasible nodes are in the set:

$$S_{feasible} = \{Node_{index} | E(P^c, P_{index}) \leq E_{bound}\} \quad (4.5)$$

where $P$ is the position of a node and $E(P_a, P_b)$ is the estimated energy cost from $P_a$ to $P_b$.

When the current energy bound $E_{bound}$ is greater than a predefined energy threshold $E_T$ ($E_{bound} > E_T$), then the algorithm is safe to explore all nodes in the feasible range $S_{feasible}$ in the next iteration loop. In other words:

$$S_{explorable} = S_{feasible}, \quad if\ E_{bound} > E_T \tag{4.6}$$

*Note*: $E(P_a, P_b)$ is generated from a dynamic model of a robot arm control system. The details of this function are developed in the following section of this paper.

4. When $E_{bound} \leq E_T$, step 2 and step 3 are applied in finding the explorable set. In the feasible set, look at all explored nodes and find the one that has the smallest number of images. In other words:

$$S_{feas\&exp} = \{Node_{index} | Node_{index} \in S_{feasible} \cap S_{explored}\} \tag{4.7}$$

$$\exp\_min = \{index | N_{\exp\_min} = \min(N_{index} | Node_{index} \in S_{feas\&exp})\} \tag{4.8}$$

where $S_{explored}$ is the set of all explored nodes.

$S_{feas\&exp}$ is the intersection set between $S_{feasible}$ and $S_{explored}$. $N_{\exp\_min}$ is the minimum number of images in the $S_{feas\&exp}$ set, and $\exp\_min$ is the index of the node that paired with $N_{\exp\_min}$.

5. The estimation function may not give accurate results for some unexplored nodes. Therefore, it is possible that the algorithm may make the camera end up in a node that has a large number of images in some iteration loops. Because of that, our algorithm needs to make sure at worst it has enough energy to go back to the best (minimum number of images) node that has been explored when the available energy is below the threshold ($E_{bound} \leq E_T$). This further reduces the feasible set $S_{feasible}$. The explorable set $S_{explorable}$ can be written as:

$$S_{explorable} = \{Node_{index} | E_1 + E_2 \leq E_{bound}\}, \quad if\ E_{bound} \leq E_T \tag{4.9}$$

$$E_1 = E(P^c, P_{index}),\ E_2 = E(P_{index}, P_{\exp\_min}) \tag{4.10}$$

Where $E_1$ denotes the energy of moving from current node location $P^c$ to another node location $P_{index}$ in space, and $E_2$ denotes the energy of moving from that node location $P_{index}$ to the node location $P_{exp\_min}$, which is paired with minimum number of images in the explored set.

Figure 4.2. illustrates the selection of explorable nodes set.

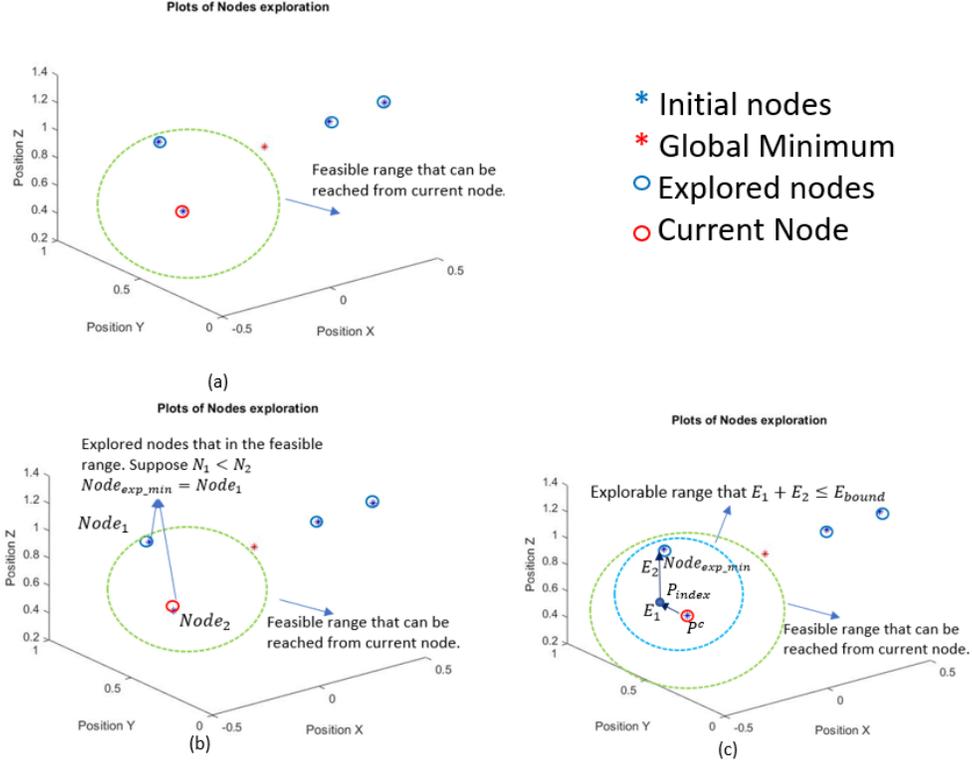

**Figure 4.2:** Find the explorable set when the global minimum is inaccessible. (a). Step 1: Obtain the feasible range $S_{feasible}$ (green dashed line). (b). Step 2: Find the explored node with a minimum picture number in the feasible range. Initial nodes (blue stars) are explored (blue circles). $Node_1$ is $Node_{exp\_min}$. (c). Step 3: Obtain explorable range when $E_{bound} \leq E_T$. The explorable range $S_{explorable}$ (blue dashed line) is a subset of $S_{feasible}$.

### 4.2.3 Stochastic Modified Evaluation

For all nodes in $S_{explorable}$, we set up a hash map $M_{explorable}$. For each node, its location and number of required pictures for that node are given as: $Node_{index}^{explorable}: (P_{index}^{explorable}, \widetilde{N}_{index}^{explorable})$. The number of pictures required for nodes in $S_{explorable}$ are calculated from the estimation function in Equation (4.1).

The algorithm tends to select nodes located around the explored node that has the minimum number of pictures because, from Equation (4.1), those nodes tend to have the smallest estimated number of pictures. However, this process does not guarantee reaching the global minimum, therefore, minimizing the number of pictures. To sufficiently explore unknown nodes as well as exploit information from already explored nodes, a stochastic process is introduced to modify the picture number estimation function given in Equation (4.1).

Assume when the locations of a camera in space deviate with a smaller amount from each other, the difference of real picture number values at those locations is also smaller. Thus, among all unexplored nodes, the estimated picture number of an unexplored node, which is closer to the explored node, is more deterministic. To emphasize this feature, we make the estimated picture number follow a normal distribution with a mean $\mu_{\widetilde{N}}$ being equal to the value from Equation (4.1) and a standard deviation $\sigma_{\widetilde{N}}$. As an unexplored node is further away from explored nodes, the larger value of $\sigma_{\widetilde{N}}$ should be assumed, which means more stochastic estimation (illustrated in Figure 4.3.). The following Equations (4.11) − (4.13) summarize the above discussion:

For any nodes $Node_{index=x}^{explorable}$ in $S_{explorable}$, its image number estimation should follow the following normal distributions:

$$P(Z_x) = \mathcal{N}(\mu_{\widetilde{N}_x}, \sigma_{\widetilde{N}_x}) \qquad (4.11)$$

$$\mu_{\widetilde{N}_x} = \widetilde{N}_x^{explorable} \qquad (4.12)$$

$$\sigma_{\widetilde{N}_x} = K_{sd} * \overline{ED}_{min} \qquad (4.13)$$

where $P(Z_x)$ is the probability of a random variable $Z$ at $Node_x^{explorable}$. $\widetilde{N}_x^{explorable}$ is the estimated value from Equation (4.1) of $Node_x^{explorable}$. $K_{sd}$ is a constant parameter, and $\overline{ED}_{min}$ is the smallest Euclidean distance among distances between that node and explored nodes.

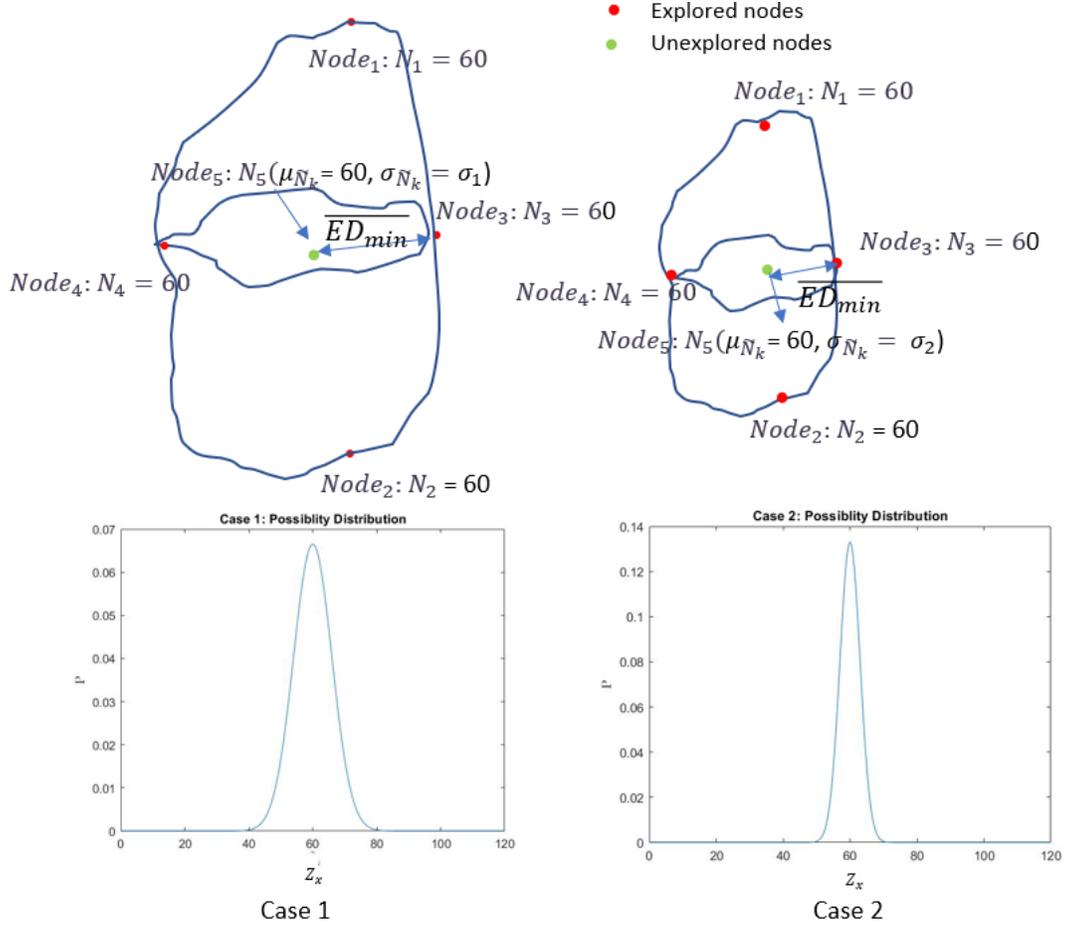

**Figure 4.3:** Both case1 and case2 have the same estimated value of an unexplored node (green point) from Equation (4.1). However, the node in case 1 has a larger distance to explored nodes (red points) compared to that in case 2. Therefore, we set different standard deviations ($\sigma_1 > \sigma_2$) to differentiate possibility distributions. This approach distinguishes nodes that have the same estimations from Equation (4.1) to have different likelihood to be explored in the next iteration.

As discussed above, the algorithm ensures that the camera at any iteration, always has enough energy to move back to the previous best node $Node_{\exp\_min}$. Therefore, a new distribution can be generated, which sets values bigger than the minimum ($N_{\exp\_min}$) in the original distribution to be the minimum value. Then Equations (4.14) − (4.16) are modified with a new random variable $Z_x^{new}$ and its expectations as follows:

$$Z_x^{new} = \begin{cases} Z_x, & if\ Z_x < N_{\exp\_min} \\ N_{\exp\_min}, & if\ Z_x \geq N_{\exp\_min} \end{cases} \quad (4.14)$$

$$P(Z_x^{new}) = P(Z_x) = \mathcal{N}(\mu_{\widetilde{N}_x}, \sigma_{\widetilde{N}_x}) \qquad (4.15)$$

$$\widehat{N}_x^{explorable} = \sum [Z_x^{new} * P(Z_x^{new})] \qquad (4.16)$$

where $\widehat{N}_x^{explorable}$ is the stochastic modified estimated value for the node $Node_x^{explorable}$.

Then a new hash map $\widehat{M}_{explorable}$ is set up with pairs as $Node_{index}^{explorable}$ : $(P_{index}^{explorable}, \widehat{N}_x^{explorable})$. The next target node to be explored is the one that has the smallest modified estimated value in the map:

$$index_{Next} = \{index | \widehat{N}_{index_{Next}}^{explorable} = \min(\widehat{N}_{index}^{explorable})\} \qquad (4.17)$$

$$P^{Next} = P_{index_{Next}}^{explorable} \qquad (4.18)$$

### 4.2.4 Estimated Energy Cost Function

As discussed above, an estimated energy cost function $E(P_a, P_b)$ calculates the estimated energy cost from $P_a$ to $P_b$. This function is used to find feasible node set $S_{feasible}$ and explorable node set $S_{explorable}$. In this section, the equations for the cost function are derived from the robot positioning-controlled system's response. Also, trade-offs between energy cost and settling time of moving are discussed.

The energy of moving the camera between two locations from $P_{initial}$ to $P_{final}$, results from the energy cost of DC motors in 6 DOFs robot arm's each joint rotating from joint angles $q_{initial}$ to $q_{final}$. Therefore, $E(P_{initial}, P_{final})$ can be described as the sum of the time integration of rotational power in each joint. That is:

$$E(P_{initial}, P_{final}) = \sum_{i=1}^{6} \int_{t_0^i}^{t_f^i} V^i \cdot I^i \cdot dt \qquad (4.19)$$

where $V^i$ is the voltage and $I^i$ is the current in the DC motor's circuit of the $i^{th}$ joint. $t_0^i$ and $t_f^i$ are initial time and final time of the $i^{th}$ joint moving from its initial angle $q_{initial}^i$ to its final angle $q_{final}^i$.

The dynamic model of a 6dofs revolutionary robotic manipulator and DC motors can be expressed as in Equation (3.110) and we can rewrite in the following:

$$\frac{1}{r_k}J_{m_k}\ddot{q}_k + \sum_{j=1}^{6}d_{j,k}\ddot{q}_j + \sum_{i,j}^{6}c_{i,j,k}(q)\dot{q}_i\dot{q}_j + \frac{1}{r_k}B\dot{q}_k + \Phi_k(q) = \frac{K_m}{r_k R}V^i \qquad (4.20)$$

$$J_m = J_a + J_g, \qquad B = B_m + \frac{K_m^2}{R} \qquad (4.21)$$

$$c_{i,j,k} = \frac{1}{2}\left(\frac{\delta d_{k,j}}{\delta q_i} + \frac{\delta d_{k,i}}{\delta q_j} - \frac{\delta d_{i,j}}{\delta q_k}\right), \quad \Phi_k = \frac{\delta V}{\delta q_k} \qquad (4.22)$$

$$i, j, k \in (1, 2, 3, 4, 5, 6) \qquad (4.23)$$

where Equations (4.20) − (4.23) express the $k^{th}$ joint dynamics equation. $q_i/q_j/q_k$ is the joint revolute variable. $J_a$ and $J_g$ represent the moment of inertia of the motor, and the gear of the model. $r_k$ is gear ratio at $k^{th}$ joint, and $B_m$ is the damping effect of the gear. $d_{i,j}$ represents the entry of inertial matrix of the robot manipulator at $i^{th}$ row and $j^{th}$ column. $c_{i,j,k}$ is Christoffel symbols and for a fixed $k$, with $c_{i,j,k} = c_{j,i,k}$. And $\Phi_k$ is the derivative of potential energy $V$ with respective to $k^{th}$ joint variance. $K_m$ is the torque constant in $N - m/amp$, and $R$ is Armature Resistance.

Take $u^i$ as the actuator input to the dynamic system (4.20) measured at $i^{th}$ joint from a designed controller. Therefore, $u^i$ equals to the right-side of Equation (4.20). That is:

$$V^i = \frac{r_k R}{K_m}u^i \qquad (4.24)$$

From Equation (3.104), the expression of current measured at $i^{th}$ joint in the Laplace domain is:

$$(Ls + R)I^i(s) = V^i(s) - \frac{K_m}{r}sq^i(s) \qquad (4.25)$$

$$\text{Then: } I^i(s) = \frac{1}{(Ls+R)}V^i(s) - \frac{K_m}{r}\frac{s}{(Ls+R)}q^i(s) \qquad (4.26)$$

$$\text{Take the inverse Laplace: } I^i(t) = \frac{1}{L}e^{-\frac{R}{L}t} * V^i(t) - \frac{K_m}{r}\left(\frac{1}{L} - \frac{R}{L^2}e^{-\frac{R}{L}t}\right) * q^i(t) \qquad (4.27)$$

where $L$ is armature inductance, $K_m$ is the torque constant, and $*$ is the convolutional multiplication. Therefore, the instant current in the time domain $I^i(t)$ is a function of the instant voltage $V^i(t)$ and the instant angle $q^i(t)$.

Various controller designs, such as PID controller [72], and Youla controller [73], of six DoFs revolute robotic manipulators have been well developed in many papers. In Chapter 5 of this dissertation, we use a Youla controller design [73] with feedback linearization (Figure 4.4.) as the position controller of the robot manipulator that holds the camera.

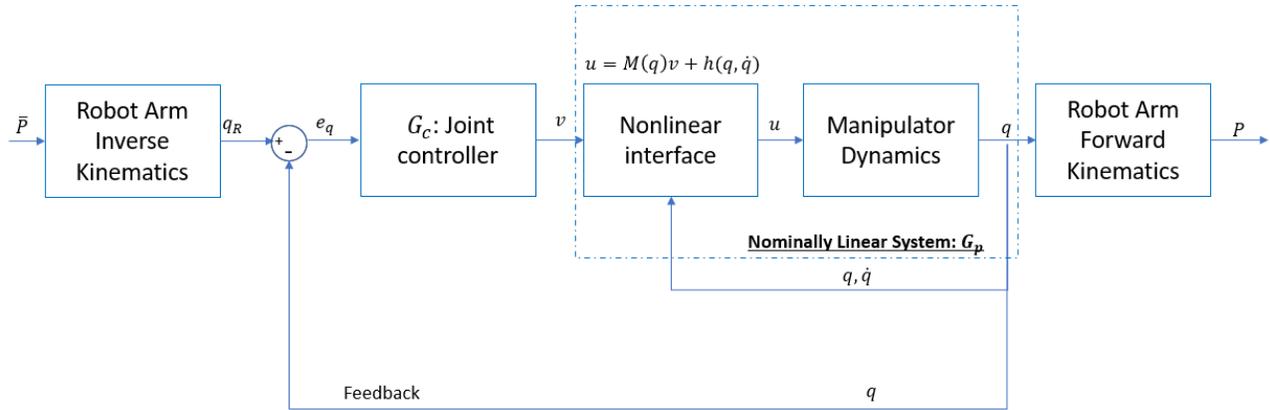

**Figure 4.4:** The block Diagram of feedback linearization Youla control design used for the joint control loop.

From Figure 4.4., the actuator input $u^i$ can be derived from a virtual input $v^i$ so that:

$$u^i = M(q^i(t))v^i + H(q^i(t), \dot{q}^i(t)) \tag{4.28}$$

where M, and H are nonlinear functions of $q^i(t)$, and $\dot{q}^i(t)$, the first derivative of $q^i(t)$.

Without showing the controller development in this chapter (which will be discussed in details in Chapter 5), the 1 DoF controller transfer function $G_c(s)$ and 1 DoF nominal plant $G_p(s)$ with $v^i$ as the virtual actuator input in the feedback linearization are expressed as follows:

$$G_c(s) = \frac{3\tau_{in}^2 s + 1}{\tau_{in}^3 s + 3\tau_{in}^2} \tag{4.29}$$

$$G_p(s) = \frac{1}{s^2} \tag{4.30}$$

where $\tau_{in}$ is a constant parameter in the controller design.

Then the following transfer functions can be calculated as:

$$\frac{v^i(s)}{q^i_{final}(s)} = \frac{G_c}{1 + G_c G_p} \tag{4.31}$$

$$\frac{q^i(s)}{q^i_{final}(s)} = \frac{G_c G_p}{1 + G_c G_p} \tag{4.32}$$

$$\frac{\dot{q}^i(s)}{q^i{}_{final}(s)} = \frac{sG_cG_p}{1+G_cG_p} \tag{4.33}$$

By taking inverse Laplace transform, and $q^i{}_{final}$ as a step input, the following Equations (4.31) – (4.33) in the time domain are:

$$v^i(t) = (q^i{}_{final} - q^i{}_{intial})(\frac{1}{\tau_{in}{}^4}t^2 e^{-\frac{t}{\tau_{in}}} + \frac{-5}{\tau_{in}{}^3}te^{-\frac{t}{\tau_{in}}} + \frac{3}{\tau_{in}{}^2}e^{-\frac{t}{\tau_{in}}}) \tag{4.34}$$

$$q^i(t) = (q^i{}_{final} - q^i{}_{intial})(\frac{1}{\tau_{in}{}^2}t^2 e^{-\frac{t}{\tau_{in}}} - \frac{1}{\tau_{in}}te^{-\frac{t}{\tau_{in}}} - e^{-\frac{t}{\tau_{in}}} + 1) + q^i{}_{intial} \tag{4.35}$$

$$\dot{q}^i(t) = (q^i{}_{final} - q^i{}_{intial})(\frac{3}{\tau_{in}{}^2}te^{-\frac{t}{\tau_{in}}} - \frac{1}{\tau_{in}{}^3}t^2 e^{-\frac{t}{\tau_{in}}}) \tag{4.36}$$

With expressions of $v^i(t), q^i(t)$, and $\dot{q}^i(t)$, Combining Equations (4.24) and (4.28), we can generate a nonlinear function $F$ so that:

$$V^i(t) = F(q^i{}_{intial}, q^i{}_{final}, t) \tag{4.37}$$

and combine Equations (4.27) and (4.37) we can generate a nonlinear function $G$ so that:

$$I^i(t) = G(q^i{}_{intial}, q^i{}_{final}, t) \tag{4.38}$$

where F, and G are nonlinear functions of $q^i{}_{initial}, q^i{}_{final}$, and $t$.

It has been shown so far that the estimated energy cost $E(P_{initial}, P_{final})$ from location $P_{initial}$ to $P_{final}$ is a function of $q_{initial}$ and $q_{final}$.

$q_{initial}$ and $q_{final}$ are derived from the inverse kinematics process (Shown in Equations (3.60) - (3.66)), that is:

$$q_{initial} = inversekinematics(P_{initial}) \tag{4.39}$$

$$q_{initial} = inversekinematics(P_{initial}) \tag{4.40}$$

And $q_{initial} = [q^i{}_{initial} | i \in (1,2,3,4,5,6)], q_{final} = [q^i{}_{final} | i \in (1,2,3,4,5,6)]$ \hfill (4.41)

Development of Equations (4.34) – (4.36) assumes that the target angle of each joint $q^i{}_{final}$ is a step input. A more realistic assumption is to set $q^i{}_{final}$ as a delayed input.

$$q^i{}_{final}(t) = (1 - e^{-\frac{t}{\tau_{delay}}})q^i{}_{final} \tag{4.42}$$

where $\tau_{delay}$ is a time constant that measures the time delay of the target in the real positioning control. $\tau_{delay} \geq 0$, and when $\tau_{delay} = 0$, it indicates no delay exists in the input.

The Laplace form of Equation (4.42) is:

$$q^i{}_{final}(s) = (\frac{1}{s} - \frac{1}{s + \frac{1}{\tau_{delay}}})q^i{}_{final} \tag{4.43}$$

With new expression of $q^i{}_{final}(s)$, Equation (4.34) – (4.36) can be developed as:

$$v^i(t) = (q^i{}_{final} - q^i{}_{intial})(\frac{A_v}{\tau_{in}{}^3}t^2 e^{-\frac{t}{\tau_{in}}} + B_v\frac{B_v}{\tau_{in}{}^2}te^{-\frac{t}{\tau_{in}}} + \frac{C_v}{\tau_{in}}e^{-\frac{t}{\tau_{in}}}$$
$$+ \frac{D_v}{\tau_{delay}}e^{-\frac{t}{\tau_{delay}}}) \tag{4.44}$$

$$A_v = \frac{1}{\tau_{in}} - \frac{-\tau_{delay}}{\tau_{in}(\tau_{in}-\tau_{delay})} \qquad B_v = \frac{-5}{\tau_{in}} - \frac{7\tau_{in}\tau_{delay}-5\tau_{delay}{}^2}{\tau_{in}(\tau_{in}-\tau_{delay})^2}$$

$$C_v = \frac{3}{\tau_{in}} - \frac{\tau_{delay}(-8\tau_{in}{}^2+9\tau_{in}\tau_{delay}-3\tau_{delay}{}^2)}{\tau_{in}(\tau_{in}-\tau_{delay})^3} \qquad D_v = -\frac{3\tau_{in}\tau_{delay}-\tau_{delay}{}^2}{(\tau_{in}-\tau_{delay})^3}$$

$$q^i(t) = (q^i{}_{final} - q^i{}_{intial})(\frac{A_Q}{\tau_{in}{}^3}t^2 e^{-\frac{t}{\tau_{in}}} + \frac{B_Q}{\tau_{in}{}^2}te^{-\frac{t}{\tau_{in}}} + \tag{4.45}$$
$$\frac{C_Q}{\tau_{in}}e^{-\frac{t}{\tau_{in}}} + \frac{D_Q}{\tau_{delay}}e^{-\frac{t}{\tau_{delay}}}+1) + q^i{}_{intial}$$

$$A_Q = \tau_{in} + \frac{\tau_{in}\tau_{delay}}{(\tau_{in}-\tau_{delay})} \qquad B_Q = -\tau_{in} - \frac{3\tau_{in}{}^2\tau_{delay}-\tau_{in}\tau_{delay}{}^2}{(\tau_{in}-\tau_{delay})^2}$$

$$C_Q = -\tau_{in} - \frac{-3\tau_{in}{}^2\tau_{delay}{}^2+\tau_{in}\tau_{delay}{}^3}{(\tau_{in}-\tau_{delay})^3} \qquad D_Q = -\frac{3\tau_{in}\tau_{delay}{}^3-\tau_{delay}{}^4}{(\tau_{in}-\tau_{delay})^3}$$

$$\dot{q}^i(t) = (q^i{}_{final} - q^i{}_{intial})(\frac{A_{\dot{Q}}}{\tau_{in}{}^4}t^2 e^{-\frac{t}{\tau_{in}}} + \frac{B_{\dot{Q}}}{\tau_{in}{}^3}te^{-\frac{t}{\tau_{in}}} + \frac{C_{\dot{Q}}}{\tau_{in}{}^2}e^{-\frac{t}{\tau_{in}}} \tag{4.46}$$
$$+ \frac{D_{\dot{Q}}}{\tau_{delay}{}^2}e^{-\frac{t}{\tau_{delay}}})$$

$$A_Q = \tau_{in} + \frac{\tau_{in}\tau_{delay}}{(\tau_{in}-\tau_{delay})} \qquad B_Q = -\tau_{in} - \frac{3\tau_{in}{}^2\tau_{delay}-\tau_{in}\tau_{delay}{}^2}{(\tau_{in}-\tau_{delay})^2}$$

$$C_Q = -\tau_{in} - \frac{-3\tau_{in}{}^2\tau_{delay}{}^2+\tau_{in}\tau_{delay}{}^3}{(\tau_{in}-\tau_{delay})^3} \qquad D_Q = -\frac{3\tau_{in}\tau_{delay}{}^3-\tau_{delay}{}^4}{(\tau_{in}-\tau_{delay})^3}$$

A simulation scenario is set up to calculate how estimated energy cost changes with varying $\tau_{delay}$. Set $P_{initial} = [-0.30, 0.05, 1.20]$, and $P_{final} = [-0.45, 0.45, 1.20]$ and use the Equations (4.39)-(4.41), (4.44)-(4.46). Figure 4.5 shows the response of one angle $q^1(t)$ with varying $\tau_{delay}$ and Table 4.1. presents the estimated energy cost: $E(P_{initial}, P_{final})$ and settling time: $t_s$ with varying $\tau_{delay}$.

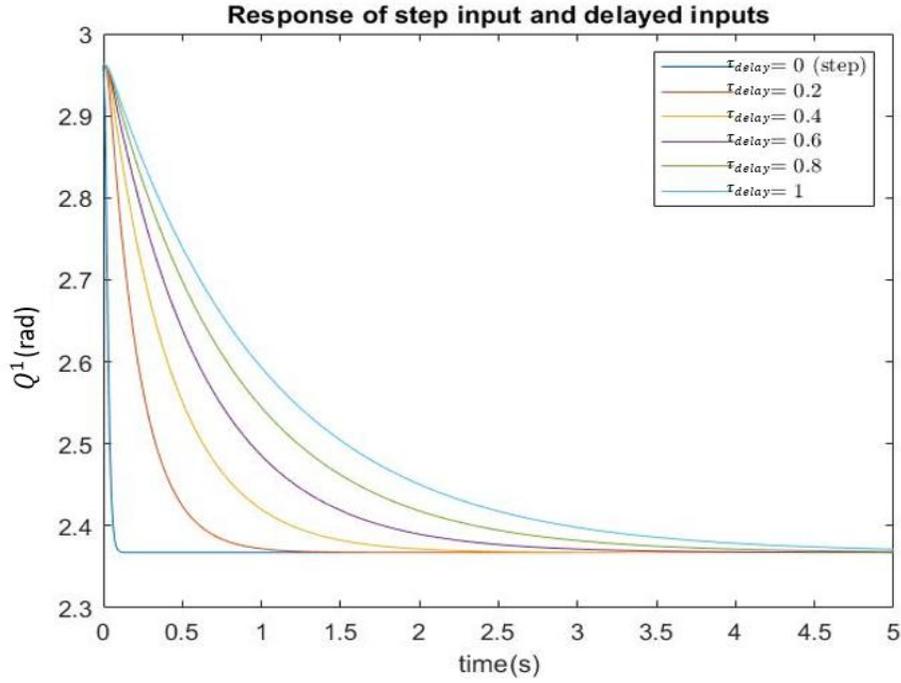

**Figure 4.5**: Response of $q^1(t)$ with varying $\tau_{delay}$.

**Table 4.1**: Estimated Energy Cost and Settling Time with Varying Delay Constant.

| $\tau_{delay}$ [s] | $E(P_{initial}, P_{final})$ [ws] | $t_s$ [s] |
|---|---|---|
| 0 | 10.895 | 0.07 |
| 0.2 | 0.102 | 0.73 |
| 0.4 | 0.030 | 1.42 |
| 0.6 | 0.016 | 2.11 |
| 0.8 | 0.011 | 2.80 |
| 1 | 0.009 | 3.49 |

Note: $E(P_{initial}, P_{final})$ is measured as [watts seconds]

Table 4.1. shows a tradeoff between the estimated energy cost and settling time; reducing the energy cost of moving the camera inevitably increases the response time. This finding matches the results in Figure 4.5.

Systematic delays are inevitable in controlled systems' design. Delays are incurred in many sources such as time required for sensors to detect and process changes, for actuators response to control signals, and for controllers to process and compute signals. In a real manufacturing environment, larger delays cause slower motion of the camera when searching the area but slower response results in less energy consumption from actuators in the manipulator.

If the delay of input is given or can be measured, the estimated energy cost can be calculated through the process in this section. However, if the value is unknown, the delay constant must be chosen and decided based on the following criteria:

- Select small $\tau_{delay}$ for a conservative algorithm that searches a small area but ensures it ends up with the minimum that has been explored.
- Select large $\tau_{delay}$ for an aggressive algorithm that searches a large area but risks not ending up at the minimum that has been explored.

In this section, the estimated energy cost function is well developed. However, the estimation function is developed from an ideal camera movement. The accuracy of the estimation can be negatively influenced by some unmodeled uncertainties such as backlash from gears in robot manipulators, unmodeled compliance components from joint vibration, etc. To tackle the problem, we can derive an online updated model of the estimation function by comparing estimated voltage and current and real-time measurements. The parameter values of motors and gears used in the simulations are summarized in Appendix A Table A4.

## 4.3 Simulation Setup

The developed algorithm has been simulated on the muti-robotic system (Figure 2.1.).

We first define the camera's operational space in this application and then provide the simulation results of a made-up scenario next.

In the above discussion, a camera operational space is gridded with nodes and each node is a potential candidate of the camera's optimal position. The camera operational space is defined by three geometric constraints below:

1. The camera can only be allowed to move to locations where fiducial markers that are attached to the tool are recognizable on the image frame. Therefore, the fiducial markers are in the angle of view of the camera, and the distance between the markers and the camera center is within a threshold.
2. The camera can only be allowed to move to locations within the reach of the robot arm.
3. The camera can only be allowed move to locations where the visual system and the tool manipulation system do not physically interfere with each other.

From the specifications of the stereo camera and the robot arm (Appendix Table A1 and A3), the geometry and dimensions of the camera operational space are analyzed in the following part of the subsections.

### 4.3.1 Reachable and Dexterous Workspace of Two-Hybrid Systems

In the spherical wrist model of the robot arm, three rotational axes represent the pitch, roll, and yaw of the end-effector independently. As shown in Figure 4.6., those three axes intersect at point $H$. Point $P$ is the location of the end-effector and it's the place where the camera or tool is mounted. For the elbow manipulator mounted with the stereo camera, the camera is placed so that its optical center line coincides with axis C. The location of camera optical center C is determined and a good estimation of it can be found in this paper [74]. For now, assume the optical center is in the middle of $\overline{PJ}$ (set $\overline{PC} = 17\ mm$). For the elbow manipulator mounted with the tool, the screwdriver is attached at the point $P$. Point J indicates the far end of the object (tool or camera) attached to the end effector. Point H is kept stationary no matter which axis rotates. Therefore, the task of positioning and rotation is decoupled in the spherical wrist robot model.

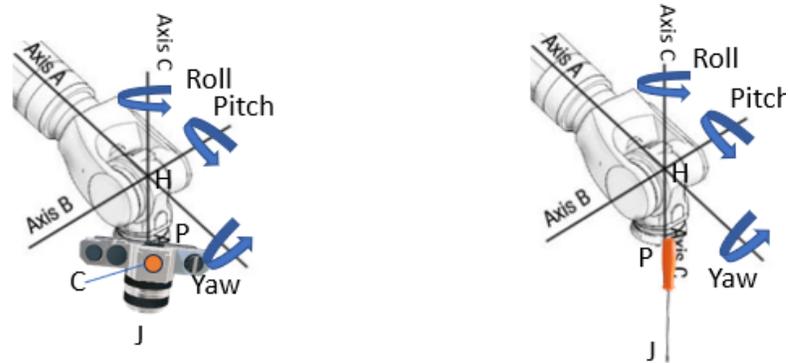

**Figure 4.6:** Spherical Wrist Model.

The idea of reachable and dexterous workspaces of an ideal elbow manipulator has been introduced in the paper [75]. An ideal elbow manipulator is a manipulator whose angles of rotation are free to move in the whole operational range $[0, 2\pi]$. However, a realistic elbow manipulator is limited to moving its joints within certain ranges of angle.

The workspace of an ideal elbow manipulator (Figure 4.7.) is a sphere centered at the joint 1 of the manipulator, denoted as point $o$. The reachable workspace concerning the center $o$ of a manipulator is the aggregate of all possible locations of the point $J$ attached with the end-effector and is denoted as $W_o(J)$. The dexterous workspace a manipulator's center $o$ is the aggregate of all possible locations that point $J$ can reach all possible orientations of the end-effector and is denoted as $W_o^d(J)$. The reachable workspace of $H$ is denoted as $W_o(H)$.

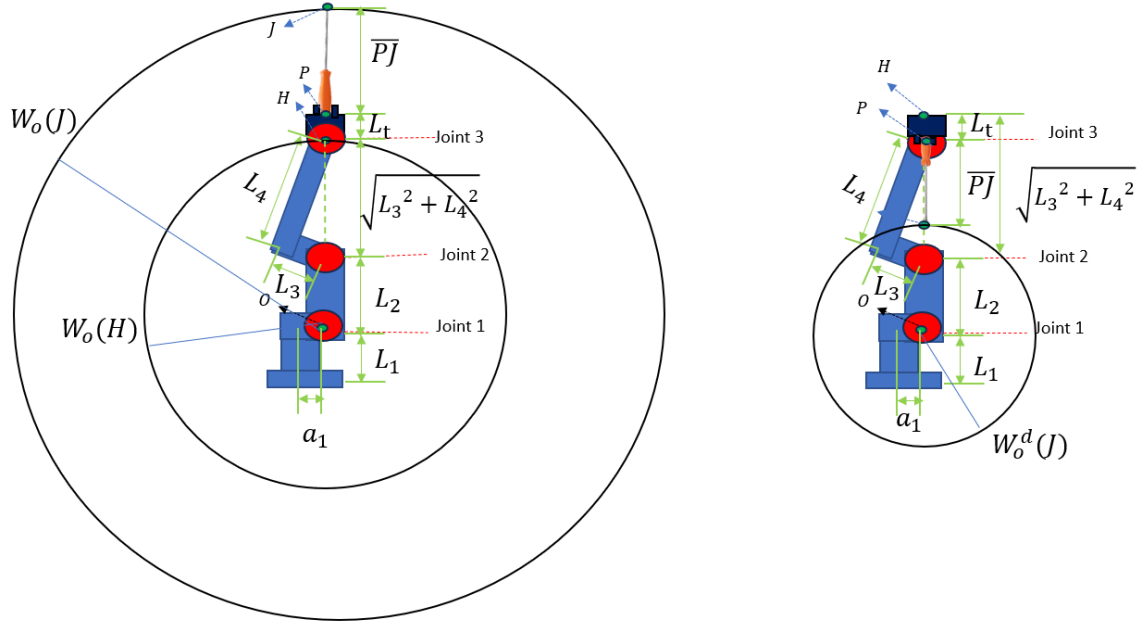

**Figure 4.7:** Workspace of an Ideal Elbow Manipulator.

It can be shown in the figure:

$$W_o^d(J) \subseteq W_o(H) \subseteq W_o(J). \tag{4.47}$$

For an ideal elbow manipulator, the radiuses of workspaces are expressed as below:

$$\text{Radius of } W_o(H) = L_2 + \sqrt{L_3^2 + L_4^2} \tag{4.48}$$

$$\text{Radius of } W_o(J) = L_2 + \sqrt{L_3^2 + L_4^2} + L_t + \overline{PJ} \tag{4.49}$$

$$\text{Radius of } W_o^d(J) = L_2 + \sqrt{L_3^2 + L_4^2} - L_t - \overline{PJ} \tag{4.50}$$

where $L_2, L_3,$ and $L_4$ are links' length of the manipulator. $L_t$ is the length of the end-effector and $\overline{PJ}$ is the length of the object that mounts on the end-effector.

Because the camera and the tool must be able to rotate in all 3 DoFs when they are at the target position. So, the dexterous workspace is used as the working space of the visual system and the tool manipulation system. Dexterous workspace of the optical center is used for the visual system and Dexterous workspace of the far endpoint of the tool is used for the tool manipulator system.

$$\text{Radius of } W_o^d(Visual) = r_V = L_2 + \sqrt{L_3^2 + L_4^2} - L_t - \overline{PC}_{camera} \qquad (4.51)$$

$$= 1.716 \, m \text{ (From Specification Appendix Table A1)}$$

$$\text{Radius of } W_o^d(Tool) = r_T = L_2 + \sqrt{L_3^2 + L_4^2} - L_t - \overline{PJ}_{tool} \qquad (4.52)$$

$$= 1.606 \, m \text{ (From Specification Appendix Table A1)}$$

Figure 4.8. shows the workspace setup for the hybrid system. The dexterous space of robot arm with camera is a sphere with radius of $r_V$ and sphere center is $O_V$ and the dextrous space of the robot arm with tools is a sphere the radius is $r_T$ and center of this sphere is $O_T$. Two systems intersect with the ground with an angle $\theta_1$ and $\theta_2$.

$$\theta_1 = \arcsin\left(\frac{L_1}{r_V}\right) = 16.77° \qquad (4.53)$$

$$\theta_2 = \arcsin\left(\frac{L_1}{r_T}\right) = 17.95° \qquad (4.54)$$

To avoid interference of the visual and the tool manipulation systems (as the third geometric constraint of the operational space), the reachable workspaces (the maximum reach) of two systems should have no overlap. The reachable workspaces of two systems are calculated from Equation (4.48) is:

$$\text{Radius of } W_o(Visual) = r_{Vr} = L_2 + \sqrt{L_3^2 + L_4^2} + L_t + \overline{PJ}_{camera} \qquad (4.55)$$

$$= 2.044 \, m \text{ (From Specification Appendix Table A1)}$$

$$\text{Radius of } W_o(Tool) = r_{Tr} = L_2 + \sqrt{L_3^2 + L_4^2} + L_t + \overline{PJ}_{tool} \qquad (4.56)$$

$$= 2.138 \, m \text{ (From Specification Appendix Table A1)}$$

Let $L_{VT}$ is the distance between $O_V$ and $O_T$. The following relationship must be satisfied:

$$L_{VT} \geq r_{Vr} + r_{Tr} = 4.182 \, m \qquad (4.57)$$

The visual system is able to detect the tool when it gets close to the target pose (Just above the bolt). Also reference points are placed near the tool's target pose. Assume a marker is placed at where the bolt is, and this marker represents as an approximated location of all interest points and reference points. Therefore, the marker should be within the dexterous space of the tool manipulation system. Set up a coordinate system with $\bar{O}_V$ (the projection of $O_V$ on the ground with $a_1$ offset) as the origin so that the origin is located at the intersection of the rotational axis of base link and the ground. The projection of $O_T$ on the ground with $a_1$ offset is $\bar{O}_T$. Assume the marker is placed on the line $\overline{\bar{O}_V \bar{O}_T}$. The distance m from marker M to $\bar{O}_V$ should be:

$$L_M \geq a_1 + L_{VT} - r_T \cos(\theta_2) \geq 2.829 \ m \tag{4.58}$$

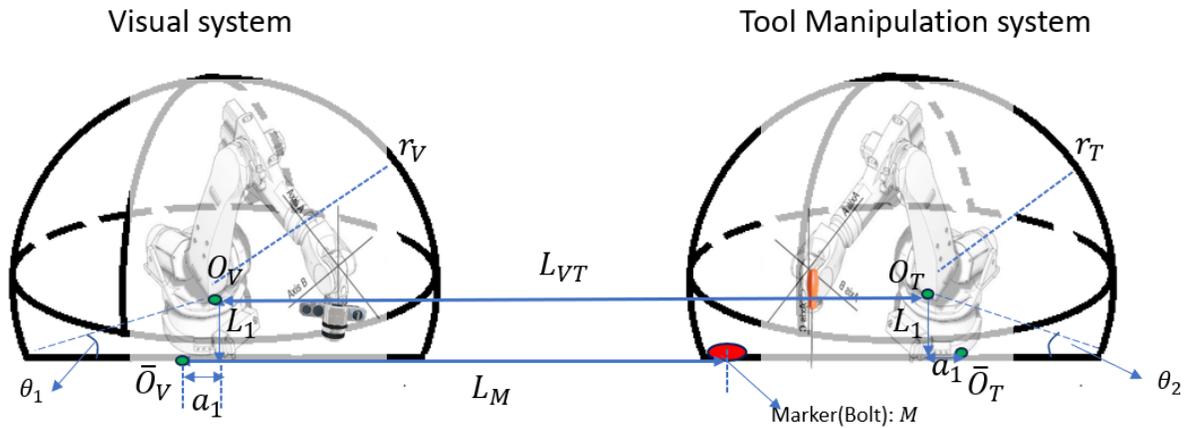

**Figure 4.8**: The workspace configuration of the multi-robotic system.

### 4.3.2 Detectable Space for the Stereo Camera

From Appendix A Table A3, the angle of view in width and height for both lens in stereo camera is $\alpha = 86.05°$, and $\beta = 55.35°$. The detectable space for each lens can be modeled as the inner area of a cone with angle $\alpha$ and $\beta$. And the overlapped space is the detectable space for the stereo camera. The model is shown below (Figure 4.9). Point C is the center of the baseline whose length is denoted as $b$ or center of the camera system. The overlap area is also a cone with angles of vertex $\alpha$ and $\beta$ with the offset $d$ from baseline. Point Q is the vertex of the cone.

$$d = \frac{b/2}{\tan(\alpha/2)} = 0.064m \tag{4.59}$$

There is an upper bound of the distance between the object to the camera center; if the object is too far away from the camera center, the dimensions of projected images are too small to be measured. Suppose to have a clear image of the fiducial markers (circle shapes), it is required that the diameter of the projected image should takes at least 5-pixel number in the image frame:

$$N_H \geq 5 \text{ resolution} \tag{4.60}$$

where $N_H$ is the pixel number in the image frame.

Select the Zed camera's mode so that its resolution is 1920*1082 and the image sensor size is 5.23mm × 2.94mm. Then numbers resolution in unit length is:

$$n = 367 \text{ resolution/mm} \tag{4.61}$$

Then the range of markers' diameter on the image frame is:

$$d_H = \frac{N_H}{n} \geq 0.0136 \text{ mm} \tag{4.62}$$

Also assume in the inertial frame, the diameter of attached fiducial marks is:

$$D_H = 12 \text{ mm} \tag{4.63}$$

Then from pinhole model of the camera, the range of distance between markers and camera center Z is:

$$Z = \frac{f_u D_H}{d_H} \leq 2.47 \text{ m} = Z_{max} \tag{4.64}$$

where $Z_{max}$ is the maximum depth of camera to detect the markers. This parameter defines the height of the cone in Figure 4.9. The detectable space (abiding the second geometric constraint) forms an elliptic cone with different angles of vertex.

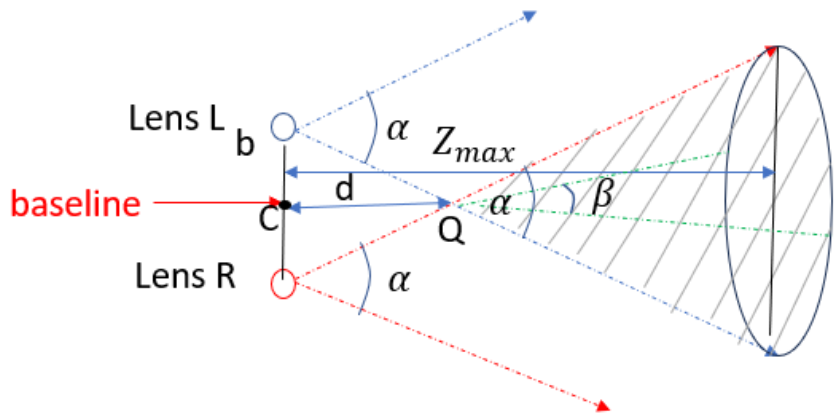

**Figure 4.9:** The Detectable Space of the Stereo Camera.

## 4.3.3 Camera Operational Space Development

In Figure 4.10., set the camera lens so that it is always parallel to the face of the marker and the baseline is parallel to $\bar{X}_V$ axis. In this case, lens is kept facing downward when detecting the marker on the ground. As discussed in the above section, the detectable space of the camera is modeled as a cone. This cone intersects with the horizontal plane and forms an oval. To have the camera detect the marker, the marker must be contained inside the oval.

An inertia coordinate system $\bar{O}_V \bar{X}_V \bar{Y}_V \bar{Z}_V$ is set up with its origin seated at the bottom center of the visual system. Make point $C$ the location of the camera center and its coordinates in the inertial frame are $(X_C^V, Y_C^V, Z_C^V)$. $Q$ is the vertex of the detectable elliptic cone, and $Q'$ is the projection of $Q$ on the horizontal plane. $M$ is the marker, and d is the distance between point $C$ and $Q$ as defined in Equation (4.59).

The elliptic cone of the camera's detectable space always intersects with the horizontal plane. Consider an extreme case when point $C$ is located at the apex of the visual system workspace as shown in Figure 4.10. Then the $Z_C^V$ coordinate of $C$ at the inertial frame is:

$$Z_C^V\{max\} = L_1 + r_V = 2.211\ m \leq Z_{max} \tag{4.65}$$

where $Z_{max}$ defined above is the maximum depth of camera to detect the markers. Therefore, the cone intersects with the horizontal plane when the camera center $C$ is placed at any location of the workspace's contour.

From expression of Cartesian coordinates in the inertia frame:

$$a_{Q'} = (Z_C^V - d)\tan(\tfrac{\alpha}{2}),\ b_{Q'} = (Z_C^V - d)\tan(\tfrac{\beta}{2}) \tag{4.66}$$

where $d$ is the same offset in Equation (4.59). $a_{Q'}$ and $b_{Q'}$ are major radius and minor radius of the oval. Also coordinates of points $Q'$ and $M$ are $(X_C^V, Y_C^V, 0)$ and $(L_M, 0, 0)$. $Q'$ is the center of the oval. To ensure the marker $M$ is inside the projected oval, it is required that:

$$\frac{(X_C^V - L_M)^2}{a_{Q'}^2} + \frac{(Y_C^V)^2}{b_{Q'}^2} \leq 1 \tag{4.67}$$

Plug Equation (4.66) into Equation (4.67):

$$\frac{(X_C^V - L_M)^2}{\tan(\tfrac{\alpha}{2})^2} + \frac{(Y_C^V)^2}{\tan(\tfrac{\beta}{2})^2} \leq (Z_C^V - d)^2 \tag{4.68}$$

**Figure 4.10:** The Detectable Elliptic Cone in the Inertial Visual System Frame.

Inequality (4.68) is an exact mathematical expression of all points $C(X_C^V, Y_C^V, Z_C^V)$ within a cone whose opening is in the positive Z direction with its vertex at $(L_M, 0, d)$ and the opening parameters as $\tan(\frac{\alpha}{2})$=0.934 and $\tan(\frac{\beta}{2})$=0.524. Therefore, this inequality defines the second geometric constraint of the camera center. It forms an elliptical cone with its vertex located in Figure 4.11. at point E, which is the offset point of the marker $M$ from the ground. As $C(X_C^V, Y_C^V, Z_C^V)$ should also be within the sphere of the workspace (abiding the first geometric constraint), the camera operational space is presented as the overlap between the cone and the sphere (the shaded area in Figure 4.11.).

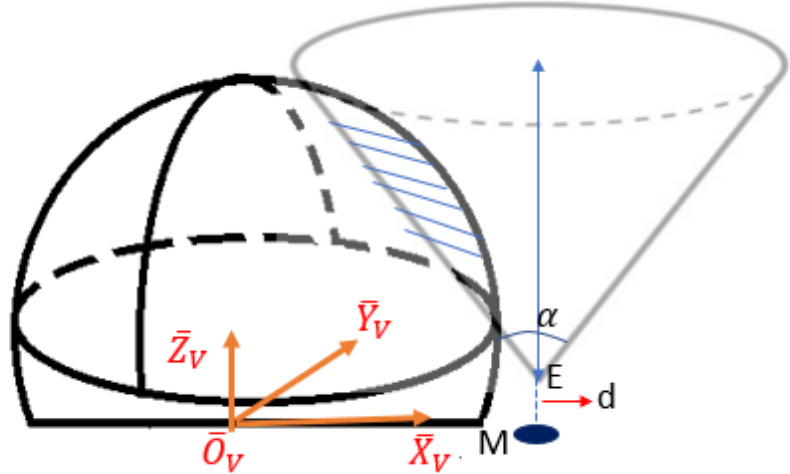

**Figure 4.11**: Illustration of Camera Operational Space (the shaded area).

From Figure 4.11., the marker should not be too far from the visual system, otherwise there is no overlap region between the cone and the sphere. The largest value of $L_M$, which is the distance from the marker to the coordinate center, occurs when the cone is tangent to the sphere as shown in Figure 4.12. To find the limit of $L_M$, draw a horizontal line from point $O_V$ so that it intersects the surface of the cone at point A and the center line of the cone at point B as shown in Figure 4.12. Let $\overline{O_V A}$ and $\overline{AB}$ are line segments' length.

Therefore, from the geometry, the upper limit of $L_M$ is:

$$L_M < \overline{O_V A} + \overline{AB} + a_1 = \frac{r_V}{\cos\left(\frac{\alpha}{2}\right)} + (L_1 - d)\tan\left(\frac{\alpha}{2}\right) + a_1 \tag{4.69}$$

Combining with the lower bound of $L_M$ in Equation (4.58), the marker M should be placed in the system so that:

$$2.829\ m \leq L_{VT} - r_T \cos(\theta_2) \leq L_M < \frac{r_V}{\cos\left(\frac{\alpha}{2}\right)} + (L_1 - d)\tan\left(\frac{\alpha}{2}\right) + a_1 = 2.925\ m \tag{4.70}$$

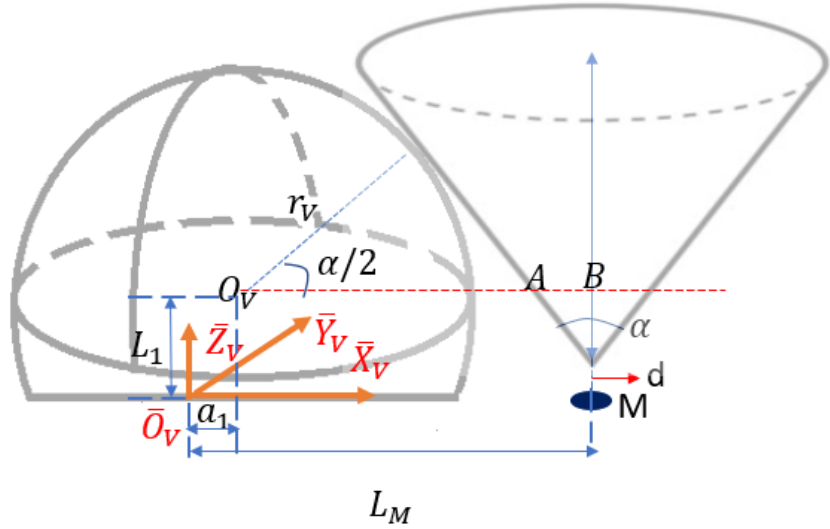

**Figure 4.12:** An Extreme Case of Marker Position.

### 4.3.4 Mathematical Expression for Node Coordinates within the Camera Operational Space

From Equation (4.70), the closer the marker's distance $L_M$ to its lower bound, the larger the operational space is. In general, larger operational space is preferred, because the camera has a larger space for searching the optimal location. Set $L_M$ equals to its lower bound:

$$L_M = 2.83m \tag{4.71}$$

Then we can obtain the cross-section area at $\bar{X}_V - \bar{Z}_V$ plane (Figure 4.13.).
The function of semi-circle and the line EFG is:

$$(\bar{Z}_V - L_1)^2 + (\bar{X}_V - a_1)^2 = r_V^2 \tag{4.72}$$

$$\bar{X}_V = -tan(\tfrac{\alpha}{2})(\bar{Z}_V - d) + L_M \tag{4.73}$$

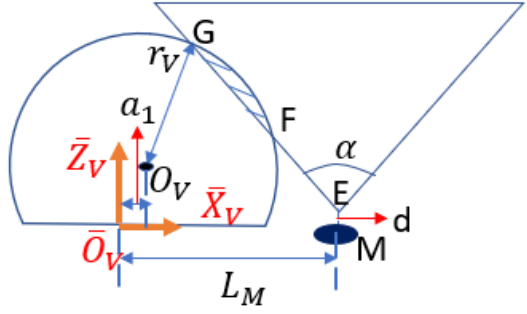

**Figure 4.13:** The Camera Operational Space Cross-section at $\bar{X}_V - \bar{Z}_V$ plane.

Combine Equations (4.72) and (4.73) and solve:

$$(1 + \tan(\tfrac{\alpha}{2})^2)\bar{Z}_V^2 - 2(\tan(\tfrac{\alpha}{2})^2 d + \tan\left(\tfrac{\alpha}{2}\right)(L_M - a_1) + L_1)\bar{Z}_V + \tan\left(\tfrac{\alpha}{2}\right)^2 d^2 + \qquad (4.74)$$
$$2\tan\left(\tfrac{\alpha}{2}\right)(L_M - a_1)d + (L_M - a_1)^2 - r_V^2 + L_1^2 = 0$$

Solve Equation (4.74) by plugging values of Zed 2 stereo camera parameters $(\alpha, d)$ in Appendix A Table A3, ABB IRB 4600 robot manipulator parameters $(a_1, L_1)$ in Appendix A Table A1, and geometrical constraints in camera operational space $(L_M, r_V)$:

$$1.872\,\bar{Z}_V^2 - 6.061\bar{Z}_V + 4.670 = 0 \qquad (4.75)$$

$$\text{Solve: } \bar{Z}_{V\,max}, \bar{Z}_{V\,min} = 1.972\,m,\ 1.265\,m \qquad (4.76)$$

$\bar{Z}_{V\,max}$ and $\bar{Z}_{V\,min}$ are the largest and smallest $\bar{Z}_V$ coordinates of the camera center $C$ that is within the camera operation space.

Take a value of $Z_C^V$ ($\bar{Z}_V$ coordinates of the camera center $C$) in the range $(\bar{Z}_{V\,min}, \bar{Z}_{V\,max}) = (1.265\,m, 1.972\,m)$. Set a plane $\bar{Z}_V = Z_C^V$, and that plane intersects the camera operational space and forms a shade area, which is the overlap between the circle and the oval, as shown in Figure 4.11. The inequality governing points $C$ within sphere at a specific $Z_C^V$ is:

$$(\bar{X}_V - a_1)^2 + \bar{Y}_V^2 \leq r_V^2 - (Z_C^V - L_1)^2 \qquad (4.77)$$

The inequality governing points $C$ within cone at $Z_C^V$ is:

$$\frac{(\bar{X}_V - L_M)^2}{\tan(\frac{\alpha}{2})^2} + \frac{(\bar{Y}_V)^2}{\tan(\frac{\beta}{2})^2} \leq (Z_C^V - d)^2 \tag{4.78}$$

By plotting those inequalities in Figure 4.14., the shaded area is where $X_C^V$ and $Y_C^V$ ($\bar{X}_V$ and $\bar{Y}_V$ coordinates of the camera center $C$) should be located within. The intersection of the two curves occurs when equality holds for Equations (4.77) and (4.78), that is:

$$(\bar{X}_V - a_1)^2 + \bar{Y}_V^{\ 2} = r_V^2 - (Z_C^V - L_1)^2 \tag{4.79}$$

$$\frac{(\bar{X}_V - L_M)^2}{\tan(\frac{\alpha}{2})^2} + \frac{(\bar{Y}_V)^2}{\tan(\frac{\beta}{2})^2} = (Z_C^V - d)^2 \tag{4.80}$$

Solve by plugging numbers:

$$(\bar{Y}_{V\,max}, \bar{Y}_{V\,min}) = \pm 0.5 * \tag{4.81}$$

$$\sqrt{3.44 * Z_C^{V^2} - 2.0241 * Z_C^V - 29.802 + 2.451 * \sqrt{-29.475 * Z_C^{V^2} + 23.707 * Z_C^V + 137.28}}$$

$\bar{Y}_{V\,max}$ and $\bar{Y}_{V\,min}$ are the largest and the smallest $\bar{Y}_V$ coordinates of the camera center $C$ in the camera operation space with specific $Z_C^V$.

Take $Y_C^V$ in the range ($\bar{Y}_{V\,min}$, $\bar{Y}_{V\,max}$), then for specific $X_C^V$:

$$\bar{X}_{V\,max} = \sqrt{r_V^2 - (Z_C^V - L_1)^2 - Y_C^{V^2}} + a_1 \tag{4.82}$$

$$\bar{X}_{V\,min} = L_M - \sqrt{(Z_C^V - d)^2 \tan(\frac{\alpha}{2})^2 - Y_C^{V^2} \frac{\tan(\frac{\alpha}{2})^2}{\tan(\frac{\beta}{2})^2}} \tag{4.83}$$

Where $\bar{X}_{V\,max}$ and $\bar{X}_{V\,min}$ are the largest and the smallest $\bar{X}_V$ coordinates of the camera center $C$ in the camera operation space with specific $Z_C^V$ and $Y_C^V$.

**Figure 4.14:** The Camera Operational Space Cross-section at $\bar{X}_V - \bar{Y}_V$ plane.

### 4.3.5 Numerical Solution of the Ideal Camera Operation Space

With known parameters from the camera and robot arm specifications, all possible location of the camera center $C(X_C^V, Y_C^V, Z_C^V)$ inside the camera operational space can be derived from the following steps:

1. Calculate $\bar{Z}_{V\,max}$ and $\bar{Z}_{V\,min}$ using Equation (4.74), and mesh grid $Z_C^V$ in the range of $(\bar{Z}_{V\,min}, \bar{Z}_{V\,max})$.
2. Take a mesh value $Z_C^V$ and calculate $\bar{Y}_{V\,max}$ and $\bar{Y}_{V\,min}$ using Equations (4.79) and (4.80), and mesh grid $Y_C^V$ in the range of $(\bar{Y}_{V\,min}, \bar{Y}_{V\,max})$.
3. Take a mesh $Y_C^V$ and calculate $\bar{X}_{V\,max}$ and $\bar{X}_{V\,min}$ using Equations (4.82) and (4.83), and mesh grid $X_C^V$ in the range of $(\bar{X}_{V\,min}, \bar{X}_{V\,max})$.

Each above computed $C(X_C^V, Y_C^V, Z_C^V)$ is inside the camera operation space. And those three steps completely account for all geometric constraints defined above. The number of mesh points (nodes available for searching) depends on the mesh grid size.

### 4.3.6 The Camera Operational Space from the Realistic Robot Manipulator

The procedures described above for finding the camera operational space assume the use of ideal robotic manipulators, where each joint can move freely within the range $[0, 2\pi]$. However, for realistic robotic manipulators, joint motion is constrained within specific angular limits. For instance, the joint limits for the ABB IRB 4600 robot are provided in Appendix A Table A2.

Unlike an ideal manipulator, whose operational space forms a simple combination of a sphere and a cone, the geometric shape of a realistic manipulator's operational space is typically irregular. Directly identifying feasible nodes within this space using mathematical methods can be computationally expensive. A more efficient approach is to begin with the ideal operational space and remove nodes that violate the joint limits. These outlier nodes can be detected using the robot's inverse kinematics model. Figure 4.15. illustrates the reduction of the ideal camera operational space to a realistic operational space after removing infeasible nodes.

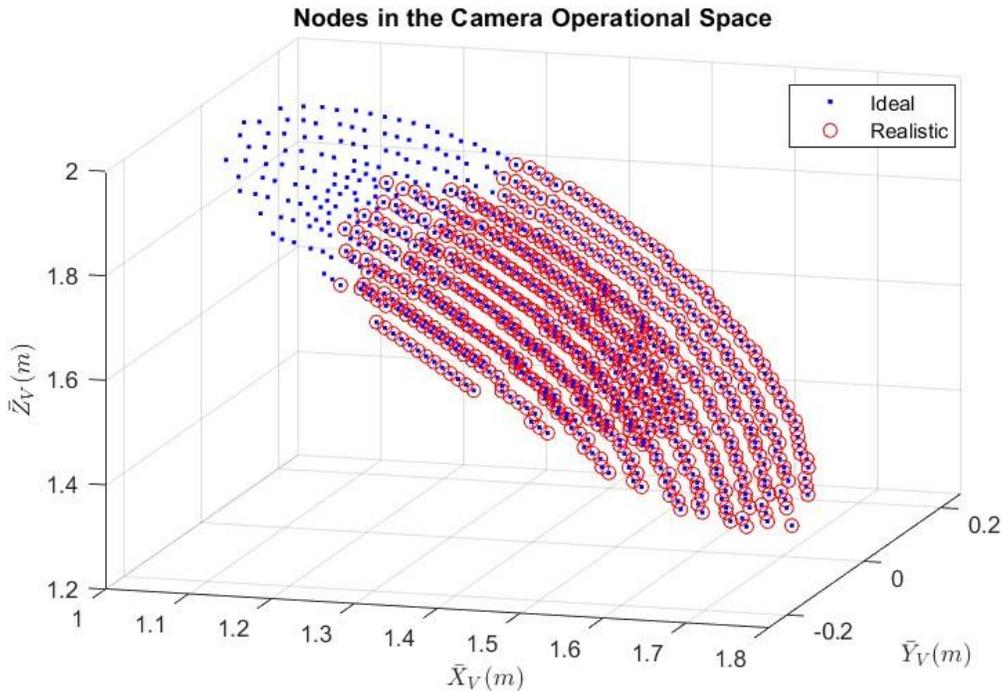

**Figure 4.15:** Reduction of Nodes from the ideal to the realistic camera operational space.

## 4.4 Simulation Results

A simulation in MATLAB is used to validate the proposed algorithm. In the simulation, the environment is set up with two minimums in the space (only one is the global minimum). $K_{est}$ and $K_{sd}$ are tunable parameters and selected as $K_{est} = 5, and\ K_{sd} = 50\ (m^{-1})$. The initial available energy bound is set to be $E_{bound} = 12$ (Watt Second (ws) or Joule). Figure 4.16. shows the sequences of nodes being explored in iteration steps by setting different energy threshold $E_T$.

By trials of different simulation runs, it can be concluded that:

- When $E_T$ is large (in this scenario, $3ws < E_T \leq 12ws$), the algorithm is conservative, so that it has only searched a small area that excludes the global minimum.
- When $E_T$ is in mid-range (in this scenario, $0.5ws \leq E_T \leq 3ws$), the algorithm drives the camera to the global minimum.

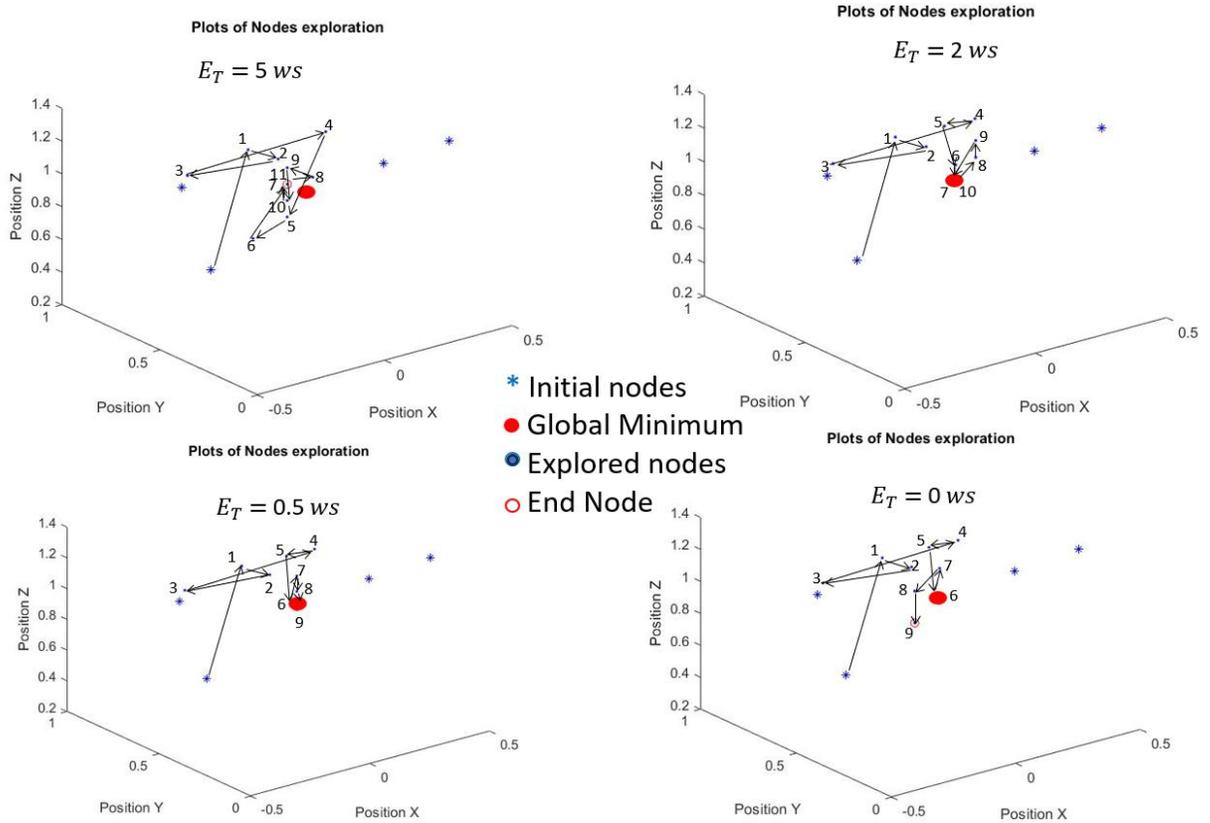

**Figure 4.16:** Plots of sequential exploration with different $E_T$.

When $E_T$ is small (in this scenario, $0ws \leq E_T < 0.5ws$), the algorithm is aggressive. Although it has searched a large amount of areas that includes the global minimum, it doesn't have enough energy to move the camera back to the best location explored in the previous stages.

The two tunable parameters $K_{est}$ and $K_{sd}$ are selected randomly in the simulation. Those parameters are used in the models to estimate the noise spatial distribution in the environment. As stated in Section 4.2.1, the positive parameter $K_{est}$ indicates that the weight of an explored node in the estimation function is based on its Euclidean distance from the node to be estimated. Larger $K_{est}$ in Equation (4.1) implies that Euclidean distance has smaller influence on the weight. In other words, with increasing $K_{est}$, the estimation result of an unexplored node is more different than that of an explored node nearby. As a result, the algorithm is more likely to search nodes around the local minimum. Similarly, $K_{sd}$, the other parameter, is proportional to the magnitude of the standard deviation in the stochastic modified process. Smaller $K_{sd}$ causes smaller standard deviation and thus more deterministic evaluation of an unexplored node, which discourages the algorithm to explore areas that are away from the local minimum.

Therefore, by increasing $K_{est}$ or decreasing $K_{sd}$, we should expect that the algorithm has higher tendency to exploit over explore. In this chapter, we have developed a parameter to present the degree of exploration in a simulation by measuring the average distance between the new explored node to the nearest previously explored node.

$$Avg\_Dis_{new} = \frac{\sum_{i=1}^{M} \min_{j \in S_{index-exp}} (\overline{ED}(P_j^{exp}, P_i^{new}))}{M} \tag{4.84}$$

Where $M$ is the total number of iterations in a simulation, $P_i^{new}$ is the position of the new explored node at $i^{th}$ iteration.

$$S_{index-exp} = \{index | Node_{index} \in S_{explored}\}, \tag{4.85}$$

$$\overline{ED}(P_a, P_b) = \sqrt{(P_{ax} - P_{bx})^2 + (P_{ay} - P_{by})^2 + (P_{az} - P_{bz})^2}, \tag{4.86}$$

with $P_a, P_b$ are 3D locations of $Node_a$ and $Node_b$:

$$P_a = (P_{ax}, P_{ay}, P_{az}) \text{ and } P_b = (P_{bx}, P_{by}, P_{bz})$$

In a simulation, a large value of the parameter $Avg\_Dis_{new}$ means new explored nodes in each search iteration are generally far away from previously explored nodes, and the algorithm succeeds in exploring a large area. A sensitivity test is given by varying $K_{est}$ and $K_{sd}$ to show how these two parameters affect the searching process in the same simulation environment. Table 4.2. presents the sensitivity analysis of $K_{est}$ by keeping a constant $K_{sd} = 50 \ (m^{-1})$ while Table 4.3. presents the sensitivity analysis of $K_{sd}$ by keeping a constant $K_{est} = 5$ (No unit). All analysis tests are based on the same scenario in Figure 4.16. with $E_{bound} = 20$ (ws) and $E_T = 2$ (ws).

**Table 4.2:** Sensitivity Analysis with Varying $K_{est}$. ($K_{sd}$ = 50 ($m^{-1}$)).

| $K_{est}$ | Iteration Numbers | $Avg\_Dis_{new}$ (m) | Reached global Minimum |
|---|---|---|---|
| 90 | 93 | 0.059 | NO |
| 70 | 83 | 0.066 | NO |
| 50 | 67 | 0.073 | YES |
| 30 | 46 | 0.085 | YES |
| 10 | 38 | 0.097 | YES |
| 5 | 47 | 0.102 | YES |
| 1 | 20 | 0.131 | NO |

**Table 4.3**: Sensitivity Analysis with Varying $K_{sd}$. ($K_{est}$ = 5 ).

| $K_{sd}(m^{-1})$ | Iteration Numbers | $Avg\_Dis_{new}$ (m) | Reached global Minimum |
|---|---|---|---|
| 90 | 36 | 0.086 | NO |
| 70 | 62 | 0.079 | YES |
| 50 | 67 | 0.073 | YES |
| 30 | 84 | 0.066 | YES |
| 10 | 127 | 0.060 | YES |
| 5 | 176 | 0.046 | YES |
| 1 | 264 | 0.037 | NO |

Table 4.2. and Table 4.3. show that by decreasing $K_{est}$ or increasing $K_{sd}$, the algorithm explores more areas as indicated by the parameter $Avg\_Dis_{new}$. The iteration numbers reduce as more areas have been explored because fewer nodes can be reached when the total energy for moving is bounded. Those conclusions can be derived from the analysis. On one hand, the algorithm is too conservative and searches few areas when $K_{est}$ is too large or $K_{sd}$ is too small. On the other hand, the algorithm is too aggressive and skips exploiting when $K_{est}$ is too small or $K_{sd}$ is too large. Both cases will make the node not end up at the global minimum. Therefore, a moderate combination of $K_{est}$ and $K_{sd}$ is preferred. Future research can focus on development of adaptive

algorithms that tune $K_{est}$ and $K_{sd}$ over iterations based on errors between real measurements and estimations.

In addition, the numerical values of $E_T$ are also picked randomly for testing. In real applications, a fixed $E_T$ should be selected before the operation of this algorithm. The selection of $E_T$ is related to the specific application scenario, size of camera's operational space, and total energy available $E_{bound}$ at beginning. However, the general thumb of the rule is the following: choose small values of $E_T$ when both $E_{bound}$ and operational space is large to encourage exploring over exploiting; otherwise, choose large $E_T$.

## 4.5 Conclusions

In this article, we have developed an algorithm that a camera can explore the workspace in a manufacturing environment and search for a location so that images' noise is minimized within the space that it can reach. This article also provides detailed development of the camera's operational space for a specific application. Results in a virtual environment have shown the algorithm succeed in bringing the camera to the optimal (or suboptimal if the optimal one is unreachable) position in this specific scenario. By reducing processing time of images, this algorithm can be used for various visual servoing applications in high-speed manufacturing.

However, challenges arise in some possible scenarios of real applications. For instance, if environmental factors vary too rapidly across space, a fine gridding of the space is required to manifest those variations in the estimation function, which increases computational complexity, resulting in poor real-time performances. The same issues exist in cases where the operational space of the camera is too large.

Other limitations include that this algorithm is incapable of differentiating contrast in images at different locations. Low contrast other than image noises occurs as another issue in visual servoing environment; for instance, low contrast in soft tissues has been shown negative effects on the performance of visual servoing controller in medical applications [76]. In addition, this algorithm does not account for cases when object is partially obstructed, which is very common in manufacturing environment. Limitations in the camera hardware that cannot account for additive noise are also not addressed in this paper. Those effects cannot be eliminated by any image denoising processes. For example, sensors with lower dynamic range may produce images where details are lost in extreme lighting conditions (bright sunlight or deep shadows). Another example is that some lenses may cause vignetting (darkening around the edges) or distortion, which can degrade the perceived image quality.

Simulation results show how energy threshold $E_T$ affects the behavior of the algorithm. A moderate threshold is suggested so that the algorithm can search for enough amount of area but enables to drive the camera back to the minimum with limited energy available. The analysis of $E_T$'s value in different applications is a focus of the future work. Future work may also include adaptive algorithms that update $K_{est}$ and $K_{sd}$ online and an adaptive function for energy estimation. Also, orientations of the camera are fixed in this paper. In the future, we would like to develop advanced algorithms that provide not only the optimal of positions but also the orientations inside the space.

In summary, this paper initially explores an adaptive algorithm of searching the optimal location in space with respective to image noises for observation in ETH or ETH/EIH cooperative configurations. Although several improvements of this algorithm can be addressed in the future, it has already shown a great potential to be applied as an upgraded feature in some of the real manufacturing tasks.

_____________________________________________________________Chapter 5

# Positioning and SISO Outer Controller Designs
___________________________________________________________________

As discussed earlier, feedback control design not only ensures the robot arm follows its target pose but also mitigates unmodeled dynamic errors, such as backlash, by treating them as disturbances within the control loop. The control architectures presented in Chapter 2 outline the sequential design process of each feedback control step. Section 5.1 details the inner feedback loop controller within the IBVS structure, while Section 5.2 focuses on the outer feedback loop controller for the visual system, referencing its SISO cases from Section 3.7.2. Section 5.3 introduces the design of the outer feedback loop controller for the SISO tool manipulation system, as introduced in Section 3.7.3. Sections 5.4 and 5.5 present simulation results and analyses for various scenarios involving the visual and tool manipulation systems, respectively. Finally, Section 5.6 derives a Lyapunov-based stability analysis for the controlled systems. This analysis provides a method to determine the optimal value of a tunable parameter in the controller design.

___________________________________________________________________

## 5.1 The Inner Loop Controller Design

This part presents the inner joint control loops for IBVS structures in both the visual system and tool manipulation system. A simulation scenario is also presented in this section.

In Section 3.5.3, a simplified form Equation (3.112) expresses an equation for the 6 DOF manipulator including the dynamics of the robot manipulator and the actuators (DC motor), rewrite it as:

$$(D(q) + J)\ddot{q} + (C(q,\dot{q}) + \frac{B}{r})\dot{q} + g(q) = u \quad (5.1)$$

where $D(q) \in \mathbb{R}^{6\times 6}$ and $C(q,\dot{q}) \in \mathbb{R}^{6\times 6}$ are inertial and Coriolis matrices respectively. $J \in \mathbb{R}^{6\times 6}$ is a diagonal matrix expressing the sum of actuator and gear inertias. $B$ is the damping factor, and $r$ is the gear ratio, $g(q) \in \mathbb{R}^{6\times 1}$ is the term for potential energy, $u \in \mathbb{R}^{6\times 1}$ is the input vector, and $q \in \mathbb{R}^{6\times 1}$ is the generalized coordinates (in this paper, $q$ is a $6 \times 1$ joint angle vector).

Simplify Equation (5.1) as follows:

$$M(q)\ddot{q} + h(q, \dot{q}) = u \quad (5.2)$$

with
$$M(q) = D(q) + J \quad (5.3)$$

$$h(q, \dot{q}) = (C(q,\dot{q}) + \frac{B}{r})\dot{q} + g(q) \quad (5.4)$$

Then, transform the control input as following:

$$u = M(q)v + h(q, \dot{q}) \quad (5.5)$$

where $v$ is a virtual input. Then, substitute for $u$ in Equation (5.1) using Equation (5.5), and since $M(q)$ is invertible, we will have a reduced system equation as follows:

$$\ddot{q} = v \quad (5.6)$$

This transformation is so-called feedback linearization technique with the new system equation given in Equation (5.6). This equation represents 6 uncoupled double integrators. The overall feedback linearization method is illustrated in Figure 5.1. In this control block diagram, the joint angle $q$ are forced to follow the target joint angle $q_R$ so that the output pose, $P$, can follow the target pose, $\bar{P}$. $P$, $\bar{P}$, $q$, and $q_R$ are all vectors with six elements (each element corresponds to a one-DoF end-effector position or joint angle). The Nonlinear interface transform the linear virtual control input $v$ to the nonlinear control input $u$ by using Equation (5.5). The output of the manipulator dynamic model, the joint angles, $q$, and their first derivatives, $\dot{q}$, are utilized to calculate $M(q)$ and $h(q, \dot{q})$ in the Nonlinear interface. The linear joint controller is designed using Youla parameterization technique [9] to control the nominally linear system in Equation (5.6).

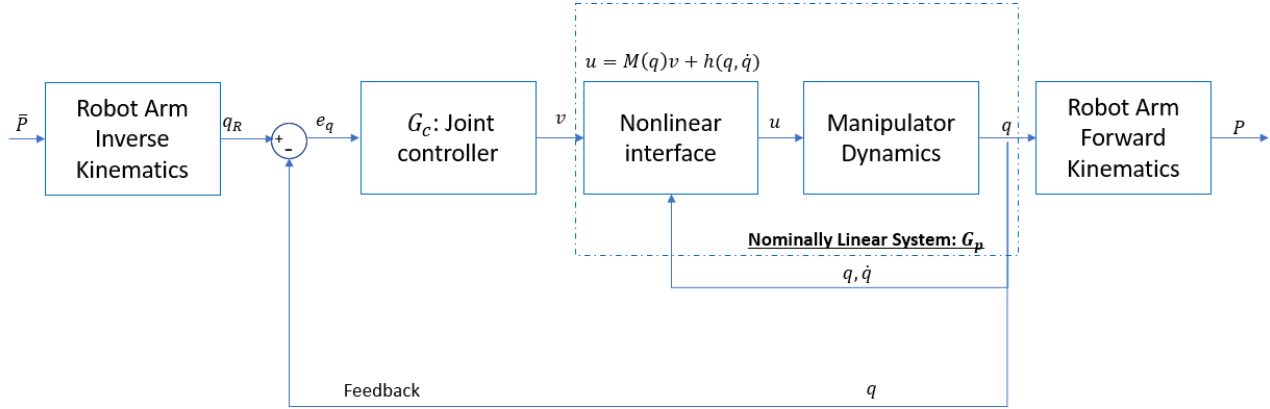

**Figure 5.1:** Feedback linearization Youla control design used for the joint control loop.

The design of a linear Youla controller with nominally linear plant is presented next.

Since the transfer functions between all inputs to outputs in Equation (5.6) are the same and decoupled, it is valid to first design a SISO controller and use the multiple of the same controller for a six-dimension to obtain the MIMO version. In other words, first design a controller $G_{C_{SISO}}^{Inner}$ that satisfies:

$$v_{SISO} = \ddot{q}_{SISO} \tag{5.7}$$

where $v_{SISO}$ is a single input to a nominally linear system and $\ddot{q}_{SISO}$ is the second order derivative of a joint angle. The controller in Figure 5.1. can be then written as:

$$G_{C_{sys}}^{Inner} = G_{C_{SISO}}^{Inner} \cdot I_{6X6} \tag{5.8}$$

where $I_{6X6}$ is a $6 \times 6$ identity matrix. In this project, Youla parameterization technique [9] is used to design the SISO controller $G_{C_{SISO}}^{Inner}$. The transfer function of the SISO nominally linear system from Equation (5.6) is:

$$G_{P_{SISO}}^{Inner} = \frac{1}{s^2} \tag{5.9}$$

Note that $G_{p_{SISO}}^{Inner}$ has two BIBO (Bounded Input Bounded Output) unstable poles at origin. To ensure internal stability of the feedback loop, the closed loop transfer function, $T_{SISO}$, should meet the interpolation conditions [77]:

$$T_{SISO}^{inner}(s=0) = 1 \tag{5.10}$$

$$\frac{dT_{SISO}^{inner}}{ds}\bigg|_{s=0} = 0 \tag{5.11}$$

Use the following relationship to compute a Youla transfer function: $Y_{SISO}$ as:

$$T_{SISO}^{inner} = Y_{SISO}^{inner} G_{p_{SISO}}^{Inner} \tag{5.12}$$

The $T_{SISO}^{inner}$ is designed so that it satisfies the conditions in Equations (5.10) and (5.11). The sensitivity transfer function, $S_{SISO}^{inner}$, is then calculated as follows:

$$S_{SISO}^{inner} = 1 - T_{SISO}^{inner} \tag{5.13}$$

Without providing the design details, the closed-loop transfer function can be in the following form to satisfy the interpolation conditions:

$$T_{SISO}^{inner} = \frac{(3\tau_{in}s + 1)}{(\tau_{in}s + 1)^3} \tag{5.14}$$

Where $\tau_{in}$ specifies the pole and zero locations and represents the bandwidth of the control system. $\tau_{in}$ can be tuned so that the response can be fast with less-overshoot.

The next step is to derive $G_{C_{SISO}}^{Inner}$ from relationships between the closed-loop transfer function, $T_{SISO}^{inner}$, the sensitivity transfer function, $S_{SISO}^{inner}$, and the Youla transfer function, $Y_{SISO}^{inner}$, in Equations (5.15) – (5.17):

$$Y_{SISO}^{inner} = T_{SISO}^{inner} G_{p_{SISO}}^{Inner^{-1}} = \frac{s^2(3\tau_{in}^2 s + 1)}{(\tau_{in}s + 1)^3} \tag{5.15}$$

$$S_{SISO}^{inner} = 1 - T_{SISO}^{inner} = \frac{s^2(\tau_{in}^3 s + 3\tau_{in}^2)}{(\tau_{in}s + 1)^3} \tag{5.16}$$

$$G_{C_{SISO}}^{Inner} = Y_{SISO}^{inner} S_{SISO}^{inner^{-1}} = \frac{3\tau_{in}^2 s + 1}{\tau_{in}^3 s + 3\tau_{in}^2} \tag{5.17}$$

From Equation (5.8), a MIMO controller can be computed as follows:

$$G_{C_{sys}}^{Inner} = \frac{3\tau_{in}^2 s + 1}{\tau_{in}^3 s + 3\tau_{in}^2} \cdot I_{6\times 6} \tag{5.18}$$

Equation (5.18) provides the expression of the desired joint controller. This configuration is precisely the inner joint control loop in both the visual and the manipulator systems as shown in Figure 2.4. and Figure 2.5.

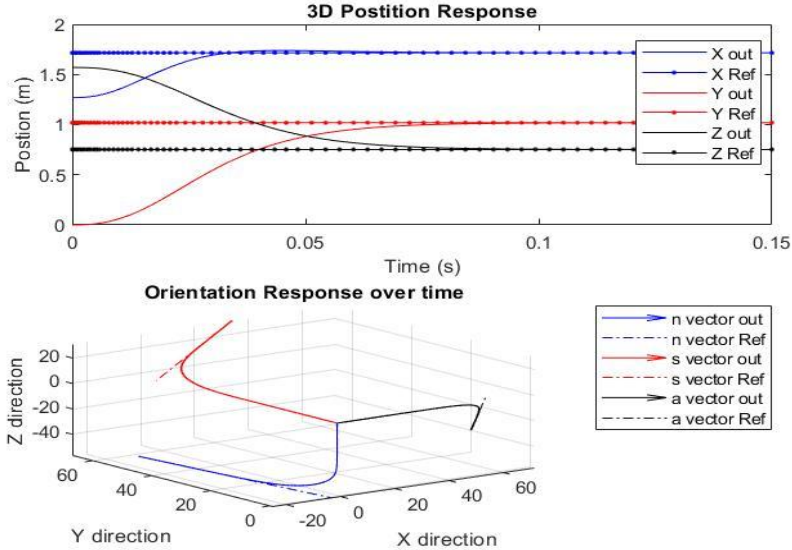

**Figure 5.2:** The simulation results for the end-effector response to an arbitrary trajectory.

Figure 5.2. shows the simulation results for a case with no disturbance. The reference position and the orientation of the end-effector are selected to be $\begin{bmatrix} \bar{X} \\ \bar{Y} \\ \bar{Z} \end{bmatrix} = \begin{bmatrix} 1.7157m \\ 1.0191m \\ 0.7518m \end{bmatrix}$ and $\begin{bmatrix} \bar{n} \\ \bar{s} \\ \bar{a} \end{bmatrix} = \begin{bmatrix} -0.425 & 0.87 & 0.25 \\ 0.8361 & 0.2714 & 0.4767 \\ 0.3469 & 0.4116 & -0.8428 \end{bmatrix}$, where $[\bar{X}, \bar{Y}, \bar{Z}]^T$ is the absolute reference position coordinate of the center of the end effector in the inertial frame and $\bar{n}, \bar{s}, \bar{a}$ represent respectively the end-effector's reference0 directional unit vector of the yaw, pitch and roll in the inertial frame. Therefore, the corresponding target angles of rotations are $q_R = [30°, 60°, -45°, 15°, 45°, 90°]$. For this simulation, we have designed the control system with the bandwidth of $100\ rad/s$; in other words, select $\tau_{in} = 0.01$s. In the following three plots, solid lines represent the responses for the end-effector position of each joint and the end-effector orientation respectively, and the dashed lines are the targets. Specifically, the orientation response of the end-effector is the vector that tangent to the curve in the second plot at each point in Figure 5.2. The simulation results show that all responses of the controlled system will be able to reach their final/steady state values within 0.1 second with no (or little) overshoots.

## 5.2 The Outer Loop Controller Designs for the SISO Visual System

The cascaded control within the visual system occurs during the camera movement adjustment phase, as discussed in Chapter 2. This control process aims to minimize positional errors of the camera remaining from the previous control stage. The overall topology of this control architecture is depicted in Figure 2.4. In the Single-Input Single-Output (SISO) scenario, previously illustrated in Figure 3.16, one coordinate of a reference point observed from the camera's optimal pose serves as the generated target for the control architecture. Figure 3.16. is reproduced here as Figure 5.3. for reference. Figure 5.4. illustrates the detailed block diagram of the SISO cascaded visual system.

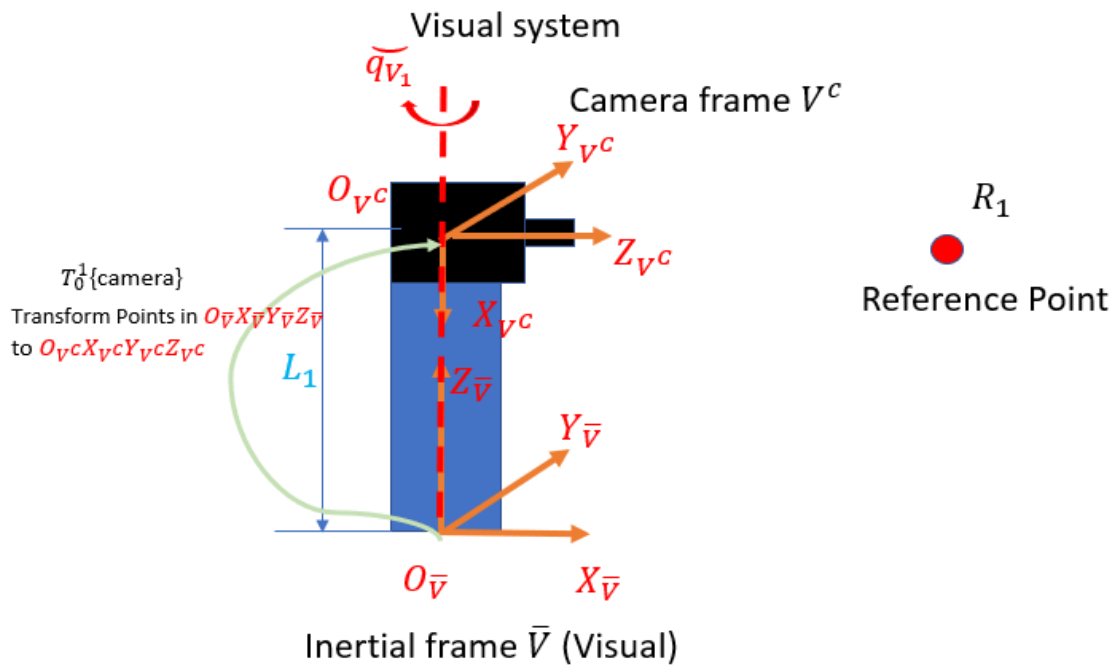

**Figure 5.3**: The SISO Camera-on-Robot model.

Section 5.1 has developed the closed loop transfer function of the inner-loop joint control system. The closed loop of SISO joint control is expressed in Equation (5.14). The SISO Camera-on-Robot Kinematics model is explained in Section 3.7.1 and the nonlinear relationship between disturbed robot joint angle $\widetilde{q_{V_1}}$ and reference point coordinate value $\widehat{v_{R_1}}$ is expressed in Equation (3.170), and it is rewritten as:

$$\widehat{v_{R_1}} = F \cdot tan(\varphi + \widetilde{q_{V_1}}) \qquad (5.19)$$

Where the angle $\varphi$ is the initial orientation of the camera with respect to the $X_{\bar{V}}$-axis in the inertia frame (Figure 5.3.), and $F$ is the focal length of the camera measured in mm.

Similarly, the target generator also relates to robot joint angle when camera is at the optimal position $\overline{q_{V_1}}$ and the corresponding reference point coordinates value $\overline{v_{R_1}}$ as the following Equation (5.20):

$$\overline{v_{R_1}} = F \cdot tan(\varphi + \overline{q_{V_1}}) \qquad (5.20)$$

The angle $\varphi$ has the same constant value in Equations (5.19) and (5.20), and its value is calculated from the reference points coordinates value in the inertial frame: $P_{R_1}^{\bar{V}} = (X_{R_1}^{\bar{V}}, Y_{R_1}^{\bar{V}}, Z_{R_1}^{\bar{V}})$:

$$\varphi = tan^{-1}(Y_{R_1}^{\bar{V}}/X_{R_1}^{\bar{V}}) \qquad (5.21)$$

The plant of outer loop control system is a combined nonlinear system composed of the inner closed loop system and camera-on-robot kinematic system. The design of the feedforward controller is first discussed in the following section.

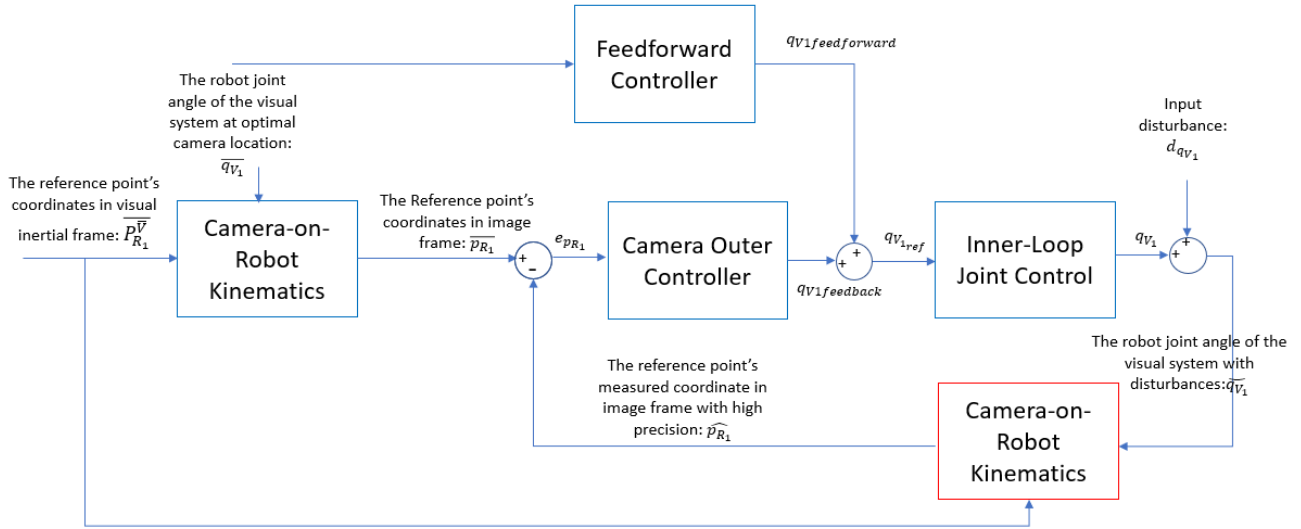

**Figure 5.4:** The camera movement adjustment SISO control block diagram.

*(Note: Blocks in red lines are referred to HIL Models)*

## 5.2.1 Feedforward Controller Design

Feedforward Controller generates a target rotational angle $q_{V1feedforward}$ based on the joint angles of the robot at camera's optimal pose. As shown in the Figure 5.5., below, Feedforward controller is the inverse process of inner-loop closed transfer function $T_{SISO}^{inner}$.

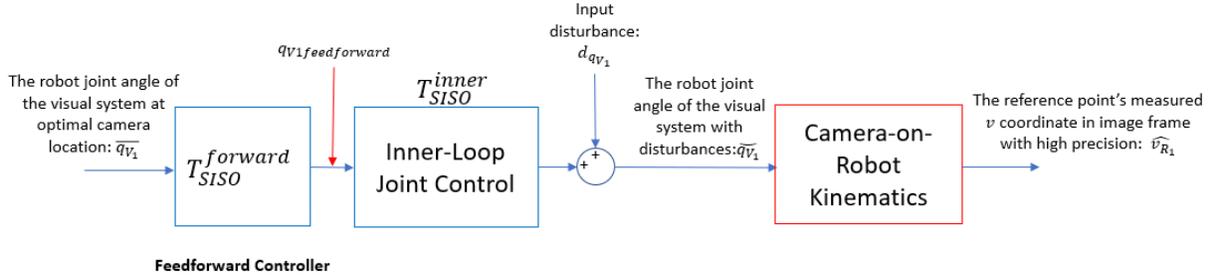

**Figure 5.5:** The SISO block diagram of the feedforward loop in the visual system.

*(Blocks in red lines are referred to HIL Models)*

The $T_{SISO}^{forward}$ can be designed as

$$T_{SISO}^{forward} = \frac{1}{T_{SISO}^{inner}} \frac{1}{(\tau_{forward}s+1)^2} = \frac{(\tau_{in}s+1)^3}{(3\tau_{in}s+1)} \frac{1}{(\tau_{forward}s+1)^2} \quad (5.22)$$

The double poles s = $-1/\tau_{forward}$ are added to make $T_{forward}$ proper. Choose $\tau_{forward}$ so that the added double poles are 10 times larger than the bandwidth of the original improper $T_{forward}$. In other words, $\tau_{forward}$ is chosen as

$$\tau_{forward} = 0.1\tau_{in} \quad (5.23)$$

Two different approaches are proposed to design the outer Feedback loop controller for the nonlinear system:

- Feedback linearization
- Model linearization

## 5.2.2 Feedback Controller Design with Feedback Linearization

The inner closed loop transfer function is already derived in Equation (5.14), and it is rewritten as:

$$T_{SISO}^{inner} = \frac{q_{V_1}(s)}{q_{V_{1feedback}}(s)} = \frac{(3\tau_{in}s + 1)}{(\tau_{in}s + 1)^3} \tag{5.24}$$

Reconstruct Equations (5.19) and (5.24) in time domain, by introducing an intermediate variable or state, $W$, as:

$$\tau_{in}^3 \dddot{W} + 3\tau_{in}^2 \ddot{W} + 3\tau_{in} \dot{W} + W = q_{V_{1feedback}} \tag{5.25}$$

$$F \cdot tan\left(\varphi + 3\tau_{in}\dot{W} + W + d_{q_{V_1}}\right) = \widehat{v_{R_1}} \tag{5.26}$$

Where

$$q_{V_1} = 3\tau_{in}\dot{W} + W \tag{5.27}$$

$$\widetilde{q_{V_1}} = d_{q_{V_1}} + q_{V_1} \tag{5.28}$$

Equations (5.25) and (5.26) describe a nonlinear third order system, where $W$ is the state, $q_{V_{1feedback}}$ is the input and $\widehat{v_{R_1}}$ is the output. Take the second order derivative of Equation (5.26) and combine with (5.25) to obtain:

$$\ddot{\widehat{v_{R_1}}} = R(W, \dot{W}, \ddot{W}) + G(W, \dot{W}, \ddot{W}) q_{V_{1feedback}} \tag{5.29}$$

where

$$\begin{aligned} R(W, \dot{W}, \ddot{W}) = {} & 2F \cdot cos^{-2}\left(\varphi + 3\tau_{in}\dot{W} + W + d_{q_{V_1}}\right) tan\left(\varphi + 3\tau_{in}\dot{W} \right.\\ & \left. + W + d_{q_{V_1}}\right)(3\tau_{in}\ddot{W} + \dot{W})^2 \\ & - F \cdot cos^{-2}\left(\varphi + 3\tau_{in}\dot{W} + W + d_{q_{V_1}}\right)\left(8\ddot{W} + \frac{9}{\tau_{in}}\dot{W} + \frac{3}{\tau_{in}^2}W\right) \end{aligned} \tag{5.30}$$

$$G(W, \dot{W}, \ddot{W}) = F \cdot cos^{-2}(\varphi + 3\tau_{in}\dot{W} + W + d_{q_{V_1}}) \frac{3}{\tau_{in}^2} \tag{5.31}$$

The derivation of Equations (5.29) – (5.31) is elaborated in Appendix E.

Then transform or map these variables so that the nonlinear system in Equations (5.25) and (5.26) can be written as an equivalent linear state-space representation as follows:

$$\varepsilon_1 = F \cdot tan\left(\varphi + 3\tau_{in}\dot{W} + W + d_{q_{V_1}}\right), \tag{5.32}$$

$$\varepsilon_2 = \dot{\varepsilon}_1 = F \cdot cos^{-2}(\varphi + 3\tau_{in}\dot{W} + W + d_{q_{V_1}}) \cdot (3\tau_{in}\ddot{W} + \dot{W}) \tag{5.33}$$

The state-space form of Equations (5.25) and (5.26) can be expressed as:

$$\dot{\varepsilon} = \begin{bmatrix} \dot{\varepsilon}_1 \\ \dot{\varepsilon}_2 \end{bmatrix} = \begin{bmatrix} 0 & 1 \\ 0 & 0 \end{bmatrix} \begin{bmatrix} \varepsilon_1 \\ \varepsilon_2 \end{bmatrix} + \begin{bmatrix} 0 \\ 1 \end{bmatrix} U \tag{5.34}$$

$$\widehat{v_{R_1}} = [1 \; 0] \begin{bmatrix} \varepsilon_1 \\ \varepsilon_2 \end{bmatrix} \tag{5.35}$$

Where

$$U = G(W, \dot{W}, \ddot{W}) q_{V_{ref_1}} + R(W, \dot{W}, \ddot{W}) \tag{5.36}$$

Transform the state-space representation back to the transfer function form:

$$Gp_{SISO}^{Visual}{}_{nominal} = \frac{\widehat{v_{R_1}}(s)}{U(s)} = C(sI - A)^{-1}B = \frac{1}{s^2} \tag{5.37}$$

Where

$$A = \begin{bmatrix} 0 & 1 \\ 0 & 0 \end{bmatrix}, B = \begin{bmatrix} 0 \\ 1 \end{bmatrix}, \text{ and } C = [1 \; 0] \tag{5.38}$$

Since the $Gp_{SISO}^{Visual}{}_{nominal}$ is the same as the plant transfer function of inner joint control system in Equation (5.9), one way to design a Youla controller for this linear system is similar to Equations (5.10) – (5.17). Another way of design is to add a new tunable variable $\zeta$ so that the system's response can be tuned. The design of the new Youla transfer function is the following:

$$Y_{SISO}^{Visual} = \frac{s^2[(\omega_n^2 \tau_{ext} + 2\zeta\omega_n)s + \omega_n^2]}{(s^2 + 2\zeta\omega_n s + \omega_n^2)(\tau_{ext} s + 1)} \tag{5.39}$$

The closed-loop transfer function of outer loop can be found by Equation (5.15):

$$T_{SISO}^{Visual} = Y_{SISO}^{Visual} Gp_{SISO}^{Visual}{}_{nominal} = \frac{(\omega_n^2 \tau_{ext} + 2\zeta\omega_n)s + \omega_n^2}{(s^2 + 2\zeta\omega_n s + \omega_n^2)(\tau_{ext} s + 1)} \tag{5.40}$$

Where $1/\tau_{ext}$ is the extra pole and is chosen to be 10 times larger than the pole of the closed loop system. In this way, the closed loop system $T_{SISO}^V$ is approximately a second order system. $\zeta$ is the damping ratio.

$\omega_n$ determines the poles and zeros' locations of outer closed-loop transfer function and therefore, represents the bandwidth of outer-loop system. It must be ensured that $1/\omega_n > \tau_{in}$ so that inner-loop responds faster than the outer-loop in the cascaded control design strategy. Set the bandwidth of outer loop is about 10 times of the inner loop, therefore, it can be given that:

$$\frac{1}{\omega_n} = 10\tau_{in} = 10\tau_{ext} \tag{5.41}$$

It can be easily proved that the designed closed outer loop $T_{SISO}^{Visual}$ satisfies the interpolation conditions in Equations (5.10) and (5.11).

Then the sensitivity transfer function, $S_{SISO}^{Visual}$ and the outer-loop controller $Gc_{SISO}^{Visual}$ can be derived from relationships (5.16) – (5.17):

$$S_{SISO}^{Visual} = 1 - T_{SISO}^{Visual} = \frac{s^2(\tau_{ext}s + 2\zeta\omega_n\tau_{ext} + 1)}{(s^2 + 2\zeta\omega_n s + \omega_n^2)(\tau_{ext}s + 1)} \tag{5.42}$$

$$Gc_{SISO}^{Visual} = Y_{SISO}^{Visual}(S_{SISO}^{Visual})^{-1} = \frac{(\omega_n^2 \tau_{ext} + 2\zeta\omega_n)s + \omega_n^2}{(\tau_{ext}s + 2\zeta\omega_n\tau_{ext} + 1)} \tag{5.43}$$

The frequency response of $T_{SISO}^{Visual}$, $S_{SISO}^{Visual}$, $Y_{SISO}^{Visual}$ are shown in Figure 5.6. (Choose $\omega_n = 10\ rad/s$). As can be seen, the frequency where $T_{SISO}^{Visual}$ and $S_{SISO}^{Visual}$ are intersected is the crossover frequency. The crossover frequency is a good estimation of the system bandwidth. The bandwidth of the closed loop transfer function is around 10 rad/s. At frequencies below the bandwidth frequency, the linear magnitude of $T_{SISO}^{Visual}$ is 1 while the linear magnitude of $S_{SISO}^{Visual}$ is very small. This ensures good estimation performance and robustness to model uncertainties. At frequencies higher than the bandwidth frequency, the magnitude of $T_{SISO}^{Visual}$, $Y_{SISO}^{Visual}$ decreases so that high frequency sensor noise can be rejected.

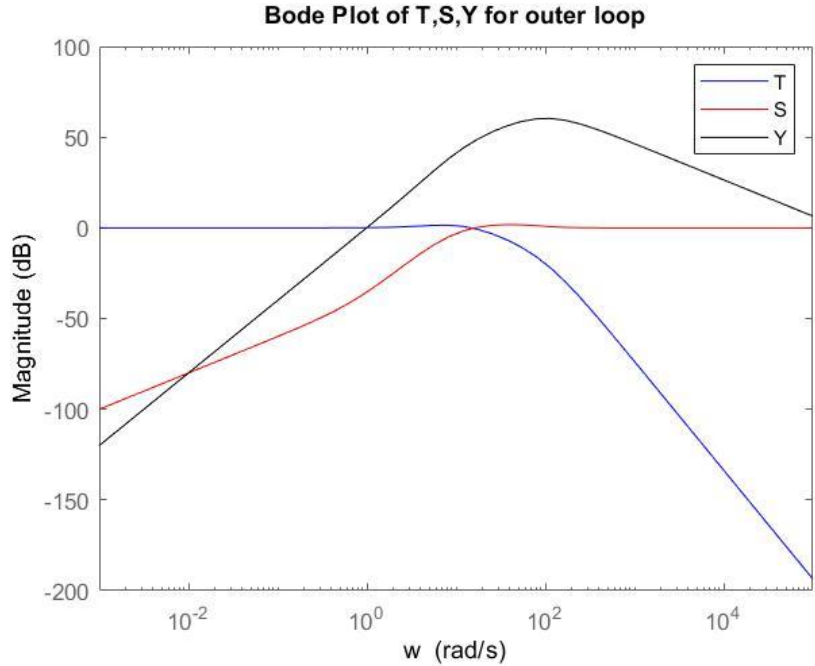

**Figure 5.6:** Frequency response of $\boldsymbol{T_{SISO}^{Visual}}$, $\boldsymbol{S_{SISO}^{Visual}}$, $\boldsymbol{Y_{SISO}^{Visual}}$ for feedback linearization design

The states $W, \dot{W}, \ddot{W}$ are computed from the rotational angle and its derivative $q_{V_1}$ and $\dot{q}_{V_1}$. From Equation (5.27), the following relationship can be obtained for the transfer function from $W$ to $q_{V_1}$:

$$W(s) = \frac{1}{3\tau_{in}s + 1} q_{V_1}(s) \tag{5.44}$$

Also, it can obtain the transfer function of $\dot{W}(s)$ and $\ddot{W}(s)$ as:

$$\dot{W}(s) = \frac{1}{3\tau_{in}s + 1} \dot{q}_{V_1}(s) \tag{5.45}$$

$$\ddot{W}(s) = \frac{s}{3\tau_{in}s + 1} \dot{q}_{V_1}(s) \tag{5.46}$$

Equations (5.44) – (5.46) are shown as the state transformation block in Figure 5.7.

It is worthwhile to note that the equation in the nonlinear interface:

$$q_{V_{ref_1}} = \frac{-R(W,\dot{W},\ddot{W})}{G(W,\dot{W},\ddot{W})} + \frac{1}{G(W,\dot{W},\ddot{W})} U \tag{5.47}$$

is only defined when $G(W, \dot{W}, \ddot{W}) = F \cdot \cos^{-2}(\varphi + 3\tau_{in}\dot{W} + W + d_{q_{V_1}}) \frac{3}{\tau_{in}^2} \neq 0$. This condition is always satisfied as function $\cos^{-2}(\varphi + 3\tau_{in}\dot{W} + W + d_{q_{V_1}})$ is in the range $[1, \infty]$.

In Figure 5.7., $q_{V_1}$ is the ideal angle of rotation from the control system.

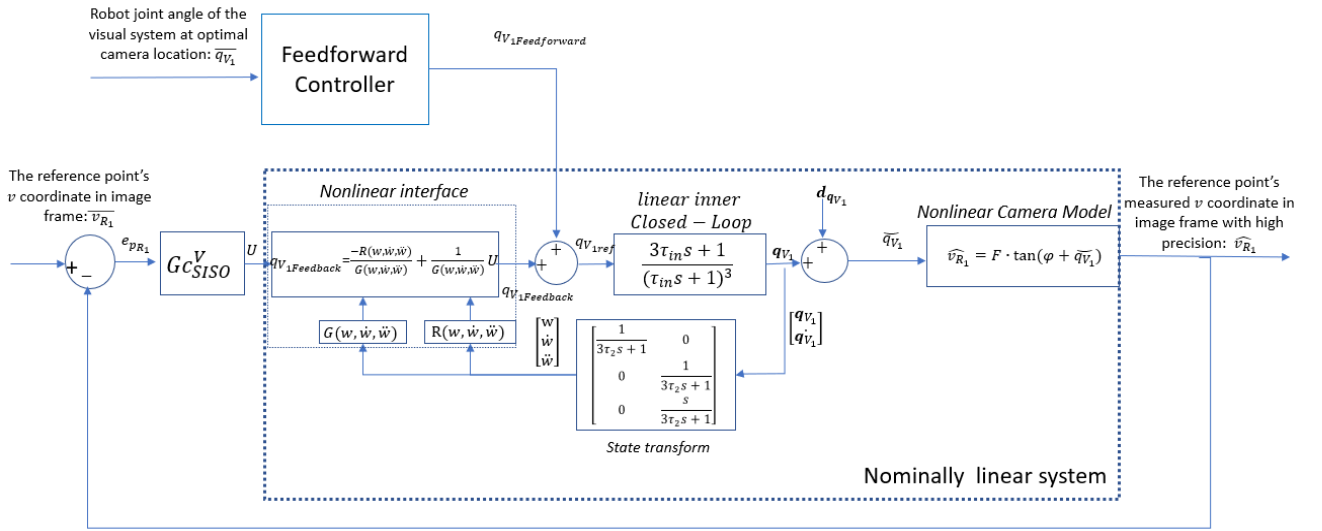

**Figure 5.7:** Block diagram of feedback linearization method for the SISO visual system.

## 5.2.3 Feedback Controller Design with Model Linearization

The nonlinear system can also be linearized first and then design a linear controller using the system transfer function. In Figure 5.8., $T_{inner-closed}$ transfer function is given in Equation (5.14) and the nonlinear form of the camera-on-robot model is provided in Equations (5.25) and (5.26). The overall dynamic system combines the inner-loop and the camera model, which will be linearized so that the combined dynamic system will then be linear. Next, we linearize, the camera model, around an equilibrium point $\widetilde{q_{V_1}}^0$:

$$\widehat{v_{R_1}} = F \cdot \cos^{-2}(\varphi + \widetilde{q_{V_1}}^0)(\widetilde{q_{V_1}} - \widetilde{q_{V_1}}^0) + F \cdot \tan(\varphi + \widetilde{q_{V_1}}^0)), \qquad (5.48)$$

If take $\widetilde{q_{V_1}}^0 = 0$, then:

$$\widehat{v_{R_1}} = F \cdot \cos^{-2}(\varphi)\widetilde{q_{V_1}} + F \cdot \tan(\varphi) \qquad (5.49)$$

Assuming $C_1^S = F \cdot \cos^{-2}(\varphi)$, $C_2^S = F \cdot \tan(\varphi)$, therefore, Equation (5.49) can be rewritten as:

$$\widehat{v_{R_1}} = C_1^S \widetilde{q_{V_1}} + C_2^S \qquad (5.50)$$

Let's define $\widehat{v_{R_1}}' = \widehat{v_{R_1}} - C_2^S$, then, the overall block diagram of the linearized system is shown in Figure 5.8.

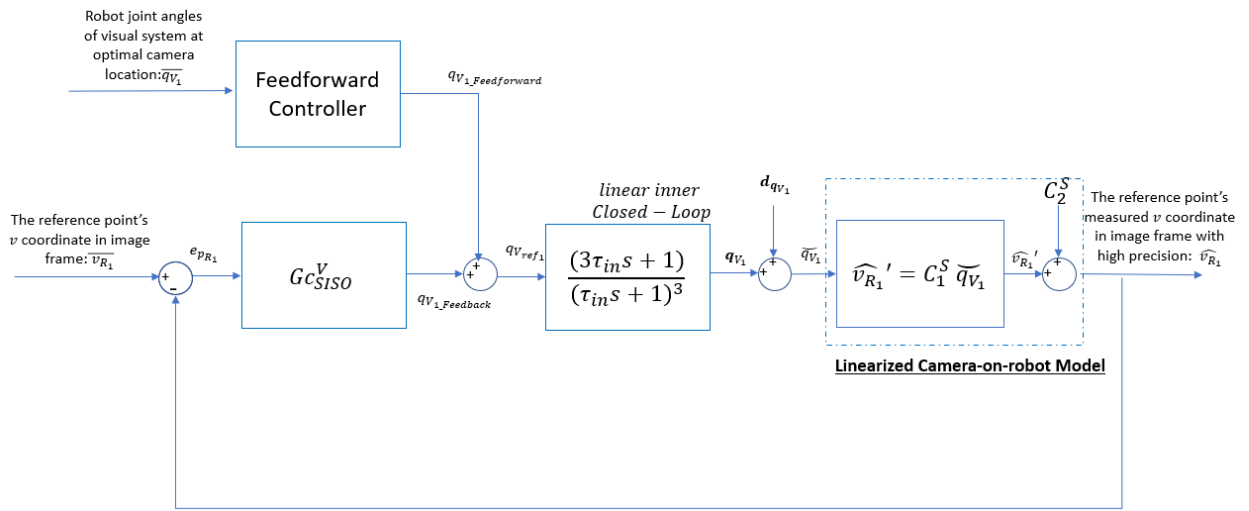

**Figure 5.8**: Block diagram of model linearization method for the SISO visual system.

The plant transfer function is derived as:

$$Gp_{SISO}^{Visual}\Big|_{linear} = \frac{\widehat{v_{R_1}}'}{q_{V_{1feedback}}} = C_1^S \frac{(3\tau_{in}s + 1)}{(\tau_{in}s + 1)^3} \tag{5.51}$$

The design of a Youla controller is trivial in this case as all poles/zeros of the plant transfer function in Equation (5.51) are located in the left half-plane, and therefore, they are stable. In this case, a selected Youla transfer function: $Y_{SISO}^{Visual}$ can shape the closed loop transfer function, $T_{SISO}^{Visual}$, by manipulating poles and zeros. All poles and zeros in the original plant can be cancelled out and new poles and zeros can be added to shape the closed-loop system. Let's select a Youla transfer function so that the closed-loop system behaves like a second order Butterworth filter, such that:

$$Y_{SISO}^{Visual} = \frac{1}{Gp_{SISO}^{Visual}\big|_{linear}} \frac{\omega_n^2}{(s^2 + 2\zeta\omega_n s + \omega_n^2)} \tag{5.52}$$

then:

$$T_{SISO}^{Visual} = \frac{\omega_n^2}{(s^2 + 2\zeta\omega_n s + \omega_n^2)} \tag{5.53}$$

where $\omega_n$ is called natural frequency and approximately sets the bandwidth of the closed–loop system. Again, it must be ensured that the bandwidth of the outer-loop is smaller than the inner-loop, i.e., $1/\omega_n > \tau_{in}$. $\zeta$ is called the damping ratio, which is another tuning parameter.

Then the sensitivity transfer function, $S_{SISO}^{Visual}$ and the outer-loop controller $Gc_{SISO}^{Visual}$ can be derived from relationships (5.15) – (5.17):

$$S_{SISO}^{Visual} = 1 - T_{SISO}^{Visual} = \frac{s^2 + 2\zeta\omega_n s}{(s^2 + 2\zeta\omega_n s + \omega_n^2)} \tag{5.54}$$

$$Gc_{SISO}^{Visual} = Y_{SISO}^{Visual}(S_{SISO}^{Visual})^{-1} = \frac{1}{C_1^S} \frac{(\tau_{in}s + 1)^3}{(3\tau_{in}s + 1)} \frac{\omega_n^2}{(s^2 + 2\zeta\omega_n s)} \tag{5.55}$$

The frequency response of $T_{SISO}^{Visual}$, $S_{SISO}^{Visual}$, $Y_{SISO}^{Visual}$ are shown in Figure 5.9. (Choose $\omega_n = 10\ rad/s$). As can be seen, the frequency where $T_{SISO}^{Visual}$ and $S_{SISO}^{Visual}$ are intersected around 10 rad/s, which is a good estimation of the system bandwidth. Same as discussion of the feedback linearization frequency responses, this controlled system has good estimation performance and robustness to model uncertainties at low frequencies ($\omega < \omega_n$), and reject noise at high frequencies ($\omega > \omega_n$)

The simulation results of both methods are presented in Chapter 7. In that chapter, it will also show that the parameter $\zeta$ affects the performance of controllers. Comparison analysis based on performance and robustness is also considered between the feedback linearization method, and the direct model linearization method.

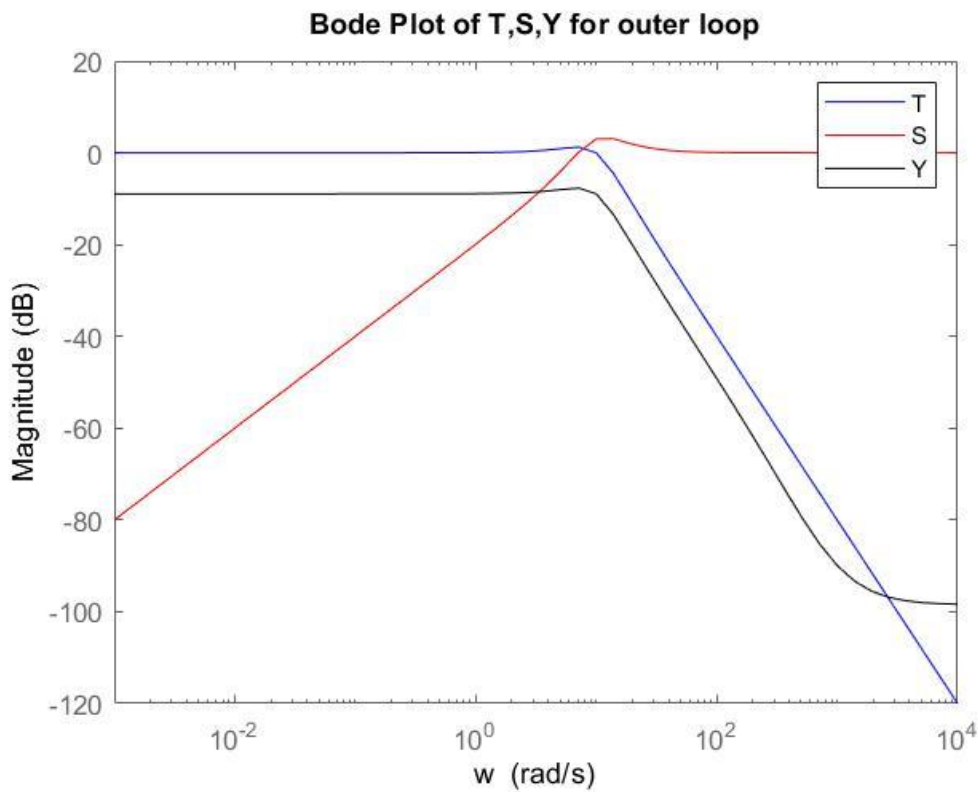

**Figure 5.9:** Frequency response of $T_{SISO}^{Visual}$, $S_{SISO}^{Visual}$, $Y_{SISO}^{Visual}$ for model linearization design.

## 5.3 The Outer Loop Controller Designs for the SISO Tool Manipulation System

The cascaded control of the tool manipulation system is aimed to provide precise tool pose control with the aid of camera's accurate observation. Figure 2.5. shows the topology of the control architecture. For the SISO case, previously illustrated in Figure 3.17., a target coordinate of an interest point on the tool is used as a generated reference in the architecture. Figure 3.17. is reproduced here as Figure 5.10. for reference. Figure 5.11. shows the block diagram of the SISO cascaded tool manipulation system.

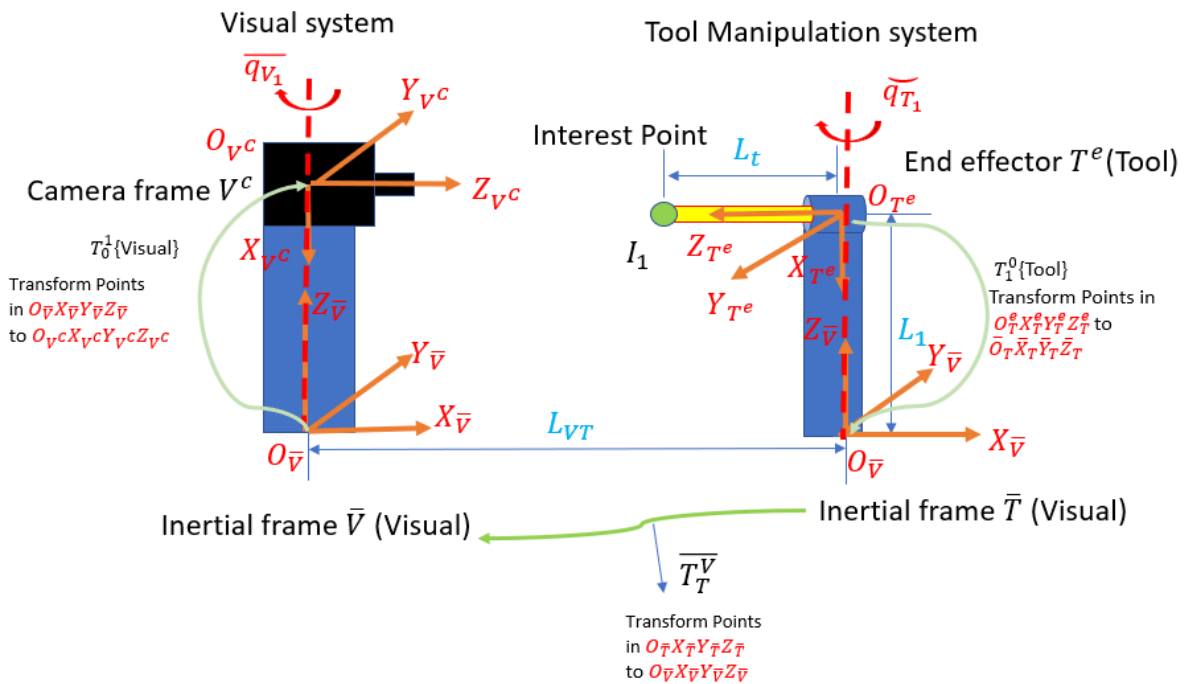

**Figure 5.10**: The SISO Camera-and-Tool combined model.

Tool-on-Robot kinematics model and Camera-on-Robot kinematics model can be combined into a single Camera-and-Tool model as shown in the Figure 5.11. below. Section 5.1 has developed the closed loop transfer function of the inner-loop joint control system. The closed loop of SISO joint control is expressed as Equation (5.14). The SISO Camera-and-Tool Model is explained in section 3.7 and the nonlinear relationship between the disturbed rotational angle of the link $\widetilde{q_{T_1}}$ and the interest point's measured image coordinate $\widehat{v_{I_1}}$ is expressed in Equation (3.151), and it is rewritten as:

$$\widehat{v_{I_1}} = F \cdot \frac{Q(\widetilde{q_{T_1}}) + \tan(\overline{q_{V_1}})}{1 - Q(\widetilde{q_{T_1}}) \tan(\overline{q_{V_1}})}$$

where
$$Q(\widetilde{q_{T_1}}) = \frac{L_t \sin(\widetilde{q_{T_1}})}{L_{VT} - L_t \cos(\widetilde{q_{T_1}})} \tag{5.56}$$

Where $L_{VT}$ is the distance between centers of visual and tool manipulation systems, $L_t$ is the length of the link that attach the interest point, and $\overline{q_{V_1}}$ is the fixed rotational angle of the visual system at the optimal pose. $L_{VT}$, $L_t$ and $\overline{q_{V_1}}$ are constant parameters.

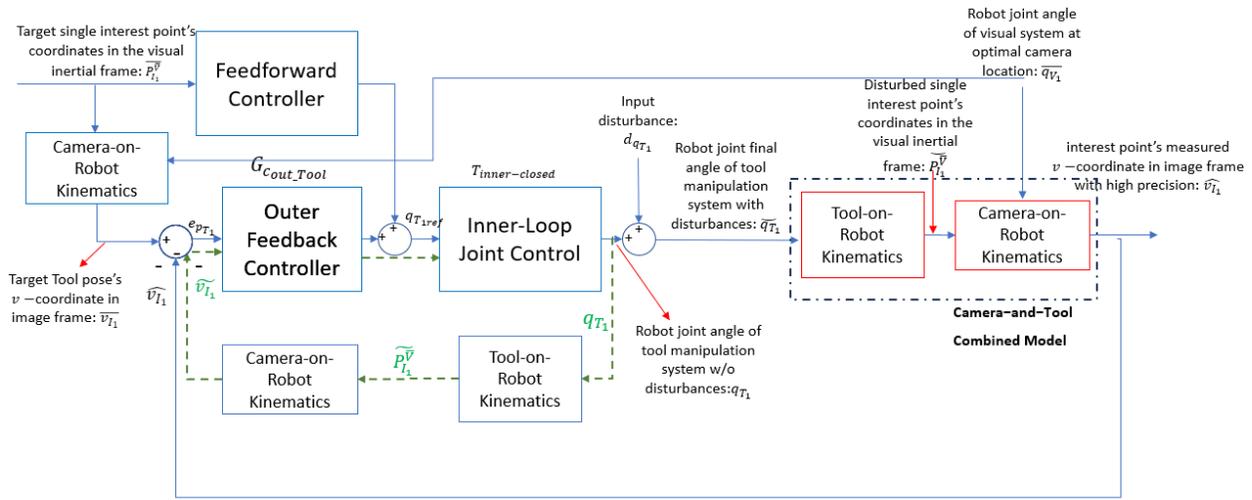

**Figure 5.11**: The SISO tool manipulation feedback control loop with feature estimation.

*(Blocks in red lines are referred to HIL Models)*

The control architecture in Figure 5.11. contains a feedforward loop and feedback loop. The design of the feedforward controller is first discussed in the following section.

### 5.3.1 Feedforward Controller Design

Feedforward Controller generates a target rotational angle $q_{T1feedforward}$ based on the tool's coordinates at the target location. As shown in the Figure 5.12. below, same as the design of the visual system, feedforward controller contains two inverse processes of the original plant in the system. The first diagram is the inverse process of the Camera-and-Tool Combined model and the second diagram $T_{SISO}^{forward}$ is the inverse process of inner-loop closed transfer function $T_{SISO}^{inner}$.

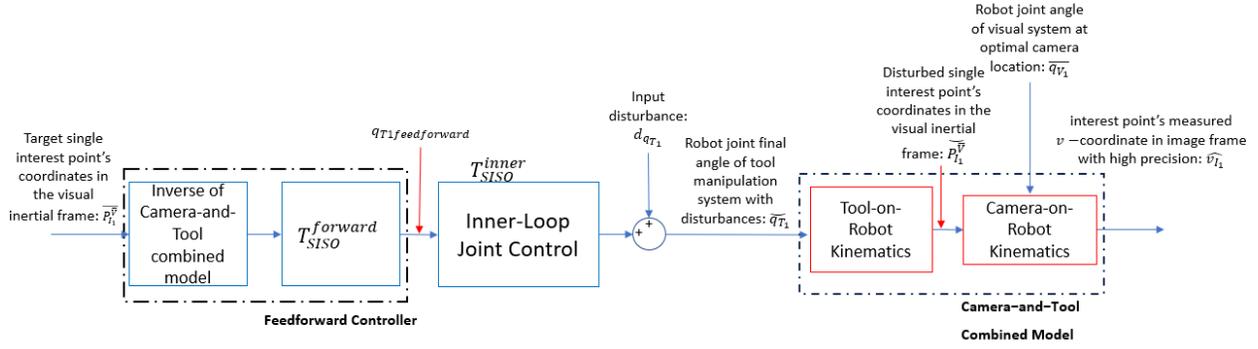

**Figure 5.12:** The SISO feedforward loop in the tool manipulation system.
*(Blocks in red lines are referred to HIL Models)*

The mathematical equation of Camera-and-Tool combined model has been shown in Equation (5.56). The inverse of Equation (5.56) is to find an expression of $\widetilde{q_{T_1}}$ from $\widehat{v_{I_1}}$. If set an angle $\sigma$ so that:

$$\tan(\sigma) = Q(\widetilde{q_{T_1}}) \tag{5.57}$$

Then from trigonometric relationship, Equation (5.56) can be transformed into the following expression:

$$\widehat{v_{I_1}} = F \cdot \tan(\sigma + \overline{q_{V_1}}) \tag{5.58}$$

Therefore,

$$\sigma = \arctan\left(\frac{\widehat{v_{I_1}}}{F}\right) - \overline{q_{V_1}} \tag{5.59}$$

Reorganize the expression of $Q(\widetilde{q_{T_1}})$ in Equation (5.56), it can be obtained:

$$L_t \sin(\widetilde{q_{T_1}}) + \tan(\sigma) L_t \cos(\widetilde{q_{T_1}}) = L_{VT} \tan(\sigma) \tag{5.60}$$

Assume the following relationship by introducing a variable $\delta$, where:

$$\cos(\delta) = L_t$$
$$\sin(\delta) = \tan(\sigma) L_t \tag{5.61}$$

Use the trigonometric relationship and reorganize Equation (5.60) from Equation (5.61):

$$\sin(\widetilde{q_{T_1}} + \delta) = L_{VT} \tan(\sigma) \tag{5.62}$$

From Equation (5.61), it can be derived that:

$$\tan(\delta) = \frac{\sin(\delta)}{\cos(\delta)} = \frac{\tan(\sigma)L_t}{L_t} = \tan(\sigma) \tag{5.63}$$

Thus,

$$\delta = \sigma \tag{5.64}$$

Therefore Equation (5.62) can be written as:

$$\sin(\widetilde{q_{T_1}} + \sigma) = L_{VT}\tan(\sigma) \tag{5.65}$$

From Equation (5.65), the expression of $\widetilde{q_{T_1}}$ can be derived as:

$$\widetilde{q_{T_1}} = \arcsin(L_{VT}\tan(\sigma)) - \sigma$$

where

$$\sigma = \arctan\left(\frac{\widehat{v_{I_1}}}{F}\right) - \overline{q_{V_1}} \tag{5.66}$$

Equation (5.66) gives the inverse model of the camera-and-tool combined model.

The plant of outer loop control system is a combined nonlinear system composed of the inner closed loop system and Camera-and-Tool combined system.

Like the visual system, two different approaches are proposed to design the outer loop controller for the nonlinear system:

- Feedback linearization
- Model linearization

### 5.3.2 Feedback Controller Design with Feedback Linearization

The development of feedback linearization controller is same as the design process for the visual system in section 5.2.2. The only difference is the parameters of nonlinear function (5.29): $R(W,\dot{W},\ddot{W})$ and $G(W,\dot{W},\ddot{W})$. The expressions of those two parameters are given in the below:

$$\begin{aligned} R(W,\dot{W},\ddot{W}) = &\ 2F \cdot \cos^{-2}(\sigma + \overline{q_{V_1}})\tan(\sigma + \overline{q_{V_1}})(3\tau_{in}\ddot{W} + \dot{W})^2 A^2 \\ &+ F \cdot \cos^{-2}(\sigma + \overline{q_{V_1}})(3\tau_{in}\ddot{W} + \dot{W})B \\ &- F \cdot \cos^{-2}(\sigma + \overline{q_{V_1}})\left(8\ddot{W} + \frac{9}{\tau_{in}}\dot{W} + \frac{3}{\tau_{in}^2}W\right)A \end{aligned} \tag{5.67}$$

$$G(W,\dot{W},\ddot{W}) = F \cdot \cos^{-2}(\sigma + \overline{q_{V_1}})\frac{3}{\tau_{in}^2}A \tag{5.68}$$

Where

$$\sigma = \arctan\left(\frac{L_t \sin\left(3\tau_{in}\dot{W} + W + d_{q_{V_1}}\right)}{L_{VT} - L_t \cos\left(3\tau_{in}\dot{W} + W + d_{q_{V_1}}\right)}\right) \quad (5.69)$$

$$A = \frac{L_{VT}L_t \cos\left(3\tau_{in}\dot{W} + W + d_{q_{V_1}}\right) - L_t^2}{L_t^2 + L_{VT}^2 - 2L_tL_{VT}\cos\left(3\tau_{in}\dot{W} + W + d_{q_{V_1}}\right)} \quad (5.70)$$

$$B = \frac{L_tL_{VT}\sin\left(3\tau_{in}\dot{W} + W + d_{q_{V_1}}\right)(L_t^2 - L_{VT}^2)}{(L_t^2 + L_{VT}^2 - 2L_tL_{VT}\cos\left(3\tau_{in}\dot{W} + W + d_{q_{V_1}}\right))^2} \quad (5.71)$$

The complete derivation of Equations (5.67) – (5.71) can be found in the Appendix E. The nominal plant of the system is same as Equation (5.37). Therefore, the same controller designed for the visual system can be used for the tool manipulation system. That is,

$$Gc_{SISO}^{Tool} = \frac{(\omega_n^2 \tau_{ext} + 2\zeta\omega_n)s + \omega_n^2}{(\tau_{ext}s + 2\zeta\omega_n\tau_{ext} + 1)} \quad (5.72)$$

Where $\omega_n$ determines the bandwidth of the outer-loop system and $1/\tau_{ext}$ is the extra pole. $\zeta$ is the damping ratio. The overall block diagram of feedback linearization method is shown in Figure 5.13. The frequency response of $T_{SISO}^{Tool}$, $S_{SISO}^{Tool}$, $Y_{SISO}^{Tool}$ is same as that shown in Figure 5.6.

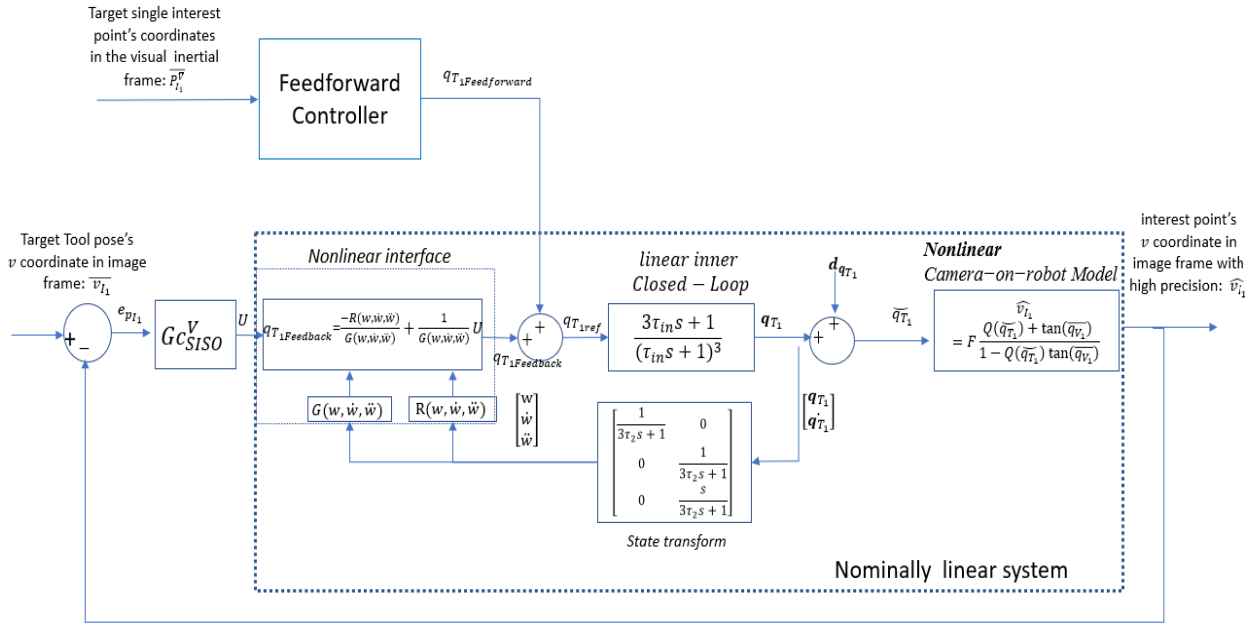

**Figure 5.13**: Block diagram of feedback linearization method for the SISO tool manipulation system.

## 5.3.2 Feedback Controller Design with Model Linearization

Same as the modeling process for the visual system, the nonlinear camera-and -tool combined model can be linearized around an equilibrium point. Set this equilibrium point to be zero, the model can be linearized as:

$$\widehat{v_{I_1}} = K_1^S \widetilde{q_{T_1}} + K_2^S \tag{5.73}$$

And

$$K_1^S = F \cdot (1 + (tan(\overline{q_{V_1}}))^2 \frac{L_{VT} L_t - L_t^2}{(L_{VT} - L_t)^2} \tag{5.74}$$

$$K_2^S = F \cdot \tan(\overline{q_{V_1}}) \tag{5.75}$$

Then, the overall block diagram of the linearized system is shown in Figure 5.14.

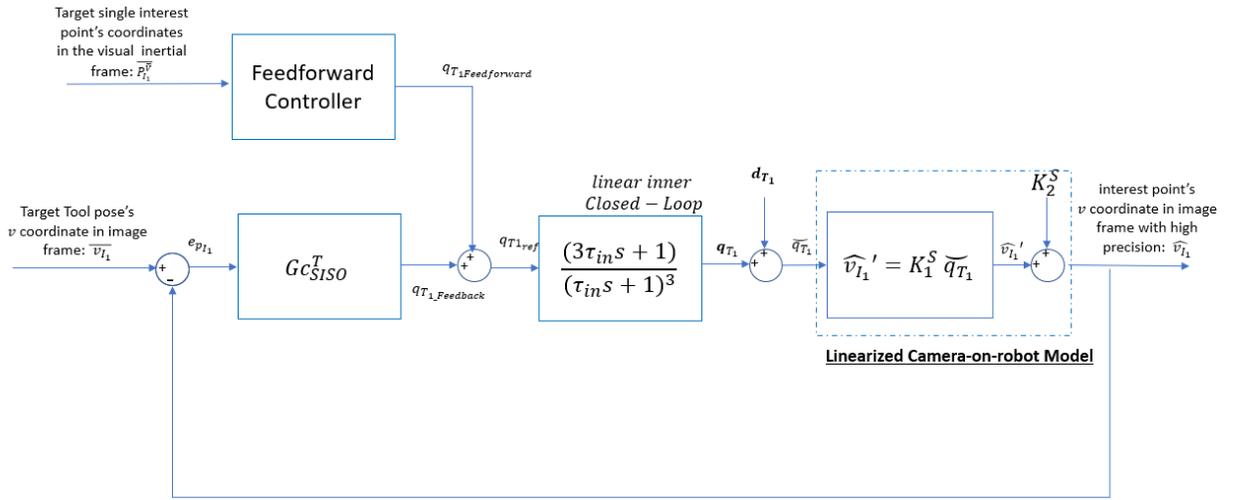

**Figure 5.14:** Block diagram of model linearization method for the SISO tool manipulation system.

With same design procedures as steps (5.52) – (5.55), the controller designed for a second order closed -loop system using the plant linearization method is given as:

$$Gc_{SISO}^{Tool} = \frac{1}{K_1^S} \frac{(\tau_{in} s + 1)^3}{(3\tau_{in} s + 1)} \frac{\omega_n^2}{(s^2 + 2\zeta \omega_n s)} \tag{5.76}$$

Where $K_1^S$ is defined in Equation (5.74) and the second order closed-loop transfer function $T_{SISO}^{Tool}$ of the overall cascaded control system is expressed as:

$$T_{SISO}^{Tool} = \frac{\omega_n^2}{s^2 + 2\zeta \omega_n s + \omega_n^2} \tag{5.77}$$

Where $\tau_{in}$ defines the bandwidth of the inner joint loop. $\omega_n$ is the natural frequency and $\zeta$ is the damping ratio of the second order system.

The frequency response of $T_{SISO}^{Tool}, S_{SISO}^{Tool}, Y_{SISO}^{Tool}$ is same as that shown in Figure 5.9.

### 5.3.4 2D Feature Estimation

As the camera is static in this control stage, the tool pose cannot be recognized and measured visually if it is outside the camera range of view. To tackle this problem, the 2D feature (image coordinates of the tool points) can be estimated from the same model in Equation (5.56) with the joint angle $q_{T_1}$ as input shown in Figure 5.11.:

$$\widetilde{p_{I_1}} = F \frac{Q(q_{T_1}) + \tan(\widetilde{q_{V_1}})}{1 - Q(q_{T_1}) \tan(\widetilde{q_{V_1}})}$$

where 
$$Q(q_{T_1}) = \frac{L_t \sin(q_{T_1})}{L_{VT} - L_t \cos(q_{T_1})} \tag{5.57}$$

As illustrated in Figure 5.11., the normal feedback loop (in blue lines) is preserved when the tool is inside the camera range of view and hence, the camera can measure the tool 2D feature $\widehat{p_{I_1}^T}$. However, when the tool is outside the range of view, the 2D feature can only be approximated as $\widetilde{p_{I_1}^T}$ (green dashed line) by the combined model as provided in Equation (5.57). A bump-less switch can be implemented to smoothly switch between these modes of operations. The switching signal changes over when the tool moves in or out of the camera range of view.

### 5.3.5 Limitation of the SISO Tool Manipulation System Control

The solution of inverse SISO camera-on-robot model is not unique. Particularly, two possible positions of the tool (denoted as rotational angles of the tool's manipulator: $\widetilde{q_{T_1}}$) will result in the same image coordinate of an interest point: $\widehat{p_{I_1}}$. As illustrated in Figure 5.15., when the tool rotates to $\widetilde{q_{T_1}^a}$ or $\widetilde{q_{T_1}^b}$, the interest point will be projected onto the image plane at the same coordinate: $\widehat{p_{I_1}}$. In other words, locations of one robot link cannot be distinguished by matching the interest point's image coordinates with the target image coordinate. However, the 6-DOF camera-on-robot model yields a unique solution for the tool's 6-DOF pose based on six distinct image coordinates corresponding to two points of interest. While a formal proof is not provided here, this uniqueness can be intuitively observed from Equations (3.123) and (3.124).

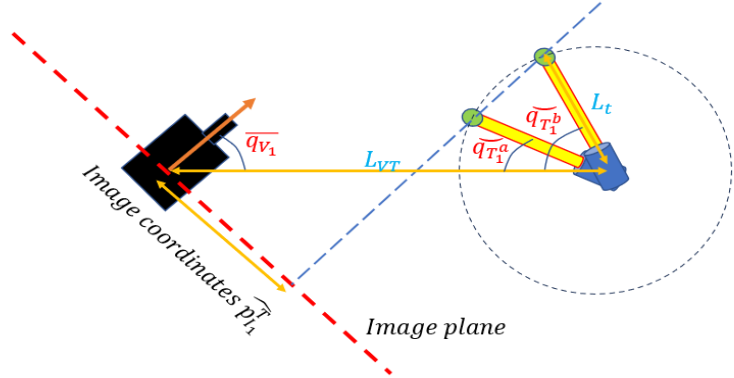

**Figure 5.15:** Top-side view of one DoF Camera-and-Tool combined model.

## 5.4 Simulation Results of the SISO Visual System

In this section, we are going to compare the closed-loop response results of the cascaded feedback and feedforward control system where the outer-loop controllers are designed using the two aforementioned methods: feedback linearization (Section 5.2.2) and model linearization (Section 5.2.3). Also, architecture of loops with feedback-only controllers is also simulated to show the performance's difference between feedback-only and feedback-feedforward control designs. The simulation results are obtained with the original nonlinear SISO camera-on-robot model: Equation (5.19). In addition, we will also illustrate how varying the damping ratio $\zeta$ affects the responses. For both methods, we chose the bandwidth of the inner-loop as $100\ rad/s$ and the bandwidth of the outer-loop as $10\ rad/s$. The camera orientation (axis $Z_V^C$) with respect to the $X_{\bar{V}}$-axis in the inertia frame is the controlled output and is defined as $O_z^C$ (measured in degrees). The target orientation of the camera is set to be aligned with X-axis in the inertia frame: $X_{\bar{V}}$; that is $\overline{O_z^C} = 0°$.

We compare the simulation responses by choosing six different damping ratios. Two are chosen as the overdamped systems ($\zeta>1$), one is chosen as a critically damped system ($\zeta=1$), and three as the underdamped systems ($\zeta<1$). We have simulated four cases and compared all controller design methods for each case. Each case is different due to varying the initial angle $\varphi$ (see Equation (5.21)) and the input disturbance $d_{q_v}$. Two cases are simulated without the input disturbance while the other two are simulated with the disturbance to compare the robustness of the controlled system.

The step responses of the camera orientation $O_z^C$ for model linearization are shown in Figures 5.16., 5.17., 5.18., and 5.19. The intrinsic camera parameters are selected to be: $F = 2.8\ mm$

(Focal length) and $\alpha = 120°$ (Angle of view). For each case, Figures (a) and (c) show results of model linearization approach with and without feedforward controllers. And Figure (b) shows results of feedback linearization without feedforward controllers. The step responses of feedback linearization with feedforward controllers are unstable so that their response cannot be shown in the plots. Without formal proof, it can be stated that any input from feedback controllers drastically alters the nonlinear interface parameters, used in feedback linearization, and hence, results in an unstable system.

The case 1 is simulated with the initial angle $\varphi < \frac{\alpha}{2}$, while the case 2 is simulated when $\varphi = \frac{\alpha}{2}$, the largest possible initial angle within the angle of view. It can be shown clearly that without any input disturbance, all approaches are able to drive the closed-loop responses to the final values.

For model linearization approach without feedforward controllers, it can be seen from the two simulation cases that there exists a damping ratio, $\zeta_{opt}$, such that

- When $\zeta \geq \zeta_{opt}$, the step responses have no overshoots and as $\zeta$ decreases, the system reaches the steady state faster.
- When $\zeta < \zeta_{opt}$, the step responses have overshoots, and the overshoots increase as $\zeta$ decreases. As $\zeta$ increases, the system reaches the steady state faster.

It can be estimated from Figures 5.16. and Figure 5.17. that $\zeta_{opt} \cong 1$ in case 1 and $\zeta_{opt} \cong 0.5$ in case 2. The most desirable system is the one without an overshoot and with the fastest response time. When $\zeta = \zeta_{opt}$, the system has the fastest response and no (or little) overshoot.

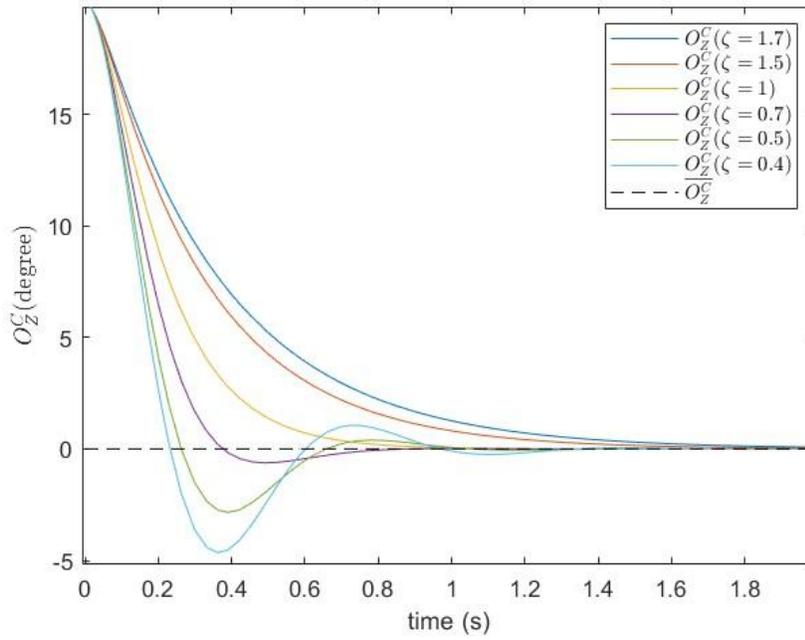

(a)

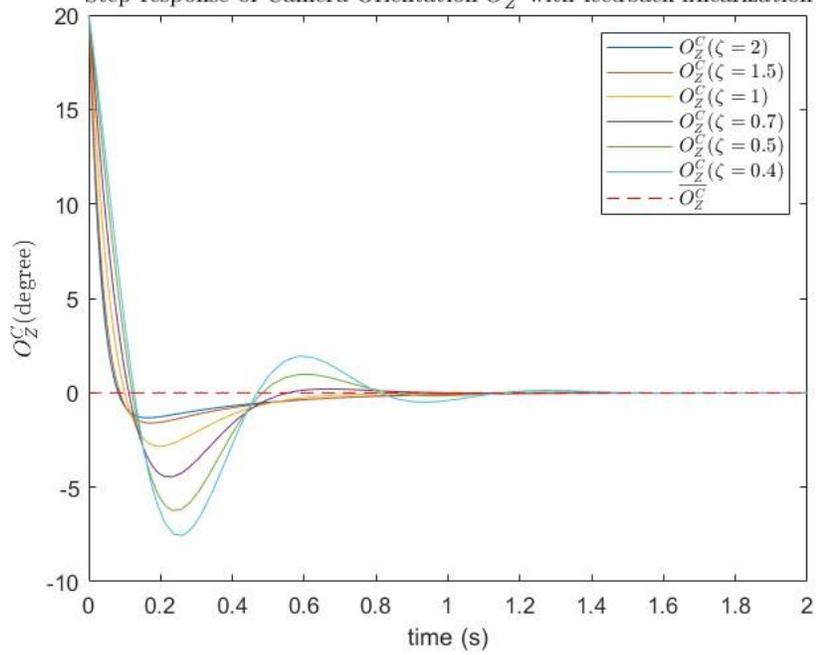

(b)

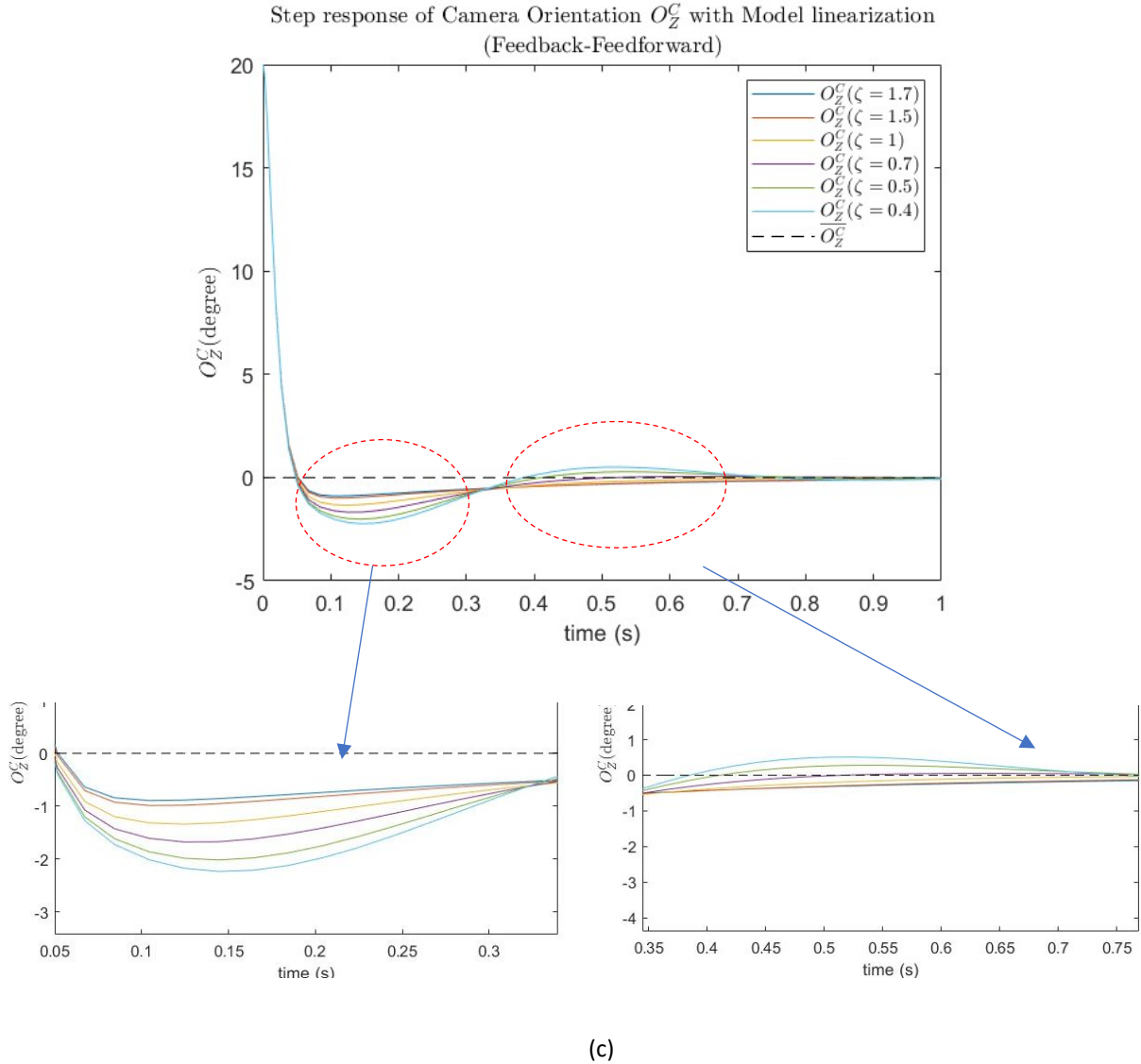

(c)

**Figure 5.16:** Step responses of $O_Z^C$ for the case 1: $\varphi < \frac{\alpha}{2} = 20°$, $d_{q_v} = 0°$. *(a) Model Linearization with only Feedback Controller. (b). Feedback Linearization (c) Model Linearization with Feedback-Feedforward Controller*

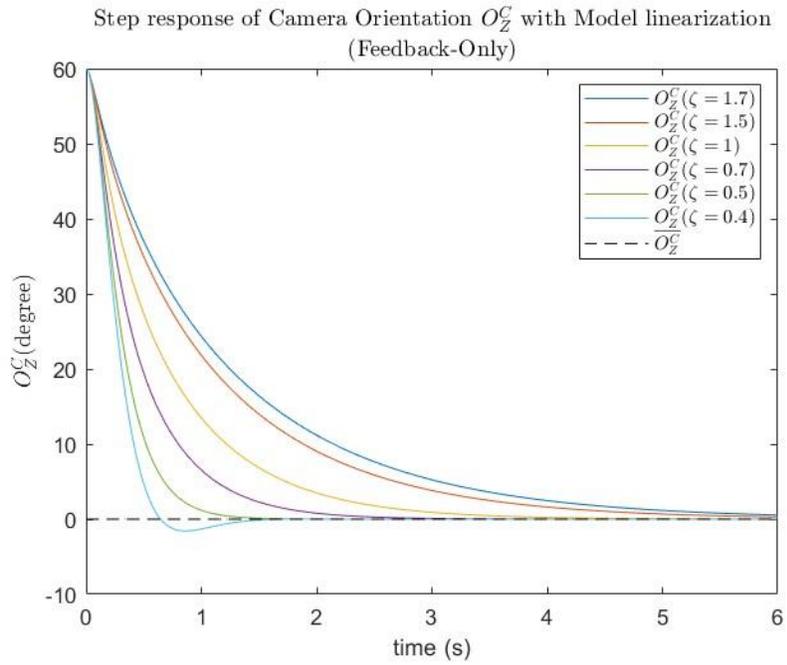

(a)

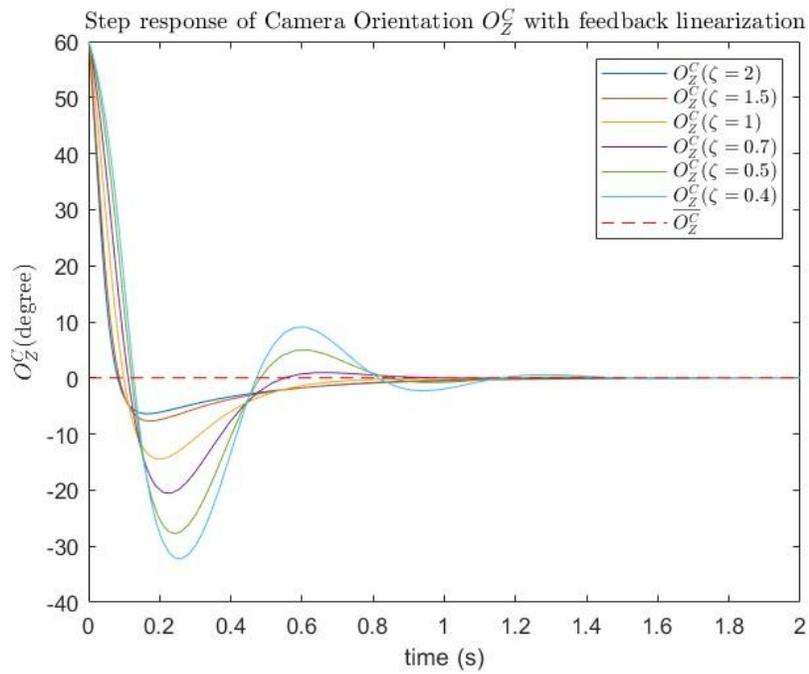

(b)

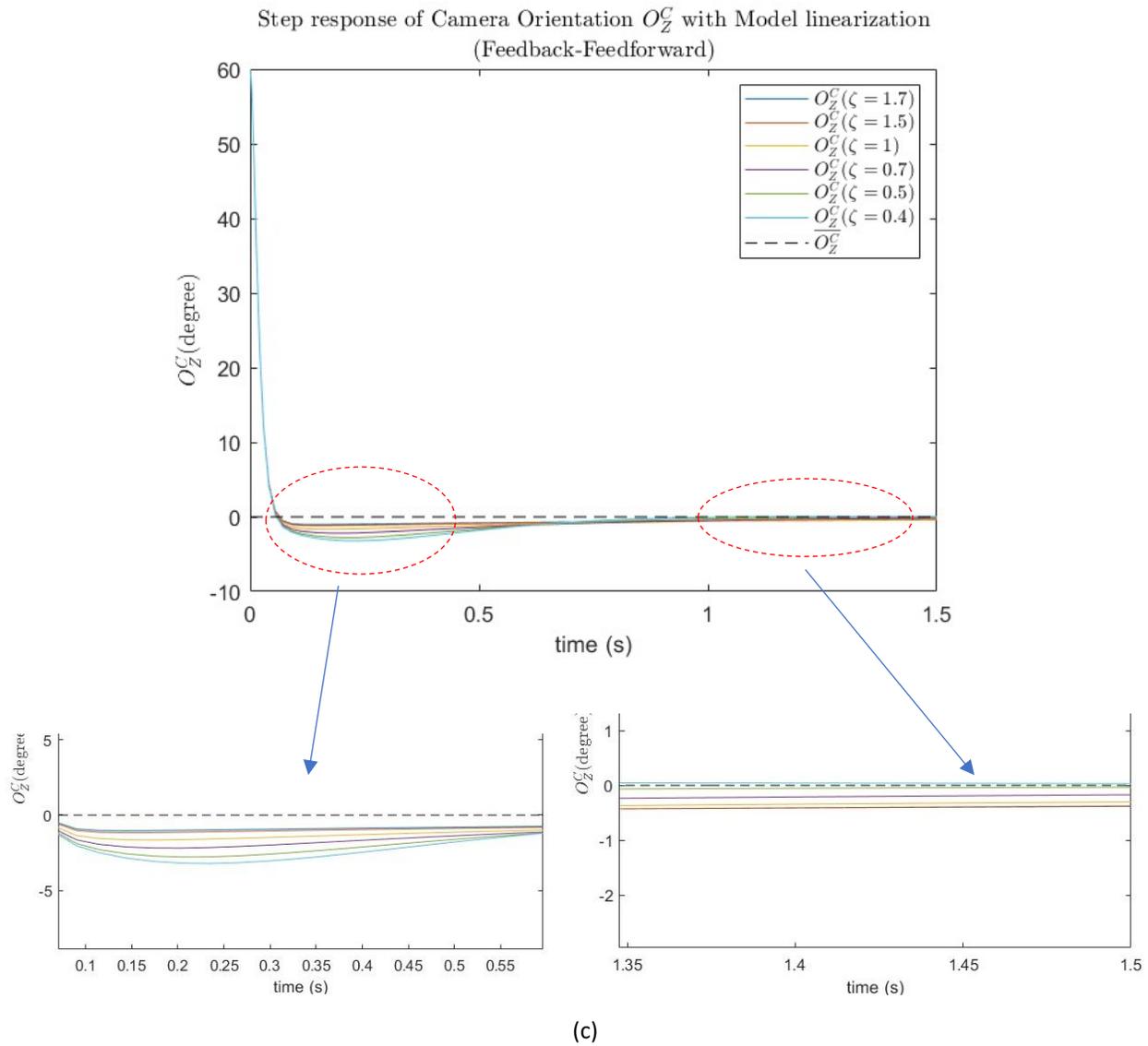

(c)

**Figure 5.17:** Step responses of $O_Z^C$ for the case 2. : $\varphi = \frac{\alpha}{2} = 60°$, $d_{q_v} = 0°$. *(a) Model Linearization with only Feedback Controller. (b). Feedback Linearization (c) Model Linearization with Feedback-Feedforward Controller*

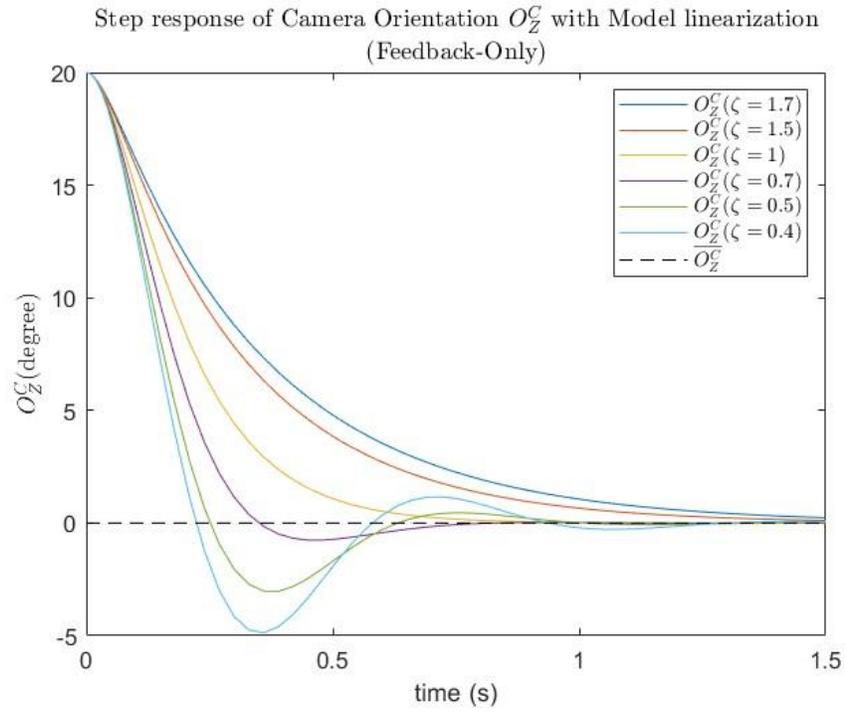

(a)

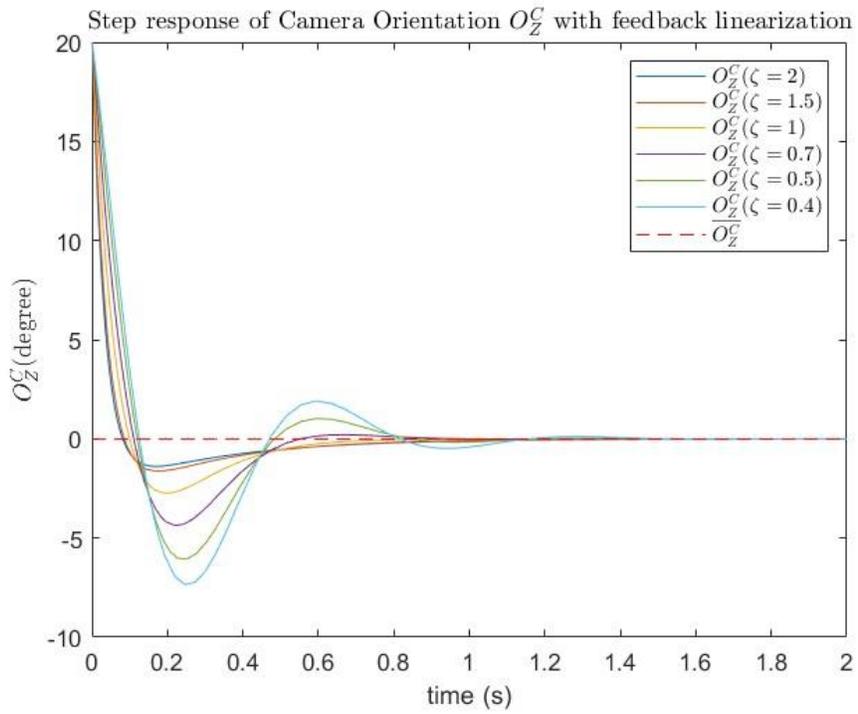

(b)

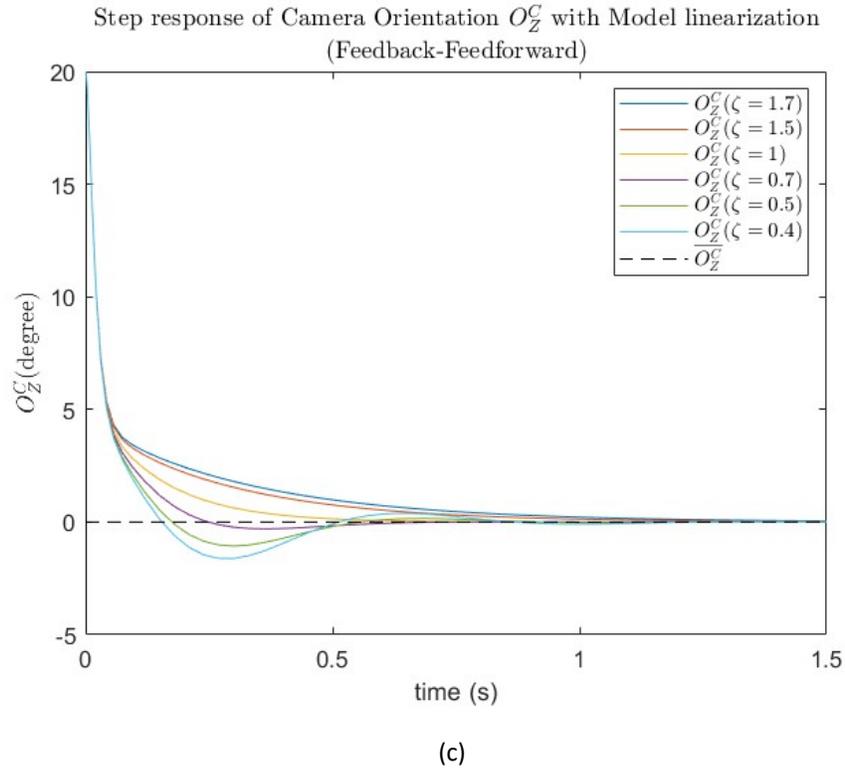

(c)

**Figure 5.18:** Step responses of $O_Z^C$ for the case 3: $\varphi < \frac{\alpha}{2} = 20°$, $d_{q_v} = 5°$. *(a) Model Linearization with only Feedback Controller. (b). Feedback Linearization (c) Model Linearization with Feedback-Feedforward Controller*

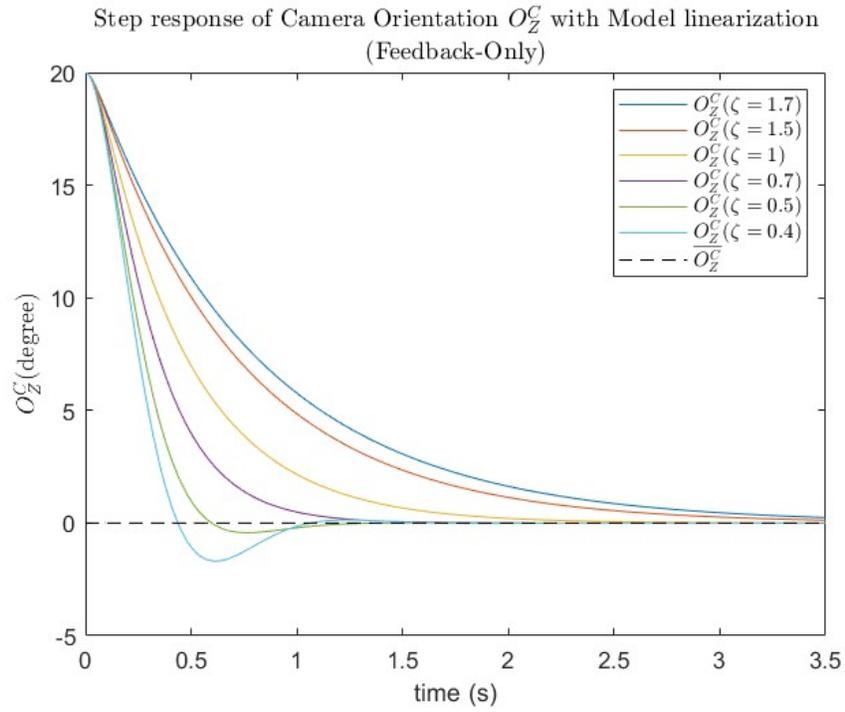

(a)

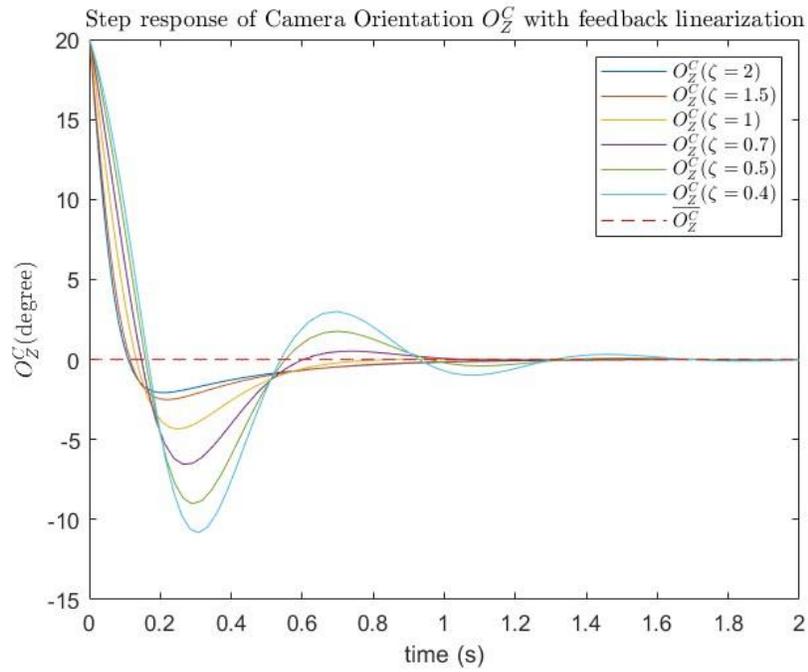

(b)

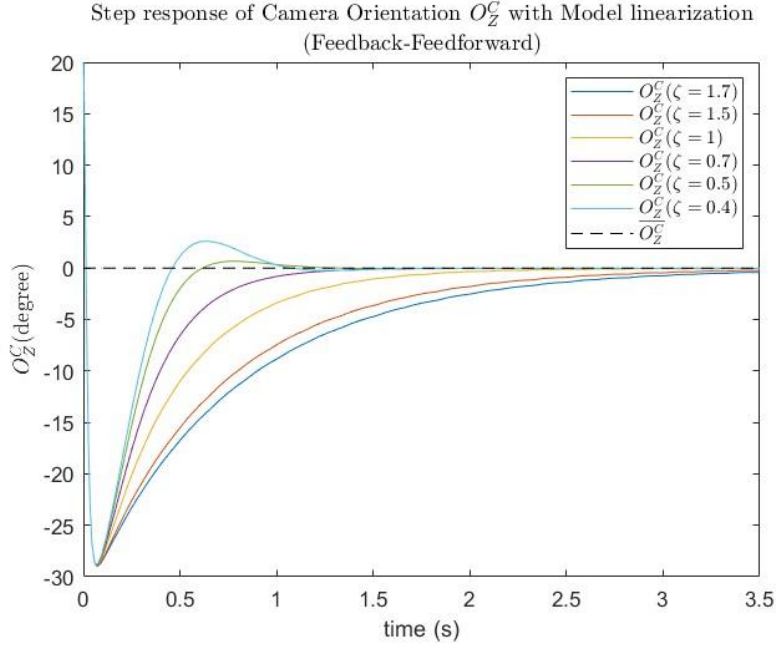

(c)

**Figure 5.19:** Step responses of $O_Z^C$ for the case 4: $\varphi < \frac{\alpha}{2} = 20°$, $d_{q_v} = -30°$. *(a) Model Linearization with only Feedback Controller. (b). Feedback Linearization (c) Model Linearization with Feedback-Feedforward Controller.*

Concluded from simulation results with feedforward controllers, overshoots increase in both cases as $\zeta$ decreases (shown in the enlarged plot on the left). In addition, small undershoot exist for some variations of $\zeta$ (shown in the enlarged plot on the right). More interestingly, a $\zeta_{opt}$ can be found in both cases so that the system reaches the steady state at the fastest rate, and it has a relatively small overshoot and no undershoot among variations of $\zeta$. $\zeta_{opt}$ is found to be approximately unchanged compared to the approach without feedforward control. ($\zeta_{opt} \cong 1$ in the case 1 and $\zeta_{opt} \cong 0.5$ in the case 2).

Therefore, it can be stated that the best performance of the controlled system for the model linearization method is when setting the damping ratio $\zeta = \zeta_{opt}$. Clearly, the value of $\zeta_{opt}$ varies with $\varphi$, the initial orientation of the camera with respect to the inertial frame, which will be further discussed in the following sections.

As for feedback linearization (as shown in part b of Figures), all responses have overall larger overshoots and undershoots than that of model linearization. The responses reach the steady-state very fast but slower than those of the model linearization method with feedforward controllers.

In cases 3 and 4, the input disturbance is introduced to the system. In case 3, a small disturbance ($d_{q_v} = 5°$) is added to the actuator input while in case 4 large disturbance ($d_{q_v} = -30°$) is added. Again, all designed controllers are robust to the input disturbances even with the significantly large disturbances (case 4). Similar to the no disturbance cases, $\zeta_{opt}$ exists for the cases with disturbances. The design incorporating feedforward controllers exhibits reduced response speed and experiences significant overshoot under conditions involving large disturbances. The observed degradation in performance likely stems from delays inherent in the feedforward controllers. The delayed signals from feedforward controllers increase the risetime and make the responses more sensitive to disturbances.

From the analysis above, the plant linearization method with feedforward controller is the preferred and the recommended method for the camera movement adjustment.

## 5.5 Scenario Simulations for the SISO Tool Manipulation System

In this section, we present the simulation results of the control system for the robot tool manipulator. Similar to the visual system, three different controlled systems, namely, the plant linearization method with only feedback, the plant linearization method with a feedforward controller, feedback linearization method, are all simulated and compared in this section. A feedforward controller is not included in the feedback linearization architecture because of the instability issue as discussed in the visual system. The controlled systems are simulated with the original nonlinear Camera-and-Tool Model as expressed in Equation (5.56). Furthermore, we illustrate how varying the damping ratio $\zeta$ affects the responses. We chose the bandwidth of the inner-loop control as $100\, rad/s$ and the bandwidth of the outer-loop control as $10\, rad/s$. The intrinsic camera parameters are chosen as following: $F = 2.8\, mm$ (Focal length) and $\alpha = 120°$ (Angle of view). Other parameters are chosen as: $L_{VT} = 4.2m$, and $L_t = 0.135m$.

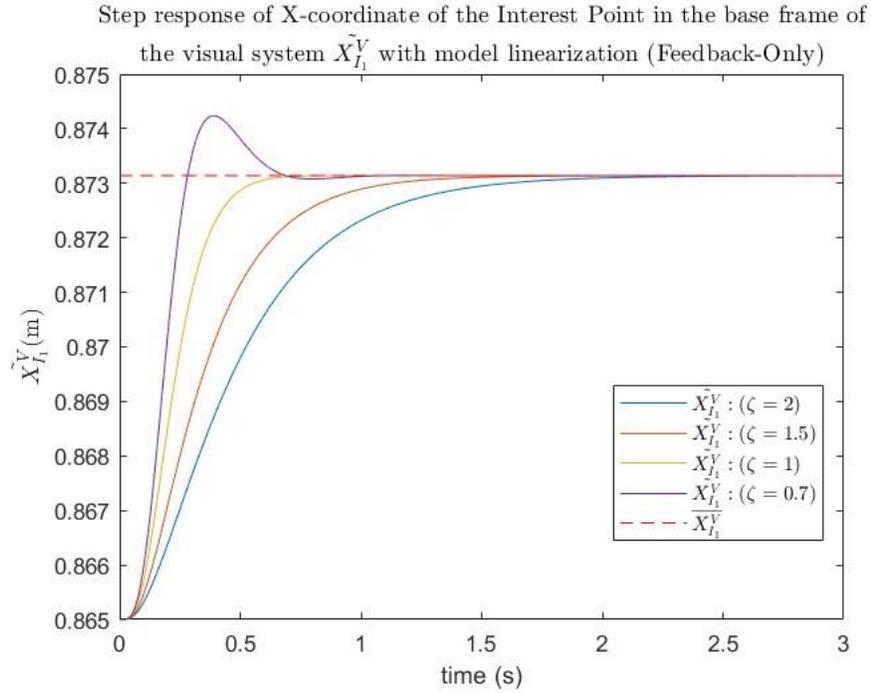

(a)

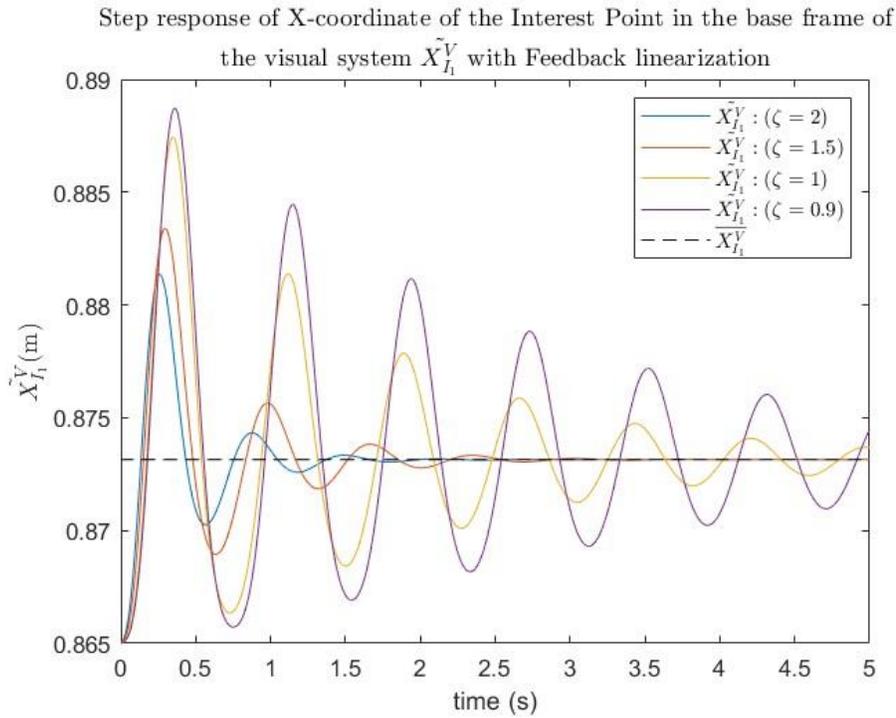

(b)

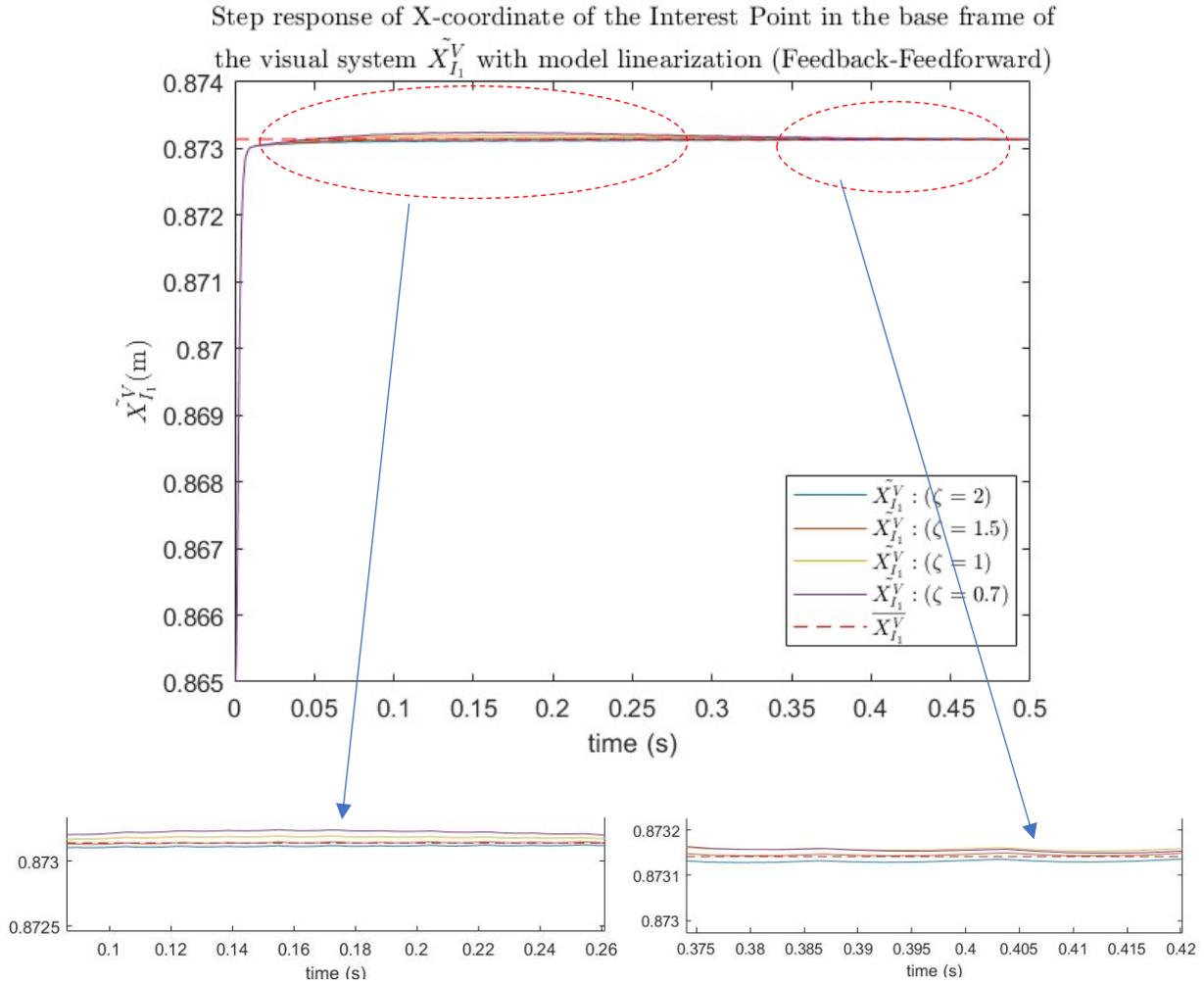

(c)

**Figure 5.20:** Step responses of $\widetilde{X_{I_1}^V}$ for the case 1: $\overline{q_v} = -5°$, $d_{q_T} = -10°$. *(a) Model Linearization with only Feedback Controller. (b). Feedback Linearization (c) Model Linearization with Feedback-Feedforward Controller*

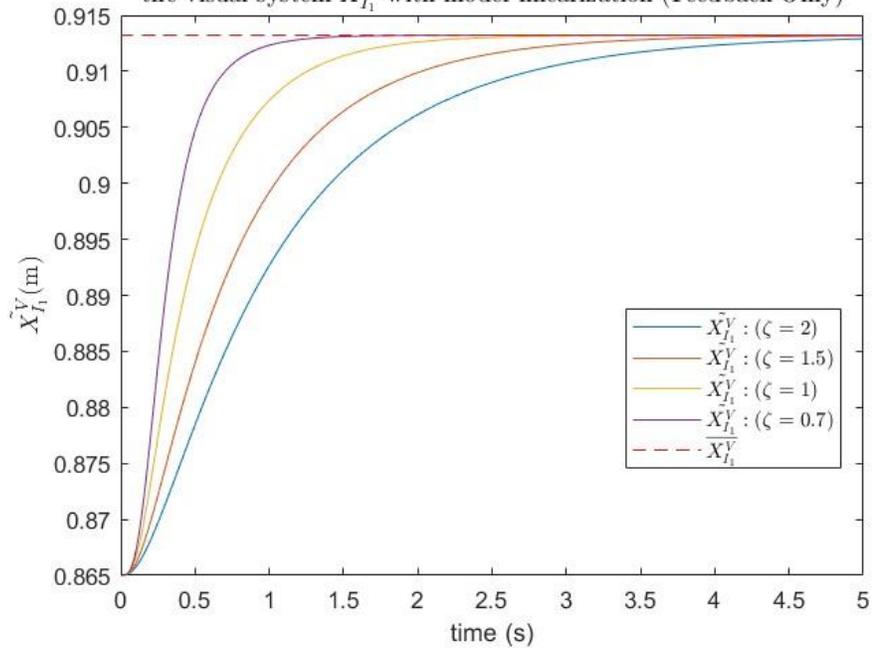

(a)

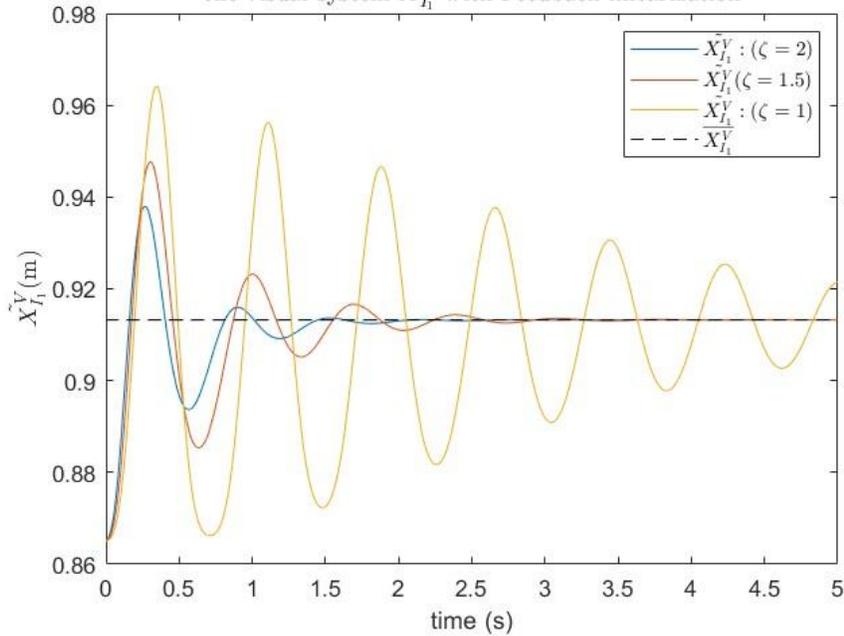

(b)

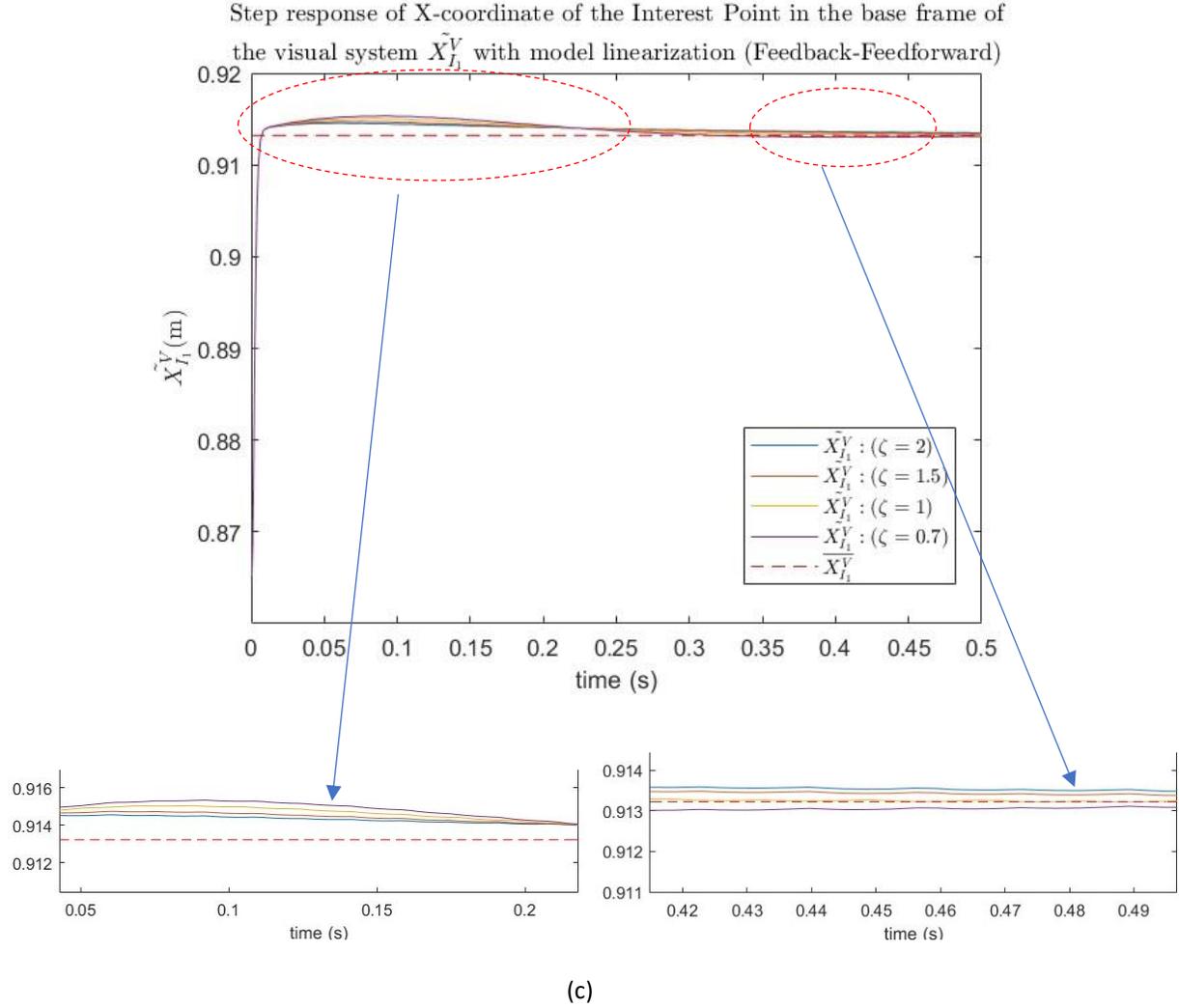

**Figure 5.21:** Step responses of $\widetilde{X_{I_1}^V}$ for the case 2: $\overline{q_v} = -65°$, $d_{q_T} = 15°$. *(a) Model Linearization with only Feedback Controller. (b). Feedback Linearization (c) Model Linearization with Feedback-Feedforward Controller.*

We compare the performance of designed control systems in two different scenarios. One scenario is simulated when the tool is kept inside the camera angle of view during the entire run time. It should be noted that both the feedback and the feedforward controllers are active during the entire simulation. The second scenario is simulated when the tool is outside the camera angle of view during the entire simulation time. When the tool pose is outside the camera angle of view, the feedback signal is replaced by the estimation signal from the model, as shown in Figure 2.6. Assume the initial angle of the tool, $q_{T_0} = 0°$ for all simulation scenarios. Figures 5.20. and 5.21.

present step responses of single output: $\widetilde{X_{I_1}^V}$ (the X coordinate tool's position in the inertial frame) in both scenarios.

In the first scenario, the camera rotates 5° counterclockwise with respect to the inertial frame. Then, the tool stays within the camera angle of view in the whole range of operation as $\overline{q_T} \in [-180°, 180°]$. In addition, a disturbance $d_{q_T} = -10°$, is added to this joint angle. Responses in Figure 5.20. illustrate that all simulated control systems can reach the steady-state and are robust to disturbances. With varying damping ratios, the feedforward-feedback system responses are faster in transient (at least six times faster) when compared to the responses of both feedback-only system in model linearization and feedback linearization. When the damping ratio is small ($\zeta$=0.7), the feedforward-feedback system results in a less overshoot when compared to the response of the feedback-only system for model linearization approach. The optimal damping ratio, $\zeta_{opt} = 1$, (as discussed in Section 5.4), results in the fastest response and no (or little) overshoots, both for the feedforward-feedback and the feedback-only control systems. In this scenario, the feedback linearization responses are the most unstable among all controllers, exhibiting large overshoots and prolonged oscillations before settling. However, their performance improves as the damping ratio $\zeta$ increases.

In the second scenario, the camera rotates 65° counterclockwise with respect to the inertial frame. Using geometry, we can calculate that the tool is out of the camera range of view when $\overline{q_T}$ is not in the range between 35.21° and 134.79°. Staring from the initial angle $q_{T_0} = 0°$ and move to the target joint angle $\overline{q_T} = 50°$ (so that the target coordinate $\overline{X_{I_1}^V} \cong 0.873m$), there is a range $\overline{q_T} \in [0°, 35.21°]$ that camera cannot detect the tool but estimation of the pose is required to drive the tool to the target. Even perturbed with disturbance $d_{q_T} = 15°$, all the simulation responses of model linearization approach shown in Figure 5.21. succeed in stabilizing and reaching the target. Feedback linearization with small damping ratio (which is not shown in the Figure) fails in stability and it still has the worst performance among all three control systems. In this scenario, the feedforward-feedback control system still converges fastest from the transient to the steady state, but generates some overshoots compared to the feedback-only control system, which has no overshoots at all. The overshoot may arise due to cumulative input disturbances that cannot be adequately eliminated by feedforward control alone, thus necessitating the intervention of feedback control.

In conclusion, feedback and feedforward control system designed from model linearization approach provides the best performance among all three controllers. Although this method tends to produce overshoots, its advantage in response speed is significant and cannot be overlooked in real-world applications. In practice, feedback-only control solutions are slower than their simulated counterparts due to additional delays—such as the time required for the camera to capture images—which are not accounted for in simulations. As a result, incorporating a feedforward controller becomes essential in real manufacturing environments, as it helps to compensate for the speed limitations of feedback-only control.

## 5.6 Lyapunov Stability Analysis for the SISO Visual System

In the following parts, Lyapunov analysis is employed to prove the stability of the controlled visual system designed with model linearization. As discussed earlier, for the visual system, $\zeta_{opt}$, the optimal damping ratio is influenced by $\varphi$, the initial angle of the camera with respect to the X-axis of the inertial frame. In this section, we will also derive the mathematical relationship between $\zeta_{opt}$ and $\varphi$ based on Lyapunov analysis.

The controller $Gc_{SISO}^V$ in Equation (5.55) is designed based on linearized camera-on-robot model. The Lyapunov stability analysis is applied to this controller using the original nonlinear model. The block diagram of the controlled system's feedback loop is shown in Figure 5.22.

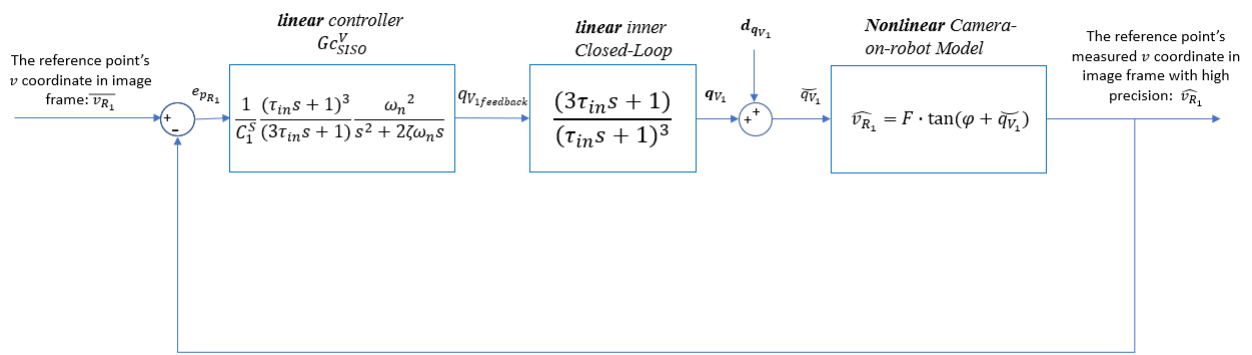

**Figure 5.22:** Block diagram of controlled visual system with model linearization.

Let us derive ordinary differential equation for this closed-loop controlled system:

$$e_{p_{R_1}}(s) = \overline{v_{R_1}}(s) - \widehat{v_{R_1}}(s) \tag{5.58}$$

$$q_{V_1}(s) = Gc_{SISO}^V \cdot T_{inner-close} \cdot e_{p_{R_1}}(s) \tag{5.59}$$

$$\widehat{v_{R_1}} = F \cdot tan(\varphi + \widetilde{q_{V_1}}) \tag{5.60}$$

Without losing generality, set the reference $\overline{v_{R_1}}(s) = 0$. And combine Equations (5.59) and (5.60):

$$q_{V_1}(s) = -Gc_{SISO}^V \cdot T_{inner-close} \cdot \widehat{v_{R_1}}(s) \tag{5.61}$$

By plugging in transfer function of $Gc_{SISO}^V(s)$ and $T_{inner-close}(s)$, Equation (5.61) can be written as:

$$\frac{q_{V_1}(s)}{\widehat{v_{R_1}}(s)} = -\frac{1}{C_1^S}\frac{\omega_n^2}{s^2 + 2\zeta\omega_n s} \tag{5.62}$$

Transform Equation (5.62) to a time domain expression:

$$\ddot{q}_{V_1} + 2\zeta\omega_n \dot{q}_{V_1} = -\frac{1}{C_1^S}\omega_n^2 \widehat{v_{R_1}} \tag{5.63}$$

From Equations (5.60) and (5.63), it can be derived:

$$\ddot{q}_{V_1} + 2\zeta\omega_n \dot{q}_{V_1} = -\frac{1}{C_1^S}\omega_n^2 F \cdot tan(\varphi + q_{V_1} + d_{q_{V_1}}) \tag{5.64}$$

Then the second order nonlinear ODE equation of the closed-loop system can be expressed as:

$$\dot{q}_{V_1} = q_{V_1}^1 = q_{V_1}^2 \tag{5.65}$$

$$\ddot{q}_{V_1} = \dot{q}_{V_1}^2 = -2\zeta\omega_n q_{V_1}^2 - \frac{1}{C_1^S}\omega_n^2 F \cdot tan(\varphi + q_{V_1} + d_{q_{V_1}}) \tag{5.66}$$

System (5.65) and (5.66) is a nonlinear system with $q_{V_1}^1$ and $q_{V_1}^2$ as states and $d_{q_{V_1}}$ as the perturbation to the system. To prove stability from Lyapunov Equations, the equilibrium point should be set to zero. Without losing generality, we can set the equilibrium point of as a new state $\widetilde{q_{V_1}} = 0$, where $\widetilde{q_{V_1}}$ is obtained by translating the state $q_{V_1}$ by a constant, $q_c$; in other words, $\widetilde{q_{V_1}} = q_{V_1} + q_c$. Set $q_c = \varphi$, then the new state is

$$\widetilde{q_{V_1}} = q_{V_1} + \varphi \tag{5.67}$$

Also,

$$\dot{\tilde{q}}_{V_1} = \dot{q}_{V_1}, \text{ and } \ddot{\tilde{q}}_{V_1} = \ddot{q}_{V_1} \tag{5.68}$$

Let us consider the new non-perturbative system by setting $d_{q_{V_1}} = 0$, substitute $C_1 = f_v \cos^{-2}(\varphi)$, and replace $q_{V_1}$, $\dot{q}_{V_1}$ and $\ddot{q}_{V_1}$ with new states $\tilde{q}_{V_1}$, $\dot{\tilde{q}}_{V_1}$ and $\ddot{\tilde{q}}_{V_1}$ in Equations (5.65) and (5.66):

$$\dot{\tilde{q}}_{V_1} = \dot{\tilde{q}}_{V_1}^1 = \tilde{q}_{V_1}^2 \tag{5.69}$$

$$\ddot{\tilde{q}}_{V_1} = \dot{\tilde{q}}_{V_1}^2 = -2\zeta\omega_n \tilde{q}_{V_1}^2 - \cos^2(\varphi)\omega_n^2 \tan(\tilde{q}_{V_1}^1) \tag{5.70}$$

The initial states at $t = 0$ is $\left(\tilde{q}_{V_1}^1, \tilde{q}_{V_1}^2\right) = (\varphi, 0)$, and the equilibriums of this system are $\left(\tilde{q}_{V_1}^1, \tilde{q}_{V_1}^2\right) = (0,0)$.

The first conclusion that can be made from system Equations (5.69) and (5.70) is that:

*Lemma 1:*

***Any Trajectory starts from initial states $\left(\tilde{q}_{V_1}^1, \tilde{q}_{V_1}^2\right) = (\varphi, 0)$ will be asymptotically stable with respective to the equilibrium states $\left(\tilde{q}_{V_1}^1, \tilde{q}_{V_1}^2\right) = (0, 0)$.***

*Proof:*

Let us consider the following Lyapunov equation:

$$V(\tilde{q}_{V_1}) = \frac{1}{2} \tilde{q}_{V_1}^T P \tilde{q}_{V_1} + \cos^2(\varphi)\omega_n^2 \ln\left(\left|\cos^{-1}(\tilde{q}_{V_1}^1)\right|\right)$$

with $\quad \tilde{q}_V = \begin{bmatrix} \tilde{q}_{V_1}^1 \\ \tilde{q}_{V_1}^2 \end{bmatrix}$ and $P = \begin{bmatrix} P_{11} & P_{21} \\ P_{12} & P_{22} \end{bmatrix} \tag{5.71}$

Consider:

$$P_{11} = 2\zeta^2 \omega_n^2, P_{12} = P_{21} = \zeta\omega_n, \text{ and } P_{22} = 1 \tag{5.72}$$

It is straightforward to demonstrate that:

$$P_{11} > 0, P_{11}P_{22} - P_{12}^2 > 0. \tag{5.73}$$

That is the quadratic matrix $P$ in Lyapunov Equation (5.71) is positive definite.

From Equations (5.71) and (5.72), the first derivative of the Lyapunov equation is derived as:

$$\dot{V}(\tilde{q}_{V_1}) = -\zeta\omega_n \cos^2(\varphi)\omega_n^2 \tan(\tilde{q}_{V_1}^1)\tilde{q}_{V_1}^1 - \zeta\omega_n \tilde{q}_{V_1}^{2\,2} \tag{5.74}$$

It can be seen that

$$\dot{V}(\tilde{q}_{V_1}) \leq 0, \text{ for } \forall \left|\tilde{q}_{V_1}^1\right| < \frac{\pi}{2} \tag{5.75}$$

Therefore, taking a set

$$D = \{\tilde{q}_{V_1} \in R^2, \|\tilde{q}_{V_1}^1\| < \frac{\pi}{2}\}, \tag{5.76}$$

It can be concluded that $V(\tilde{q}_{V_1})$ is positive definite and $\dot{V}(\tilde{q}_{V_1})$ is negative definite over $D$ which is the entire operational range of the states.

For any initial state $\left(\tilde{q}_{V_1}^1(0), \tilde{q}_{V_1}^2(0)\right) = (\varphi, 0)$, where $|\varphi| \leq \left|\frac{\alpha}{2}\right| < \frac{\pi}{2}$, and $\alpha$ is the angle of view:

$$V(\tilde{q}_{V_1})|_{(\tilde{q}_{V_1}^1, \tilde{q}_{V_1}^2)} \leq V(\tilde{q}_{V_1})|_{(\tilde{q}_{V_1}^1, \tilde{q}_{V_1}^2)=(\frac{\alpha}{2},0)} \tag{5.77}$$

from the condition (5.75).

Define a set:

$$\Omega_c = \{\tilde{q}_{V_1} \in R^2 | V(\tilde{q}_{V_1}) \leq c\}$$

where
$$c = V(\tilde{q}_{V_1})|_{(\tilde{q}_{V_1}^1, \tilde{q}_{V_1}^2)=(\frac{\alpha}{2},0)} \tag{5.78}$$

The set $\Omega_c$ is bounded and contained in the set $D$, which has been defined in set (5.76). Also, Lyapunov equation $V(\tilde{q}_{V_1})$ satisfies the asymptotically stability conditions over $D$. Then every trajectory starts in $\Omega_c$ remains in $\Omega_c$ and approaches $(0,0)$ as $t \to \infty$.

Thus, by the theorem of Lyapunov stability [78], it can be proven that the origin is globally asymptotically stable Equations (5.65) – (5.66).

***Proof Concluded.***

It can be further stated that for the system (5.65) − (5.66):

*Lemma 2:*

**Any Trajectory starts from initial state $\left(\widetilde{q_{V_1}^1}, \widetilde{q_{V_1}^2}\right) = (\varphi, 0)$ will be exponentially stable with respective to equilibrium state $\left(\widetilde{q_{V_1}^1}, \widetilde{q_{V_1}^2}\right) = (0, 0)$, if $\left|\widetilde{q_{V_1}^1}(0)\right| \leq 76.8°$.**

*Proof:*

To prove Exponential stability, we need to illustrate that Lyapunov equation satisfies:

$$k_1 \|\widetilde{q_{V_1}}\|_2^2 \leq V(\widetilde{q_{V_1}}) \leq k_2 \|\widetilde{q_{V_1}}\|_2^2 \quad (5.79)$$

$$\dot{V}(\widetilde{q_{V_1}}) \leq - k_3 \|\widetilde{q_{V_1}}\|_2^2 \quad (5.80)$$

Where $k_1$, $k_2$, $k_3$ are positive constants, and $\|\widetilde{q_{V_1}}\|_2^2$ is two-norm square of vector $[\widetilde{q_{V_1}^1}, \widetilde{q_{V_1}^2}]$; i.e., $\|\widetilde{q_{V_1}}\|_2^2 = \widetilde{q_{V_1}^1}^2 + \widetilde{q_{V_1}^2}^2$.

It is valid to apply the same Lyapunov Equations (5.71) and (5.72):

$$V(\widetilde{q_{V_1}}) = \frac{1}{2} \widetilde{q_{V_1}}^T P \widetilde{q_{V_1}} + \cos^2(\varphi) \omega_n^2 \ln\left(\left|\cos^{-1}(\widetilde{q_{V_1}^1})\right|\right)$$

where
$$P = \begin{bmatrix} 2\zeta^2 \omega_n^2 & \zeta \omega_n \\ \zeta \omega_n & 1 \end{bmatrix} \quad (5.81)$$

The eigenvalue of $P$ can be expressed as:

$$\lambda_{min} = \frac{1 + 2\zeta^2 \omega_n^2 - \sqrt{1 + 4\zeta^4 \omega_n^4}}{2} \quad (5.82)$$

$$\lambda_{max} = \frac{1 + 2\zeta^2 \omega_n^2 + \sqrt{1 + 4\zeta^4 \omega_n^4}}{2} \quad (5.83)$$

Thus, the first term in $V(\widetilde{q_{V_1}})$:

$$\frac{1}{2} \lambda_{min} \|\widetilde{q_{V_1}}\|_2^2 \leq \frac{1}{2} \widetilde{q_{V_1}}^T P \widetilde{q_{V_1}} \leq \frac{1}{2} \lambda_{max} \|\widetilde{q_{V_1}}\|_2^2 \quad (5.84)$$

Moreover, it is straightforward to determine that the second term of $V(\widetilde{q_{V_1}})$:

$$0 \leq \cos^2(\varphi) \omega_n^2 \ln\left(\left|\cos^{-1}(\widetilde{q_{V_1}^1})\right|\right) \leq 0.825 \cos^2(\varphi) \omega_n^2 \widetilde{q_{V_1}^1}^2, \quad (5.85)$$

$$for\ \forall \left|\widetilde{q_{V_1}^1}\right| \leq 76.8°$$

Combine Equations (5.84) and (5.85):

$$\frac{1}{2}\lambda_{min}\left\|\widetilde{q_{V_1}}\right\|_2^2 \leq V(\widetilde{q_{V_1}}) \leq (\frac{1}{2}\lambda_{max} + 0.825cos^2(\varphi)\omega_n^2)\left\|\widetilde{q_{V_1}}\right\|_2^2, \quad (5.86)$$
$$for\ \forall \left|\widetilde{q_{V_1}^1}\right| \leq 76.8°$$

Consider $k_1 = \frac{1}{2}\lambda_{min}$, $k_2 = \frac{1}{2}\lambda_{max} + cos^2(\varphi)\omega_n^2$. Then the condition in Equation (5.79) is satisfied.

The first derivative of Lyapunov equation is:

$$\dot{V}(\widetilde{q_{V_1}}) = -\zeta\omega_n cos^2(\varphi)\omega_n^2\ tan\left(\widetilde{q_{V_1}^1}\right)\widetilde{q_{V_1}^1} - \zeta\omega_n \widetilde{q_{V_1}^2}^2 \quad (5.87)$$

As $-tan\left(\widetilde{q_{V_1}^1}\right)\widetilde{q_{V_1}^1} \leq -\widetilde{q_{V_1}^1}^2$, we can write:

$$\dot{V}(\widetilde{q_{V_1}}) \leq -\omega_n^2\zeta\omega_n\widetilde{q_{V_1}^1}^2 - \zeta\omega_n\ \widetilde{q_{V_1}^1}^2 \quad (5.88)$$

Then:

$$\dot{V}(\widetilde{q_{V_1}}) \leq - min\ (\zeta\omega_n cos^2(\varphi)\omega_n^2, \zeta\omega_n)\left\|\widetilde{q_{V_1}}\right\|_2^2 \quad (5.89)$$

Therefore, consider $k_3 = - min\ (\zeta\omega_n cos^2(\varphi)\omega_n^2, \zeta\omega_n)$, then condition (5.80) satisfied. Because both conditions in Inequalities (5.79) and (5.80) are satisfied for $\forall\left|\widetilde{q_{V_1}}\right| \leq 76.8°$. Lemma 2 is proved.

***Proof Concluded.***

Notice the condition of Lemma 2 requires that $\left|\widetilde{q_{V_1}}\right| \leq 76.8°$. It can be shown that the condition $\left|\widetilde{q_{V_1}}\right| \leq \frac{\alpha}{2} \leq 76.8°$ must be held. Therefore, the viewing angle is $\alpha \leq 152.6°$, which holds true for the Zed stereo camera used in Appendix A Table A3.

From the proof of lemma 2, the bounds of Inequalities (5.79) and (5.80) are:

$$k_1 = \frac{1}{2}\lambda_{min}$$

$$k_2 = \frac{1}{2}\lambda_{max} + 0.825\cos^2(\varphi)\omega_n^2$$
$$k_3 = \min\,(\zeta\omega_n\cos^2(\varphi)\omega_n^2, \zeta\omega_n) \tag{5.90}$$

From Inequalities (5.79) and (5.80), it can be found:

$$\frac{1}{k_3}\dot{V}(\widetilde{q_{V_1}}) \leq -\|\widetilde{q_{V_1}}\|_2^2 \tag{5.91}$$

$$-\frac{1}{k_2}V(\widetilde{q_{V_1}}) \geq -\|\widetilde{q_{V_1}}\|_2^2 \tag{5.92}$$

Therefore:

$$\dot{V}(\widetilde{q_{V_1}}) \leq -\frac{k_3}{k_2}V(\widetilde{q_{V_1}}) \tag{5.93}$$

Also:

$$\|\widetilde{q_{V_1}}\|_2^2 \leq \frac{1}{k_1}V(\widetilde{q_{V_1}}) \tag{5.94}$$

From Inequality (5.93):

$$\frac{1}{k_1}V(\widetilde{q_{V_1}}) \leq \frac{1}{k_1}e^{-\frac{k_3}{k_2}t}V(\widetilde{q_{V_1}}(0)) \tag{5.95}$$

From Inequalities (5.94) and (5.95):

$$\|\widetilde{q_{V_1}}\|_2^2 \leq \frac{1}{k_1}e^{-\frac{k_3}{k_2}t}V(\widetilde{q_{V_1}}(0)) \leq \frac{k_2}{k_1}e^{-\frac{k_3}{k_2}t}\|\widetilde{q_{V_1}}(0)\|_2^2$$

Then
$$\|\widetilde{q_{V_1}}(t)\|_2 \leq \sqrt{\frac{k_2}{k_1}}\,e^{\frac{-k_3}{2k_2}t}\|\widetilde{q_{V_1}}(0)\|_2 \tag{5.96}$$

It can be observed from Inequality (5.96) that there is an estimated exponential decaying rate:

$$\beta_{est} = \frac{k_3}{2k_2} = \frac{\min\,(\zeta\omega_n\cos^2(\varphi)\omega_n^2, \zeta\omega_n)}{\lambda_{max} + 1.71\cos^2(\varphi)\omega_n^2} \tag{5.97}$$

The unit is $(s^{-1})$.

Notice Equation (5.97) is only an estimation and if the bound in Inequalities (5.79) and (5.80) are tighter, the estimated decaying rate $\beta_{est}$ will become more accurate. $\beta_{est}$ describes the rate at which the decays occur (response speed from initial states to steady states). If the parameter $\omega_n$ is selected to define the bandwidth, the damping ratio $\zeta$ will also influence the behavior of $\beta_{est}$. The simulation results have already shown the impact of different $\zeta$ on the closed-loop system response.

Equation (5.97) shows the convergence rate $\beta_{est}$ is a function of damping ratio $\zeta$ and initial angle $\varphi$, if we keep other parameters constant. We can take the first derivative of Equation (5.97) with respective to $\zeta$ to find $\zeta_{opt}$ as following,

$$\zeta_{opt} = \arg_{\zeta} \left( \frac{\delta \beta_{est}}{\delta \zeta} \bigg|_{\varphi=\varphi_s, \omega_n=10 rad/s} = 0 \right) \quad (5.98)$$

where $\varphi_s$ is the specific $\varphi$ value. Assuming a constant $\omega_n$, $\zeta_{opt}$ is the solution of Equation (5.98) with a specific $\varphi_s$. The true optimal damping ratio $\zeta_{opt}$ can be found by brute-force search using simulation. The simulation results are used as the benchmark for verification of Equation (5.98). The benchmark simulation is performed by varying $\varphi$ in the range [1° : 1° : 60°] and $\zeta$ in the range [0.1:0.01:1]. Figure 5.23. shows how $\zeta_{opt}$ and $\zeta_{opt}$ estimated from Equation (5.98) varies as a function of $\varphi$.

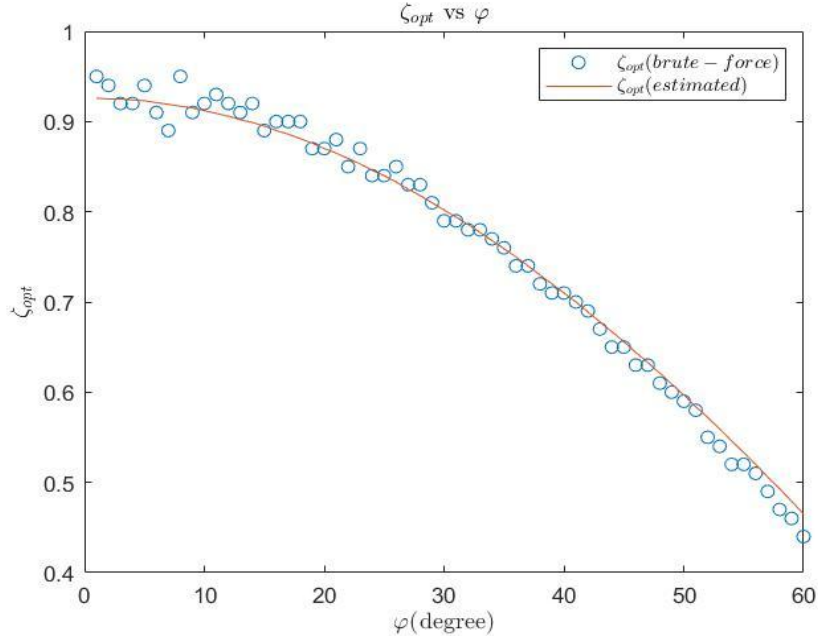

**Figure 5.23**: Comparison between $\zeta_{opt}$ and $\zeta_{opt}$ **estimated** from Equation (5.98).

Figure 5.23. verifies that Equation (5.98) seems to be an accurate method to find an estimation of $\zeta_{opt}$ for varying initial image angle ($\varphi$). This numerical method of estimating $\zeta_{opt}$ can be used in the design of the Youla controller described in Equation (5.55) so that it provides the best performance compared to other designs with damping ratios.

## 5.7 Conclusion

In this chapter, we first developed a 6-DOF inner-loop controller to guide the robotic manipulator's joints to their desired positions. This controller was designed using Youla parameterization in combination with feedback linearization. Subsequently, a SISO outer-loop controller was developed for both the visual and tool manipulation systems. Three control strategies were explored: (1) feedforward and feedback with model linearization, (2) feedback-only with model linearization, and (3) feedback linearization. Each method was mathematically formulated and evaluated based on its performance. Among these, the feedforward and feedback approach with model linearization yielded the best results, demonstrating minimal overshoot and the fastest response in both systems. Additionally, for the visual system, the optimal damping ratio—ensuring the fastest response without overshoot—was derived through Lyapunov analysis. The development of these SISO outer-loop controllers establishes a solid foundation for the design of MIMO outer-loop controllers in the following chapters.

_______________________________________________________________Chapter 6

# MIMO Outer Controller Designs

As discussed in Chapter 5, feedforward-feedback control design from model linearization provides optimal performance and robustness against disturbances in SISO cases. To utilize the analysis of SISO simulations and move on to MIMO systematic control design, only model linearization method with the feedforward-feedback control architecture will be developed and discussed in this chapter for both the visual system and the tool manipulation system. Due to the underactuated nature of the MIMO system, the discussion of controllers' design for the visual system is more complex than that of the tool manipulation system. For that reason, section 6.1 first discusses the MIMO controller design of IBVS structure in the tool manipulation system, and then section 6.2 addresses the MIMO controller design of IBVS structure in the visual system. Additionally, Section 6.3 addresses the issue of kinematic singularities, which can lead to loss of controllability and instability near rank-deficient configurations of the manipulator. A damped least-squares-based inverse kinematic solution with smooth switching is proposed to ensure stable controller behavior even in the vicinity of singular poses.

_______________________________________________________________

# 6.1 The Outer Loop Controller Design for MIMO Tool Manipulation System

In control system block diagram (Figure 2.6.), the linear closed loop transfer function of inner-loop joint control is derived in Equation (5.14). The Tool-on-Robot kinematics model and the Camera-on-Robot kinematics, which can be combined as a new model: Camera-and-Tool Combined Model, compose of the nonlinear part of the plant in outer feedback loop. Assume a screwdriver with a length of $L_{tool}$ is grasped by the gripper at the end effector, and the body of screwdriver is parallel to the roll axis of the end-effector of the robot manipulator. Figure 3.15. illustrates the combined Camera-and-Tool model, and for reference, it is reproduced in Figure 6.1. to support the discussions in this section.

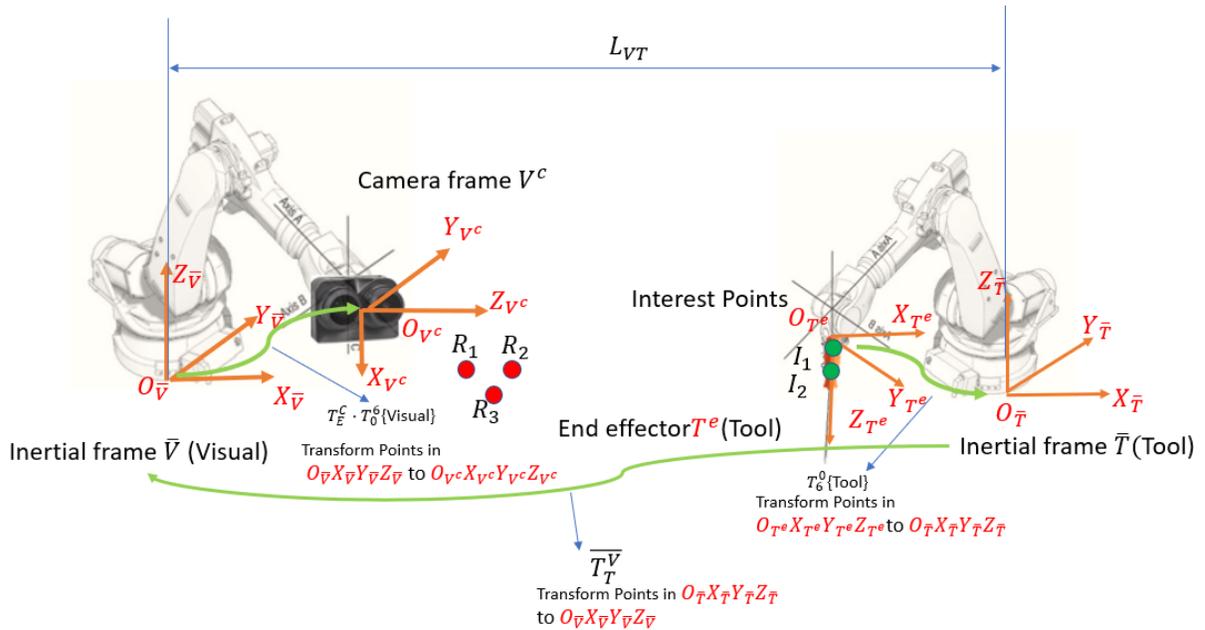

**Figure 6.1**: Camera-and-Tool combined model.

Two fiducial feature points attached to it are placed on the screwdriver where their coordinates measured in the end-effector frame as $P_{I_1}^{T^e} = [0,0,0]^T$, and $P_{I_2}^{T^e} = [0,0,L_{tool}/2]^T$. Equations (3.157) and (3.155) express a nonlinear function denoted as $\mathcal{H}$ that relates the perturbed robot joint angles $\widetilde{q_T}$ and 2D image coordinates of two fiducial points $\widehat{p_I}$, and we can rewrite those as follows:

$$\text{For any time } t \geq 0, \widehat{p_I} = \mathcal{H}(\widetilde{q_T}(t), P_{a_{tool}}, P_I^{T^e}) \tag{6.1}$$

$$P_{a_{tool}} = [P_{a_{camera}}, P_{a_{robot}}, \overline{T_E^C}, \overline{T_T^V}, \overline{q_V}] \tag{6.2}$$

Parameters $P_{a_{tool}}$, includes stereo camera parameter $P_{a_{camera}}$, robot geometric parameter $P_{a_{robot}}$, and transformation matrix $\overline{T_E^C}, \overline{T_T^V}$ and the optimal joint angles $\overline{q_V}$.

### 6.1.1 Feedback Controller Design with Model Linearization

We can use model linearization method by linearizing the nonlinear function (6.1) at different linearized points and design linear controllers utilizing these linearized models. By chosen a specific linearized point $\widetilde{q_T}^0$, the nonlinear function (6.1) can be linearized using the Jacobian matrix, expressed as:

$$\widehat{p_I} = J_{\mathcal{H}}(\widetilde{q_T}^0)\widetilde{q_T} + \mathcal{H}(\widetilde{q_T}^0) \tag{6.3}$$

Where $J_{\mathcal{H}}(\widetilde{q_T}^0) \in \mathbb{R}^{6 \times 6}$ is the Jacobian matrix of $\mathcal{H}(\widetilde{q_T})$ evaluated as $\widetilde{q_T} = \widetilde{q_T}^0$.

Assuming $K_1^M = J_{\mathcal{H}}(\widetilde{q_T}^0)$, $K_2^M = \mathcal{H}(\widetilde{q_T}^0)$, therefore, Equation (6.3) can be rewritten as:

$$\widehat{p_I} = K_1^M \widetilde{q_T} + K_2^M \tag{6.4}$$

Define $\widehat{p_I}' = \widehat{p_I} - K_2^M$, treating $K_2^M$ as an input disturbance, then, the overall block diagram of the linearized system is shown in Figure 6.2. It should be noted that at singular configurations of the robot, the Jacobian matrix $J_{\mathcal{H}}(\widetilde{q_T}^0)$ becomes rank-deficient, resulting in a loss of degrees of freedom and potential instability in controller design. This issue will be examined in detail in Section 6.3, where we also propose a method to address the control challenges arising from such singularities.

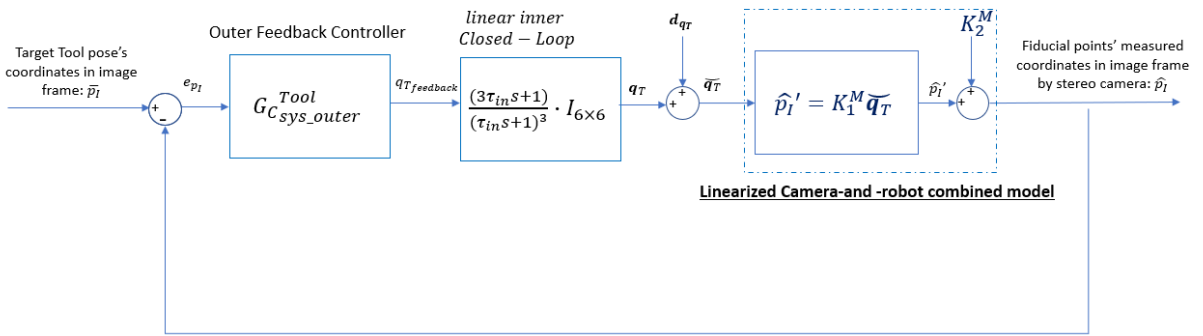

**Figure 6.2:** Feedback loop with linearized camera-and-robot combined model.

The linearized plant transfer function is derived as:

$$G_{p_{sys_{outer}}}^{Tool}(Linear) = \frac{\widehat{p_I}'}{q_{T_{ref}}} = K_1^M \frac{(3\tau_{in}s + 1)}{(\tau_{in}s + 1)^3} \cdot I_{6\times 6} \tag{6.5}$$

As $K_1^M$ is coupled, the first step to derive a controller for the multivariable system using model linearization is to find the Smith-McMillan form of the plant [77]. This form allows the multivariable system to be decoupled through similarity transformations:

$$G_{p_{sys_{outer}}}^{Tool}(Linear) = U_L^{Tool} M_p^{Tool} U_R^{Tool} \tag{6.6}$$

where $U_L^{Tool} \in \mathbb{R}^{6 \times 6}$ and $U_R^{Tool} \in \mathbb{R}^{6 \times 6}$ are the left and right unimodular matrices, and $M_p^{Tool} \in \mathbb{R}^{6 \times 6}$ is the Smith-McMillan form of $G_{p_{sys_{outer}}}^{Tool}(Linear)$.

To compute this, we first perform a singular value decomposition (SVD) [77] of the matrix $K_1^M$:

$$K_1^M = U^{Tool} \Sigma^{Tool} (V^{Tool})^T \tag{6.7}$$

Here, $U^{Tool} \in \mathbb{R}^{6 \times 6}$ is an orthogonal matrix of left singular vectors, $V^{Tool} \in \mathbb{R}^{6 \times 6}$ is an orthogonal matrix of right singular vectors, and $\Sigma^{Tool} \in \mathbb{R}^{6 \times 6}$ is a diagonal matrix containing the singular values of $K_1^M$.

Substituting (6.7) into (6.5) yields:

$$G_{p_{sys_{outer}}}^{Tool}(Linear) = U^{Tool} \Sigma^{Tool} (V^{Tool})^T \frac{(3\tau_{in}s + 1)}{(\tau_{in}s + 1)^3} \cdot I_{6\times 6} \tag{6.8}$$

Since both $\Sigma^{Tool}$ and $\frac{(3\tau_{in}s+1)}{(\tau_{in}s+1)^3} \cdot I_{6\times 6}$ are diagonal matrices, they commute and we can rewrite the expression as:

$$G_{p_{sys_{outer}}}^{Tool}(Linear) = U^{Tool} (\Sigma^{Tool} \frac{(3\tau_{in}s + 1)}{(\tau_{in}s + 1)^3} \cdot I_{6\times 6})(V^{Tool})^T \tag{6.9}$$

Identifying this with the Smith-McMillan form yields:

$$U_L^{Tool} = U^{Tool} \tag{6.10}$$

$$M_p^{Tool} = \Sigma^{Tool} \frac{(3\tau_{in}s + 1)}{(\tau_{in}s + 1)^3} \cdot I_{6\times 6} \tag{6.11}$$

$$U_R^{Tool} = (V^{Tool})^T \tag{6.12}$$

Equations (6.9) – (6.12) demonstrate the development of the Smith-McMillan form the singular value decomposition.

Each nonzero element in the diagonal matrix $M_p^{Tool}$ is in the form:

$$M_p^{Tool}(i,i) = \Sigma^{Tool}(i,i) \cdot \frac{(3\tau_{in}s + 1)}{(\tau_{in}s + 1)^3}, i\epsilon(1,2,3,4,5,6) \quad (6.13)$$

The design of a Youla controller for each nonzero entry in $M_p^{Tool}$ is trivial in this case as all poles/zeros of the plant transfer function matrix are in the left half-plane, and therefore, they are stable. In this case, the selected decoupled Youla transfer function matrix, $M_Y^{Tool}$, can be selected to shape the decoupled closed loop transfer function matrix, $M_T^{Tool}$. All poles and zeros in the original plant can be cancelled out and new poles and zeros can be added to shape the closed-loop system. Let's select a Youla transfer function matrix so that the decoupled closed-loop SISO system behaves like a second order Butterworth filter, such that:

$$M_T^{Tool} = \frac{\omega_n^2}{(s^2 + 2\zeta\omega_n s + \omega_n^2)} \cdot I_{6\times 6} \quad (6.14)$$

where $\omega_n$ is called natural frequency and approximately sets the bandwidth of the closed–loop system. It must be ensured that the bandwidth of the outer-loop is smaller than the inner-loop, i.e., $1/\omega_n > \tau_{in}$. Parameter, $\zeta$, is called the damping ratio, which is another tuning parameter.

We can then compute the decoupled diagonalized Youla transfer function matrix $M_Y^{Tool}$. The diagonal entry of $i^{th}$ row is denoted as $M_Y^{Tool}(i,i)$:

$$M_Y^{Tool}(i,i) = \frac{M_T^{Tool}(i,i)}{M_p^{Tool}(i,i)}$$

$$= \frac{1}{\Sigma^{Tool}(i,i)} \frac{\omega_n^2}{(s^2 + 2\zeta\omega_n s + \omega_n^2)} \frac{(\tau_{in}s + 1)^3}{(3\tau_{in}s + 1)}, i\epsilon(1,2,3,4,5,6) \quad (6.15)$$

The final coupled Youla, closed loop, sensitivity, and controller transfer function matrices are computed as:

$$Y_{sys\_outer}^{Tool} = U_R^{Tool} M_Y^{Tool} U_L^{Tool} \quad (6.16)$$

$$T_{y_{sys\_outer}}^{Tool} = G_{p_{sys\_outer}}^{Tool} \cdot Y_{sys\_outer}^{Tool} \quad (6.17)$$

$$S_{y_{sys\_outer}}^{Tool} = 1 - T_{y_{sys\_outer}}^{Tool} \quad (6.18)$$

$$G_{C_{sys\_outer}}^{Tool} = Y_{sys\_outer}^{Tool} \cdot (S_{y_{sys\_outer}}^{Tool})^{-1} \quad (6.19)$$

The controller developed in the above section is based on the linearization of the combined model at a particular linearized point $\widetilde{q_T}^0$. This controller can only stabilize at certain range of joint angles around $\widetilde{q_T}^0$. To address this issue, we propose an adaptive controller that is updated online based on local linearization at the current joint configuration. This adaptive control framework is illustrated in Figure 6.3.

The first step in this adaptive strategy is to estimate the current joint angles $\widetilde{q_T}$ from the current measured images coordinates $\hat{p}_I$. Using the stereo vision model and robot forward kinematics presented in earlier chapters, the combined nonlinear model has been formulated in Equation (6.1). The inverse model, denoted as $\hbar$, defined in Equation (3.164), allows us to compute the estimated joint angles as:

$$\forall\, t \geq 0, \widetilde{q_T}(t) = \hbar(\hat{p}_I(t), P_{a_{tool}}, P_I^{T^e}) \quad (6.20)$$

Where $\hbar$ is the inverse function of Equation (6.1), $P_{a_{tool}}$ is defined in Equation (6.2), and $P_I^{T^e}$ are fiducial points measured in the end-effector frame of the tool manipulation system. As discussed in Section 3.6.4, the estimated joint configuration $\widetilde{q_T}$ exists, but is generally not unique, due to the non-uniqueness of the inverse kinematics problem. However, a specific solution can be selected based on the disambiguation strategy outlined in Table 3.2.

Once the current estimate $\widetilde{q_T}$, is obtained, we compute the Jacobian matrix of the nonlinear model at this configuration. Applying singular value decomposition (SVD) to the Jacobian allows us to extract the left and right unimodular matrices and derive the Smith-McMillan form, as discussed in Equations (6.6) – (6.13). This decomposition enables construction of a locally linearized controller $G_{C_{sys\_outer}}^{Tool}$ that adapts in real-time to the system's changing dynamics.

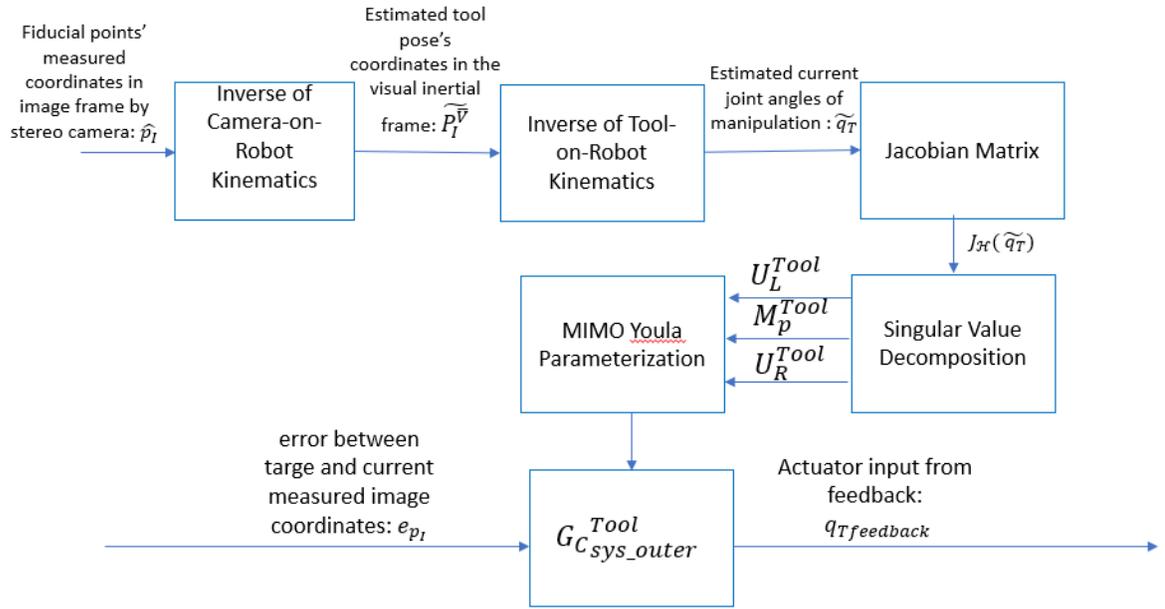

**Figure 6.3**: Block diagram of adaptive control design for the tool manipulation system.

In this framework, the adaptive outer-loop controller is computed online using the current linearized plant model, which is evaluated at the current estimated joint angles of the robot. This ensures that the controller continuously adapts to the system's configuration. The mapping between the nonlinear plant and its linearized representation is valid across the entire joint space, thereby enabling consistent linearization at any configuration. As a result, the adaptive feedback controller maintains robust performance throughout the full range of robot operations.

### 6.1.2 Feedforward Controller Design

The Feedforward Controller generates a target joint angle, $q_{Tfeedforward}$ based on the desired tool coordinates at the target location. Similar to the SISO case, the feedforward controller incorporates two inverse processes derived from the original plant model. The first inverse process corresponds to the tool-on-robot kinematics, which maps the desired end-effector pose to the required joint configuration. The second inverse process, denoted as $T_{MIMO}^{forward}$, represents the inverse of the inner-loop closed-loop transfer function, $T_{MIMO}^{inner}$. These processes are illustrated in the corresponding block diagrams Figure 6.4. and form the core of the feedforward controller's predictive action.

Equations (3.113)–(3.116) define the mathematical formulation of the tool-on-robot kinematics model. The inverse of this model, which is essential for computing the required joint configurations from a desired tool pose, is expressed as:

$$\widetilde{q_T} = \Gamma(\overline{P_I^V}, \overline{T_T^V}, P_{a_{robot}}, \overline{P_I^{T^e}}) \tag{6.21}$$

Where $\Gamma$ is a nonlinear function representing the inverse of the tool-on-robot kinematics model. $\overline{P_I^V}$ denotes the estimated coordinates of visual markers in the visual inertial frame. $\overline{T_T^V}$ is the constant transformation matrix from the tool inertial frame to the visual inertial frame. $\overline{P_I^{T^e}}$ are the coordinates of fiducial markers measured in the end-effector (tool) frame, and $P_{a_{robot}}$ represents the parameter set describing the robot's kinematic structure.

This inverse computation yields the joint angle configuration $\widetilde{q_T}$ necessary for achieving the desired tool pose in the visual frame, forming a critical component of the feedforward control strategy.

The forward transfer function in the feedforward control path, denoted as $T_{MIMO}^{forward}$, is constructed as the inverse of the inner-loop closed-loop transfer function $T_{MIMO}^{inner}$, augmented with a second-order low-pass filter to ensure properness. It is defined as:

$$T_{MIMO}^{forward} = \frac{1}{T_{MIMO}^{inner}} \frac{1}{(\tau_{forward}s+1)^2} = \frac{(\tau_{in}s+1)^3}{(3\tau_{in}s+1)} \frac{1}{(\tau_{forward}s+1)^2} \cdot I_{6\times 6} \tag{6.22}$$

To make $T_{forward}^{MIMO}$ proper, two additional poles at s = $-1/\tau_{forward}$ are introduced through the second-order filter. The time constant $\tau_{forward}$ is selected to be ten times smaller than the original system's time constant $\tau_{in}$, ensuring that the added poles lie significantly beyond the system's bandwidth and do not interfere with primary system dynamics. This is given by:

$$\tau_{forward} = 0.1\tau_{in} \tag{6.23}$$

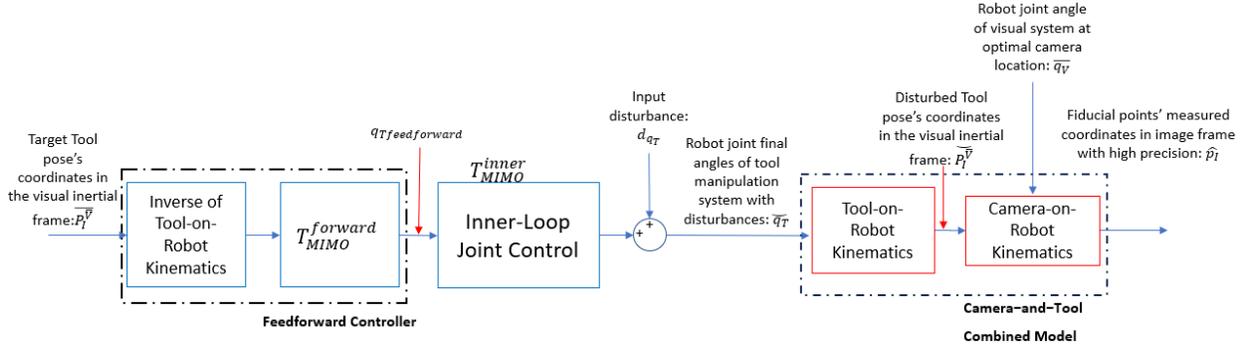

**Figure 6.4:** Feedforward control design for the MIMO tool manipulation system.

### 6.1.3 2D Feature Estimation

Since the camera is stationary and does not move with the robot arm, the pose of the tool cannot be directly measured when it moves outside the camera's field of view. To address this limitation, the 2D image features (i.e., the image-plane coordinates of the tool's fiducial points) can be estimated using the same combined kinematic-visual model defined in Equation (6.1), with the joint angles $q_T$ as input. This estimation process is illustrated in Figure 2.6., and its mathematical formulation, previously given in Equation (3.156), can be restated as:

$$\forall\, t \geq 0,\ \widetilde{p}_I(t) = \mathcal{H}(q_T(t), P_{a_{tool}}, \overline{P_I^{T^e}}) \tag{6.24}$$

Where $\widetilde{p}_I(t)$ denotes the estimated image coordinates of the tool markers at time $t$. $\mathcal{H}(\cdot)$ is the Camera-and-Tool Combined model that maps the joint angles of robot manipulators to the estimated image coordinates. $P_{a_{tool}}$ contains all parameters in the Camera-and-Tool Combined model, and $\overline{P_I^{T^e}}$ refers to the known positions of the markers in the end-effector frame.

This model-based estimation allows visual feedback to continue even when direct visual tracking is temporarily unavailable.

## 6.2 The Outer Loop Controller Design for MIMO Visual System

Figure 2.4. illustrates the complete block diagram of the control architecture for the visual servoing system. Within this framework, Equation (5.14) defines the inner-loop closed-loop transfer function of the joint-level controller, which represents the linear component of the plant. In contrast, the camera-on-robot kinematics model introduces a nonlinear component, accounting for the transformation between joint motion and observed image features.

Assume the 3D positions of three reference points $R_1$, $R_2$, and $R_3$ in the visual inertial frame are known and denoted as: $\overline{P_{R_1}^V} = (\overline{X_{R_1}^V}, \overline{Y_{R_1}^V}, \overline{Z_{R_1}^V}, 1)$, $\overline{P_{R_2}^V} = (\overline{X_{R_2}^V}, \overline{Y_{R_2}^V}, \overline{Z_{R_2}^V}, 1)$, and $\overline{P_{R_3}^V} = (\overline{X_{R_3}^V}, \overline{Y_{R_3}^V}, \overline{Z_{R_3}^V}, 1)$. By combining Equations $(3.113) - (3.130)$, the Camera-on-robot Kinematics model can be formulated as a nonlinear function $\Psi$, which maps the robot's joint angles to the predicted image coordinates of these reference points. Using Equations (3.130) and (3.132), the model is expressed as:

$$\forall\, t \geq 0,\; \widehat{p_R}(t) = \Psi(\widetilde{q_V}(t), P_{a_{visual}}, P_R^{\overline{V}}) \qquad (6.25)$$

Where $\widehat{p_R}(t)$ denotes the measured image coordinates of the reference points at time $t$. $\widetilde{q_V}(t)$ is the vector of perturbed joint angles in the visual robot manipulator. $P_R^{\overline{V}}$ is the set of known 3D coordinates of the reference points in the inertial frame, and parameters $P_{a_{visual}}$ is the visual system parameter set, defined as:

$$P_{a_{visual}} = [P_{a_{camera}}, P_{a_{robot}}, \overline{T_E^C}] \qquad (6.26)$$

Here, $P_{a_{camera}}$ contains the intrinsic and extrinsic parameters of the stereo camera, $P_{a_{robot}}$ contains the geometric parameters of the robot, and $\overline{T_E^C}$ is the transformation matrix from the robot's end-effector frame to the camera frame.

## 6.2.1 Feedback Controller Design with Model Linearization

Similar to the tool manipulation system, the visual system described by Equation (6.19) can be locally linearized at various operating points. Figure 3.16. illustrates the Camera-on-Robot model, and for reference, it is reproduced in Figure 6.5.

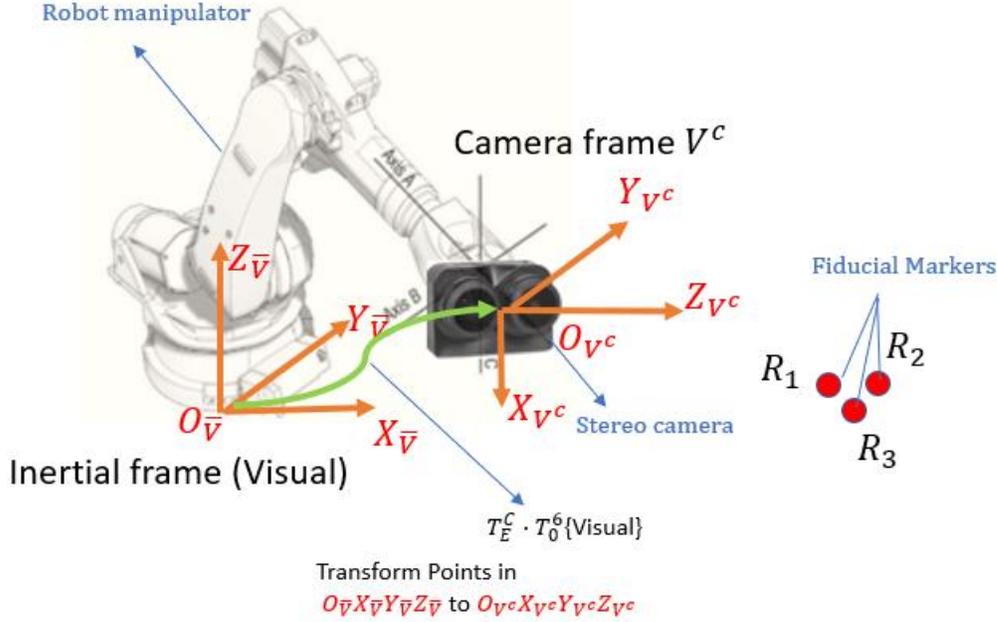

**Figure 6.5:** Camera-on-Robot model.

By selecting a specific linearization point $\widetilde{q}_V^{\,0}$, the nonlinear function Equation (6.25) can be approximated using a first-order Taylor expansion, yielding the Jacobian form:

$$\widehat{p_R} = J_\Psi(\widetilde{q}_V^{\,0})\widetilde{q}_V + \Psi(\widetilde{q}_V^{\,0}) \tag{6.27}$$

Where $J_\Psi(\widetilde{q}_V^{\,0}) \in \mathbb{R}^{6 \times 6}$ is the Jacobian matrix of $\Psi(\widetilde{q}_V)$, evaluated at the nominal joint configuration $\widetilde{q}_V^{\,0}$, and $\widehat{p_R}$ is the measured image coordinates of reference points. Please note that the Jacobian matrix $J_\Psi(\widetilde{q}_V^{\,0})$ becomes rank-deficient when the robot is in a singular configuration. In Section 6.3, we propose a method to address the challenges associated with such singularities and to maintain stability and continuity in the control process.

For notational convenience, let us define: $C_1^M = J_\Psi(\widetilde{q}_V^{\,0})$, $C_2^M = \Psi(\widetilde{q}_V^{\,0})$, Substituting these definitions into Equation (6.27), the linearized model becomes:

$$\widehat{p_R} = C_1^M \widetilde{q_V} + C_2^M \qquad (6.28)$$

Let's define $\widehat{p_R}' = \widehat{p_R} - C_2^M$, as an input disturbance, then, the overall block diagram of the linearized system is shown in Figure 6.6.

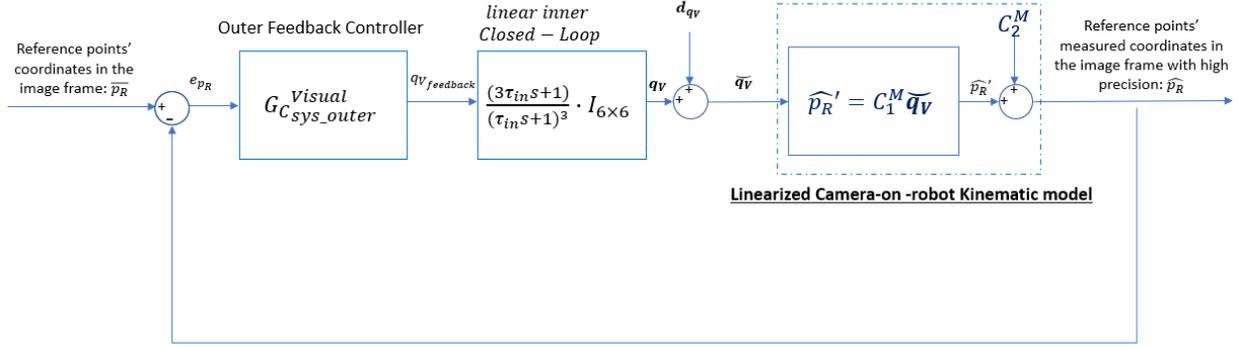

**Figure 6.6**: Feedback loop with linearized Camera-on-Robot Kinematic model.

As discussed in Section 2.1, the Location Determination Problem (LDP) for a stereo camera corresponds to the Perspective-Three-Point (P3P) problem. That is, to uniquely determine the pose of the camera in 3D space, at least three distinct reference points must be observed in the image plane. For each point, the stereo camera provides three image coordinates (e.g., left and right camera pixel locations in $u$ axis and common pixel locations in $v$ axis), resulting in a total of nine measurements in the system output: ($\widehat{p_R} \in \mathbb{R}^{9X1}$). In contrast, the visual robot manipulator considered here is a 6-degree-of-freedom (6-DOF) elbow manipulator, which consists of six independently actuated joints. Therefore, the control input to the plant is: ($\widetilde{q_V} \in \mathbb{R}^{6X1}$). This leads to an important structural characteristic: the plant $G_{p_{sys_{outer}}}^{Visual}$ is an underactuated system, meaning that the number of outputs (9) exceeds the number of control inputs (6). As a result, both the plant and the controller transfer functions in the outer-loop visual control system are non-square matrices, and their dimensions are:

$$G_{c_{sys_{outer}}}^{Visual} \in \mathbb{R}^{6X9}, \qquad G_{p_{sys_{outer}}}^{Visual} \in \mathbb{R}^{9X6} \qquad (6.29)$$

The linearized plant transfer function for the visual system, as shown in Figure 6.4., is given by:

$$G_{p_{sys_{outer}}}^{Visual}(Linear) = \frac{\widehat{p_V'}}{q_{V_{ref}}} = C_1^M \frac{(3\tau_{in}s + 1)}{(\tau_{in}s + 1)^3} \cdot I_{9\times 6} \tag{6.30}$$

To decouple and simplify the control design for this multivariable system, we derive the Smith-McMillan form using Singular Value Decomposition (SVD), similar to the procedure used for the tool manipulation system. The transfer function can be decomposed as:

$$G_{p_{sys_{outer}}}^{Visual}(Linear) = U_L^{Visual} M_p^{Visual} U_R^{Visual} \tag{6.31}$$

where $U_L^{Visual} \in \mathbb{R}^{9 \times 9}$ and $U_R^{Visual} \in \mathbb{R}^{6 \times 6}$ are the left and right unimodular matrices, and $M_p^{Visual} \in \mathbb{R}^{9 \times 6}$ is the Smith-McMillan form of $G_{p_{sys_{outer}}}^{Visual}(Linear)$, a diagonal matrix structure containing the system's decoupled dynamic modes. Each nonzero element in $M_p^{Visual}$ is a scalar gain multiplied by the common transfer function $\frac{(3\tau_{in}s+1)}{(\tau_{in}s+1)^3}$, resulting in the structure:

$$M_p^{Visual} = \frac{3\tau_{in}s + 1}{(\tau_{in}s + 1)^3} * \begin{bmatrix} Gain_1 & \cdots & 0 \\ \vdots & \ddots & \vdots \\ 0 & & Gain_6 \\ 0 & \cdots & 0 \\ 0 & \cdots & 0 \\ 0 & & 0 \end{bmatrix} \tag{6.32}$$

Where $Gain_i$ represents the numerical scaling for the $i^{th}$ row decoupled channel. The last three output coordinates are uncontrollable, as the 6-DOF manipulator cannot actuate more than six independent outputs.

The decoupled closed-loop transfer function, $M_T^{Visual}$, is selected to match a second-order Butterworth filter, similar to Equation (6.14), but padded to match the system's output dimensions:

$$M_T^{Visual} = \frac{\omega_n^2}{(s^2 + 2\zeta\omega_n s + \omega_n^2)} \cdot \begin{bmatrix} I_{6\times 6} & 0_{3\times 3} \\ 0_{3\times 3} & 0_{3\times 3} \end{bmatrix} \tag{6.33}$$

where $\omega_n$ is the natural frequency that defines the bandwidth of the outer-loop controller, which must satisfy i.e., $1/\omega_n > \tau_{in}$. $\zeta$ is the damping ratio. $0_{3\times 3}$ is a 3×3 matrix matrix of zeros, and $I_{6\times 6}$ is the identity matrix for the controllable outputs.

Following the same steps outlined in Equations (6.15) − (6.19), the coupled controller transfer function, $G_{c_{sys_{outer}}}^{Visual}$ can be constructed at a specific linearization point $\widetilde{q_V}^0$. However, because the plant dynamics vary with joint configuration, instability may arise when the robot deviates from this nominal point. To address this, an adaptive controller is developed that computes the Jacobian matrix online, enabling real-time updates of the linearized model and the corresponding controller parameters.

The first step in the adaptive control process is to estimate the current joint angles $\widetilde{q_V}$ from the measured image coordinates $\widehat{p_R}$. This estimation corresponds to the inverse of the camera-on-robot kinematic model, and is mathematically defined as:

$$\widetilde{q_V} = \varphi(\widehat{p_R}, P_{a_{visual}}, P_R^{\bar{V}}) \tag{6.28}$$

Where $\varphi$ is the inverse function of the nonlinear kinematic mapping $\Psi$ introduced in Equation (6.25). $P_{a_{visual}}$ defined in Equation (6.26), includes the parameters of the stereo camera, robot geometry, and transformation matrix $\overline{T_E^C}$, and $P_R^{\bar{V}}$ represents the known 3D coordinates of the reference points in the visual inertial frame. As noted in Section 3.6.2, the estimated joint configuration $\widetilde{q_V}$ is guaranteed to exist; however, it is typically non-unique due to the inherent ambiguities of the inverse kinematics problem. To resolve this, a specific solution can be systematically chosen using the disambiguation strategy presented in Table 3.2.

Once the current estimated $\widetilde{q_V}$ is obtained, the system is locally linearized at this configuration. The Jacobian matrix of the nonlinear model is computed, followed by the Singular Value Decomposition (SVD) to obtain the Smith-McMillan form. This allows construction of the corresponding linear controller $G_{C_{sys\_outer}}^{Visual}$ tailored to the current operating point. The overall process is depicted in Figure 6.7., which illustrates the structure of the adaptive visual control algorithm, including the estimation, linearization, and real-time controller update stages.

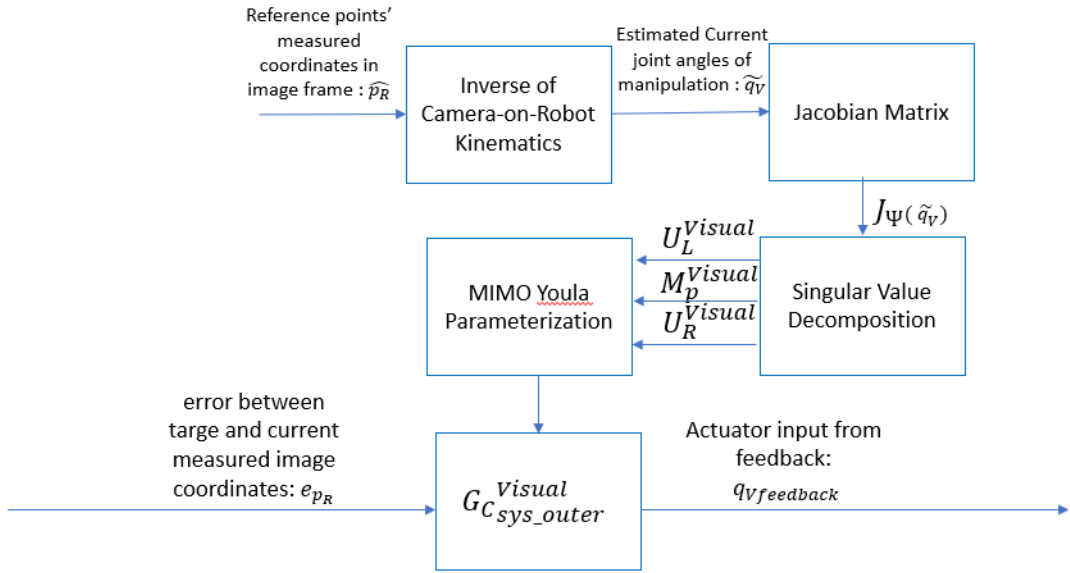

**Figure 6.7**: Block diagram of adaptive control design for the visual system.

The adaptive outer-loop controller is continuously evaluated at the current estimated joint angles of the robot, rather than relying on a fixed linearization point. This dynamic evaluation results in a more accurate representation of the system's behavior across different configurations. In contrast to a single controller designed at a specific operating point, the adaptive controller maintains model fidelity throughout the entire joint space. As a result, the adaptive feedback controller exhibits robust performance over the full range of operating conditions, effectively compensating for nonlinearities and variations in the system dynamics.

### 6.2.2 Limitation of Traditional Controller Design for Overdetermined Systems

PnP (Perspective-*n*-Points)-based Image-Based Visual Servoing (IBVS) presents well-known challenges for visual control in robotics, especially when the system becomes overdetermined—that is, when the number of visual features exceeds the number of available control inputs. For instance, in monocular camera systems, at least four 2D–3D point correspondences are required to determine a unique camera pose [66]. However, since a camera's pose consists of six degrees of freedom (DOF), a 6-DOF robot manipulator may need to track eight or more visual features to ensure sufficient spatial constraint.

In traditional IBVS [7], the interaction matrix (also known as the image Jacobian) relates feature motion in the image to the joint velocities. When the system is overdetermined, this Jacobian becomes tall (more rows than columns), introducing under-actuation. It can instead cause convergence to local minima, preventing the robot from reaching the desired pose. Although local asymptotic stability is guaranteed, global asymptotic stability is not—especially under overdetermined configurations [7].

Recent research has proposed several methods to mitigate these limitations:

Hybrid switched control by Gans et al. [79] combines IBVS and Position-Based Visual Servoing (PBVS), switching between them using Lyapunov functions. This approach improves robustness and prevents loss of features or divergence. However, it introduces discontinuities near switching surfaces and is only locally stable.

2½D visual servoing by Chaumette et al. [80] fuses image and pose information in a block-triangular Jacobian, enabling smoother trajectories and avoiding singularities. However, it is sensitive to depth estimation errors and requires careful feature selection.

MPC-based IBVS developed by Roque et al. [81] applies Model Predictive Control (MPC) to IBVS for underactuated systems like quadrotors. This method handles constraints and guarantees formal stability, but it relies on hover-based linearization and lacks explicit integration of visual dynamics, limiting generalizability.

While these methods improve convergence and robustness, they also introduce significant computational costs. The switched controller requires multiple tailored strategies [82]; the 2½D approach demands real-time visual and positional fusion [83]; and MPC requires solving optimization problems at every control step, limiting real-time feasibility [84].

In this dissertation, a stereo camera system is used to observe 3D reference points in the workspace. As detailed in Appendix B, a minimum of three 2D–3D correspondences is required to determine the stereo camera pose. In the proposed system, the control loop tracks three reference points, each contributing three image-plane coordinates, for a total output dimension of: $\widehat{p_R} \in \mathbb{R}^{9X1}$. Meanwhile, the robot manipulator has only six actuated joints, and the control input is: $\widetilde{q_V} \in \mathbb{R}^{6X1}$. This results in an overdetermined plant, where only six out of nine outputs can be controlled. The remaining three outputs (image coordinates of reference point $R_3$) are uncontrollable.

Figure 6.8. presents the frequency responses of the closed-loop transfer function $T_{MIMO}^V$ and the corresponding sensitivity function $S_{MIMO}^V$, using $\omega_n = 10\ rad/s$. As shown, the closed-loop and sensitivity functions intersect at approximately 10 rad/s for the first six outputs (image features of $R_1$ and $R_2$). However, the sensitivity function remains 0 dB and the magnitude of $T_{MIMO}^V$ remains very low for the remaining three outputs, confirming their uncontrollability.

To address this partial controllability, a feedforward controller is introduced. It computes target joint configurations $\overline{q_V}$ corresponding to the desired camera pose. These target values are generated from inverse kinematics and used as reference inputs. The feedforward controller enables fast system response and helps avoid convergence to local minima, especially in overdetermined configurations.

The design methodology for this feedforward controller follows the same approach used in tool position control and is detailed in Section 6.2.3. Comparative analysis between the feedforward-feedback hybrid system and a feedback-only system is presented in Chapter 7, demonstrating the performance benefits of incorporating predictive motion commands in visually guided control.

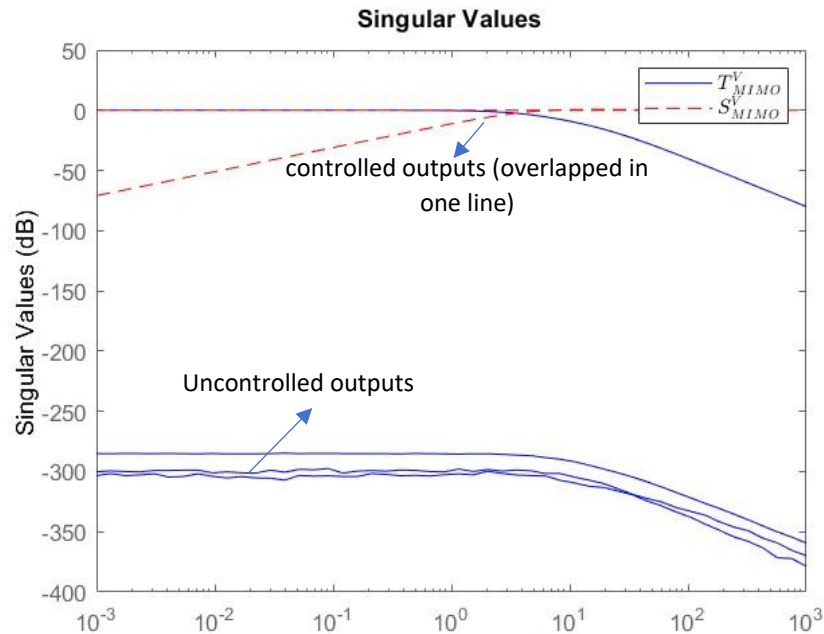

**Figure 6.8:** Frequency response of $T_{MIMO}^V$, $S_{MIMO}^V$ for the MIMO visual system.

### 6.2.3 Feedforward controller design

The Feedforward Controller generates a target joint angle $q_{Vfeedforward}$ based on the known coordinates of the reference points when the camera is at its optimal pose. As in the SISO case, the feedforward controller in the visual system incorporates only the inverse of the inner-loop closed-loop transfer function, denoted as $T_{MIMO}^{forward}$.

The structure of $T_{MIMO}^{forward}$ follows the same formulation as given in Equations (6.22) and (6.23). This ensures that the feedforward path remains proper while shaping the system dynamics for fast convergence toward the target pose.

Figure 6.9. illustrates the open-loop architecture of the feedforward controller, highlighting how the reference points' coordinates in image frame are generated from the predicted target joint configuration, independent of feedback correction.

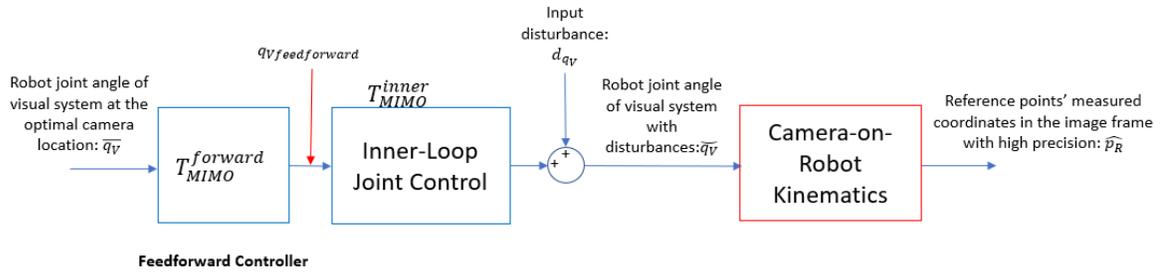

**Figure 6.9:** Block diagram of feedforward control design for the MIMO visual system.

## 6.3 Handling Singularity-Induced Instability in Controller Design

Sections 6.1 and 6.2 have demonstrated the use of Jacobian matrices in the design of MIMO controllers through the Jacobian linearization method. In general, we can write an expression that relates the joint angles $q$ and image coordinates $p$:

$$p = J(q) \cdot q \qquad (6.29)$$

In this design process, the Jacobian matrix $J$ is decomposed using singular value decomposition as follows:

$$J = U \cdot \Sigma \cdot (V)^T \qquad (6.30)$$

where:

$J$ is the Jacobian matrix of the manipulator,

$U$ is an orthogonal matrix containing the left singular vectors,

$V$ is an orthogonal matrix containing the right singular vectors, and

$\Sigma$ is a diagonal matrix containing the singular values $\sigma_i \geq 0$.

If the robot is at or near a singular configuration, the Jacobian matrix $J$ becomes rank-deficient, meaning that one or more singular values in $\Sigma$ approach zero. This poses a significant problem for controller design—especially when using Youla parameterization, where the controller structure involves cancelling the original linearized plant. In such cases, we are taking the inverse of the Jacobian matrix and the singular values $\sigma_i$ appear in the denominator of the resulting control law. Since the Jacobian is potentially rank-deficient, we can take the Moore-Penrose pseudoinverse:

$$J^\dagger = V \cdot \Sigma^\dagger \cdot (U)^T \tag{6.31}$$

Consequently, when the manipulator approaches a singular configuration, one or more singular values $\sigma_i \to 0$, and the corresponding terms in $\Sigma^\dagger$: $\frac{1}{\sigma_i} \to \infty$, the controller attempts to apply very large control efforts (approaching infinity) to produce only small end-effector movements, leading to poor numerical conditioning, amplified actuator commands, and potential instability in the system.

To overcome this, the Damped Least-Squares (DLS) solution [85] modifies the inverse of each singular value as:

$$\frac{1}{\sigma_i} \to \frac{\sigma_i}{\sigma_i^2 + \lambda^2} \tag{6.32}$$

where $\lambda > 0$ is the damping factor that regularizes the inversion. This modification keeps the inverse well-defined and bounded near singularities.

To achieve both stability near singularities and accuracy in regular configurations, a smooth switching mechanism is integrated into the DLS inverse of the Jacobian. Instead of applying a constant damping factor, the damping coefficient $\lambda$ is adjusted based on the smallest singular value $\sigma_{min}$ of the Jacobian matrix $J$. When the robot is far from singularities, i.e., when $\sigma_{min} \gg \sigma_{th}$, a threshold, the standard pseudoinverse is used as Equation (6.31).

However, as $\sigma_{min}$ approaches the singularity threshold $\sigma_{th}$ damping is gradually introduced using the DLS formulation:

$$J^{\dagger}_{DLS} = V \cdot diag(\frac{\sigma_i}{\sigma_i^2 + \lambda^2}) \cdot (U)^T \tag{6.33}$$

To enable a smooth transition between these regimes, the damping factor $\lambda$ is defined as a piecewise function:

$$\lambda = \begin{cases} 0 & \sigma_{min} > \sigma_{th} + \delta \\ \lambda_{max} \cdot (1 - \frac{\sigma_{min} - \sigma_{th}}{\delta}), & \sigma_{th} < \sigma_{min} \leq \sigma_{th} + \delta \\ \lambda_{max} & \sigma_{min} \leq \sigma_{th} \end{cases} \tag{6.34}$$

Where is a $\delta$ transition margin used to create a smooth interpolation between the undamped and damped regions, rather than switching abruptly.

This formulation ensures that the controller uses the undamped inverse in well-conditioned regions, while gradually introducing damping as the manipulator approaches a singularity. This prevents abrupt control transitions, reduces numerical instability, and ensures consistent joint behavior across the workspace.

While the smooth switching strategy described above provides a practical and robust solution for singularity avoidance, the selection of the damping parameters—particularly the maximum damping coefficient $\lambda_{max}$ and the transition bandwidth $\delta$ —remains largely heuristic. In this dissertation, fixed values of $\lambda_{max}$ and $\delta$ were selected based on empirical tuning to balance numerical stability and tracking accuracy. However, the optimal choice of these parameters can vary significantly depending on the manipulator's kinematic structure, the task space trajectory, and performance objectives such as energy efficiency or smoothness.

In future research, we intend to explore adaptive and learning-based methods for selecting these parameters in real time. One promising approach is to adjust damping based on manipulability measures [86], which quantify the closeness to singularity using the determinant or condition number of the Jacobian. Other works have also proposed adaptive damping algorithms that scale $\lambda$ continuously based on singular value trajectories [85]. Moreover, machine learning techniques, such as reinforcement learning or Gaussian process regression, have been investigated for tuning IK solvers and nullspace behaviors in redundant manipulators [87-88]. These methods offer potential for developing context-aware damping strategies that improve robustness without sacrificing responsiveness. Integrating such approaches into the controller design could yield smoother, safer, and more efficient robot performance across a wider range of tasks and configurations.

In all simulations and experimental tests conducted in this dissertation, the following constant values were used:

$$\lambda_{max} = \delta = 0.01 \tag{6.35}$$

## 6.4 Conclusion

In this chapter, we developed MIMO outer-loop controllers for both the visual system and the tool manipulation system, employing a feedforward-feedback control architecture based on model linearization. The adaptive feedback controllers are designed using Youla parameterization, applied to linearized models evaluated at the current joint angles. This approach ensures robust stability across the full range of the robot's operational configurations.

To enhance responsiveness, feedforward controllers are integrated into both systems. These controllers not only improve dynamic performance, which is particularly important for high-speed manufacturing tasks, but also help mitigate the risk of convergence to local minima, especially in visual positioning tasks involving the camera.

Special attention was given to the treatment of kinematic singularities, which pose a critical challenge to controller stability and actuator effort near rank-deficient configurations. A smooth damped least-squares solution was developed to regularize the inverse kinematics in these cases, ensuring consistent and stable control performance even as the robot approaches singular poses.

In Chapter 7, we present simulation results that validate the performance of the proposed control systems across a variety of fastening and unfastening alignment scenarios, highlighting both accuracy and robustness under realistic conditions.

_________________________________________________________________Chapter 7

# Results and Analysis
___________________________________________________________________

The MIMO control architectures developed in this dissertation are evaluated through simulations that incorporate nonlinear models of both the stereo camera and the robot manipulator. Each control architecture is tested under scenarios with and without input disturbances to assess the robustness of the controllers. Furthermore, model uncertainties are introduced to evaluate the performance of the control systems under imperfect modeling conditions.

Section 7.1 presents and analyzes the simulation results for the tool manipulation system, while Section 7.2 focuses on the visual system, including a comparative study of the system's performance with and without the feedforward controller designed in Chapter 6.

All simulations are conducted in MATLAB Simulink, utilizing the ZED 2 stereo camera system and the ABB IRB 4600 elbow-type robotic manipulator. The technical specifications of both the camera and the manipulator are provided in tabulated form in the appendix for reference.

___________________________________________________________________

## 7.1 Simulation Analysis of the MIMO Tool Manipulation System

### 7.1.1 Simulation Results for the MIMO Tool Manipulation System

In this section, we simulate the closed-loop adaptive control system under various scenarios to evaluate the performance and stability of the proposed MIMO controller. In all cases, the tool (end-effector) is guided to a predefined target pose specified in the inertial frame of the visual system $\{\bar{V}\}$. The target position is given by: $\overline{P_I^{\bar{V}}} = [-1m, 0.2m, 0.3m]^T$, where $m$ represents meters. The desired orientation of the tool is defined by the rotation matrix $[\bar{n}, \bar{s}, \bar{a}]$, corresponding to the unit vectors of yaw, pitch, and roll directions in the inertial frame: $[\bar{n}, \bar{s}, \bar{a}] = \begin{bmatrix} 0 & -1 & 0 \\ -1 & 0 & 0 \\ 0 & 0 & -1 \end{bmatrix}$, which requires the tool to be aligned vertically in the inertial frame of the visual system $\{\bar{V}\}$.

The bandwidth of actuators in industrial robot manipulators typically ranges from 5 Hz to 100 Hz, corresponding approximately to 31 rad/s to 628 rad/s. For the purposes of this simulation, we assume that the inner loop—representing joint-level control—operates at a fixed bandwidth of 100 rad/s across all scenarios. This corresponds to a time constant of $\tau_{in} = 0.01$ s. In cascaded control architectures, it is standard practice to design the inner loop to be at least 10 times faster than the outer loop to ensure proper time-scale separation and prevent interference. Accordingly, the outer-loop bandwidth is selected as one-tenth of the inner-loop bandwidth: $\omega_n$ = 10 rad/s.

Additional parameters for the controller are either computed analytically or selected based on design guidelines. In the feedforward controller, the time constant $\tau_{forward}$ is chosen to be one-tenth of the inner-loop time constant, as described in Equation (6.23): $\tau_{forward} = 0.1\, \tau_{in} = 0.001$ s. To ensure the outer-loop response is overdamped, and thus suppresses overshoot, the damping ratio $\zeta$ is selected to be relatively large: $\zeta = 1.5$.

In the simulation plots:

- Blue lines represent the response trajectories of the end-effector position.
- Red lines indicate the target positions.
- Circles mark the initial pose of the end-effector.
- Stars indicate the target pose.
- Black arrows visualize the changing orientation along the trajectory.

- Red arrows represent the target orientation at the goal pose.

Three distinct scenarios are visualized in Figures 7.1, 7.2, and 7.3. In the first scenario, the tool (end-effector) starts at the pose: $P_{I_0}^{\bar{V}} = [1.404m, 0.228m, 1.171m]^T$, and $[n_0, s_0, a_0] = \begin{bmatrix} -0.4893 & -0.0262 & 0.8717 \\ 0.2427 & 0.9560 & 0.1650 \\ -0.8377 & 0.2932 & -0.4614 \end{bmatrix}$, where two fiducial points are within the camera's field of view, allowing their coordinates to be tracked throughout the entire control period. In the second and third scenario, the tool (end-effector) starts at the pose: $P_{I_0}^{\bar{V}} = [1.285m, 0m, 1.57m]^T$, and $[n_0, s_0, a_0] = \begin{bmatrix} 0 & 0 & 1 \\ 0 & 1 & 0 \\ -1 & 0 & 0 \end{bmatrix}$, where two of the fiducial points are initially outside the camera's field of view. Their image coordinates must be estimated during the initial phase of the control loop until the camera aligns itself with the tool.

To assess the robustness of the controller, joint angle disturbances are introduced in Scenarios 1 and 3: $d_q = [0.1°, 0.5°, 0.2°, 0.3°, -0.1°, 0.3°]$. Table 7.1. summarizes the initial and target poses, along with the corresponding joint angle configurations for each scenario.

**Table 7.1**: Tool pose and robot joint angles at initial and final state of simulation scenarios.

| | Tool Pose | Robot Joint Angles |
|---|---|---|
| Format | $\begin{bmatrix} n_x & s_x & a_x & d_x \\ n_y & s_y & a_y & d_y \\ n_z & s_z & a_z & d_z \end{bmatrix}$ Where $[n_x, n_y, n_z]^T$, $[s_x, s_y, s_z]^T$ and $[a_x, a_y, a_z]^T$ are Yaw, Pitch, and Roll, and $[d_x, d_y, d_z]^T$ (in meters) is the position, measured in inertial frame {O}. | $\begin{bmatrix} q_{T_1} & q_{T_2} \\ q_{T_3} & q_{T_4} \\ q_{T_5} & q_{T_6} \end{bmatrix}$ Where $[q_{T_1}, q_{T_2}, q_{T_3}, q_{T_4}, q_{T_5}, q_{T_6}]$ in (degrees) are robot joint angles in the tool manipulation system. |

| | Scenario 1 | | Scenario 2 & 3 | |
|---|---|---|---|---|
| | Tool Pose | Robot Joint Angles | Tool Pose | Robot Joint Angles |
| Initial State | $\begin{bmatrix} -0.489 & -0.026 & 0.872 & 1.404 \\ 0.243 & 0.956 & 0.165 & 0.228 \\ -0.838 & 0.293 & -0.461 & 1.717 \end{bmatrix}$ | $\begin{bmatrix} 9.09° & 10.43° \\ 8.55° & 9.70° \\ 8.63° & 8.79° \end{bmatrix}$ | $\begin{bmatrix} 0 & 0 & 1 & 1.285 \\ 0 & 1 & 0 & 0 \\ -1 & 0 & 0 & 1.57 \end{bmatrix}$ | $\begin{bmatrix} 0° & 19.16° \\ -22.16° & 0° \\ 177.01° & 0° \end{bmatrix}$ |
| Final State | $\begin{bmatrix} 0 & -1 & 0 & -1 \\ -1 & 0 & 0 & 0.2 \\ 0 & 0 & -1 & 0.3 \end{bmatrix}$ | $\begin{bmatrix} 168.69° & 26.22° \\ 48.84° & 0° \\ 16.94° & 78.69° \end{bmatrix}$ | $\begin{bmatrix} 0 & -1 & 0 & -1 \\ -1 & 0 & 0 & 0.2 \\ 0 & 0 & -1 & 0.3 \end{bmatrix}$ | $\begin{bmatrix} 168.69° & 26.22° \\ 48.84° & 0° \\ 16.94° & 78.69° \end{bmatrix}$ |

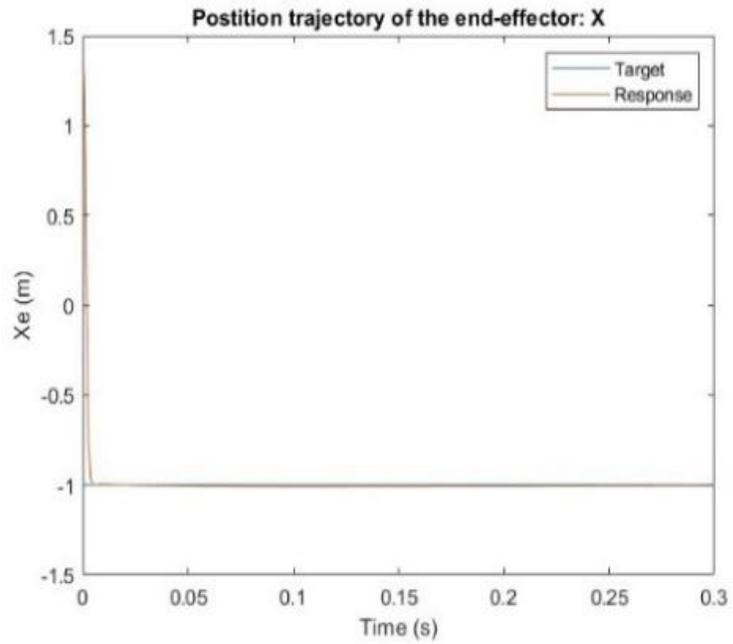

(a)

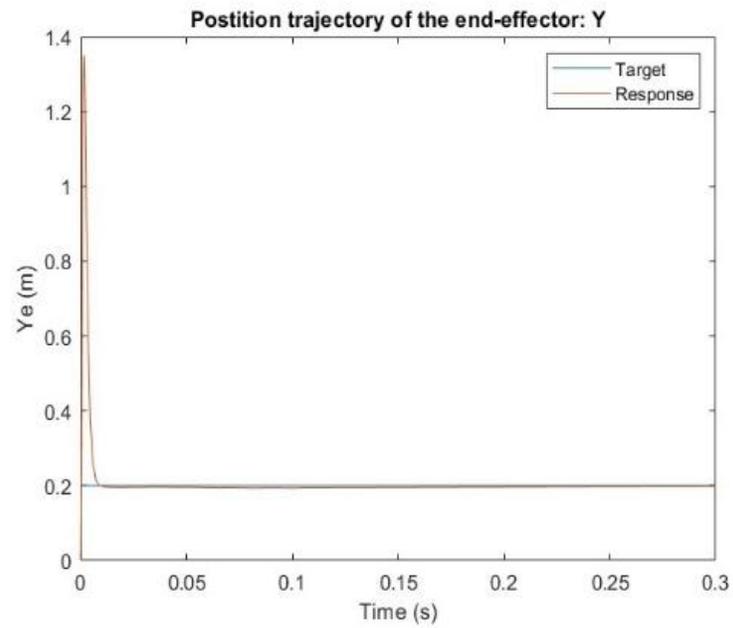

(b)

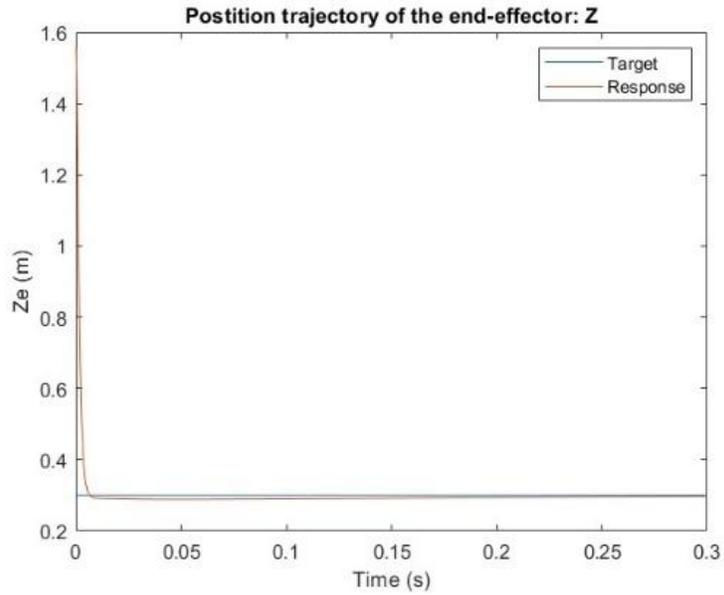

(c)

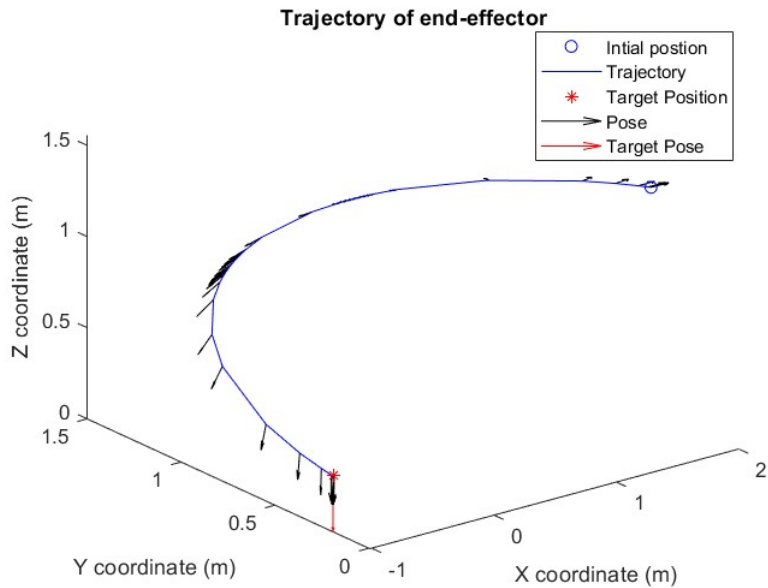

(d)

**Figure 7.1:** Response of the tool's (end-effector's) pose in scenario 1: Start in field view of camera with disturbances. *(a) Trajectory of X-coordinates of the end-effector. (b) Trajectory of Y-coordinates of the end-effector. (c) Trajectory of Z-coordinates of the end-effector. (d) Trajectory of the end-effector and tool poses.*

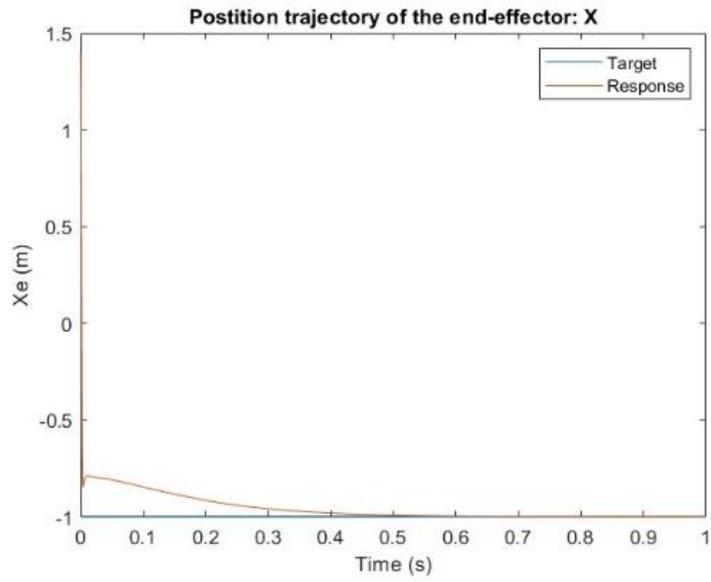

(a)

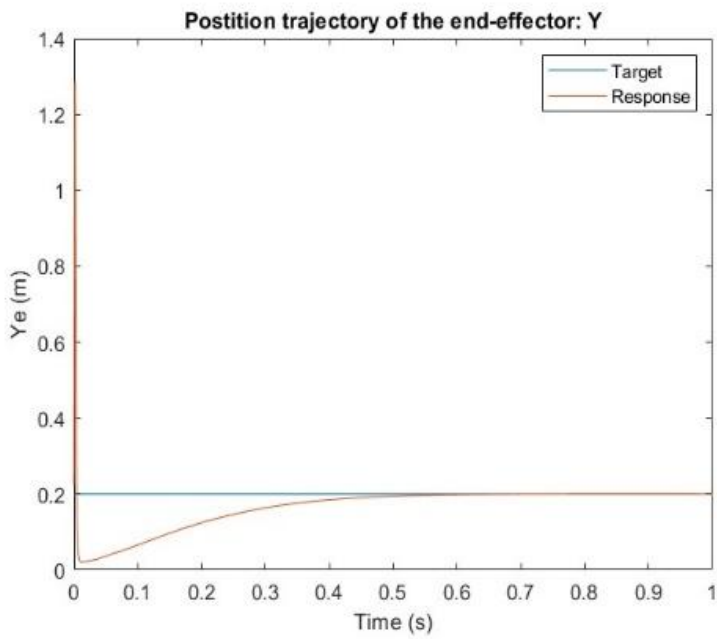

(b)

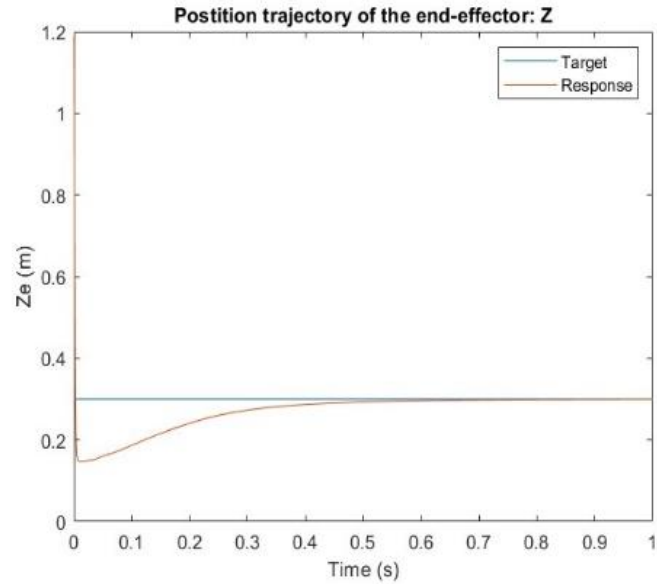

(c)

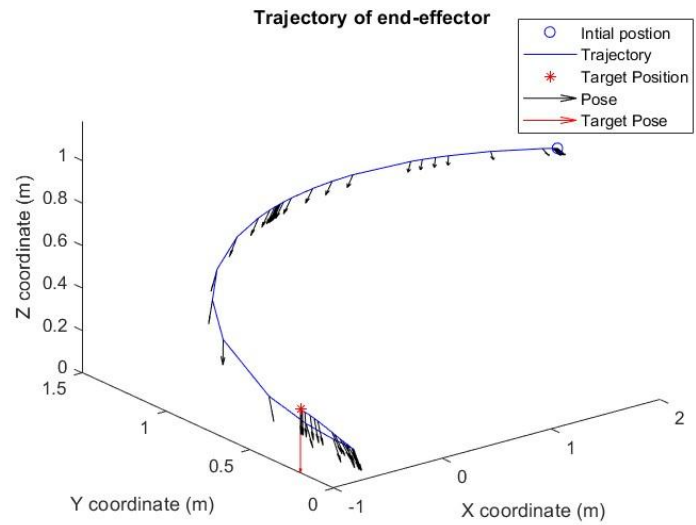

(d)

**Figure 7.2:** Response of the tool's (end-effector's) pose in scenario 2: Start outside field view of camera with no disturbances. *(a) Trajectory of X-coordinates of the end-effector. (b) Trajectory of Y-coordinates of the end-effector. (c) Trajectory of Z-coordinates of the end-effector. (d) Trajectory of the end-effector and tool poses.*

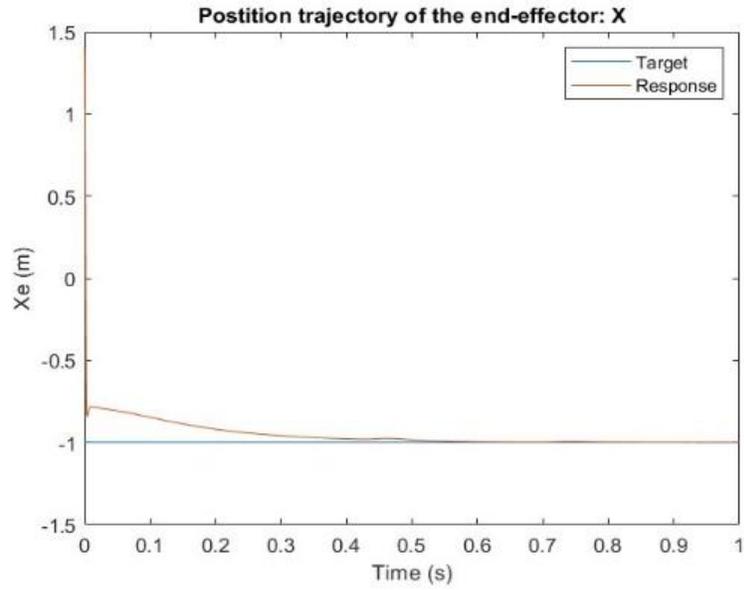

(a)

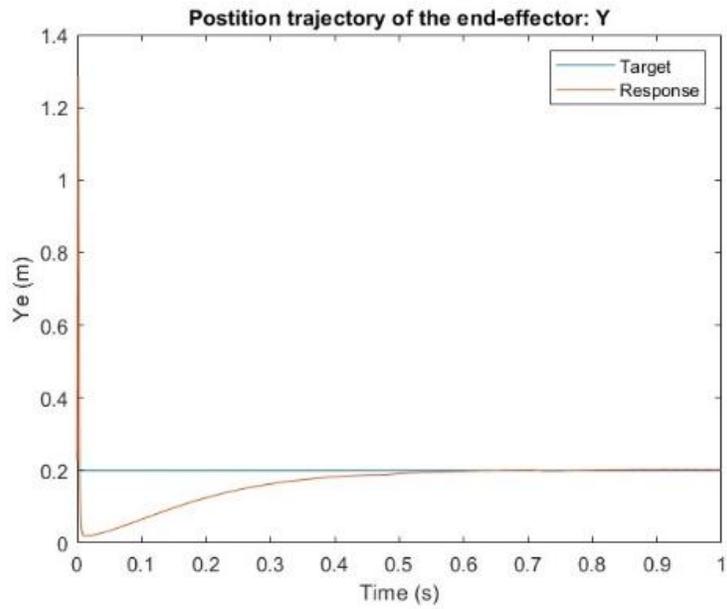

(b)

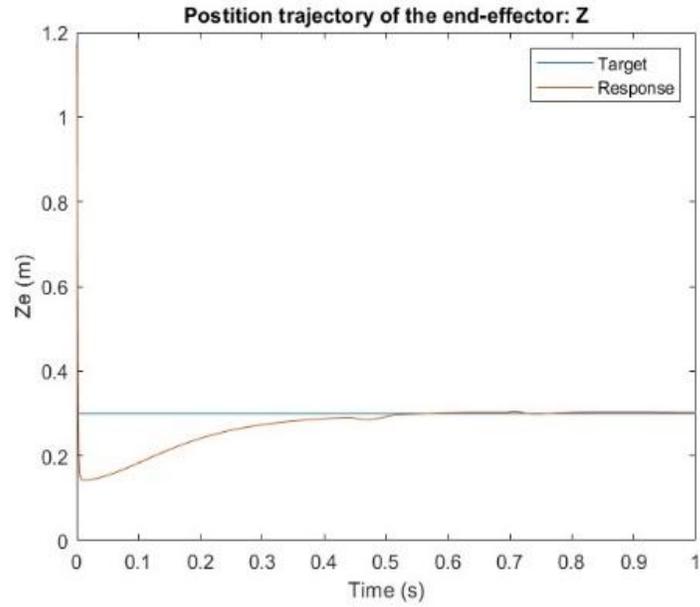

(c)

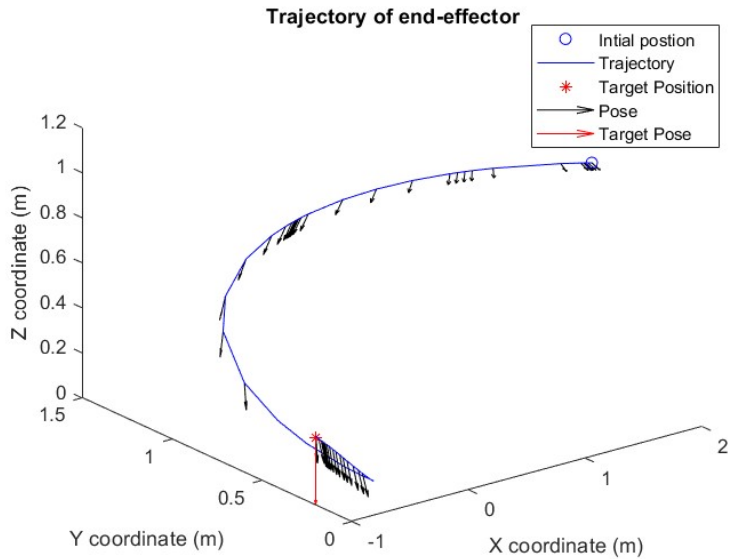

(d)

**Figure 7.3:** Response of the tool's (end-effector's) pose in scenario 3: Start outside field view of camera with disturbances. *(a) Trajectory of X-coordinates of the end-effector. (b) Trajectory of Y-coordinates of the end-effector. (c) Trajectory of Z-coordinates of the end-effector. (d) Trajectory of the end-effector and tool poses.*

The simulation results in Figures 7.1. and 7.3. demonstrate that the proposed adaptive feedforward-feedback controller is capable of guiding the tool to its target pose in a stable and accurate manner, even in the presence of input disturbances. Comparing the orientation trajectories in Figures 7.2. and 7.3., it is evident that disturbances introduce instability in the transient phase, but the controller is able to restore stability in the steady-state response.

When the tool's coordinates are measurable by the camera (as in Figure 7.1.), the combined action of the feedforward and feedback controllers results in fast and precise convergence, with response times under 0.3 seconds and minimal overshoot. In contrast, when the tool is initially outside the camera's field of view (as in Figures 7.2. and 7.3.), the system still achieves convergence in less than one second. However, larger overshoots are observed in some position trajectories.

These overshoots are attributed to accumulated disturbances that the feedforward controller alone cannot suppress, particularly in the absence of visual feedback. The feedback component becomes critical once the tool enters the field of view, at which point the system corrects the residual errors.

In conclusion, while the continuous feedforward-feedback control architecture performs well overall, it exhibits greater sensitivity to disturbances when the tool starts from a position far from the target, especially outside the visual range. Although these disturbances are eventually compensated once feedback becomes available, the resulting overshoots can be significant. Future research will investigate strategies to mitigate these overshoots, such as:

- Expanding the camera's field of view,
- Employing switching control algorithms that transition from feedforward to feedback modes based on visual availability, rather than relying on a continuous blended controller.

## 7.1.2 Robustness Analysis of Model Uncertainties for the MIMO Tool Manipulation System

In this section, we investigate the impact of parameter variations on the steady-state error of the end-effector's X-coordinate position. Specifically, we examine uncertainties in the lengths of link 2 and link 4 of the robot manipulator, denoted as $L_2$ and $L_4$, respectively. These parameters are selected because they possess the largest numerical values among the manipulator's geometric dimensions and thus are expected to have a significant influence on positional accuracy.

Figure 7.4. presents the results of a robustness test conducted under Simulation Scenario 1, where each parameter is varied independently over a range from 50% to 110% of its nominal value. Across the entire range of variations, the resulting steady-state error in the end-effector's X position remains below 1%, indicating that the proposed control system maintains high accuracy despite substantial geometric uncertainty. However, when parameter variations fall outside this range, the steady-state error exceeds 1%, indicating a degradation in control accuracy beyond the tested robustness bounds.

This result can be used as a quantitative benchmark demonstrating the robust performance of the controller in the presence of model variations. From the plot, we conclude that the adaptive feedforward-feedback controller is robust to geometric parameter uncertainty, particularly in the key structural dimensions of the robot arm.

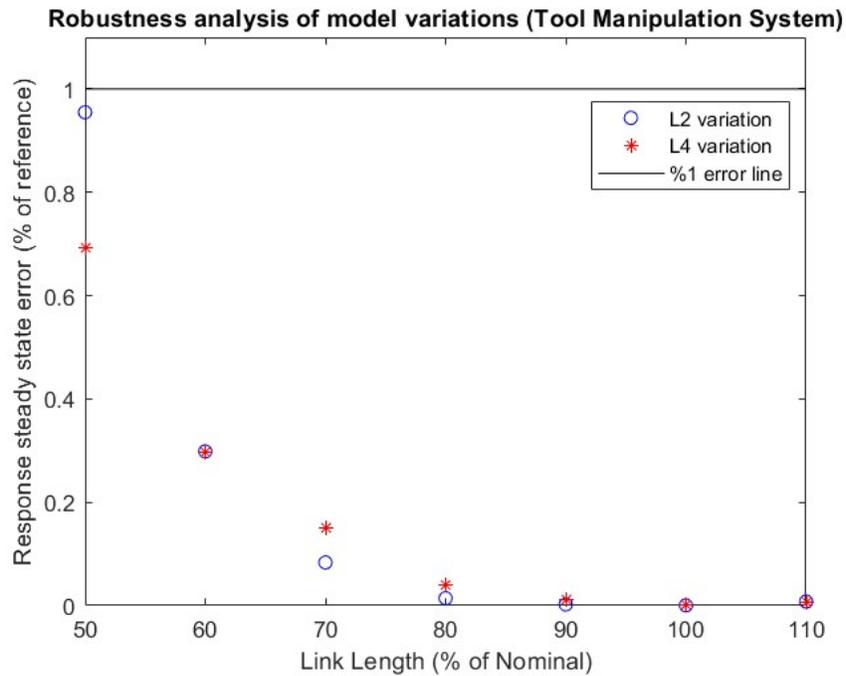

**Figure 7.4:** Robustness analysis of model variations for the tool manipulation system.

## 7.2 Simulation Analysis of the MIMO Visual System

### 7.2.1 Simulation Results for the MIMO Visual System

To evaluate the performance of MIMO controller design for the visual system, we simulated two scenarios. All simulations are conducted entirely in MATLAB Simulink, with hardware-in-the-loop components replaced by their corresponding mathematical models. The camera system performs 2D feature estimation of three virtual points in space, with their coordinates in the inertial frame selected as: $[-0.5m\ 0\ 0]^T$, $[0\ 0\ 0.5m]^T$, $[2m, -2m, 0]^T$.

Many camera noise removal algorithms have been proposed in Section 1.3.4. For this dissertation, we assume that the images captured by the camera have been preprocessed using image averaging technique, and the noise has been almost perfectly attenuated. In other words, the only remaining disturbances in the system are due to unmodeled joint dynamics, such as compliance and flexibility, which are modeled as input disturbances in the controlled system.

Two scenarios were simulated:

Scenario 1: Without input disturbances.

Scenario 2: With a 1° step input disturbance added to each joint of the robot arms for the entire simulation time.

In both scenarios, the camera system starts from an initial pose in the inertial frame of the visual system, denoted as $Pose_{initila}^{\bar{V}}$, and maneuvers target pose, denoted as $Pose_{final}^{\bar{V}}$. Table 7.2. summarizes the initial and final poses for each scenario, along with the corresponding joint configurations.

Figure 7.5. and Figure 7.7. present the responses of the six joint angles for the two scenarios, respectively. Figure 7.6. and Figure 7.8. show the responses of the nine image coordinates over time for each scenario. In each case, the feedback-only controlled system (left plot) is compared to the feedforward-and-feedback controlled system (right plot). These comparisons focus on overshoot, response time, and target tracking performance.

Consistent with the simulation setup for the MIMO tool manipulation control in Section 7.1.1, all scenarios in this simulation assume an inner-loop bandwidth of 100 rad/s ($\tau_{in} = 0.01s$) and an outer-loop bandwidth of 10 rad/s ($\omega_n = 10\ \text{rad/s}$), ensuring appropriate separation between fast joint-level dynamics and slower outer-loop control.

**Table 7.2**: Camera pose and robot joint angles at initial and final state of simulation scenarios.

| | Camera Pose | Robot Joint Angles |
|---|---|---|
| Format | $\begin{bmatrix} n_x & s_x & a_x & d_x \\ n_y & s_y & a_y & d_y \\ n_z & s_z & a_z & d_z \end{bmatrix}$ Where $[n_x, n_y, n_z]^T$, $[s_x, s_y, s_z]^T$ and $[a_x, a_y, a_z]^T$ are Yaw, Pitch, and Roll, and $[d_x, d_y, d_z]^T$ (in meters) is the position, measured in inertial frame {O}. | $\begin{bmatrix} q_{V_1} & q_{V_2} \\ q_{V_3} & q_{V_4} \\ q_{V_5} & q_{V_6} \end{bmatrix}$ Where $[q_{V_1}, q_{V_2}, q_{V_3}, q_{V_4}, q_{V_5}, q_{V_6}]$ in (degrees) are robot joint angles in the visual system. |

| | *Scenario* 1 | | *Scenario* 2 | |
|---|---|---|---|---|
| | Camera Pose | Robot Joint Angles | Camera Pose | Robot Joint Angles |
| Initial State | $\begin{bmatrix} 0 & 0 & 1 & 1.27 \\ 0 & 1 & 0 & 0 \\ -1 & 0 & 0 & 1.57 \end{bmatrix}$ | $\begin{bmatrix} 0° & 0° \\ 0° & 0° \\ 0° & 0° \end{bmatrix}$ | $\begin{bmatrix} -0.070 & -0.998 & 0.002 & 1.064 \\ -0.996 & 0.070 & -0.060 & 0.964 \\ -0.060 & -0.007 & -0.998 & 0.939 \end{bmatrix}$ | $\begin{bmatrix} 42.40° & 21.20° \\ 4.58° & -2.86° \\ 66.46° & -42.40° \end{bmatrix}$ |
| Final State | $\begin{bmatrix} -0.11 & 0.14 & 0.98 & 1.31 \\ -0.09 & 0.98 & -0.15 & -0.01 \\ -0.99 & -0.10 & -0.09 & 1.49 \end{bmatrix}$ | $\begin{bmatrix} 0.48° & 2.21° \\ 2.05° & -82.68° \\ 9.46° & 77.47° \end{bmatrix}$ | $\begin{bmatrix} 0 & -1 & 0 & 1 \\ -1 & 0 & -1 & 1 \\ 0 & 0 & 0 & 1 \end{bmatrix}$ | $\begin{bmatrix} 45° & 18.59° \\ 4.35° & 0° \\ 67.06° & -45° \end{bmatrix}$ |

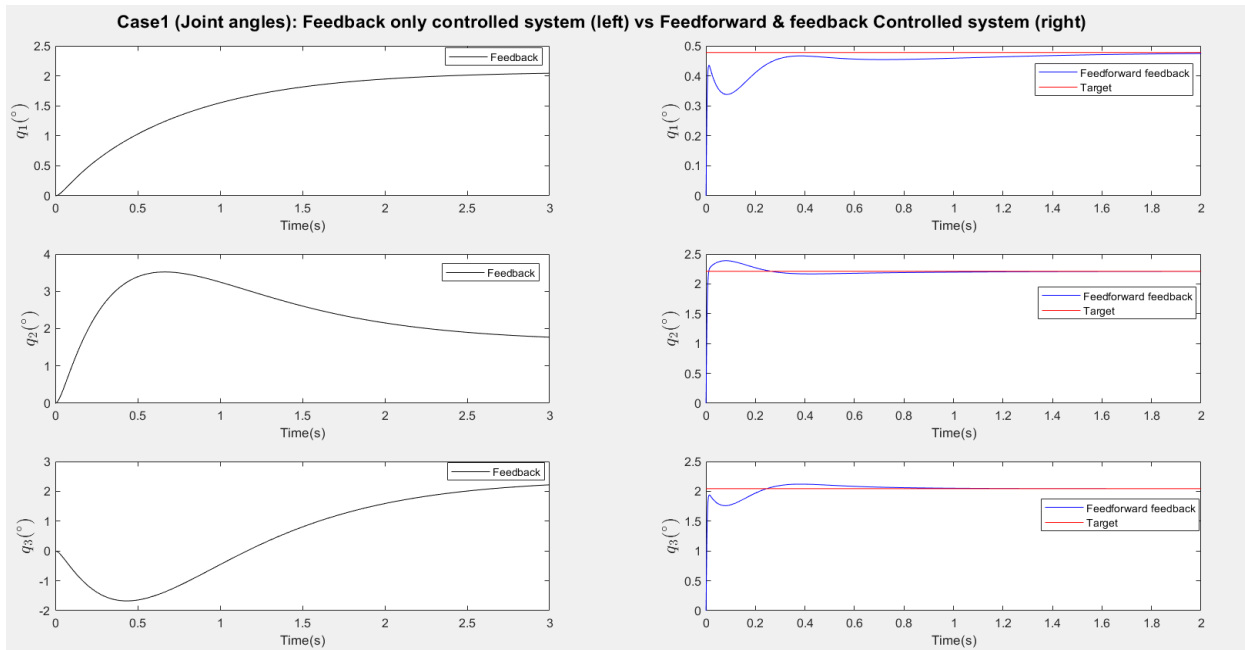

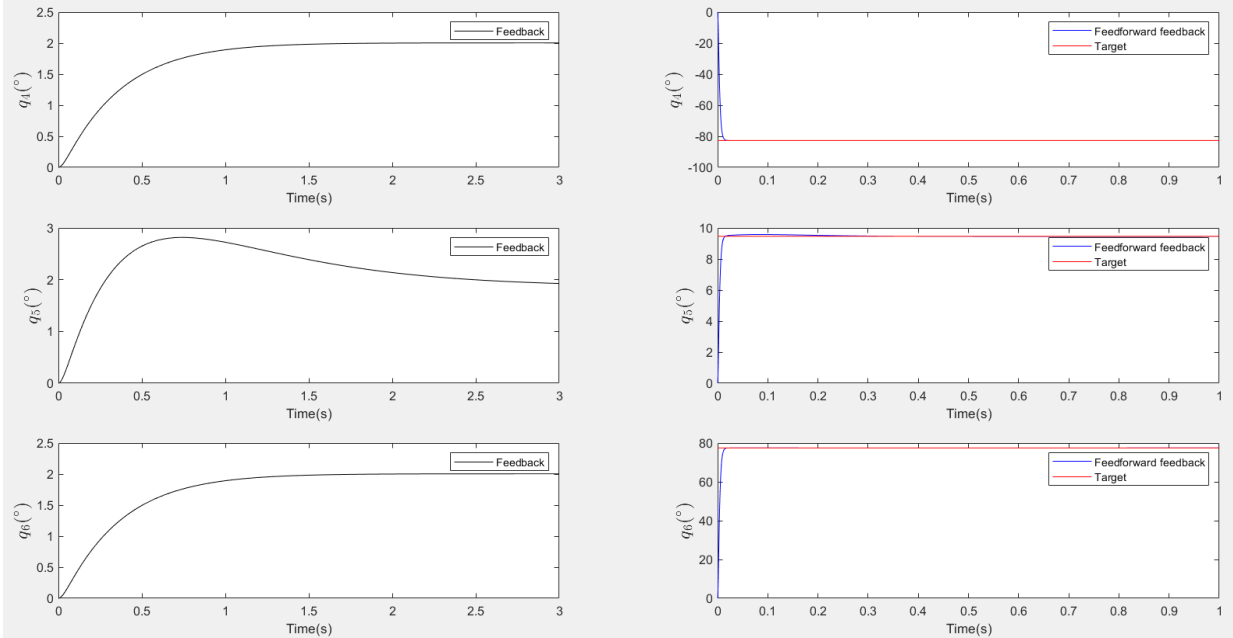

**Figure 7.5:** Scenario one (no disturbances): Response of robot joint angles.

(*Left: Feedback-only responses, Right: Feedforward-feedback responses*).

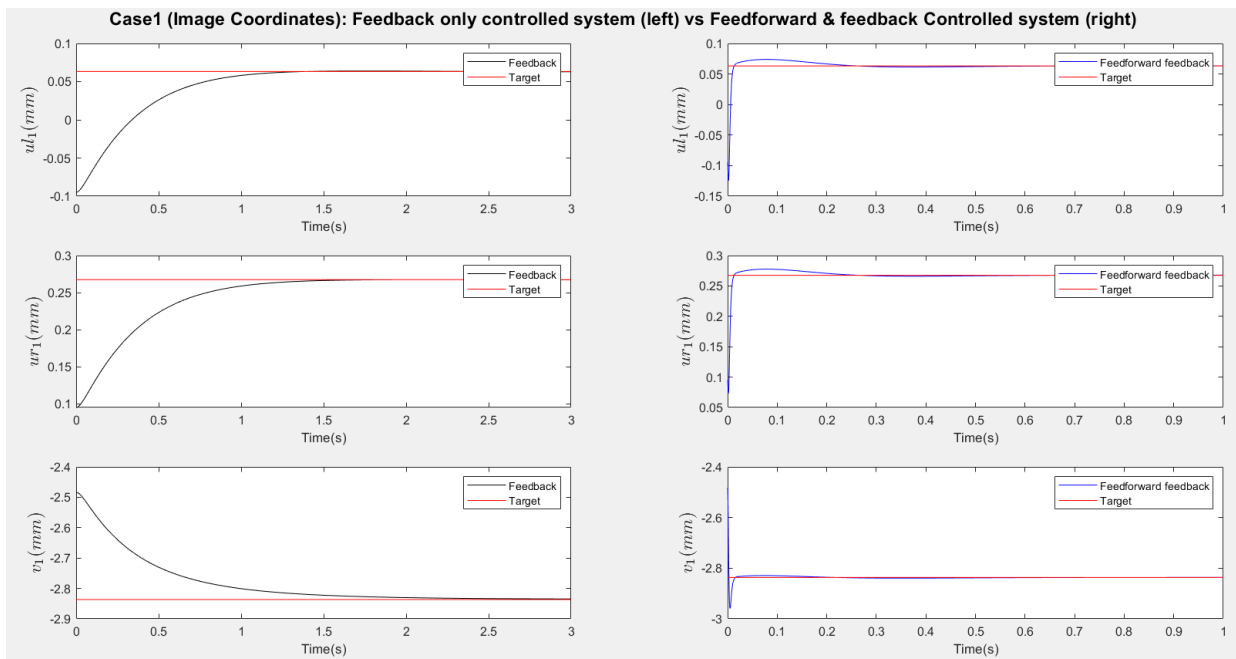

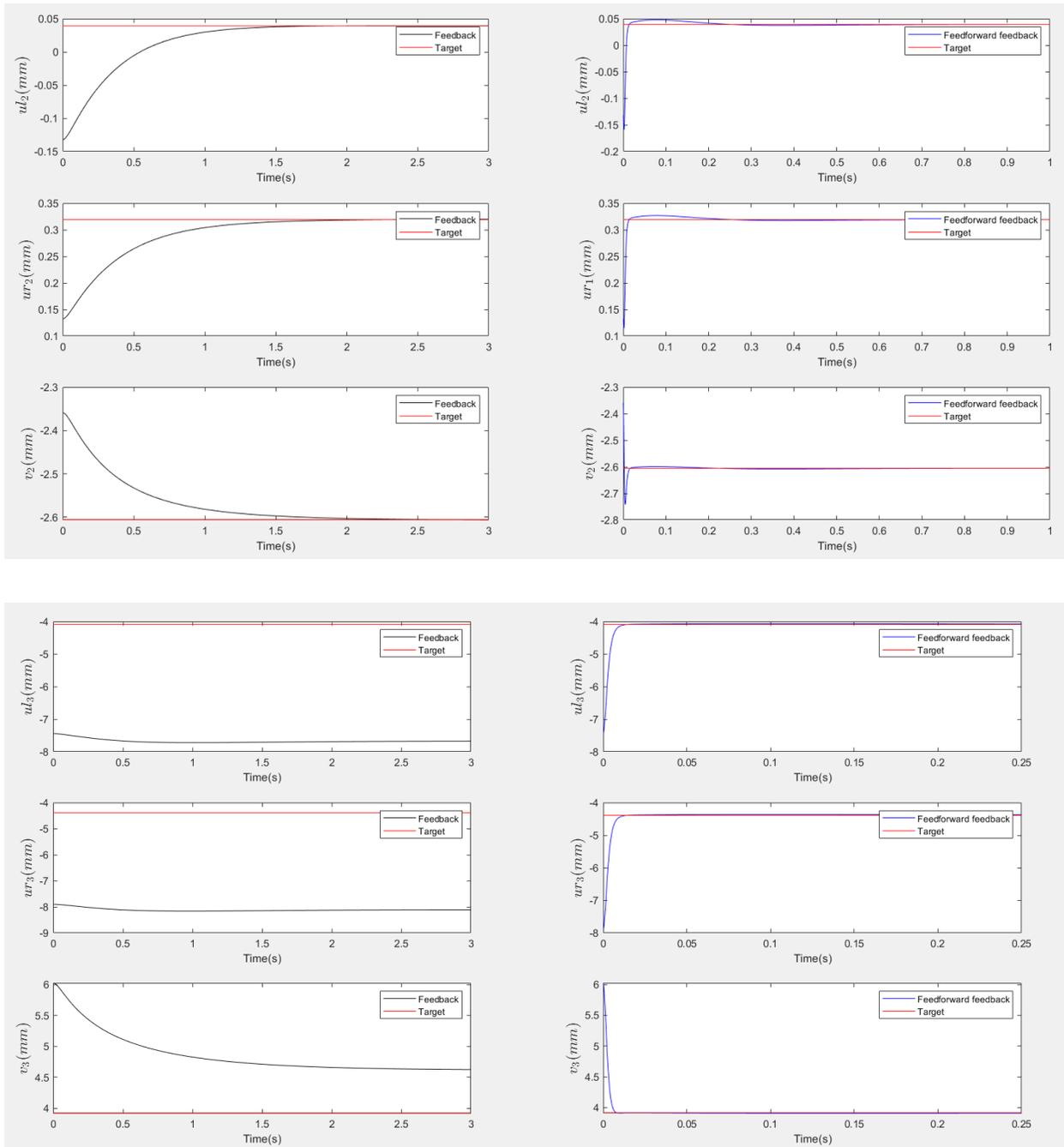

**Figure 7.6**: Scenario one (no disturbances): Response of image coordinates.

*(Left: Feedback-only responses, Right: Feedforward-feedback responses). The three coordinates of the third point are only matched in the feedforward-feedback approach.*

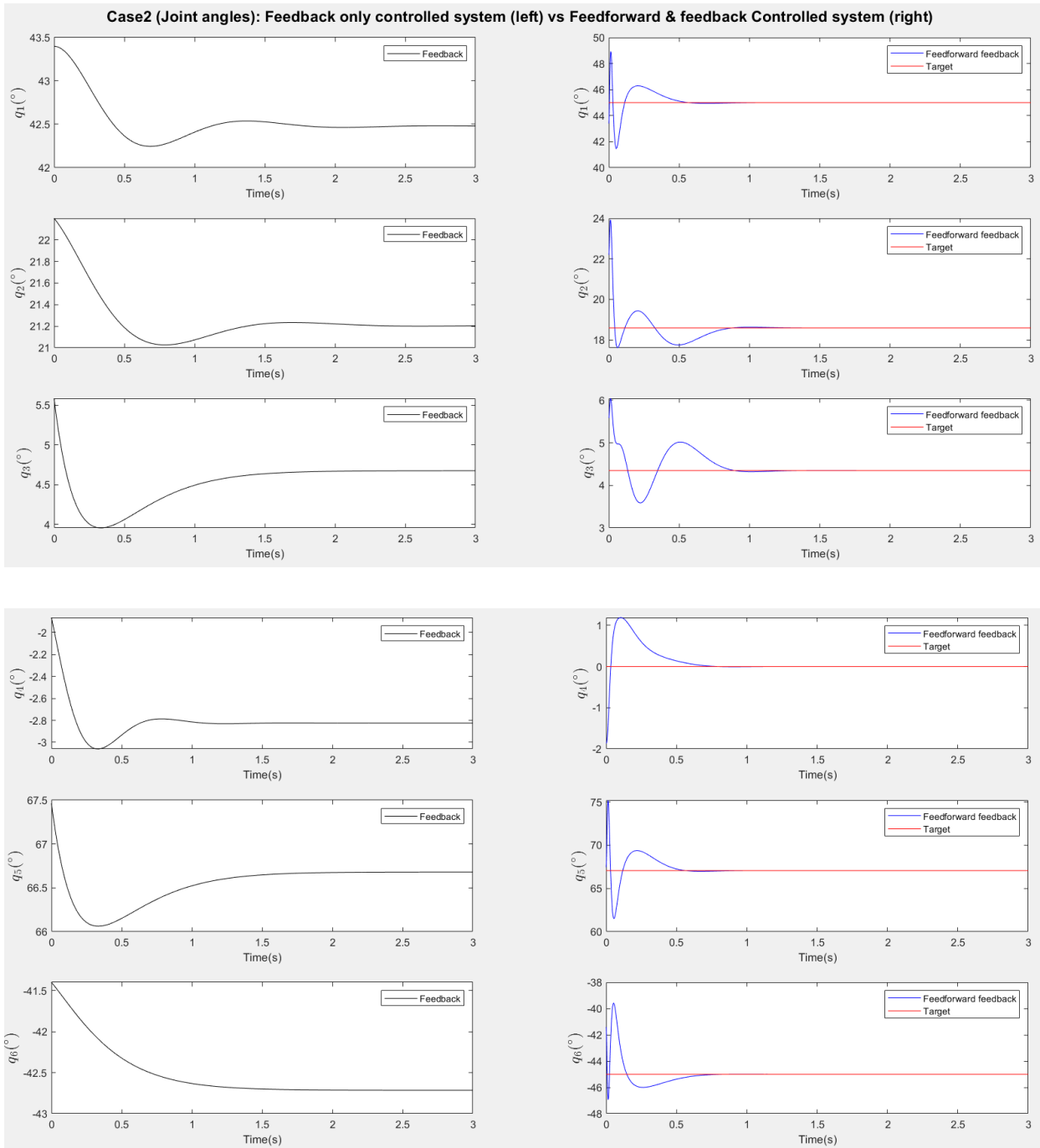

**Figure 7.7**: Scenario two (add disturbances): Response of robot joint angles.

(*Left: Feedback-only responses, Right: Feedforward-feedback responses).*

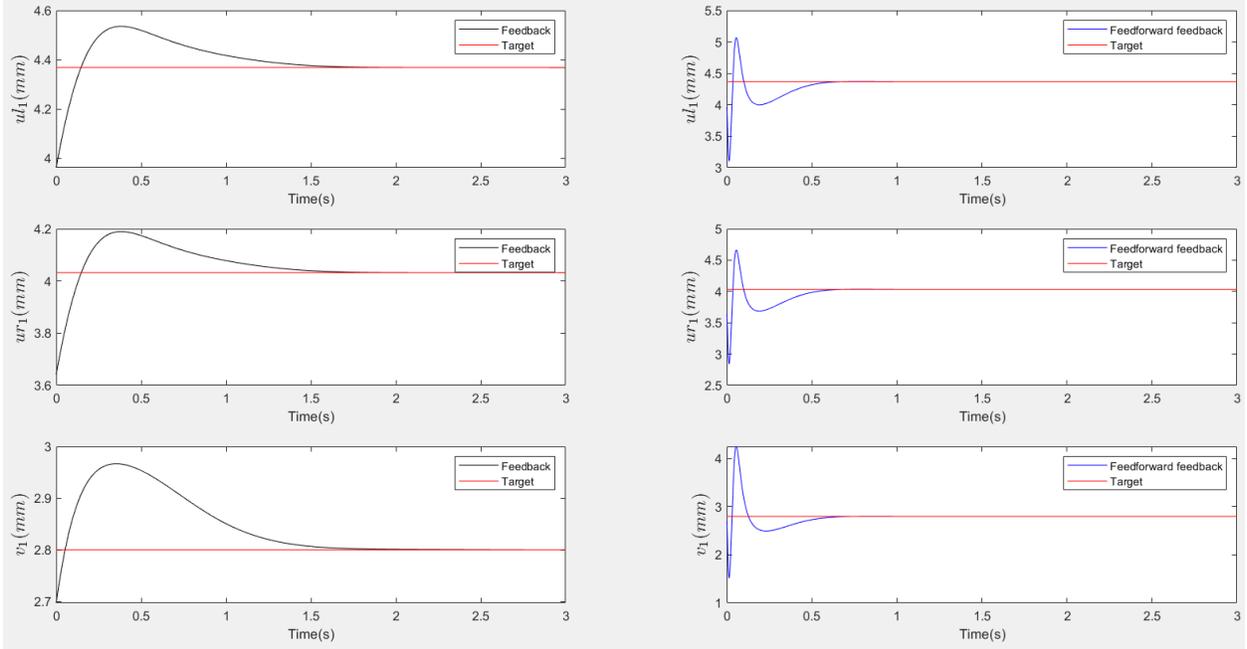
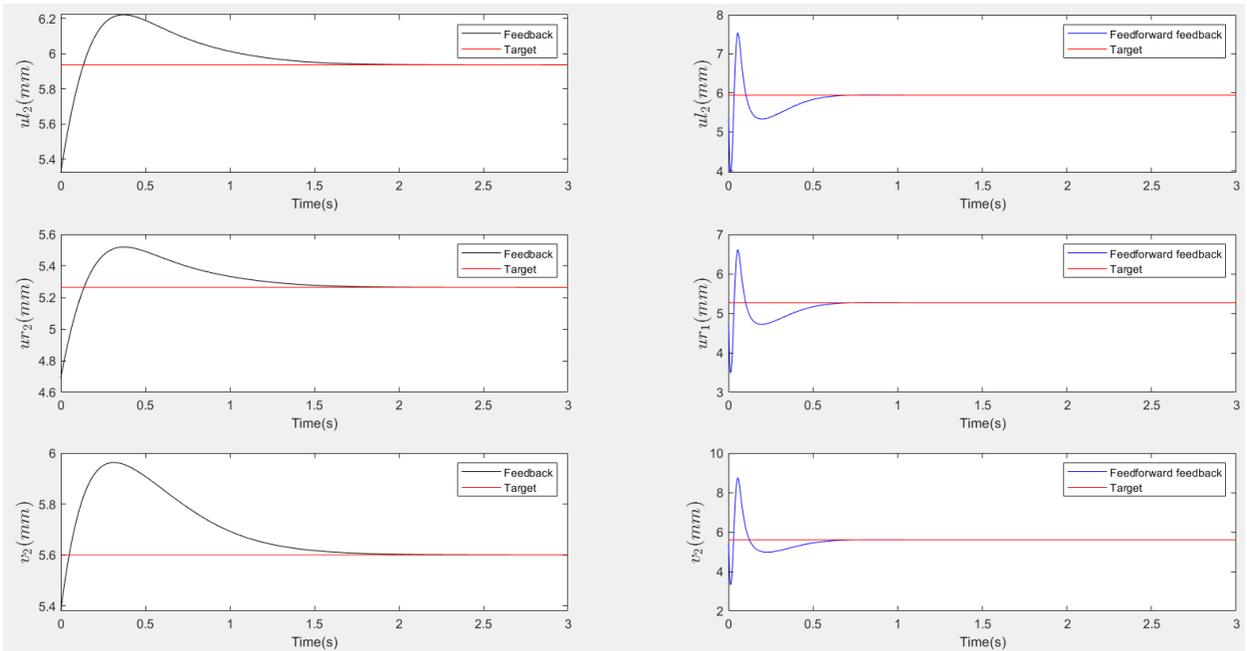

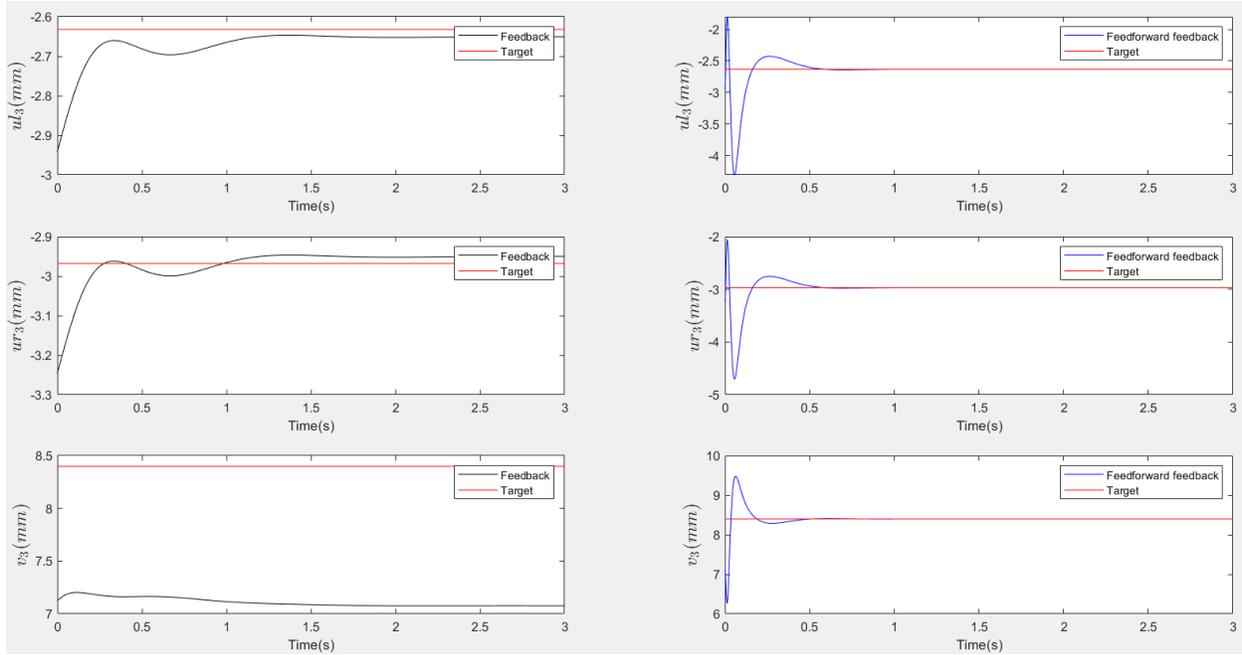

**Figure 7.8**: Scenario two (add disturbances): Response of image coordinates.

(*Left: Feed-back only responses, Right: Feedforward-feedback responses). The three coordinates of the third point are only matched in the feedforward-feedback approach.*

The response plots indicate that both the feedback-only controller and the combined feedforward-and-feedback controller successfully stabilize the system and reach a steady state within three seconds, even in the presence of small input disturbances (Scenario Two). However, the feedback-only controller fails to guide the camera to its desired pose and falls into local minima, as evident from the third point's coordinates ($ul_3, ur_3, v_3,$), which do not match the target at steady state. This issue arises from the overdetermined nature of the stereo-based visual servoing system, where the number of output constraints (9) exceeds the degrees of freedom (DoFs) available for control (6). As a result, the feedback controller can only match six out of nine image coordinates, leaving the rest coordinates unmatched.

In contrast, the feedforward-and-feedback controller avoids local minima and accurately moves the camera to the target pose. This is because the feedforward component directly controls the robot's joint angles rather than image features. Since the joint angles (6 DoFs) uniquely correspond to the camera's pose (6 DoFs), the feedforward controller helps the system reach the global minimum by using the desired joint configurations as inputs.

When comparing performance, the system with the feedforward controller exhibits a shorter transient period (less than 2 seconds) compared to the feedback-only system (less than 3 seconds). However, the feedforward controller can introduce overshoot, particularly in the presence of disturbances. This occurs because feedforward control provides an immediate control action based on desired setpoints, resulting in significant initial actuator input that causes overshoot. Additionally, a feedforward-only system is less robust against disturbances and model uncertainties. Fine-tuning the camera's movement under these conditions requires a feedback controller.

Therefore, the combination of feedforward and feedback control ensures fast and accurate camera positioning. The feedforward controller enables rapid convergence toward the desired pose, while the feedback controller improves robustness and corrects errors due to disturbances or uncertainties. Together, they work cooperatively to achieve optimal performance.

### 7.2.2 Robustness Analysis of Model Uncertainties for the MIMO Visual System

In this section, we investigate the impact of parameter variations on the steady-state error of the image coordinates of reference points in the MIMO visual system control. Specifically, we examine uncertainties in two image coordinates along the $v$-axis:

- $v_1$: the v-coordinate of reference point $R_1$, which is controllable in the feedback-only control system,
- $v_3$: the v-coordinate of reference point $R_3$, which is not controllable in the feedback-only control system.

This analysis is designed to assess whether the controllability of an output affects the system's sensitivity to model parameter uncertainties. As introduced in Section 7.1.2, the parameters selected for variation are the lengths of link 2 and link 4 of the robot manipulator, denoted $L_2$ and $L_4$, respectively. These parameters are chosen due to their relatively large numerical magnitudes, which make them likely contributors to geometric modeling errors in the kinematic chain. Variations in $L_2$ and $L_4$ directly affect the pose of the end-effector, which in turn alters the projected image coordinates of reference points $R_1$ and $R_3$ through the camera's kinematic mapping.

Figures 7.9. and 7.10. present the results of the robustness tests performed by varying the lengths of link 2 and link 4, respectively. Both figures show the percentage steady-state errors of the image coordinates $v_1$ and $v_3$, evaluated under Simulation Scenario 1, in which no input disturbances are introduced.

A threshold of 1% steady-state error is adopted to define an acceptable output variation. From the results, it is evident that the proposed control system maintains acceptable performance for $L_2$ variations in the range of 90% to 110% of its nominal value. In contrast, the system demonstrates greater robustness to variations in $L_4$, with acceptable steady-state errors observed across a wider range—from 60% to 130% of the nominal value. Interestingly, the lack of controllability of certain image coordinates does not degrade robustness. As shown in the figures, the uncontrollable coordinate $v_3$ exhibits lower steady-state errors than the controllable coordinate $v_1$ at equivalent parameter variation levels.

In summary, the adaptive feedforward-feedback control system exhibits strong robustness to geometric parameter uncertainties, particularly in the lengths of key manipulator links, for both controllable and uncontrollable image coordinates.

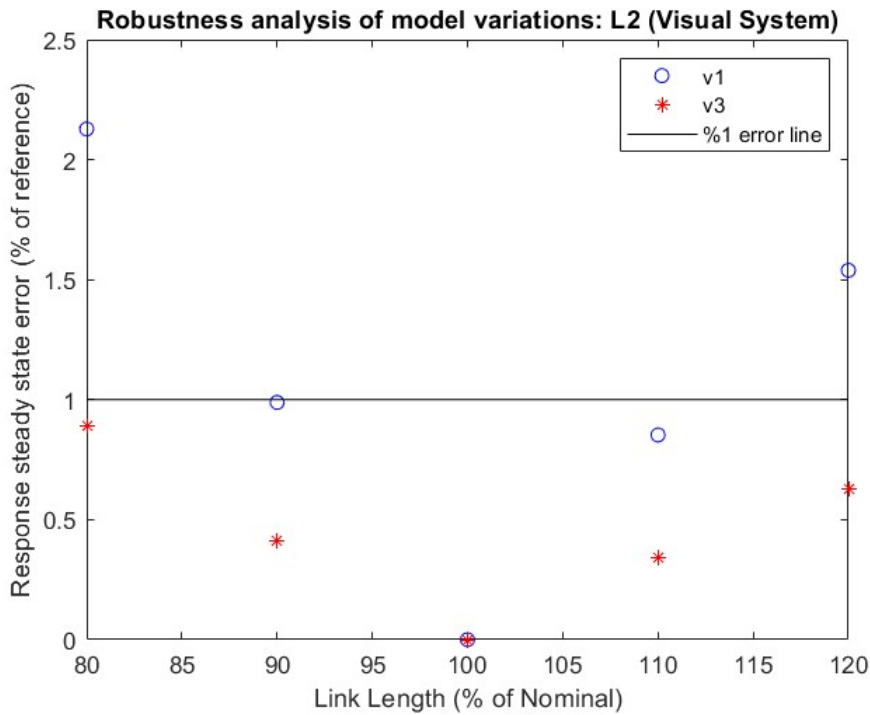

**Figure 7.9:** Robustness analysis of model variations ($L_2$) for the visual system.

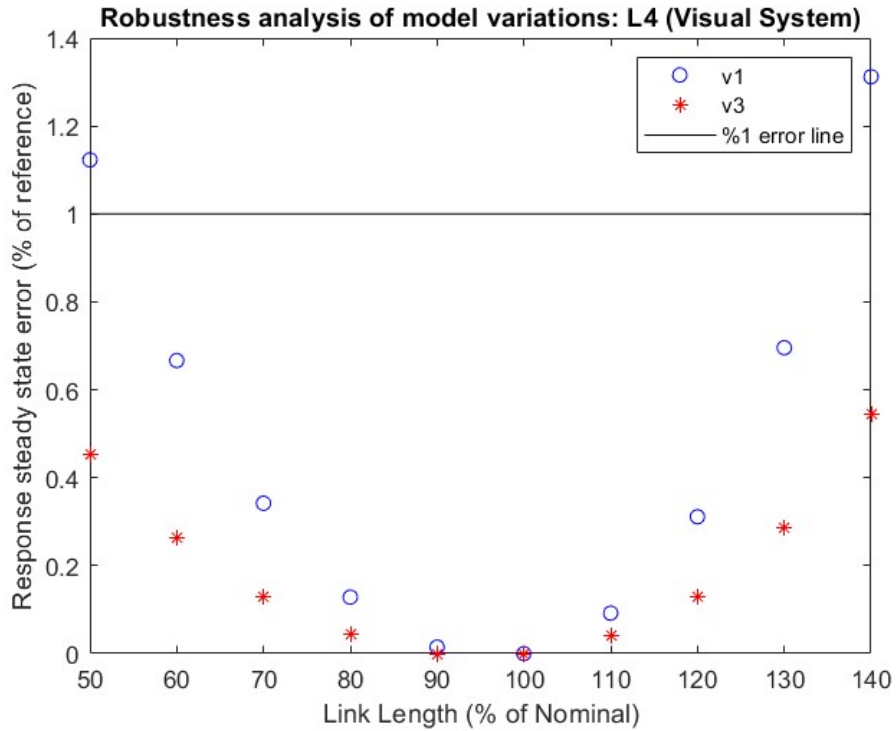

**Figure 7.10:** Robustness analysis of model variations ($L_4$) for the visual system.

## 7.3 Conclusion

In this chapter, we simulated the MIMO outer-loop controllers for both the visual and tool manipulation systems. The feedforward controllers enable rapid transitions from initial to target states, providing a fast dynamic response. Meanwhile, the feedback controllers utilize visual measurements to accurately guide the camera or tool to their desired poses. Notably, the feedforward controller also addresses the overdetermined nature of the visual system by generating reference trajectories that reduce reliance on feedback alone. The combined feedforward-feedback architecture demonstrates strong potential for deployment in high-speed, high-precision manufacturing applications, where both responsiveness and accuracy are critical.

_________________________________________________Chapter 8

# Conclusion and Future work

## 8.1 Conclusions

This research develops a precise control system for a hybrid camera and robot manipulator in automated manufacturing. By modeling both components, we analyze their dynamics and propose a sequence of control methodologies.

First, we introduce an algorithm that enables the camera to explore the workspace and identify positions that minimize image noise within its reach. Simulation results demonstrate the impact of the energy threshold $E_T$ on the algorithm's behavior, suggesting a moderate threshold to balance exploration and energy efficiency.

Next, we design a visual servoing architecture to mitigate measurement and dynamic noise during camera movement. Image averaging is employed to reduce noise, with a Single-Input Single-Output (SISO) case analyzed first. Results indicate that a feedforward-feedback design with Youla model linearization offers the best stability and settling time against input disturbances. The optimal damping ratio $\zeta_{opt}$ is linked to initial joint angles and can be approximated using a function introduced in Section 5.6. In the Multi-Input Multi-Output (MIMO) case, six actuators control nine image coordinate outputs, leaving three coordinates uncontrollable in the feedback loop. A feedforward-feedback approach, introduced in Chapter 6, effectively avoids local minima as shown in Chapter 7.

Finally, we develop a control system to guide the tool to its target while compensating for dynamic errors. SISO control designs are introduced in Chapter 5, followed by a discussion of the MIMO tool manipulation system in Chapter 6. A feedforward controller and feature estimation loop stabilize the tool's pose even outside the camera's view. To enhance performance, an adaptive Youla-based feedback controller ensures stability across the robot's operational range. Simulations in Chapter 7 validate the system's robustness against disturbances and model uncertainties.

Overall, our approach extends classical Image-Based Visual Servoing (IBVS) by improving robustness and performance under uncertainties. The fast response times demonstrated in various scenarios suggest strong potential for real-world implementation in automated manufacturing.

## 8.2 Future Work

Future research will focus on refining the camera movement algorithm. A key area is the analysis of $E_T$ across different applications. Adaptive algorithms that update $K_{est}$ and $K_{sd}$ online will also be explored. Additionally, while this study keeps camera orientations fixed, future work aims to optimize both position and orientation for improved efficiency.

In SISO camera adjustment control, Lyapunov stability is established, and an estimation function for $\zeta_{opt}$ is derived. Extending this analysis to MIMO systems will be a priority.

For tool manipulation, feedforward controllers may drive the tool away from its target under disturbances. A potential solution is a switching algorithm that dynamically transitions between feedforward and feedback control. Another avenue is using $H_\infty$ control techniques [89] to simultaneously design both controllers for enhanced stability and disturbance rejection.

In camera movement control, the feedforward controller ensures global rather than local minima in overdetermined systems. An alternative approach is designing a prefilter to reallocate the system's null space for precise control of all outputs. We aim to compare this method with feedforward control to determine the most suitable approach for high-speed manufacturing.

Finally, we plan to implement our methods in real manufacturing settings by setting up experimental instruments and validating our proposed controllers.

# Appendix A

# Tables of Specification and Figures of Geometric Dimensions

In this section, we will show the geometric model of a specific robot manipulator ABB IRB 4600 45/2.05 and a figure of a camera model: Zed 2 with dimensions. This section also contains specification tables of robots' dimensions, camera, and motor installed inside the joints of manipulators.

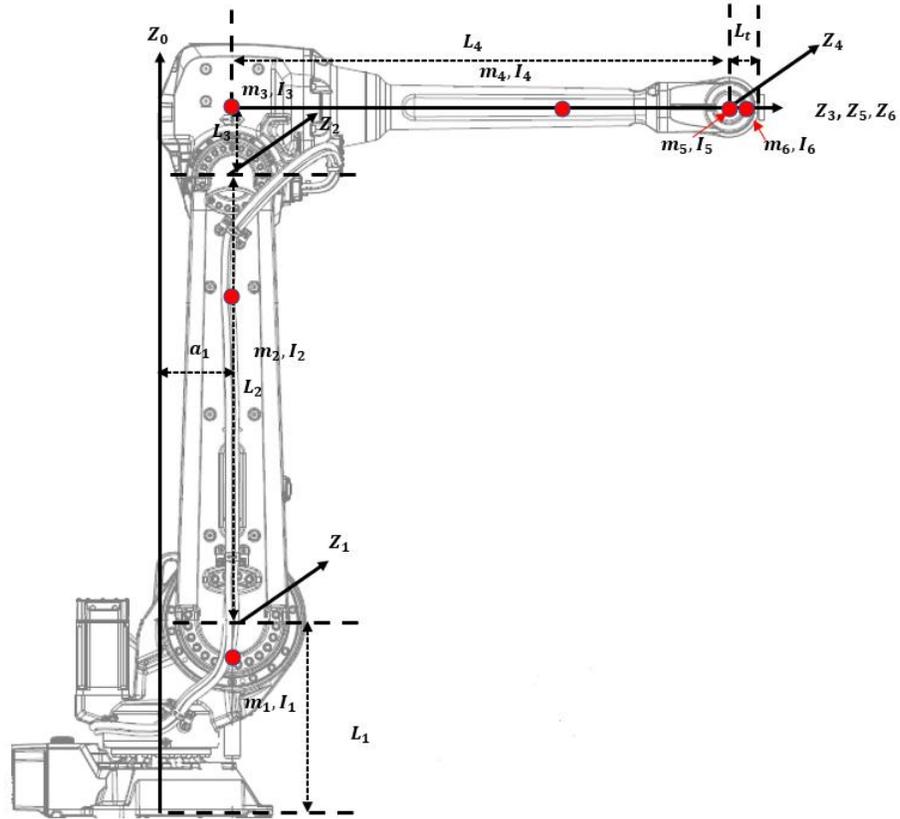

**Figure A.1:** IRB ABB 4600 model

Dimensions are in mm

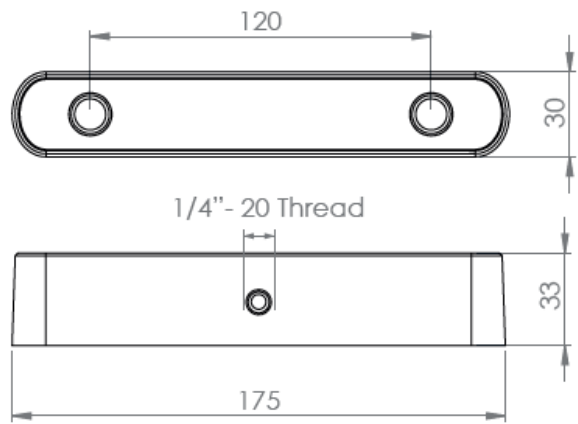

**Figure A.2:** Zed 2 stereo camera model with dimensions.

**Table A.1:** Specification table of ABB IRB 4600 45/2.05 model (dimensions).

| Dimension | Value(m) | Moment of inertial Matrix diagonal value ($I_{xx}, I_{yy}, I_{zz}$) | Value ($kg \cdot m^2$) |
|---|---|---|---|
| Link 1 Length: $a_1$ | 0.175 | Link 1 Moment of inertial: $I_1$ | (0, 0.35, 0) |
| Link 1 Offset: $L_1$ | 0.495 | Link 2 Moment of inertial: $I_2$ | (0.13, 0.524, 0.539) |
| Link 2 Length: $L_2$ | 0.9 | Link 3 Moment of inertial: $I_3$ | (0.066, 0.086, 0.0125) |
| Link 3 Length: $L_3$ | 0.175 | Link 4 Moment of inertial: $I_4$ | (1.8e-3, 1.3e-3, 1.8e-3) |
| Link 4 Offset: $L_4$ | 0.96 | Link 5 Moment of inertial: $I_5$ | (3e-4, 4e-4, 3e-4) |
| Link 6 Offset: $L_t$ | 0.135 | Link 6 Moment of inertial: $I_6$ | (1.5e-4, 1.5e-4, 4e-5) |
| Tool length: $\overline{PJ}_{tool}$/L$_{tool}$ | 0.127 | | |
| Mass | Value(kg) | Center of Mass in its own Reference Frame: ($r_x, r_y, r_z$) | Value(m) |
| Total mass: $m_t$ | 425 | | |
| Link 1 mass: $m_1$ | 3.1 | Link 1 Center of mass: $C_1$ | (0, 0.05, 0) |
| Link 2 mass: $m_2$ | 277.5 | Link 2 Center of mass: $C_2$ | (-0.55, 0, 0.02) |
| Link 3 mass: $m_3$ | 61.9 | Link 3 Center of mass: $C_3$ | (0, 0, 0) |
| Link 4 mass: $m_4$ | 31 | Link 4 Center of mass: $C_4$ | (0, -0.3755, 0) |
| Link 5 mass: $m_5$ | 20.5 | Link 5 Center of mass: $C_5$ | (0, 0, 0) |
| Link 6 mass: $m_6$ | 1 | Link 6 Center of mass: $C_6$ | (0, 0.675, 0) |

**Table A.2:** Specification table of ABB IRB 4600 45/2.05 model (axis working range).

| Axis Movement | Working range |
|---|---|
| Axis 1 rotation | +180° to -180° |
| Axis 2 arm | +150° to -90° |
| Axis 3 arm | +75° to -180° |
| Axis 4 wrist | +400° to -400° |
| Axis 5 bend | +120° to -125° |
| Axis 6 turn | +400° to -400° |

**Table A.3:** Specification table of stereo camera Zed 2.

| Parameters | Values |
|---|---|
| Focus length: f | 2.8 mm |
| Baseline: B | 120 mm |
| Weight: W | 170g |
| Depth range: | 0.5m-25m |
| Diagonal Sensor Size: | 6mm |
| Sensor Format: | 16:9 |
| Sensor Size: W X H | 5.23mm × 2.94mm |
| Angle of view in width: $\alpha$ | 86.09° |
| Angle of view in height: $\beta$ | 55.35° |

**Table A.4:** Specification table of motors and gears.

| Parameters | Values |
| --- | --- |
| **DC Motor** | |
| Armature Resistance: $R$ | 0.03 Ω |
| Armature Inductance: $L$ | 0.1 mH |
| Back emf Constant: $K_b$ | 7 mv/rpm |
| Torque Constant: $K_m$ | 0.0674 N/A |
| Armature Moment of Inertia: $J_a$ | 0.09847 kg$m^2$ |
| **Gear** | |
| Gear ratio: $r$ | 200:1 |
| Moment of Inertia: $J_g$ | 0.05 kg$m^2$ |
| Damping ratio: $B_m$ | 0.06 |

# Appendix B

# Analysis and Proof of LDP for the Stereo Camera

In the cascaded control of camera movement, we need to uniquely determine the orientation and position of the camera in the space. It is a perspective-n-point (PnP) problem: how many $n$ correspondences between 2D projections in images and 3D points in the space are needed to know the perspective that the image is obtained? Paper [66] shows at least 4 correspondences are needed to uniquely determine the pose of a monocular camera (P4P problem).

Take an example of P3P case of a monocular camera. Three points *A, B*, and *C* are in space. $O_1$, $O_2$ and $O_3$ are different perspective centers. The angle ∠ AOB, ∠ AOC and ∠ BOC are the same for all three perspectives. With fixed focal length, we will get same image coordinates of points *A, B*, and *C* if they are observed with a monocular camera from those three perspectives. In other words, we cannot uniquely identify the pose of camera with image coordinates' measurements of just three points.

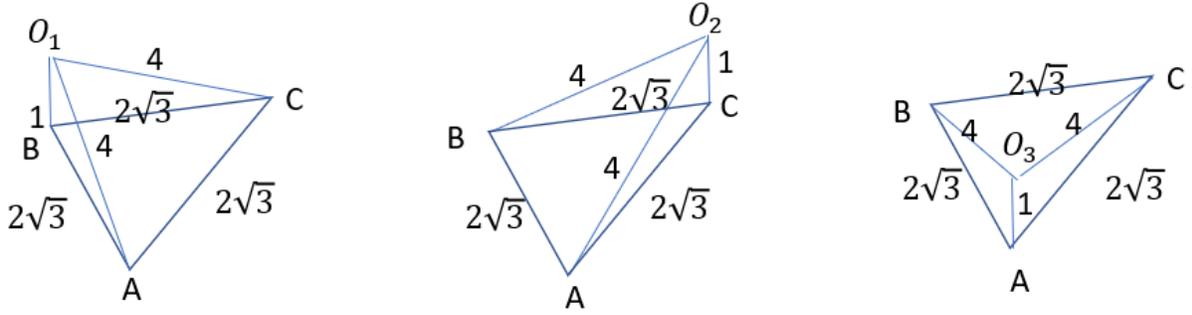

**Figure B.1:** P3P case of a monocular camera.

LDP for a stereo camera is not discussed in research work. A stereo camera can detect three image coordinates of a point in space. It is proposed in this paper that a complete solution of LDP for a stereo camera is a P3P problem. Below is the complete proof of this proposition.

**Proof:**

For a stereo camera shown in Figure 3.2., if the baseline and focal length of the stereo camera are fixed, given the image coordinates $(v, ur, ul)$ of an object point, we can easily compute the 3D coordinates of the object point $(X^C, Y^C, Z^C)$ by setting up a 3D cartesian coordinate at center of baseline:

$$[X^C, Y^C, Z^C]^T = \left[\frac{vb}{ul-ur}, \frac{b(ul+ur)}{2(ul-ur)}, \frac{fb}{ul-ur}\right]^T \tag{B1}$$

Equation (B1) gives a unique mapping from image coordinates of a point to its 3D coordinates in the cartesian frame. The orientation and position of the camera system determines the origin position and axis orientations of the cartesian system in space. Then the LDP becomes, given $n$ points whose 3D coordinates are measured in an unknown cartesian coordinate system in space, what is the smallest number $n$ so that we can determine the position and orientation of the 3D cartesian coordinate frame that has been set up in the space?

**1. P1P** problem for the stereo camera:

If we know one point in space, such as this point's coordinates are given by a 3D Cartesian coordinate system, then there are infinity number of correspond coordinates systems can be set up with.

Any satisfied coordinates can set its origin on the surface of the sphere with its center at the point and the radius $R = \sqrt{X^{C2} + Y^{C2} + Z^{C2}}$, where $[X^C, Y^C, Z^C]^T$ are the coordinates measured by the cartesian system (Figure B2.).

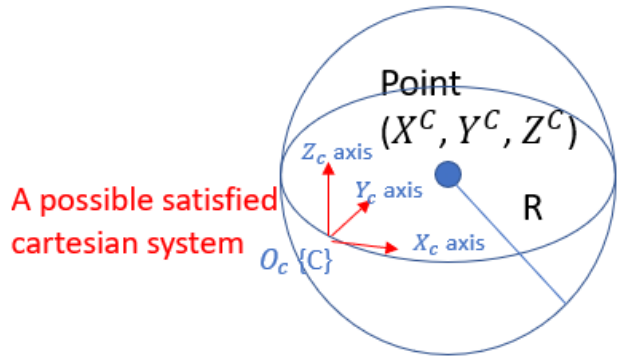

**Figure B.2: P1P** Problem for a Stereo Camera System.

**2. P2P** problem for the stereo camera:

If we know two points' coordinates in space, there are infinity number of corresponding coordinates systems can be set up with. Any satisfied coordinates system can set its origin on the circle with its center at the point $O$ and the radius $R$ shown in Figure B3. The circle is constrained by a triangle $O_C, P_1, P_2$, whose sides' lengths are denoted as $R_1, R_2$ and $D$, where

$R_1 = \sqrt{X_1^{C2} + Y_1^{C2} + Z_1^{C2}}$, $R_2 = \sqrt{X_2^{C2} + Y_2^{C2} + Z_2^{C2}}$, and $D = \sqrt{(X_1^C - X_2^C)^2 + (Y_1^C - Y_2^C)^2 + (Z_1^C - Z_2^C)^2}$. Then the radius of the circle $R$ is the height of the base $D$ of the triangle. The center of the circle $O$ is the cross-section of the height and the base.

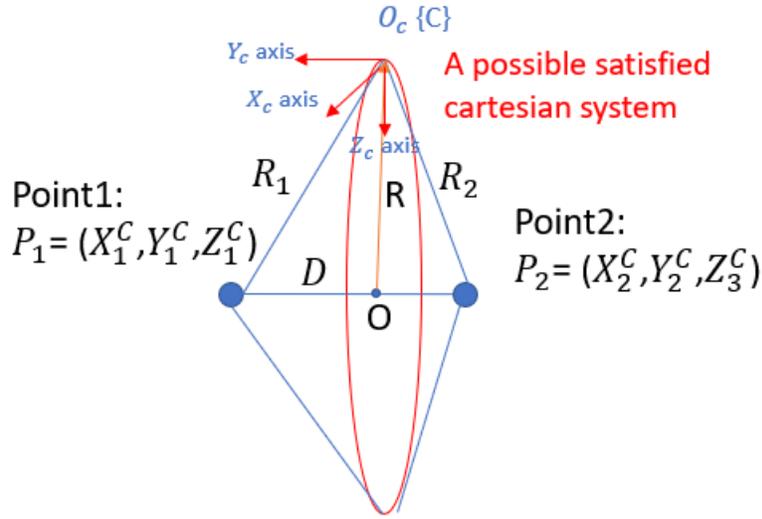

**Figure B.3: P2P** problem for a stereo camera system.

### 3. P3P problem for the stereo camera:

If we know three points in the space, where the lines connecting these points are not co-linear, then we can uniquely set up one coordinate system. As shown in Figure B4. below, three points which are not co-linear can define a plane in space with its normal unit vector $\vec{n}$ uniquely defined. With coordinates of three points pre-known, the vector $\vec{n}$ can be calculated as the following. Vector $\overrightarrow{P_1P_2} = (X_2^C - X_1^C, Y_2^C - Y_1^C, Z_2^C - Z_1^C)$, $\overrightarrow{P_1P_3} = (X_3^C - X_1^C, Y_3^C - Y_1^C, Z_3^C - Z_1^C)$. The unit vector is in the direction perpendicular to the plane and can be expressed as $\vec{n} = \frac{\overrightarrow{P_1P_2} \times \overrightarrow{P_1P_3}}{|\overrightarrow{P_1P_2} \times \overrightarrow{P_1P_3}|}$, where $X$ indicates the cross product. The angles between $\vec{n}$ and XYZ axis of the coordinate system are expressed respectively in the following:

$$\cos(\theta_x) = \vec{n} \cdot \vec{i} \quad \text{(B2)}$$

$$\cos(\theta_y) = \vec{n} \cdot \vec{j} \quad \text{(B3)}$$

$$\cos(\theta_z) = \vec{n} \cdot \vec{k} \quad \text{(B4)}$$

Where $\theta_x, \theta_y$ and $\theta_z$ are angles between $\vec{n}$ and unit vectors in X, Y, and Z directions denoted as $\vec{i}$, $\vec{j}$, and $\vec{k}$. Therefore, with $\vec{n}$ direction is fixed in space, the orientations of each axis of the coordinates system can be computed uniquely above.

**P1P** problem already shows that origin of the coordinate system must be on the surface of sphere with center at $P_1$ and radius $R = \sqrt{X_1^{C2} + Y_1^{C2} + Z_1^{C2}}$ as shown in Figure B2. Each satisfied coordinate system built with a different origin point on the surface of the sphere results in a unique configuration of the axis' orientations. Therefore, as the axis' orientations are defined in space, the position of the frame (or the position of the origin) is also uniquely defined in space. In conclusion, **P3P** problem is sufficient for the LDP of a stereo camera system.

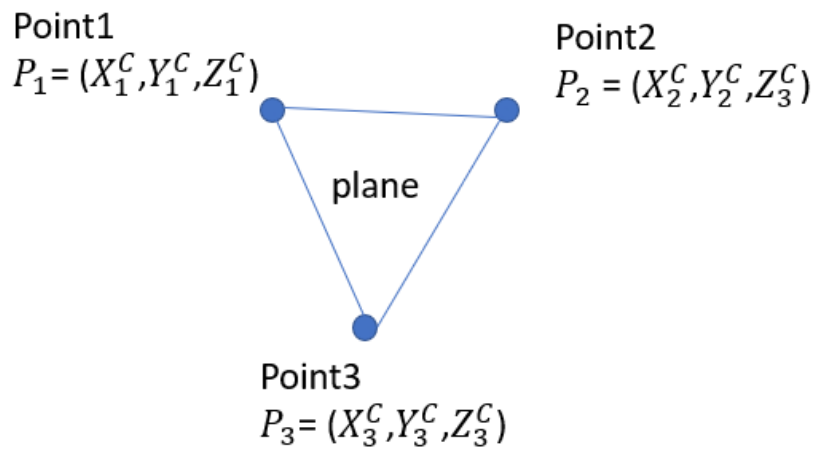

**Figure B.4: P3P** Problem for a Stereo Camera System.

***Prove concluded***

# Appendix C

# Steps of D-H Convention

Transformation matrix between joints can be obtained by D-H convention. Here the following are steps that one can follow and generate matrix: Equation (3.48) on any type of robot manipulators.

Step 1: Locate and label the joint axes $z_0, \ldots z_{n-1}$ where $n$ is number of joints.

Step 2: Establish the base frame. Set the origin anywhere on the $z_0$-axis. The $x_0$ and $y_0$ axes are chosen conveniently to form a right-hand frame.

For $i = 1, \ldots n - 1$, perform Steps 3 to 5.

Step 3: Locate the origin $o_i$ where the common normal to $z_i$ and $z_{i-1}$ intersects $z_i$. If $z_i$ intersects $z_{i-1}$ locate $o_i$ at this intersection. If $z_i$ and $z_{i-1}$ are parallel, locate $o_i$ at joint $i$.

Step 4: Establish $x_i$ along the common normal between $z_{i-1}$ and $z_i$ through $o_i$, or in the direction normal to the $z_{i-1}$-$z_i$ plane if $z_{i-1}$ and $z_i$ intersect.

Step 5: Establish $y_i$ to complete a right-hand frame.

Step 6: Establish the end-effector frame $o_n x_n y_n z_n$. Assuming the $n$-th joint is revolute, set $k_n = a$ along the direction $z_{n-1}$. Establish the origin $o_n$ conveniently along $z_n$, preferably at the center of the gripper or at the tip of any tool that the manipulator may be carrying. Set $j_n = s$ the direction of the gripper closure and set $i_n = n$ as $s \times a$. If the tool is not a simple gripper, set $x_n$ and $y_n$ conveniently to form a right-hand frame.

Step 7: Create a table of link parameters $a_i, d_i, \alpha_i, \theta_i$.

$a_i$ = distance along $x_i$ from $o_i$ to the intersection of the $x_i$ and $z_{i-1}$ axes.

$d_i$ = distance along $z_{i-1}$ from $o_{i-1}$ to the intersection of the $x_i$ and $z_{i-1}$ axes. $d_i$ is variable if joint $i$ is prismatic.

$\alpha_i$ = the angle between along $z_{i-1}$ and $z_i$ measured about $z_{i-1}$ and $x_i$ (See Figure C1.).

$\theta_i$ = the angle between $x_{i-1}$ and $x$ measured about $z_{i-1}$ (See Figure C1.). $\theta_i$ is variable is joint $i$ is revolute.

Step 8: Form the homogeneous transformation matrices $A_i$ by substituting the above parameters into Equation (3.48).

Step 9: Form $T_0^n = A_1 \ldots A_n$. This then gives the position an orientation of the tool frame expressed in base coordinates.

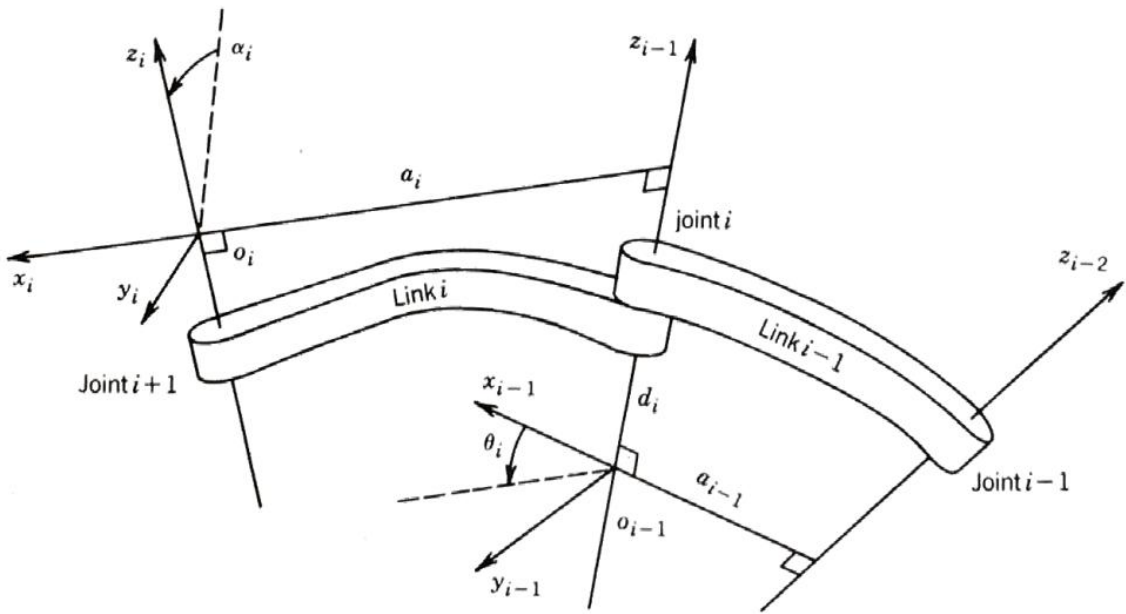

**Figure C.1:** D-H frame assignment.

# Appendix D

## Derivation of Inverse Kinematics of Non-Singularity Pose in ABB IRB 4600 Six DOFs Robot Manipulators

In this section, we derive the mathematical expressions for the inverse kinematics of a manipulator in a non-singular configuration, as presented in Equations (3.60) through (3.66).

As illustrated in Figure D1., the end-effector $E$ is offset from the wrist center $P$ by a distance $L_t$, measured along the rotational axis of joint 6, denoted as $Z_6$ This axis is also aligned with the end-effector's roll axis, represented by the unit vector $\vec{a}$. The position of the end-effector in the base frame $\{O_0 X_0 Y_0 Z_0\}$ is given by $[d_x, d_y, d_z]^T$. Likewise, the position of the wrist center $P$, also expressed in the base frame is $P = [P_x, P_y, P_z]^T$. The orientation vector $\vec{a}$ is decomposed in the base frame as $\vec{a} = [a_x, a_y, a_z]^T$.

We obtain the relationship between the end-effector and the wrist center as:

$$d = P + L_t \cdot \vec{a} \qquad (D1)$$

This can be rewritten component-wise along each coordinate axis:

$$P_x = d_x - L_t a_x \qquad (D2)$$

$$P_y = d_y - L_t a_y \qquad (D3)$$

$$P_z = d_z - L_t a_z \qquad (D4)$$

The position control is isolated from the orientation control, as the position of wrist center is only determined by first three joint angles $(q_1, q_2, q_3)$.

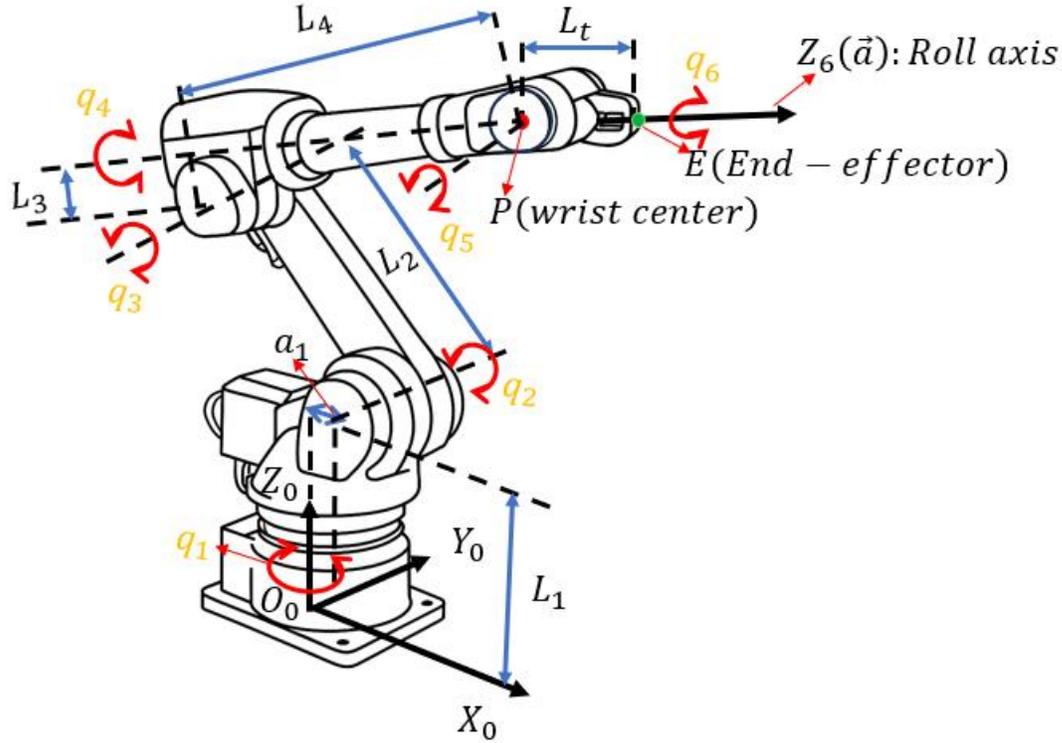

**Figure D.1:** Kinematic Decoupling: Illustration of the geometric separation between the wrist center (position, controlled by joints $q_1, q_2, q_3$) and the end-effector orientation (controlled by joints $q_4, q_5, q_6$).

As shown in Figure D2., the wrist center $P$ is projected onto the $X_0 - Y_0$ plane as $P'$. The first joint angle $q_1$ is defined as the angle between the base frame's $X_0$-axis and the projection vector $\overline{O_0 P'}$. This corresponds to a standard planar rotation and is computed as:

**Right**: 
$$q_1 = arctan(\frac{P_y}{P_x}) \quad (D5.1)$$

This solution corresponds to the right-arm configuration. The alternative left-arm configuration differs by $\pi$, yielding:

**Left**: 
$$q_1 = arctan\left(\frac{P_y}{P_x}\right) - \pi \quad (D5.2)$$

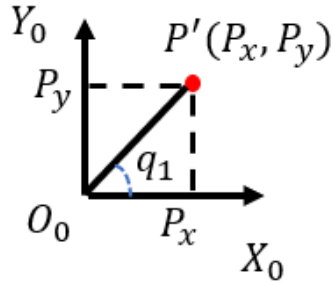

**Figure D.2:** Determination of joint angle $q_1$ by projecting the wrist center $P$ onto the $X_0 - Y_0$ plane.

To solve for the joint angles $q_2$ and $q_3$, we analyze the geometry of the manipulator's first three links in the sagittal plane — the vertical plane formed by the shoulder, elbow, and wrist center.

As shown in Figure D3., this plane contains the wrist center $P$, and the projection of the vector from joint 2 to the wrist center can be decomposed as:

$r$: the horizontal (base plane) distance from the shoulder to the wrist center

$Z$: the vertical displacement between the shoulder and wrist center

These components are given by:

$$r = \sqrt{P_x^2 + P_y^2} - a_1 \tag{D6}$$

$$Z = P_z - L_1 \tag{D7}$$

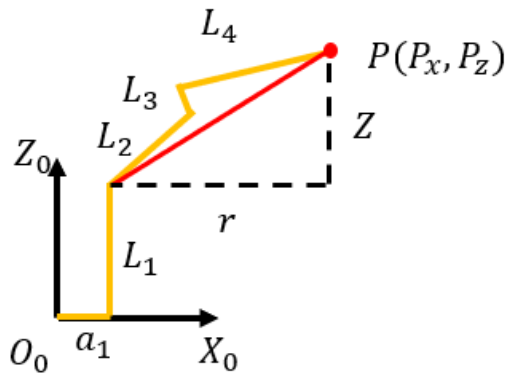

**Figure D.3:** Example of the sagittal plane..

Figure D4. illustrates the sagittal-plane triangle for right-arm configurations. The triangle is formed by:

$|O_1P|$: the distance from joint 2 (shoulder) to the wrist center

$|O_1A|=L_2$: the upper arm link

$|AP|=L_3^2 + L_4^2$: the forearm link length

Applying the law of cosines, the angle $\theta$ at joint 2 is:

$$\theta = \arccos\left(\frac{|O_1P|^2 + |O_1A|^2 - |AP|^2}{2|O_1P||O_1A|}\right) \tag{D8}$$

$$= \arccos\left(\frac{(r^2 + Z^2) + L_2^2 - (L_3^2 + L_4^2)}{2 \cdot \sqrt{r^2 + Z^2} \cdot L_2}\right)$$

Similarly, applying the law of cosines again to solve for angle $\delta$ at the elbow:

$$\delta = \arccos\left(\frac{|AP|^2 + |O_1A|^2 - |O_1P|^2}{2|AP||O_1A|}\right) \tag{D9}$$

$$= \arccos\left(\frac{(L_3^2 + L_4^2) + L_2^2 - (r^2 + Z^2)}{2 \cdot \sqrt{L_3^2 + L_4^2} \cdot L_2}\right)$$

**Right above Configuration:**

For the right-above (elbow-up) configuration, the joint angles are:

$$q_2 = \frac{\pi}{2} - \theta - \arctan\left(\frac{Z}{r}\right) \tag{D10.1}$$

Then, the third joint angle $q_3$ is

$$q_3 = \pi - \delta - \arctan\left(\frac{L_4}{L_3}\right) \tag{D10.2}$$

**Right down Configuration:**

For the right-down (elbow-down) configuration:

$$q_2 = \frac{\pi}{2} + \theta - \arctan\left(\frac{Z}{r}\right) \tag{D11.1}$$

$$q_3 = \pi + \delta - \arctan\left(\frac{L_4}{L_3}\right) \tag{D11.2}$$

Figure D5. shows the sagittal-plane triangles for the left-arm configurations. The angle derivation follows the same geometric structure.

**Left above Configuration:**

$$q_2 = -\frac{\pi}{2} + \theta + \arctan\left(\frac{Z}{r}\right) \tag{D12.1}$$

$$q_3 = -\pi + \delta - \arctan\left(\frac{L_4}{L_3}\right) \tag{D12.2}$$

**Left down Configuration:**

$$q_2 = -\frac{\pi}{2} - \theta + \arctan\left(\frac{Z}{r}\right) \tag{D13.1}$$

$$q_3 = -\pi - \delta - \arctan\left(\frac{L_4}{L_3}\right) \tag{D13.2}$$

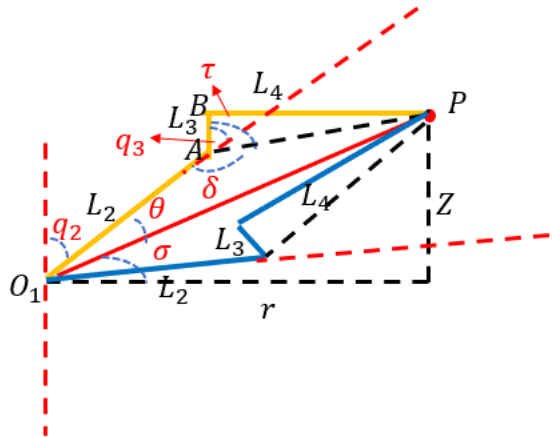

**Figure D.4**: Determination of joint angles $q_2$ and $q_3$: right above and right down configurations.

**Figure D.5**: Determination of joint angles $q_2$ and $q_3$: left above and left down configurations.

Given joint angles $q_1$, $q_2$, and $q_3$, we can compute the rotation matrix that transforms coordinates from frame 3 (attached to the fourth joint) to the base frame (frame 0). This rotation matrix is denoted:

$$R_3^0 \in \mathbb{R}^{3\times3} = R_3^2 \cdot R_2^1 \cdot R_1^0 \tag{D14}$$

Where $R_3^2 \in \mathbb{R}^{3\times3}$, $R_2^1 \in \mathbb{R}^{3\times3}$, and $R_1^0 \in \mathbb{R}^{3\times3}$ are the rotation submatrices of the homogeneous transformation matrices $A_3^2 \in \mathbb{R}^{4\times4}$, $A_2^1 \in \mathbb{R}^{4\times4}$, and $A_1^0 \in \mathbb{R}^{4\times4}$ in Equation (3.37) – (3.39), respectively.

By explicitly multiplying out the product in Equation (D14), we obtain the following expression for $R_3^0$:

$$R_3^0 = \begin{bmatrix} c_1 s_{2,3} & s_1 & c_1 c_{2,3} \\ s_1 s_{2,3} & -c_1 & s_1 c_{2,3} \\ c_{2,3} & 0 & -s_{2,3} \end{bmatrix} \tag{D15}$$

The rotation matrix from the base frame to frame 3, $R_0^3$, is the inverse of $R_3^0$, and since the rotation matrix is orthogonal, this inverse equals its transpose:

$$R_0^3 = (R_3^0)^{-1} = (R_3^0)^T = \begin{bmatrix} c_1 s_{2,3} & s_1 s_{2,3} & c_{2,3} \\ s_1 & -c_1 & 0 \\ c_1 c_{2,3} & s_1 c_{2,3} & -s_{2,3} \end{bmatrix} \tag{D16}$$

Note:
$$c_i \equiv \cos(q_i), s_i \equiv \sin(q_i)$$
$$c_{i,j} \equiv \cos(q_i + q_j), s_{i,j} \equiv \sin(q_i + q_j) \tag{D17}$$
$$i, j \in \{1,2,3\}$$

Figure D6. illustrates the portion of the manipulator associated with the orientation of the end-effector, which is governed by the final three joint angles: $q_4, q_5, q_6$.

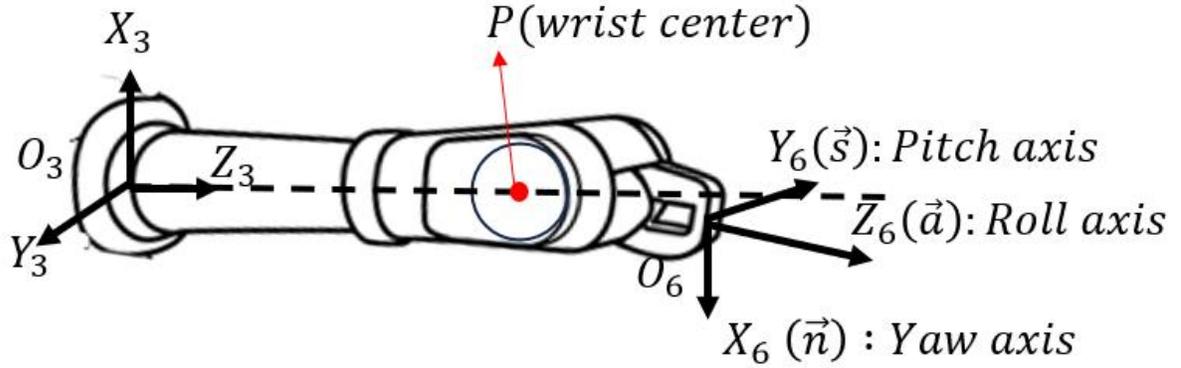

**Figure D.6**: Robot arm joint and link configuration influencing end-effector orientation.

Given the joint angles $q_1, q_2$, and $q_3$, the orientation of the wrist (i.e., the final three joints $q_4, q_5, q_6$) can be determined using the end-effector axes and the frame transformation.

The joint angle $q_5$ corresponds to the angular deviation between the unit vector $\widehat{Z_6}$ (or equivalently, the approach vector $\vec{a}$ and the axis $\widehat{Z_3}$, both expressed in frame 3. The unit vector $\vec{a} = [a_x, a_y, a_z]^T$ is originally measured in the base frame (frame 0). Using the rotation matrix $R_0^3$ from Equation (D16), $\vec{a}$ is transformed into frame 3 as:

$$\vec{a}^{(3)} = R_0^3 \cdot \vec{a}^{(3)} = \begin{bmatrix} c_1 s_{2,3} & s_1 s_{2,3} & c_{2,3} \\ s_1 & -c_1 & 0 \\ c_1 c_{2,3} & s_1 c_{2,3} & -s_{2,3} \end{bmatrix} \cdot \begin{bmatrix} a_x \\ a_y \\ a_z \end{bmatrix} \qquad (D18)$$

The unit vector $\widehat{Z_3}$ in its own frame is trivial:

$$\widehat{Z_3}^{(3)} = \begin{bmatrix} 0 \\ 0 \\ 1 \end{bmatrix} \qquad (D19)$$

Thus, the cosine of the joint angle $q_5$ is the projection of $\vec{a}^{(3)}$ onto the $\widehat{Z_3}^{(3)}$ axis:

$$q_5 = \arccos\left((\widehat{Z_3}^{(3)})^T \cdot \vec{a}^{(3)}\right) = \arccos\left(c_1 c_{2,3} a_x + s_1 c_{2,3} a_y - s_{2,3} a_z\right) \qquad (D20)$$

As shown in Figure D7., the angle $q_4$ is the azimuthal angle of $\vec{a}^{(3)}$ projected onto the $X_3 - Y_3$ plane. It is computed as:

$$q_4 = \mathrm{artan}\left(\frac{\vec{a}_Y^{(3)}}{\vec{a}_X^{(3)}}\right) = \mathrm{artan}\left(\frac{s_1 a_x - c_1 a_y}{c_1 s_{2,3} a_x + s_1 s_{2,3} a_y + c_{2,3} a_z}\right) \qquad (D21)$$

The joint angle $q_6$ represents the rotational alignment of the end-effector's x-axis $\vec{n}$ and y-axis $\vec{s}$ about the aligned $\widehat{Z}_3^{(3)}$ axis. As shown in Figure D8., when $q_6= 0$, the projection of $\widehat{Z}_3^{(3)}$ lies along $\vec{n}^{(3)}$. As joint 6 rotates, $\vec{n}^{(3)}$ and $\vec{s}'^{(3)}$ rotate accordingly.

We compute their transformed representations in frame 3 as:

$$\vec{n}'^{(3)} = R_0^3 \cdot \vec{n} = \begin{bmatrix} c_1 s_{2,3} & s_1 s_{2,3} & c_{2,3} \\ s_1 & -c_1 & 0 \\ c_1 c_{2,3} & s_1 c_{2,3} & -s_{2,3} \end{bmatrix} \cdot \begin{bmatrix} n_x \\ n_y \\ n_z \end{bmatrix} \tag{D22}$$

$$\vec{s}'^{(3)} = R_0^3 \cdot \vec{s} = \begin{bmatrix} c_1 s_{2,3} & s_1 s_{2,3} & c_{2,3} \\ s_1 & -c_1 & 0 \\ c_1 c_{2,3} & s_1 c_{2,3} & -s_{2,3} \end{bmatrix} \cdot \begin{bmatrix} s_x \\ s_y \\ s_z \end{bmatrix} \tag{D23}$$

The joint angle $q_6$ is given by:

$$q_6 = artan\left(\frac{-(\widehat{Z}_3^{(3)})^T \cdot \vec{s}'^{(3)}}{(\widehat{Z}_3^{(3)})^T \cdot \vec{n}'^{(3)}}\right) = -artan\left(\frac{c_1 c_{2,3} s_x + s_1 c_{2,3} s_y - s_{2,3} s_z}{c_1 c_{2,3} n_x + s_1 c_{2,3} n_y - s_{2,3} n_z}\right) \tag{D24}$$

Due to the symmetry of the wrist mechanism, an equivalent set of joint angles $\{q_4 + \pi, -q_5, q_6 + \pi\}$ yields the same end-effector orientation as the original solution $\{q_4, q_5, q_6\}$. Thus, the alternate solution can be written as:

$$q_4 = artan\left(\frac{s_1 a_x - c_1 a_y}{c_1 s_{2,3} a_x + s_1 s_{2,3} a_y + c_{2,3} a_z}\right) + \pi \tag{D25}$$

$$q_5 = -c_1 c_{2,3} a_x - s_1 c_{2,3} a_y + s_{2,3} a_z \tag{D26}$$

$$q_6 = -artan\left(\frac{c_1 c_{2,3} s_x + s_1 c_{2,3} s_y - s_{2,3} s_z}{c_1 c_{2,3} n_x + s_1 c_{2,3} n_y - s_{2,3} n_z}\right) + \pi \tag{D27}$$

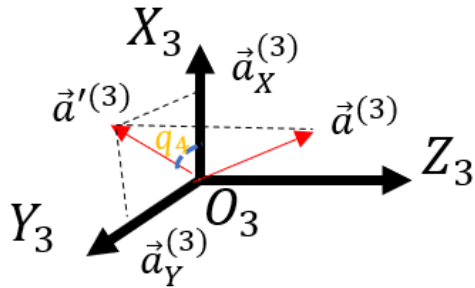

**Figure D.7:** Determination of joint angle $q_4$: Projection of vector $\vec{a}^{(3)}$ onto the $X_3 - Y_3$ plane.

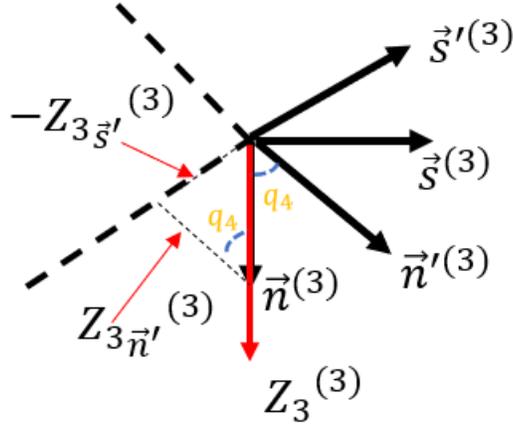

**Figure D.8:** Determination of joint angle $q_6$: Projection of vector $\widehat{Z_3}^{(3)}$ onto the $\vec{n}^{(3)} - \vec{s}^{(3)}$ plane.

To summarize, the joint configurations for the non-singularity pose can be expressed as follows:

**Joint angles that relate to the wrist center locations $(q_1, q_2, q_3)$:**

**Right above Configuration:**

$$q_1 = arctan(\frac{P_y}{P_x}) \tag{D28.1}$$

$$q_2 = \frac{\pi}{2} - \arccos\left(\frac{(r^2 + Z^2) + L_2{}^2 - (L_3{}^2 + L_4{}^2)}{2 \cdot \sqrt{r^2 + Z^2} \cdot L_2}\right) - \arctan(\frac{Z}{r}) \tag{D29.1}$$

$$q_3 = \pi - \arccos\left(\frac{(L_3{}^2 + L_4{}^2) + L_2{}^2 - (r^2 + Z^2)}{2 \cdot \sqrt{L_3{}^2 + L_4{}^2} \cdot L_2}\right) - \arctan(\frac{L_4}{L_3}) \tag{D30.1}$$

**Right down Configuration:**

$$q_1 = arctan(\frac{P_y}{P_x}) \tag{D28.2}$$

$$q_2 = \frac{\pi}{2} + \arccos\left(\frac{(r^2 + Z^2) + L_2{}^2 - (L_3{}^2 + L_4{}^2)}{2 \cdot \sqrt{r^2 + Z^2} \cdot L_2}\right) - \arctan(\frac{Z}{r}) \tag{D29.2}$$

$$q_3 = \pi + \arccos\left(\frac{(L_3{}^2 + L_4{}^2) + L_2{}^2 - (r^2 + Z^2)}{2 \cdot \sqrt{L_3{}^2 + L_4{}^2} \cdot L_2}\right) - \arctan\left(\frac{L_4}{L_3}\right) \qquad \text{(D30.2)}$$

**Left above Configuration:**

$$q_1 = \arctan\left(\frac{P_y}{P_x}\right) - \pi \qquad \text{(D28.3)}$$

$$q_2 = -\frac{\pi}{2} + \arccos\left(\frac{(r^2 + Z^2) + L_2{}^2 - (L_3{}^2 + L_4{}^2)}{2 \cdot \sqrt{r^2 + Z^2} \cdot L_2}\right) + \arctan\left(\frac{Z}{r}\right) \qquad \text{(D29.3)}$$

$$q_3 = -\pi + \arccos\left(\frac{(L_3{}^2 + L_4{}^2) + L_2{}^2 - (r^2 + Z^2)}{2 \cdot \sqrt{L_3{}^2 + L_4{}^2} \cdot L_2}\right) - \arctan\left(\frac{L_4}{L_3}\right) \qquad \text{(D30.3)}$$

**Left down Configuration:**

$$q_1 = \arctan\left(\frac{P_y}{P_x}\right) - \pi \qquad \text{(D28.3)}$$

$$q_2 = -\frac{\pi}{2} - \arccos\left(\frac{(r^2 + Z^2) + L_2{}^2 - (L_3{}^2 + L_4{}^2)}{2 \cdot \sqrt{r^2 + Z^2} \cdot L_2}\right) + \arctan\left(\frac{Z}{r}\right) \qquad \text{(D29.3)}$$

$$q_3 = -\pi - \arccos\left(\frac{(L_3{}^2 + L_4{}^2) + L_2{}^2 - (r^2 + Z^2)}{2 \cdot \sqrt{L_3{}^2 + L_4{}^2} \cdot L_2}\right) - \arctan\left(\frac{L_4}{L_3}\right) \qquad \text{(D30.3)}$$

**Joint angles that relate to the end-effector orientations ($q_4, q_5, q_6$):**

**The first solution set:**

$$q_4 = \arctan\left(\frac{s_1 a_x - c_1 a_y}{c_1 s_{2,3} a_x + s_1 s_{2,3} a_y + c_{2,3} a_z}\right) \qquad \text{(D31.1)}$$

$$q_5 = \arccos\left(c_1 c_{2,3} a_x + s_1 c_{2,3} a_y - s_{2,3} a_z\right) \qquad \text{(D32.1)}$$

$$q_6 = -\arctan\left(\frac{c_1 c_{2,3} s_x + s_1 c_{2,3} s_y - s_{2,3} s_z}{c_1 c_{2,3} n_x + s_1 c_{2,3} n_y - s_{2,3} n_z}\right) \qquad \text{(D33.1)}$$

**The second solution set:**

$$q_4 = arctan\left(\frac{s_1 a_x - c_1 a_y}{c_1 s_{2,3} a_x + s_1 s_{2,3} a_y + c_{2,3} a_z}\right) + \pi \tag{D31.2}$$

$$q_5 = -\arccos\left(c_1 c_{2,3} a_x + s_1 c_{2,3} a_y - s_{2,3} a_z\right) \tag{D32.2}$$

$$q_6 = -arctan\left(\frac{c_1 c_{2,3} s_x + s_1 c_{2,3} s_y - s_{2,3} s_z}{c_1 c_{2,3} n_x + s_1 c_{2,3} n_y - s_{2,3} n_z}\right) + \pi \tag{D33.2}$$

# Appendix E

## Supplemental Derivation of Feedback Linearization Design for Outer Loops

### Section D1. Feedback Linearization for the SISO Visual System

In this section we will show explicit derivation from nonlinear systems expressed in Equations (5.25) – (5.26), to the transformed expression in Equations (5.29) – (5.31).

Equations (5.25) and (5.26) can be rewritten in the following as:

$$\tau_{in}^3 \dddot{W} + 3\tau_{in}^2 \ddot{W} + 3\tau_{in}\dot{W} + W = q_{V_1 feedback} \tag{E1}$$

$$F \cdot tan\left(\varphi + 3\tau_{in}\dot{W} + W + d_{q_{V_1}}\right) = \widehat{v_{R_1}} \tag{E2}$$

Equations (E1) and (E2) describe a nonlinear third order system, where $W$ is the state, $q_{V_1 feedback}$ is the input and $\widehat{v_{R_1}}$ is the output. $F, \tau_{in}, \varphi$ and $d_{q_{V_1}}$ are constant parameters.

Take the first derivative of Equation (E2), we have:

$$\dot{\widehat{v_{R_1}}} = F \cdot \frac{1}{\cos\left(\varphi + 3\tau_{in}\dot{W} + W + d_{q_{V_1}}\right)^2} \cdot (3\tau_{in}\ddot{W} + \dot{W}) \tag{E3}$$

where

$$\frac{d(tan\theta)}{dt} = \frac{1}{\cos^2(\theta)}, for\ any\ \theta \in [0, 2\pi] \tag{E4}$$

Take the second derivative of Equation (E2) or equivalently take the first derivative of Equation (E3), we have

$$\ddot{\widehat{v_{R_1}}} = 2F \cdot \frac{\sin\left(\varphi + 3\tau_{in}\dot{W} + W + d_{q_{V_1}}\right)}{\cos\left(\varphi + 3\tau_{in}\dot{W} + W + d_{q_{V_1}}\right)^3} \cdot (3\tau_{in}\ddot{W} + \dot{W})^2$$
$$+ F \cdot \frac{1}{\cos\left(\varphi + 3\tau_{in}\dot{W} + W + d_{q_{V_1}}\right)^2} \cdot (3\tau_{in}\dddot{W} + \ddot{W}) \tag{E5}$$

Rearrange (E1), we can obtain an expression of $\dddot{W}$ as:

$$\dddot{W} = -\frac{3}{\tau_{in}}\ddot{W} - \frac{3}{\tau_{in}^2}\dot{W} - \frac{1}{\tau_{in}^3}W + \frac{1}{\tau_{in}^3}q_{V_1 feedback} \tag{E6}$$

Plug (E6) into (E5), we can generate:

$$\ddot{\widehat{v_{R_1}}} = 2F \cdot \frac{\sin\left(\varphi + 3\tau_{in}\dot{W} + W + d_{q_{V_1}}\right)}{\cos\left(\varphi + 3\tau_{in}\dot{W} + W + d_{q_{V_1}}\right)^3} \cdot (3\tau_{in}\ddot{W} + \dot{W})^2$$
$$+ F \cdot \frac{1}{\cos\left(\varphi + 3\tau_{in}\dot{W} + W + d_{q_{V_1}}\right)^2} \cdot (-8\ddot{W} - \frac{9}{\tau_{in}}\dot{W} - \frac{3}{\tau_{in}^2}W \tag{E7}$$
$$+ \frac{3}{\tau_{in}^2} q_{V_1 feedback})$$

Rearrange (E7), it can be shown that:

$$\ddot{\widehat{v_{R_1}}} = 2F \cdot tan\left(\varphi + 3\tau_{in}\dot{W} + W + d_{q_{V_1}}\right) cos^{-2}\left(\varphi + 3\tau_{in}\dot{W} + W + d_{q_{V_1}}\right)(3\tau_{in}\ddot{W} + \dot{W})^2$$
$$-F \cdot cos^{-2}\left(\varphi + 3\tau_{in}\dot{W} + W + d_{q_{V_1}}\right)(8\ddot{W} + \frac{9}{\tau_{in}}\dot{W} + \frac{3}{\tau_{in}^2}W) \tag{E8}$$

$$+F \cdot \cos^{-2}\left(\varphi + 3\tau_{in}\dot{W} + W + d_{q_{V_1}}\right)\frac{3}{\tau_{in}^2}q_{V_1 feedback}$$

If we write (E8) in the form:

$$\ddot{\widehat{v_{R_1}}} = R(W, \dot{W}, \ddot{W}) + G(W, \dot{W}, \ddot{W})q_{V_1 feedback} \tag{E9}$$

Then:

$$\begin{aligned}R(W, \dot{W}, \ddot{W}) =\ & 2F \cdot \cos^{-2}\left(\varphi + 3\tau_{in}\dot{W} + W + d_{q_{V_1}}\right)\tan\left(\varphi + 3\tau_{in}\dot{W}\right.\\ & \left. + W + d_{q_{V_1}}\right)(3\tau_{in}\ddot{W} + \dot{W})^2 \\ & -F \cdot \cos^{-2}\left(\varphi + 3\tau_{in}\dot{W} + W + d_{q_{V_1}}\right)\left(8\dddot{W} + \frac{9}{\tau_{in}}\ddot{W} + \frac{3}{\tau_{in}^2}\dot{W}\right)\end{aligned} \tag{E10}$$

$$G(W, \dot{W}, \ddot{W}) = F \cdot \cos^{-2}(\varphi + 3\tau_{in}\dot{W} + W + d_{q_{V_1}})\frac{3}{\tau_{in}^2} \tag{E11}$$

## Section D2. Feedback Linearization for the SISO Tool Manipulation System

In this section we will show explicit derivation from nonlinear systems expressed similar as the SISO visual system, to the transformed expression in Equations (5.67) – (5.68).

The transfer function of SISO manipulation system can be expressed as:

$$\tau_{in}^3 \dddot{W} + 3\tau_{in}^2\ddot{W} + 3\tau_{in}\dot{W} + W = q_{T_1 feedback} \tag{E12}$$

$$\widehat{v_{I_1}} = F \cdot \frac{Q(W, \dot{W}) + \tan(\overline{q_{V_1}})}{1 - Q(W, \dot{W})\tan(\overline{q_{V_1}})} \tag{E13}$$

$$Q(W, \dot{W}) = \frac{L_t \sin\left(3\tau_{in}\dot{W} + W + d_{q_{T_1}}\right)}{L_{VT} - L_t \cos\left(3\tau_{in}\dot{W} + W + d_{q_{T_1}}\right)} \tag{E14}$$

Equations (E12) – (E14) describe a nonlinear third order system, where $W$ is the state, $q_{T_1 feedback}$ is the input and $\widehat{v_{I_1}}$ is the output. $F, L_{VT}, L_t, \overline{q_{V_1}}$ and $d_{q_{T_1}}$ are constant parameters.

Lets first simplify (E13) by introducing a variable $\sigma$, so that:

$$\tan(\sigma) = Q(W, \dot{W}) \tag{E15}$$

And
$$\sigma = \arctan \left( \frac{L_t \sin\left(3\tau_{in}\dot{W} + W + d_{q_{T_1}}\right)}{L_{VT} - L_t \cos\left(3\tau_{in}\dot{W} + W + d_{q_{T_1}}\right)} \right) \tag{E16}$$

Equation (E13) can be rewritten as:

$$\widehat{v_{I_1}} = F \cdot \frac{\tan(\sigma) + \tan(\overline{q_{V_1}})}{1 - \tan(\sigma)\tan(\overline{q_{V_1}})} = F \cdot \tan(\sigma + \overline{q_{V_1}}) \tag{E17}$$

Take the first derivative of Equation (E17), we have:

$$\dot{\widehat{v_{I_1}}} = F \cdot \cos^{-2}(\sigma + \overline{q_{V_1}}) \frac{d\sigma}{dt} \tag{E18}$$

Where $\frac{d\sigma}{dt}$ is the first derivative of Equation (E16), and it can be expressed as:

$$\frac{d\sigma}{dt} = \frac{1}{1 + \left(\dfrac{L_t \sin\left(3\tau_{in}\dot{W} + W + d_{q_{T_1}}\right)}{L_{VT} - L_t \cos\left(3\tau_{in}\dot{W} + W + d_{q_{T_1}}\right)}\right)^2}$$

$$\frac{L_t \cos\left(3\tau_{in}\dot{W} + W + d_{q_{T_1}}\right)\left(3\tau_{in}\ddot{W} + \dot{W}\right)\left(L_{VT} - L_t \cos\left(3\tau_{in}\dot{W} + W + d_{q_{T_1}}\right)\right)}{\left(L_{VT} - L_t \cos\left(3\tau_{in}\dot{W} + W + d_{q_{T_1}}\right)\right)^2}$$

$$- \frac{L_t^2 \sin^2\left(3\tau_{in}\dot{W} + W + d_{q_{T_1}}\right)\left(3\tau_{in}\ddot{W} + \dot{W}\right)}{\left(L_{VT} - L_t \cos\left(3\tau_{in}\dot{W} + W + d_{q_{T_1}}\right)\right)^2} \tag{E19}$$

By taking:

$$L_t^2 \sin^2\left(3\tau_{in}\dot{W} + W + d_{q_{T_1}}\right) + L_t^2 \cos^2\left(3\tau_{in}\dot{W} + W + d_{q_{T_1}}\right) = L_t^2 \tag{E20}$$

We can simplify (D19) as:

$$\frac{d\sigma}{dt} = \frac{L_t \cos\left(3\tau_{in}\dot{W} + W + d_{q_{T_1}}\right)\left(3\tau_{in}\ddot{W} + \dot{W}\right)L_{VT} - L_t^2\left(3\tau_{in}\ddot{W} + \dot{W}\right)}{L_{VT}^2 + L_t^2 - 2L_{VT}L_t \cos\left(3\tau_{in}\dot{W} + W + d_{q_{T_1}}\right)} \tag{E21}$$

Take the second derivative of Equation (E13), which is equivalent to take the first derivative of Equation (E18):

$$\ddot{\widehat{v_{I_1}}} = 2F \cdot \tan(\sigma + \overline{q_{V_1}}) \cos^{-2}(\sigma + \overline{q_{V_1}}) \left(\frac{d\sigma}{dt}\right)^2 + F \cdot \cos^{-2}(\sigma + \overline{q_{V_1}}) \frac{d\left(\frac{d\sigma}{dt}\right)}{dt} \tag{E22}$$

Where $\frac{d\sigma}{dt}$ is expressed in Equation (D21), and $\frac{d\left(\frac{d\sigma}{dt}\right)}{dt}$ is the first derivative of Equation (E21) as:

$$\begin{aligned}\frac{d\left(\frac{d\sigma}{dt}\right)}{dt} &= \frac{-L_t L_{VT} \sin\left(3\tau_{in}\dot{W} + W + d_{q_{T_1}}\right)\left(3\tau_{in}\ddot{W} + \dot{W}\right)^2}{L_{VT}^2 + L_t^2 - 2L_{VT}L_t\cos\left(3\tau_{in}\dot{W} + W + d_{q_{T_1}}\right)} \\ &+ \frac{L_t L_{VT}\cos\left(3\tau_{in}\dot{W} + W + d_{q_{T_1}}\right)\left(3\tau_{in}\dddot{W} + \ddot{W}\right) - L_t^2\left(3\tau_{in}\dddot{W} + \ddot{W}\right)}{L_{VT}^2 + L_t^2 - 2L_{VT}L_t\cos\left(3\tau_{in}\dot{W} + W + d_{q_{T_1}}\right)} \\ &- \frac{L_t\cos\left(3\tau_{in}\dot{W} + W + d_{q_{T_1}}\right)\left(3\tau_{in}\ddot{W} + \dot{W}\right)L_{VT} - L_t^2\left(3\tau_{in}\ddot{W} + \dot{W}\right)}{\left(L_{VT}^2 + L_t^2 - 2L_{VT}L_t\cos\left(3\tau_{in}\dot{W} + W + d_{q_{T_1}}\right)\right)^2} \\ &\cdot 2L_{VT}L_t\sin\left(3\tau_{in}\dot{W} + W + d_{q_{T_1}}\right)\left(3\tau_{in}\ddot{W} + \dot{W}\right)\end{aligned} \quad \text{(E23)}$$

We can simplify (E23) as:

$$\begin{aligned}\frac{d\left(\frac{d\sigma}{dt}\right)}{dt} &= \frac{L_t L_{VT}\left(L_t^2 - L_{VT}^2\right)\sin\left(3\tau_{in}\dot{W} + W + d_{q_{T_1}}\right)\left(3\tau_{in}\ddot{W} + \dot{W}\right)}{\left(L_{VT}^2 + L_t^2 - 2L_{VT}L_t\cos\left(3\tau_{in}\dot{W} + W + d_{q_{T_1}}\right)\right)^2} \\ &+ \frac{\left(L_t L_{VT}\cos\left(3\tau_{in}\dot{W} + W + d_{q_{T_1}}\right) - L_t^2\right)\left(3\tau_{in}\dddot{W} + \ddot{W}\right)}{L_{VT}^2 + L_t^2 - 2L_{VT}L_t\cos\left(3\tau_{in}\dot{W} + W + d_{q_{T_1}}\right)}\end{aligned} \quad \text{(E24)}$$

Rearrange (E12), we can obtain an expression of $\dddot{W}$ as:

$$\dddot{W} = -\frac{3}{\tau_{in}}\ddot{W} - \frac{3}{\tau_{in}^2}\dot{W} - \frac{1}{\tau_{in}^3}W + \frac{1}{\tau_{in}^3}q_{T_1 feedback} \quad \text{(E25)}$$

Plug (E25) into (E22), we can generate:

$$\ddot{\hat{v}}_{I_1} = R(W, \dot{W}, \ddot{W}) + G(W, \dot{W}, \ddot{W})q_{T_1 feedback} \quad \text{(E26)}$$

where

$$\begin{aligned}R(W, \dot{W}, \ddot{W}) &= 2F \cdot \cos^{-2}(\sigma + \overline{q_{V_1}})\tan(\sigma + \overline{q_{V_1}})\left(3\tau_{in}\ddot{W} + \dot{W}\right)^2 A^2 \\ &\quad + F \cdot \cos^{-2}(\sigma + \overline{q_{V_1}})\left(3\tau_{in}\ddot{W} + \dot{W}\right)B\end{aligned} \quad \text{(D27)}$$

$$-F \cdot \cos^{-2}(\sigma + \overline{q_{V_1}})\left(8\ddot{W} + \frac{9}{\tau_{in}}\dot{W} + \frac{3}{\tau_{in}^2}W\right)A$$

$$G(W,\dot{W},\ddot{W}) = F \cdot \cos^{-2}(\sigma + \overline{q_{V_1}})\frac{3}{\tau_{in}^2}A \tag{E28}$$

And:

$$\sigma = \arctan\left(\frac{L_t \sin\left(3\tau_{in}\dot{W} + W + d_{q_{T_1}}\right)}{L_{VT} - L_t \cos\left(3\tau_{in}\dot{W} + W + d_{q_{T_1}}\right)}\right) \tag{E29}$$

$$A = \frac{L_{VT}L_t \cos\left(3\tau_{in}\dot{W} + W + d_{q_{T_1}}\right) - L_t^2}{L_t^2 + L_{VT}^2 - 2L_tL_{VT}\cos\left(3\tau_{in}\dot{W} + W + d_{q_{T_1}}\right)} \tag{E30}$$

$$B = \frac{L_tL_{VT}\sin\left(3\tau_{in}\dot{W} + W + d_{q_{T_1}}\right)(L_t^2 - L_{VT}^2)}{(L_t^2 + L_{VT}^2 - 2L_tL_{VT}\cos\left(3\tau_{in}\dot{W} + W + d_{q_{T_1}}\right))^2} \tag{E31}$$